\documentclass[12pt,oneside]{cmuthesis}
\usepackage[utf8]{inputenc}
\usepackage[T1]{fontenc}
\usepackage[rmargin=1in,lmargin=1in,bmargin=1.5in]{geometry}
\usepackage{hyperref}
\usepackage{natbib}
\usepackage{graphicx}
\usepackage{amssymb}
\usepackage{amsmath}
\usepackage{url}
\usepackage{caption}
\usepackage{subcaption}
\usepackage{listings}
\usepackage[normalem]{ulem}
\usepackage{csquotes}
\usepackage{bbm}

\usepackage{lmodern}
\usepackage{booktabs}
\usepackage{enumitem}
\usepackage{setspace}
\usepackage{wrapfig}
\usepackage{mathtools}
\usepackage{epigraph}

\usepackage{mleftright}
\usepackage{xparse}

\usepackage[hang,flushmargin]{footmisc}

\usepackage[dvipsnames]{xcolor}

\usepackage{multirow}

\usepackage{CJKutf8}

\usepackage[acronym,shortcuts,acronymlists={hidden}]{glossaries}

\usepackage{times}
\usepackage{latexsym}
\newlength\longest

\newcommand{\eg}{\textit{e.g.}}
\newcommand{\ie}{\textit{i.e.}}
\newcommand{\etc}{\textit{etc}}
\newacronym{nlp}{NLP}{Natural Language Processing}
\newacronym{nmt}{NMT}{Neural Machine Translation}
\newacronym{smt}{SMT}{Statistical Machine Translation}
\newacronym{mt}{MT}{Machine Translation}
\newacronym{lm}{LM}{Language Modeling}
\newacronym{bpe}{BPE}{Byte-Pair Encoding}
\newacronym{dataset}{MTNT}{Machine Translation of Noisy Text}
\newacronym{mtnt}{MTNT}{Machine Translation of Noisy Text}
\newacronym{enfr}{\texttt{en-fr}}{English-French}
\newacronym{fren}{\texttt{fr-en}}{French-English}
\newacronym{enja}{\texttt{en-ja}}{English-Japanese}
\newacronym{jaen}{\texttt{ja-en}}{Japanese-English}
\newacronym{oov}{OOV}{out-of-vocabulary words}
\newacronym{rdb}{RDchrF}{relative decrease in chrF}
\newacronym{seq2seq}{seq2seq}{sequence-to-sequence}
\newacronym{nll}{NLL}{negative log likelihood}
\newacronym{dro}{DRO}{distributionally robust optimization}
\newacronym{mlp}{MLP}{multi-layer perceptron}
\newacronym{erm}{ERM}{empirical risk minimization}
\newacronym{kl}{KL}{Kullback-Leibler}
\newacronym{p-dro}{P-DRO}{Parametric DRO}
\newcommand{\emoji}[1]{\includegraphics[width=1em]{emoji_images/#1.png}}
\newcommand{\ja}[1]{\begin{CJK}{UTF8}{min}#1\end{CJK}}

\definecolor{Code}{rgb}{0,0,0}
\definecolor{Decorators}{rgb}{0.5,0.5,0.5}
\definecolor{Numbers}{rgb}{0.5,0,0}
\definecolor{MatchingBrackets}{rgb}{0.25,0.5,0.5}
\definecolor{Keywords}{rgb}{0,0,1}
\definecolor{self}{rgb}{0,0,0}
\definecolor{Strings}{rgb}{0,0.63,0}
\definecolor{Comments}{rgb}{0,0.63,1}
\definecolor{Backquotes}{rgb}{0,0,0}
\definecolor{Classname}{rgb}{0,0,0}
\definecolor{FunctionName}{rgb}{0,0,0}
\definecolor{Operators}{rgb}{0,0,0}
\definecolor{Background}{rgb}{0.98,0.98,0.98}
\lstdefinelanguage{Python}{numbers=left,numberstyle=\footnotesize,numbersep=1em,
xleftmargin=1em,framextopmargin=2em,framexbottommargin=2em,showspaces=false,
showtabs=false,showstringspaces=false,frame=l,
tabsize=4,
basicstyle=\ttfamily\small\setstretch{1},backgroundcolor=\color{Background},
commentstyle=\color{Comments}\slshape,
stringstyle=\color{Strings},morecomment=[s][\color{Strings}]{"""}{"""},morecomment=[s][\color{Strings}]{'''}{'''},morecomment=[l][\color{Strings}]{\#},
morekeywords={import,from,class,def,for,while,if,is,in,elif,else,not,and,or,print,break,continue,return,True,False,None,access,as,,del,except,exec,finally,global,import,lambda,pass,print,raise,try,assert},keywordstyle={\color{Keywords}\bfseries},
morekeywords={[2]@invariant,pylab,numpy,np,scipy},keywordstyle={[2]\color{Decorators}\slshape},emph={self},emphstyle={\color{self}\slshape},
}

\newcommand\fren{\texttt{fr-en}}
\newcommand\deen{\texttt{de-en}}
\newcommand\csen{\texttt{cs-en}}
\newcommand\unk{\texttt{<unk>}}

\DeclareMathOperator*{\argmax}{arg\,max}
\DeclareMathOperator*{\argmin}{arg\,min}

\newcommand\codeurl{\url{https://github.com/pmichel31415/translate/tree/paul/pytorch_translate/research/adversarial/experiments}}
\newcommand\teapoturl{\url{https://github.com/pmichel31415/teapot-nlp}}

\newcommand\unconstrained{Unconstrained}
\newcommand\knn{kNN}
\newcommand\unkonly{CharSwap}

\newcommand{\KL}[2]{\text{KL}({#1} \Vert {#2})}

\DeclareMathOperator*{\w}{\mathbf{w}}
\DeclareMathOperator{\Ell}{\mathcal{L}}
\DeclareMathOperator{\E}{\mathbb{E}}
\DeclareMathOperator*{\Ladv}{\mathcal L_{\text{adv}}}
\newcommand{\bigO}[1]{\mathcal{O}(#1)}
\DeclareMathOperator*{\stgt}{s_{\text{tgt}}}
\DeclareMathOperator*{\ssrc}{s_{\text{src}}}
\DeclareMathOperator*{\dtgt}{d_{\text{tgt}}}

\NewDocumentCommand{\evalat}{sO{\big}mm}{%
  \IfBooleanTF{#1}
   {\mleft. #3 \mright|_{#4}}
   {#3#2|_{#4}}%
}

\newcommand{\davidson}{\textsc{DWMW17}}
\newcommand{\founta}{\textsc{FDCL18}}

\usepackage[titles]{tocloft}
\title{ 
{\bf Learning Neural Models\\for Natural Language Processing\\in the Face of Distributional Shift}}
\author{Paul Michel}
\date{July 2021}
\Year{2021}
\trnumber{CMU-LTI-21-013}

\committee{
\begin{tabular}{rl}
Graham Neubig (Chair) & Carnegie Mellon University\\
Zico Kolter & Carnegie Mellon University \\
Zachary Lipton & Carnegie Mellon University \\
Tatsunori Hashimoto & Stanford University
\end{tabular}
}

\begin{document}

\maketitle

\clearpage

\thispagestyle{empty}
\null\vfill

\settowidth\longest{\large\itshape Mais, pour voir, il convient d’abord de participer.}
{\centering
\parbox{\longest}{%
  \raggedright{\large\itshape%
    Connaître, ce n’est point démonter, ni expliquer. \\
    C’est accéder à la vision. \\
    Mais, pour voir, il convient d’abord de participer. \\
    Cela est dur apprentissage\ldots\par\bigskip
  }   
  \raggedleft\normalsize\sc{Antoine de Saint Exupery}\par%
}
}

\vfill\vfill

\clearpage

\vfill
\section*{Abstract}
The dominating NLP paradigm of training a strong neural predictor to perform one task on a specific dataset has led to state-of-the-art performance in a  variety of applications (eg. sentiment classification, span-prediction based question answering or machine translation). However, it builds upon the assumption that the data distribution is stationary, ie. that  the data is sampled from a fixed distribution both at training and test time. This way of training is inconsistent with how we as humans are able to learn from and operate within a constantly changing stream of information. Moreover, it is ill-adapted to real-world use cases where the data distribution is expected to shift over the course of a model's lifetime.

The first goal of this thesis is to characterize the different forms this shift can take in the context of natural language processing, and propose benchmarks and evaluation metrics to measure its effect on current deep learning architectures. We then proceed to take steps to mitigate the effect of distributional shift on NLP models. To this end, we develop methods based on parametric reformulations of the distributionally robust optimization framework. Empirically, we demonstrate that these approaches yield more robust models as demonstrated on a selection of realistic problems. In the third and final part of this thesis, we explore ways of efficiently adapting existing models to new domains or tasks. Our contribution to this topic takes inspiration from information geometry to derive a new gradient update rule which alleviate catastrophic forgetting issues during adaptation.
\vfill

\clearpage
\section*{Acknowledgements}

This thesis would not have been made possible without the continuous support, involvement and help from my amazing PhD supervisor Graham Neubig. He spent countless hours discussing research with me, helping me with paper writing and was overall very supportive throughout my PhD. I realize now that I was extremely lucky to have such a thoughtful and conscientious advisor. It was a great pleasure to be working with him all these years.

I also want to thank the members of my committee, Zico Kolter, Zachary Lipton and Tatsunori Hashimoto for their insightful feedback which helped shape this document into its current form. I am particularly grateful for Tatsu's involvement, as he became a close collaborator over the last year or so. I must also thank the many anonymous reviewers who gave feedback on the work presented here. Our relationship over the years had its ups and downs, but overall the discussions we had had a net positive impact on my work, and for that I am thankful.

Of course, the work presented here was the product of collaboration with all my co-authors Xian Li and Juan Pino (both of whom I had the pleasure to intern with at Facebook in 2018), Liz Salesky (our discussions at the JSALT workshop in 2019 helped spur the work presented in Chapter \ref{ch:regularizing_trajectories}) and Lucia Specia (also met at JSALT) and the entire team behind the two WMT robustness shared tasks.

I am also very lucky to have had the opportunity to collaborate with a host of brilliant researchers on other papers which didn't make it in this document: in no particular order Jaime Carbonell, Xinyi Wang, Hieu Pham, Omer Levy, Danish, Ameet Talwalkar, Antonios Anastasopoulos, Ziyi Dou and Junjie Hu. I was also very happy to work with Keita Kurita and Lucio Dery, both brilliant students and who made our collaborations together enjoyable first experiences in advising more junior researchers. I am also thankful to the Deepmind language team (and in particular my hosts Dani Yogatama and Sebastian Ruder) who hosted me for a remote internship in the middle of a global pandemic.

I attribute the main reason for my pleasurable PhD experience to the fantastic environment at CMU and at the LTI. I was happy to be surrounded by other remarkable students within our research group (NeuLab), in particular Chunting Zhou, Pengcheng Yin, Chaitanya Malaviya and Shruti Rijhwani, as well as all the others who joined us over the years. I was also lucky to make long lasting friends at CMU: Abhilasha Ravichander, Siddarth Dalmia, Rajat Kulshreshtha, Chirag Nagpal, Evangelia Spiliopoulou, Artidoro Pagnoni, Mariia Ryskina, Shruti Palaskar, Raphael Olivier, Xinya Li, James Fiacco and all the others. I'm also grateful for the wonderful people I met at various conferences (back in the olden days of in-person conferences), in particular Kris Cao, Kazuya Kawakami, Roee Aharoni, Gail Weiss and Jad Kabbara.

These acknowledgments would not be complete without a mention of the Pittsburgh food and drinks scene. Many a research discussion was had around a drink at Peter's pub, Stack'd, Fuel and Fuddle, Mario's or Hemingway's. I also have fond memories of lunches/dinners/coffee breaks at the Chipotle on Forbes, Sushi Fuku, Quiznos, the indian foodtruck, Porch, Tazza de Oro and Arriviste to mention only a few. They kept my belly full and my wallet flat. Finally, a special mention goes to the Carnegie Museum of Natural History whose incredible dinosaur exhibit (and good coffee) was a welcome breath of fresh air while I was putting the finishing touches to this thesis.

I must also acknowledge the continued support of my parents, Matthieu and Véronique. They have given me every opportunity I needed to go as far as I could in my studies. I realize that this is a privilege that many people I know (and many more I don't) did not have.

Last, but perhaps foremost, I would not have been able to go through these PhD years without the unconditional love and support from my then girlfriend, now wife, Clémence. You endured five long years of distance to make this possible, yet somehow you were always close by my side. This thesis is, of course, dedicated to you.

\clearpage

\tableofcontents

\chapter{Introduction}
\label{sec:introduction}

The ability to use and comprehend language as a mean of communication is a unique characteristic of human intelligence. Natural language is fundamentally non-stationary: the way we produce utterances is modulated by a myriad of factors that are ever changing based on our interlocutors, social context, emotions, or communication medium used. Our capacity to operate in and adapt to such a dynamic environment is another defining feature of intelligence. Indeed from their youngest age, infants are seldom placed in static situations, yet they are still able to learn emerging patterns from a constant flow of information (see \eg{} \citet{dupoux2018cognitive}). Hence, it is a natural goal for artificial intelligence research to develop agents that are capable of acquiring natural language understanding and generation abilities in the face of a changing environment.

The development of computer systems which can understand and produce natural language is realized through \ac{nlp}. In recent years, neural-network based machine learning models have been the source of rapid progress in \ac{nlp}, and now represent the state-of-the-art in a large number of real-world applications (such as machine translation \citep{wu2016google}, named entity recognition \citep{lample2016neural} or speech recognition \citep{xiong2017microsoft}), reaching human parity on a variety of benchmarks \citep{glueleaderboard2020,squadleaderboard2020}\footnote{Meaning that they outperform human annotated baselines, we make no claim that these systems have achieved any kind of human-like intelligence.}. Yet, these models exhibit a pattern of failures when confronted with changes to their training environment:
\begin{itemize}
    \item They are highly sensitive to \emph{domain shift}: they perform poorly when confronted with text from a different domain, for example text concerning different topics \citep{gururangan-etal-2020-dont}, demographics \citep{blodgett2016demographic,amodei2016deep,hovy2015tagging}, and even using different data collection processes \citep{gururangan2018annotation}. In particular, these models often perform poorly when evaluated on sub-populations,\footnote{Domains that exist, but are under-represented in their training data \citep{sagawa2019distributionally}.} and they can latch on to spurious correlations \citep{mccoy2019right}.
    \item They are vulnerable to specific \emph{word or character-level perturbations} \citep{papernot2016crafting,ebrahimi2018hotflip,belinkov2018synthetic}. This phenomenon poses security concerns, as this opens up trained models to various types of attacks by nefarious agents: \textit{e.g.} bypassing toxicity detection models \citep{hosseini2017deceiving}.
    \item When they are trained on new tasks or domains, they rapidly forget previously acquired knowledge, a phenomenon known as \emph{catastrophic forgetting} \citep{mccloskey1989catastrophic,kirkpatrick2017overcoming}. This makes adaptation to new data distributions an arduous and unreliable process necessitating significant amounts of domain expertise and regularization \citep{micelibarone-EtAl:2017:EMNLP2017}.
\end{itemize}

These shortcomings are not only at odds with the natural behaviour observed in intelligent agents mentioned above, but represents a liability for practical users and designers of such systems. Indeed, machine learning models trained on static datasets can fail spectacularly in the face of real-world situations, sometimes with dire consequences (for example, this mis-translation of ``good morning'' into ``attack them'' due to dialectic variations; \citet{hern2017facebook}). More insidiously, the representation disparity between training datasets and the real world can, for example, unfairly affect and further disenfranchise minority groups \citep{hashimoto2018fairness,dixon2018measuring}.

The principal desired characteristic of the machine learning models used in \ac{nlp} is the ability to learn and generalize from a finite number of training examples. The most common learning paradigm, empirical risk minimization, relies on the assumption that this training data is sampled from a fixed distribution, and generalization is construed as the ability to perform well on unseen samples from this same distribution. The failure cases described above occur when the model operates in a different distribution than the one it was trained on, in which case standard generalization guarantees break down. In this context, the discrepancy between a machine learning model's learning and operating environment can be described as \emph{distributional shift}. The research presented in this thesis focuses on what we believe to be the three major axes of progress towards addressing distributional shift in the context of \ac{nlp}: \emph{Evaluation}, \emph{Robustness} and \emph{Adaptation}.

We start with \textbf{evaluation} because distributional shift is particularly difficult to characterize and measure, especially in regard to natural language. This is partly due to the absence of a canonical metric structure of the data. In other words, it is not clear how to efficiently measure the semantic similarity between two sentences and as a result there is no straight-forward way to measure the discrepancy between two samples, let alone two distributions. Therefore, as a natural first step in addressing distributional shift, we propose a new benchmark (Chapter \ref{ch:mtnt}) and evaluation metric (Chapter \ref{ch:on_evaluation_of_adversarial}) to evaluate domain shift and robustness to adversarial perturbations respectively. With these tools in hand, we set out to construct \textbf{robust} models that are trained to be less sensitive to distributional shift, even in the absence of explicit information on the nature of shift. This is done by leveraging the diversity of data within the training distribution to ensure uniform performance over a variety of domains present in the training data (sub-populations). Specifically we formulate a parametric version of the \acrlong{dro} framework which allows for training models that are more robust to sub-population shift (Chapters \ref{ch:modeling_the_second_player_dro} and \ref{ch:r_pdro}). Finally, learning in a static environment is fundamentally sub-optimal: we cannot expect our models to perform well on each and every possible future setting, and we must be able to \textbf{adapt} them to any new scenario we encounter. Consequently, we research mechanisms by which we are able to fine-tune a trained model on new evidence, without forgetting previously acquired knowledge (Chapter \ref{ch:regularizing_trajectories}).

The detailed outline of our contributions in this thesis is presented below:

\begin{itemize}
    \item \textbf{Chapter \ref{ch:background}} lays out a description of the distributional shift phenomenon and discusses previous and current work in this general area.
    \item \textbf{Part I: Measuring and Evaluating Robustness}
    \begin{itemize}
        \item In \textbf{Chapter \ref{ch:mtnt}}, we introduce a benchmark for evaluating machine translation models against distributional shift. Specifically, the proposed dataset --- dubbed MTNT --- provides a testbed for translation of noisy social media text, a challenging domain for common state-of-the-art models trained on news and parliamentary proceedings. This work was published as \cite{michel18mtnt} at EMNLP 2018\footnote{\url{https://www.aclweb.org/anthology/events/emnlp-2018/}}, \cite{li2019findings} at WMT 2019\footnote{\url{https://www.statmt.org/wmt19/robustness.html}} and \citep{specia2020findings} at WMT 2020.\footnote{\url{https://www.statmt.org/wmt20/robustness.html}}
        \item In \textbf{Chapter \ref{ch:on_evaluation_of_adversarial}}, we examine the difficulty of evaluating adversarial perturbations on \ac{nlp} models, with an emphasis on sequence-to-sequence models. We propose an evaluation framework centered around the idea of evaluating the perturbations' effect on the semantics of the input sentences. The content of this chapter originally appeared as \cite{michel19evaluation} at NAACL 2019.\footnote{\url{https://www.aclweb.org/anthology/events/naacl-2019/}}
    \end{itemize}
    \item \textbf{Part II: Making Robust Models}
    \begin{itemize}
        \item \textbf{Chapter \ref{ch:modeling_the_second_player_dro}} describes an approach for training models that are robust to distributional shift, using a parametric version of the \ac{dro} framework. Specifically, we train a model to minimize its expected loss under the distribution defined by another generative model (the ``adversary''), which is itself trained to represent the worst-case distribution for the model. In experiments, we find that this approach (called \acl{p-dro}) yields models that are more robust compared to other baselines in the \ac{dro} literature. This chapter was previously published at ICLR 2021\footnote{\url{https://iclr.cc/Conferences/2021}} as \citet{michel2021modeling}.
        \item \textbf{Chapter \ref{ch:r_pdro}} presents an improvement over the P-DRO approach whereby a parametric \emph{re-weighting} of the training data is learned. With appropriate constraints, and a simple but crucial minibatch-level renormalization trick, we find that this results in a simple \ac{dro} approach which reliably outperforms other baselines and necessitates little hyper-parameter tuning. The work presented in this chapter is currently undergoing peer review.
    \end{itemize}
    \item \textbf{Part III: Adaptation}
    \begin{itemize}
        \item In \textbf{Chapter \ref{ch:regularizing_trajectories}}, we develop a technique for mitigating the catastrophic forgetting issue that arises when adapting a model to a new domain or a new task. The approach, inspired by ideas in information geometry, is demonstrated to drastically reduce forgetting compared to regular gradient descent, and enjoys complementarity with common continual-learning baselines. This chapter contains work that was released as a preprint.
    \end{itemize}
\end{itemize}
\clearpage

\chapter{Background and Literature Review}
\label{ch:background}

From a broad perspective, the goal of machine learning is to produce a model that is able to perform a given task, such as predicting an output variable $y$ from an input, $x$ (\eg{} sentiment given a movie review, or French translation given an English sentence\ldots) given finite amounts of training data (in our example, $(x,y)$ pairs). In this chapter, we formalize this setting for the rest of this thesis.

\section{Terminology and Definitions}
\label{sec:terminology}

In machine learning, we are confronted with learning to perform a task given a finite number of examples. In the context of this thesis, a task $T$ will be referring to a triplet containing two measurable spaces $\mathcal X$ and $\mathcal Y$, as well as a probability distribution $p$ with support on the Cartesian product $\mathcal X\times \mathcal Y$. $\mathcal X$ represents the space of all possible inputs (for example the space of all $32\times 32$ images, or the space of all English sentences), while $\mathcal Y$ stands for the space of all admissible outputs (for example the set $\{\text{positive}, \text{negative}\}$ in sentiment classification, or all sentences of a target language in translation). Finally, $p$ will represent the true distribution of the data in nature (for example it will assign high probability to a pair of English and French sentences that are translations of each other).

In general, learning a task will consist of training a model to approximate the conditional distribution $p(y\mid x)$, based only on a finite number of training examples $x,y\sim p$. Specifically, we will focus on parametric models, \ie{} models that are fully determined by a finite, constant (with respect to the size of the training data) number of parameters described by a vector $\theta$ that can take its values in $\Theta$, the set of all possible parameters (in practice $\mathbb R^d$ or a subset thereof, where $d$ is the total number of parameters).

Traditionally, given a loss function $\ell: \mathcal X\times\mathcal{Y}\times\Theta\mapsto \mathbb R^+$ measuring the discrepancy between model $\theta$'s prediction on $x$ and the reference output $y$, the modeler will estimate parameters $\theta^*$ that better fit the distribution $p$ data by minimizing the expected risk
\begin{equation}
    \mathcal{L}_{\text{Expected Risk}}(\theta)=\mathbb E_{(x,y)\sim p}\ell_\theta(x,y).
\end{equation}
Note that we abuse notation and write $\ell_\theta(x,y)$ for $\ell(x,y,\theta)$ to distinguish between the dependency over $\theta$, the model parameters, and $(x, y)$, the model's inputs. As implied earlier, one cannot compute the total expectation $\mathbb E_{x,y\sim p}$ and we must resort to minimizing the empirical risk computed over a finite training set ${(x_1,y_1),\ldots,(x_n, y_n)}$
\begin{equation}
    \mathcal{L}_{\text{Empirical Risk}}(\theta)=\frac 1 n \sum_{i=1}^n\ell_\theta(x_i,y_i)
\end{equation}

\section{Distributional Shift}
\label{sec:distributional_shift}

\subsection{What is Distributional Shift}

In the \acf{erm} paradigm, machine learning models are trained to minimize their expected loss under a single, static distribution. However, this assumption is not always valid in practical applications. For example, a machine translation system trained on a majority of news article might come to be used to translate social media comments. While the overall task is can arguably be considered the same at a high level from the point of view of a human, there are significant differences in the data distribution (for instance, social media comments may contain more informal language).

In the most general sense, distributional shift (or distribution shift) is simply an instance of the test distribution $q$ being different from the training data distribution $p$. Of course, this definition is too broad to be realistic, let alone useful in practice.\footnote{Consider, in a binary classification task, setting $q(y\mid x)=1-p$. Only the distribution changes, yet the two tasks are directly at odds with each other.} Precisely defining and categorizing what characterizes an acceptable shift is still very much an active area of research.

\subsection{Categorizing Distributional Shift}

While we will not attempt to formulate a complete taxonomy of distributional shifts, we will paint a broad strokes picture of how they are categorized in the literature. We will distinguish two main schools of thought for distinguishing distributional shift: a ``mechanistic'' and a ``causal'' approach.

\subsubsection{Mechanistic Categorization}

The first approach is to distinguish shifts by describing, in mathematical terms, the mechanics of how the change manifests in the data generation process. The appeal of this categorization is that it allows for a more principled, theoretical approach to tackling distributional shift. A summary of the examples described in this section can be found in Table \ref{tab:mechanistic_categorization}.

\begin{table}
\begin{center}
\caption{\label{tab:mechanistic_categorization} Examples of ``mechanistic'' categorizations of distributional shift.}
\begin{tabular}{llcl}
 & $p(x,y)$ & $\longrightarrow$ & ${\color{red}q(x,y)}$ \\
\midrule
Covariate shift & $p(x)~p(y\mid x)$ & $\longrightarrow$ & ${\color{red}q(x)}~p(y\mid x)$ \\
Label shift & $p(x\mid y)~p(y)$ & $\longrightarrow$ & $p(x\mid y)~{\color{red}q(y)}$ \\
Concept shift & $p(x)~p(y\mid x)$ & $\longrightarrow$ & $p(x~){\color{red}q(y\mid x)}$ \\
Sub-population shift & $\sum_{i=1}^K\alpha_i~p_i(x, y)$ & $\longrightarrow$ & $\sum_{i=1}^K{\color{red}\beta_i}~p_i(x, y)$ \\
\end{tabular}
\end{center}
\end{table}

The most well known instance is \emph{covariate shift} \citep{shimodaira2000improving,sugiyama2005input,storkey2009training,bickel2009discriminative}, where only the distribution over the input space $\mathcal X$ (sometimes called the space of ``covariates'') changes, and the conditional distribution $p(y\mid x)$ is assumed to stay the same. Covariate shift is a rather intuitive assumption, especially in such cases where there is a clear causal relationship between the covariates $x$ and the targets $y$. It holds an important place within the distribution shift literature because its particular characteristics justify the use of specific algorithms, some of which enjoy appealing properties. For instance, it can be tackled by means of re-weighting with the covariate likelihood ratio $q(x)/p(x)$ \citep{shimodaira2000improving}, and the covariate shift assumption can yield tighter guarantees \citep{duchi2019distributionally}.

Another, diametrically opposed type of shift is \emph{label-shift} (or \emph{prior probability shift} in \citet{storkey2009training}), where only the marginal on $\mathcal Y$ changes, and the reverse conditional $p(x\mid y)$ stays fixed. This is most relevant when the causal relationship between $x$ and $y$ is reversed, for example when diagnosing an illness ($y$) from its symptoms ($x$). Specific methods have been developed to tackle this scenario \citep{zhang2013domain,lipton2018detecting}, although their application is generally less common in the \ac{nlp} literature. Finally, a comparatively less well studied occurrence is \emph{concept-shift}, in which the marginal $p(x)$ is assumed fixed, but the conditional changes to another distribution $q(y\mid x)$ \citep{quinonero2009dataset}. This can result from a mismatch between annotators: for instance in machine translation, different professional annotators may use slightly different translation conventions. \citet{widmer1996learning} postulates that this is due to the presence of some ``unknown context'', such as the two translators having had a different education, which leads to them adhering to two mildly different definitions of the translation problem. A concrete example of this phenomenon in toxicity detection (in the English language) is the ``n-word'' problem. This racial slur and other historically charged terms such as ``queer'' are often used disparagingly towards certain underrepresented minorities. However some of them were re-appropriated by minority speakers within their respective communities \citep{rahman2012n}. Without information about the speaker, it can be difficult to determine the toxicity of an utterance containing such words, even for human annotators \citep{sap2019risk}.

Another, more elaborate example of a ``mechanistically defined'' category of distribution shift is what \citet{koh2020wilds} calls ``sub-population shift''. In sub-population shift, it is assumed that the training distribution $p$ can be decomposed as a mixture of $K$ different domains $p=\alpha_1q_1+\ldots+\alpha_Kq_K$, $\sum_i\alpha_i=1$. This can be the case by design: for example the training data for the WMT machine translation competition\footnote{\url{http://statmt.org/wmt21/}} is aggregated from many different machine translation corpora, or the MultiNLI textual entailment dataset \citep{williams18multinli} was created specifically to contain data from different domains. But in many more cases the training data may be an aggregate of various domains, unbeknownst to the dataset's creator: for instance text datasets often contain data from different dialects (\eg{} American vs. British English, French vs. Canadian French, etc\ldots). The benchmarks aginst which the methods described Part II of this thesis will be evaluated are all examples of sub-population shifts.

\subsubsection{Causal Categorization}

An alternate view is to distinguish the distribution shifts through the real world phenomena that cause them. This more empirical approach has the merit of describing realistic cases of distribution shift which may not clearly (or provably) fit into the mathematically defined categories presented in the previous section, but are prevalent enough to warrant the development of bespoke solutions.

A predominant cause of distributional shift is change over time \citep{kelly1999impact,hand2006classifier}. For instance, \citet{bloomfield1933language} documents nine categories of lexical semantic drift that words may undergo over time\footnote{The various causes of this drift are elaborated upon in later studies \citep{blank2013historical,grzega2007english}}. While not laying out a precise typology, \citet{mcculloch2020because} also touches on a variety of semantic shift that occurred more recently (and at a comparatively faster pace) with the advent of the internet. The effect of these changes in meaning on word embedding models such as Word2Vec (\citet{mikolov2013efficient}, a staple of the \ac{nlp} pipeline at the time) were summarized in \citet{kutuzov2018diachronic}. Identifying time as the underlying cause for this phenomenon has the benefit of suggesting solutions tailored to the problem at hand: for example, one could use dynamic word-embedding models \citep{rudolph2018dynamic} to predict upcoming drift, and use this knowledge to augment the training data. In a more recent example, \citet{lazaridou2021pitfalls} highlights the specific importance of ``dynamic evaluation''\footnote{The process of continuously training a language model at test time.} \citep{mikolov2010recurrent} for mitigating loss in performance due to time shift in language models.

In other examples, distributional shift can be described in terms of data collection: a prime example is what \citet{storkey2009training} calls \emph{biased sampling}. Subtle design choices at the time of the creation of the training dataset (such as data source, pre-processing, or annotation scheme) can introduce biases into the training distribution. A recent illustration of biased sampling is the case of crowd-sourced natural language inference\footnote{Natural language inference is another name for the task of recognizing textual entailment. In its most common formulation, the model is given two sentences, the premise and the hypothesis, and is asked to determine whether the first entails or contradict the second (or whether there is no clear entailment relationship).} datasets \citep{gururangan2018annotation}: annotators tasked with crafting a contradictory hypothesis from a premise would frequently resort to introducing negation markers (in English: ``not'', ``never'', etc\ldots). This introduced a spurious correlation between the presence of such negative words and the ``contradiction'' label. When confronted with test data that doesn't conform to this pattern (\eg{} cases of entailment where the hypothesis does include a negation marker), models trained with \ac{erm} break down \citep{sagawa2019distributionally}.

\subsubsection{Local and Global Shifts}

Aside from the two types of categorization described above, we believe that it is worth making another distinction between what we will call ``local'' and ``global'' shifts, purely by virtue of the fact that they are both the subject of considerable, yet distinct areas of the literature. We use the term \emph{local} to refer to cases where the shifted distribution $q$ can be obtained by applying a transformation (deterministic or non-deterministic) to data points in $p$. Of course, this isn't a very clearly defined category since any distribution can be transported into another one via an optimal transport map. In this case, we refer specifically to shifts that are best explicitly defined by such ``local'' transformations. This category encompasses phenomena such as character or word-level noise (of the kind described in \citet{belinkov2018synthetic} for instance) or adversarial perturbations \citep{papernot2016crafting,ebrahimi2018hotflip,Ebrahimi2018OnAE,cheng2018seq2sick}, small modifications to the input of a machine learning model that cause dramatic changes to its output.

The notion of \emph{global} shift will then be used to refer to those type of distributional shift that cannot easily be modeled as local shifts. We make this distinction because, as will be discussed in Chapter \ref{ch:on_evaluation_of_adversarial}, local shift can --- and should --- be addressed specifically.

\section{Robustness to Distributional Shift}
\label{sec:robustness}

\subsection{Distributionally Robust Optimization}

There is a rather large body of literature devoted to training models that are robust to distributional shift, under the name of \acf{dro}. In \ac{dro}, models are trained not to minimize their expected risk under any one distribution $p$, but rather their worst case risk under a pre-determined family of distributions $\mathcal{Q}$, called the ``uncertainty set''
\begin{equation}\label{eq:dro_problem}
    \min_{\theta\in\Theta}\max_{q\in\mathcal Q}\underbrace{\E_{(x,y)\sim q}\ell_\theta(x,y)}_{\mathcal L_{\text {DRO}}(\theta)}.
\end{equation}

The \ac{dro} loss in Equation \ref{eq:dro_problem} provides an upper bound on the expected loss of the model under any distribution in the uncertainty set $\mathcal Q$, which motivates the use the solution of the min-max problem as a robust model. However this objective is only useful insofar that $\mathcal Q$ (1) covers test distributions of interest (corresponding to different domains, demographics, etc.) and (2) is not overly pessimistic. To fulfil this second condition, there should exist some model $\theta^*$ that achieves low loss simultaneously on the test distribution as well as $\mathcal Q$. This often requires that $\mathcal Q$ only contain distributions that are ``close'' to the training distribution under some divergence metric, or that are restricted to covariate shifts.

In some cases, it is conceivable that $\mathcal Q$ is available at training time. For instance, we might have access to some amount of data in $K$ domains of interest, in which case the 
\ac{dro} loss can be written as a maximum over the finite set $\mathcal Q=\{q_1,\ldots,q_K\}$. This setting is sometimes referred to as Group-\ac{dro} and has been studied in \citet{oren2019distributionally,sagawa2019distributionally,zhou2021examining}.

In the absence of explicit information about the domains of interest, it is up to the practitioner to carefully define this uncertainty set. There is substantial existing work on nonparametric formulations of \ac{dro}, where $\mathcal{Q}$ is expressed as a divergence ball centered at the training distribution. Popular examples in the literature are:
\begin{itemize}
    \item {\bf $f$-divergence balls}, also called $\phi$-divergences or R\'{e}nyi divergences \citep{cover1999elements,van2014renyi} are a family of divergence metrics of the form $d_f(p,q)=\E_qf(\frac p q)$ where $f: \mathbb R_+\rightarrow \mathbb R$ is a convex function mapping 1 to 0, and $q$ is assumed to be absolutely continuous to $p$ ($p(x, y) = 0 \Rightarrow q(x, y)$). Well-known representatives include the \ac{kl} divergence ($f(u)=u\log u$), Pearson's $\chi_2$ divergence ($f(u)=\frac 1 2(u-1)^2$) or the total variation distance ($f(u)=\frac 1 2 |u-1|$). They have seen extensive use in the \ac{dro} literature  \citep{ben2013robust,hu2013kullback,faury2020distributionally}.
    \item {\bf Wasserstein/IPM balls}. The 1-Wasserstein distance or earth mover's distance between two probability distributions $p$, $q$ on a metric space $\mathcal Z$ is defined as $W_1=\inf_{\pi\in\Pi}\E_{z, z'\sim\pi}\rho(z, z')$ where $\Pi$ is the set of all joint distributions on $\mathcal Z \times \mathcal Z$ with marginals $p$ and $q$, and $\rho$ is $\mathcal Z$'s canonical metric. Contrary to $f$-divergences, the Wasserstein metric makes no assumptions of absolute continuity, and it allows practitioners to design problem specific uncertainty sets by choosing the most relevant distance metric $\rho$. This has made it a popular choice for \ac{dro} problems where such metrics are well defined (\eg{} for real valued inputs), particularly in the context of adversarial robustness \citep{gao2016distributionally,esfahani2018data,sinha2017certifying}. Integral probability metrics (IPMs) are another common class of divergences between probability distributions to which (under some conditions, see Theorem 11.8.2 in \citet{dudley2018real}) the Wasserstein distance belongs. They take the form $\gamma_{\mathcal F}(p, q) = \sup_{h\in \mathcal F}\left|\E_ph - \E_q h\right|$ where $\mathcal F$ is a set of real valued bounded measurable functions on $\mathcal X\times \mathcal Y$. Although Wasserstein-DRO is are by far more common in the literature, there has been some recent work on general IPM bounded uncertainty sets \citep{husain2020distributional}.
    \item \textbf{Moment constraints} consists in uncertainty sets where admissible distribution have to satisfy some (generally linear or quadratic) constraint on their moments, \eg{} $\Vert \E_{q}-\mu\Vert\leq M$ for some pre-specified values $\mu$ and $M$. They are perhaps one of the oldest formulations of \ac{dro}, with roots in operations research \citep{scarf1957minmax,dupavcova1987minimax,delage2010distributionally} and with some exceptions \citep{nguyen2020robust}, they have seen relatively little use in the modern machine learning context.
    \item \textbf{CVaR constraints} (or $\alpha$-coverage). Originating from the financial risk management literature \citep{rockafellar2000optimization}, the conditional value at risk (CVaR) is concerned with guaranteeing good performance under all distributions $q$ that $\alpha$-covered by $p$, meaning that for all $(x,y)$, $q(x, y)\leq \alpha p(x, y)$ with $\alpha\in]0,1]$. This formalizes in the language of probability distributions the idea that $q$ is a subset that covers only a fraction $\alpha$ of the training data. CVaR-constrained uncertainty sets underpin a variety of recently proposed approaches for robust machine learning, even if they are not always explicitly acknowledged as such \citep{fan2017learning,curi2020adaptive,levy2020large}.
\end{itemize}

All these nonparametric approaches are appealing as they require very little domain-specific knowledge, have well-understood theory \citep{duchi2018learning}, and optimization procedures (\eg{} \citet{hu2013kullback} for KL constraints and \citet{levy2020large} for $\chi_2$ and CVaR constraints). We refer to \citep{rahimian2019distributionally} for a more in-depth overview.

Unfortunately, nonparametric \ac{dro} algorithms suffer from being overly pessimistic. Their uncertainty sets tend to include distributions that are exceedingly difficult to learn, or not representative of real-world distribution shifts. Furthermore, they often cannot enforce even basic constraints such as covariate shift \citep{DuHa20dis, hu2018does}. Group-structured \ac{dro} uncertainty sets \citep{sagawa2019distributionally} overcome some of these challenges, but as alluded to before, they require significant domain expertise to pre-specify target sub-populations that a model should be robust to.

\subsection{Measuring Robustness}

Over the years, various works have attempted to come up with specific measures of model robustness in order to provide out-of-distribution generalization guarantees. Unfortunately, common divergence metrics such as the KL divergence ($\KL{p}{q}=\E_p\log \frac q p$; \citet{kullback1951information}) or the total variation distance ($d_{\text{TV}}(p, q)=\sup_x \vert p(x) - q(x)\vert$; \citet{saks1937theory}) generally yield vacuous bounds.

\citet{ben2010theory} (following a line of work initiated in \citet{kifer2004detecting} and \citet{ben2007analysis}) introduced a new distance metric called the $\mathcal H$-divergence, specifically geared towards yielding out-of-distribution generalization bounds. The central insight of the $\mathcal H$-divergence is to induce a distance that is specific to the family of models being trained (the titular $\mathcal H$). A key feature of the $\mathcal H$-divergence is that it can be approximated with another metric, the proxy $\mathcal{A}$-distance. In broad strokes, the proxy $\mathcal{A}$-distance is proportional to how easy it is to discriminate between the two domains $p$ and $q$ using classifiers from $\mathcal H$. More precisely, given two datasets $U_p$ and $U_q$ sampled from $p$ and $q$ respectively, it reads:
\begin{equation}
    \hat d_{\mathcal A}(p, q)=2(1-2\epsilon(U_p, U_q))
\end{equation}
with
\begin{equation}
\epsilon(U_p, U_q)=\min_{h\in\mathcal H}\frac 1 {|U_p| + |U_q|}\sum_{x\in U_p\cup U_q}|h(x)-\mathbbm{1}_{x\in U_p}|
\end{equation}
In other words, $\epsilon$ is the best error-rate (within $\mathcal H$) for distinguishing $U_p$ from $U_q$. Naturally the exact minimum is in general intractable, but it can be estimated by training a classifier on finite amounts of training data sampled from $p$ and $q$, making it more conducive to applied research.

Although the $\mathcal H$-divergence literature has inspired some work in ``unsupervised domain adaptation''\footnote{Training models to be robust to domain shift given unlabeled samples from the target domain} (\citet{ganin15unsupervised} and the follow-up literature on domain adversarial neural networks), it has seen relatively little use in \ac{nlp}, with some notable exceptions \citep{blitzer2007domain,rai2010domain,glorot2011domain}. More recent empirical comparisons \citep{elsahar2019annotate,kashyap2021domain} have shown the proxy $\mathcal{A}$-Distance to be competitive when it comes to predicting performance drop in the face of domain shift, possibly heralding a resurgence of the metric. We refer the interested reader to \citet{kashyap2021domain} for a complete taxonomy and a more exhaustive overview of domain divergences (written with an \ac{nlp} audience in mind).

With regards to adversarial perturbation specifically, there is a considerable amount of work devoted to providing certifiable robustness guarantees (see \citet{katz2017reluplex,sinha2017certifying, wong2018provable,raghunathan2018certified} \textit{inter alia}). Often, these guarantees rely on ideas related to Lipschitz continuity bounds: limiting the change in a model's output as a function of small perturbations in its inputs. These ideas are difficult to transpose to \ac{nlp} as they rely on some notion of a ``small change'' in the input, which is usually described in terms of $\ell_2$ norm for continuous data such as images, but is harder to formalize in natural languages. \citet{jia2019certified} circumvent this issue by adapting an approach from the computer vision literature (Interval Bound
Propagation (IBP); \citet{dvijotham2018training}) to specific neural architectures to yield robustness guarantees to word-level perturbations at the embedding level. Overall, guaranteed adversarial robustness in \ac{nlp} is still an active area of research, with recent work extending these ideas to more modern architectures \citep{shi2020robustness} for example.

\clearpage

\part{Measuring and Evaluating Robustness}
\label{sec:measuring}

\chapter{MTNT: A Testbed for Machine Translation of Noisy Text}
\label{ch:mtnt}

\section{Introduction}

After this overview of the distributional shift problem, in this chapter we present our first contribution: a benchmark dataset for \acf{mtnt}, consisting of noisy user-generated content such as the following comment:

\begin{displayquote}
\#nlproc is actualy f*ing hARD tbh \emoji{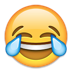}
\end{displayquote}

This handcrafted sentence showcases several types of noise that are commonly seen on social media: abbreviations (``\#nlproc''), typographical errors (``actualy''), obfuscated profanities (``f*ing''), inconsistent capitalization (``hARD''), Internet slang (``tbh'' for ``to be honest'') and emojis (\emoji{832}).  Although machine translation has achieved significant quality improvements over the past few years due to the advent of \ac{nmt} \cite{kalchbrenner-blunsom:2013:EMNLP,sutskever2014sequence,bahdanau2014neural,wu2016google}, systems are still not robust to noisy input like this \cite{belinkov2017synthetic,khayrallah2018noise}. For example, Google Translate\footnote{\url{translate.google.com} as of May 2018} translates the above example into French as:

\begin{displayquote}
\#nlproc est en train de f * ing dur hb
\end{displayquote}
which translates back into English as ``\#nlproc is in the process of [f * ing] hard hb''. This shows that noisy input can lead to erroneous translations that can be misinterpreted or even offensive.

Noise in social media text is a known issue that has been investigated in a variety of previous work \cite{eisenstein:2013:NAACL-HLT,baldwin-EtAl:2013:IJCNLP}. Most recently, \citet{belinkov2017synthetic} 
have focused on the difficulties that character based \ac{nmt} models have translating text with character level noise within individual words (from scrambling to simulated human errors such as typos or spelling/conjugation errors).
This is a good first step towards noise-robust \ac{nmt} systems, but as we demonstrate in Section \ref{sec:noise}, word-by-word replacement or scrambling of characters doesn't cover all the idiosyncrasies of language on the Internet.

At this point, despite the obvious utility of creating noise-robust MT systems, and the scientific challenges contained therein, there is currently a bottleneck in that there is no standard open benchmark for researchers and developers of \ac{mt} systems to test the robustness of their models to these and other phenomena found in noisy text on the Internet. 
In this chapter, we introduce \ac{mtnt}, a new, realistic dataset aimed at testing robustness of \ac{mt} systems to these phenomena.
The dataset contains naturally created noisy source sentences with professionally sourced translations both in a pair of typologically close languages (English and French) and distant languages (English and Japanese).
We collect noisy comments from the Reddit\footnote{\url{www.reddit.com}} online discussion website (Section \ref{sec:collection}) in English, French and Japanese, and commissioned professional translators to translate to and from English, resulting in approximately 1000 test samples and from 6k to 36k training samples in four language pairs (\ac{enfr}, \ac{fren}, \ac{enja} and \ac{jaen}).
In addition, we release additional small monolingual corpora in those 3 languages to both provide data for semi-supervised adaptation approaches as well as noisy \ac{lm} experiments.
We test standard translation models (Section \ref{sec:mt_exp}) on our data to understand their failure cases and to provide baselines for future work.\footnote{Additional language modeling experiments can be found in Appendix \ref{sec:lm_exp}} Finally, we describe the 2019 and 2020 WMT Robustness shared task in which \ac{mtnt} was used as a benchmark for robust machine translation (Section \ref{sec:mtnt_shared_task}). The data is publicly available at \url{https://pmichel31415.github.io/mtnt/}.

\section{Noise and Input Variations in Language on the Internet}
\label{sec:noise}

\subsection{Examples from Social Media Text}

The term ``noise'' can encompass a variety of phenomena in natural language, with variations across languages (\eg what is a typo in logographic writing systems?) and type of content \cite{baldwin-EtAl:2013:IJCNLP}. To give the reader an idea of the challenges posed to \ac{mt} and \ac{nlp} systems operating on this kind of text, we provide a non-exhaustive list of types of noise and more generally input variations that deviate from standard \ac{mt} training data we've encountered in Reddit comments:

\begin{itemize}
\item \textbf{Spelling/typographical errors}: ``across'' $\rightarrow$ ``accross'', ``receive'' $\rightarrow$ ``recieve'', ``could have'' $\rightarrow$ ``could of'', ``temps'' $\rightarrow$ ``tant'', ``\ja{除く}'' $\rightarrow$ ``\ja{覗く}''
\item \textbf{Word omission/insertion/repetition}: ``je n'aime pas'' $\rightarrow$ ``j'aime pas'',``je pense'' $\rightarrow$ ``moi je pense'' 
\item \textbf{Grammatical errors}: ``a ton of'' $\rightarrow$ ``a tons of'', 
``There are fewer people'' $\rightarrow$ ``There are less people''
\item \textbf{Spoken language}: ``want to'' $\rightarrow$ ``wanna'', ``I am'' $\rightarrow$ ``I'm'', ``je ne sais pas'' $\rightarrow$ ``chais pas'', ``\ja{何を笑っているの}'' $\rightarrow$ ``\ja{何わろてんねん}'', 
\item \textbf{Internet slang}: ``to be honest'' $\rightarrow$ ``tbh'', ``shaking my head'' $\rightarrow$ ``smh'', ``mort de rire'' $\rightarrow$ ``mdr'', ``\ja{笑}'' $\rightarrow$ ``w''/``\ja{草}''
\item \textbf{Proper nouns} (with or without correct capitalization): ``Reddit''$\rightarrow$ ``reddit''
\item \textbf{Dialects}: African American Vernacular English, Scottish, Provençal, Québécois, Kansai, Tohoku...
\item \textbf{Code switching}: ``This is so cute'' $\rightarrow$ ``This is so kawaii'', ``C'est trop conventionel'' $\rightarrow$ ``C'est trop mainstream'', ``\ja{現在捏造中…}'' $\rightarrow$ ``Now \ja{捏造}ing...''
\item \textbf{Jargon}: on Reddit: ``upvote'', ``downvote'', ``sub'', ``gild''
\item \textbf{Emojis and other unicode characters}: \emoji{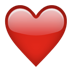},\emoji{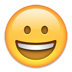},\emoji{832},\emoji{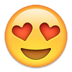},\emoji{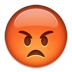}, \emoji{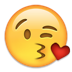}, \emoji{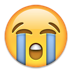}
\item \textbf{Profanities/slurs} (sometimes masked) ``f*ck'', ``m*rde'' \ldots
\end{itemize}

\subsection{Is Translating Noisy Text just another Adaptation Problem?}

To a certain extent, translating noisy text is a type of \emph{adaptation}, which has been studied extensively in the context of both \ac{smt} and \ac{nmt} \cite{axelrod-he-gao:2011:EMNLP,li-EtAl:2010:PAPERS3,luong2015stanford,chu-dabre-kurohashi:2017:Short,micelibarone-EtAl:2017:EMNLP2017,wang-EtAl:2017:EMNLP20174,michel2018extreme}. However, it presents many differences with previous domain adaptation problems, where the main goal is to adapt from a particular topic or style. In the case of noisy text, it will not only be the case that a particular word will be translated in a different way than it is in the general domain (e.g. as in the case of ``sub''), but also that there will be increased lexical variation (e.g. due to spelling or typographical errors), and also inconsistency in grammar (e.g. due to omissions of critical words or mis-usage).
The sum of these differences warrants that noisy \ac{mt} be treated as a separate instance than domain adaptation, and our experimental analysis in \ref{sec:mt:analysis} demonstrates that even after performing adaptation, \ac{mt} systems still make a large number of noise-related errors.

\section{Collection Procedure}
\label{sec:collection}

\begin{figure*}[!t]
\centering
\includegraphics[width=\textwidth]{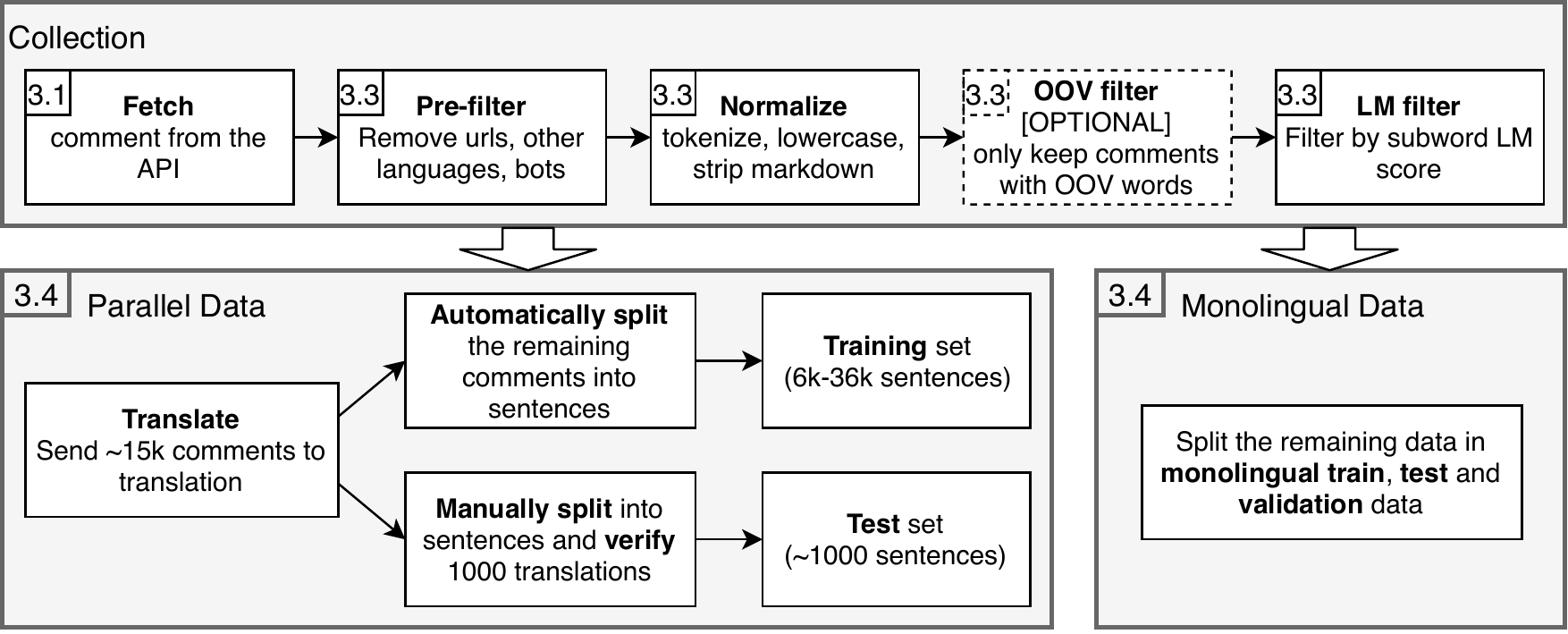}
\caption{\label{fig:mtnt_collection_diagram}Summary of our collection process and the respective sections addressing them. We apply the same procedure for each language.}
\end{figure*}

We first collect noisy sentences in our three languages of interest, English, French and Japanese. We refer to Figure \ref{fig:mtnt_collection_diagram} for an overview of the data collection and translation process.

We choose Reddit as a source of data because (1) its content is likely to exhibit noise, (2) some of its sub-communities are entirely run in different languages, in particular, English, French and Japanese, and  (3) Reddit is a popular source of data in curated and publicly distributed NLP datasets \cite{tan2016winning}. 
We collect data using the public Reddit API.
\footnote{In particular, we use this implementation: \url{praw.readthedocs.io/en/latest}, and our complete code is available at \url{http://www.cs.cmu.edu/~pmichel1/mtnt/}.} 

Note that the data collection and translation is performed at the comment level. We split the parallel data into sentences as a last step.

\subsection{Data Sources}

For each language, we select a set of communities (``subreddits'') that we know contain many comments in that language:

\begin{description}
\item[English:] Since an overwhelming majority of the discussions on Reddit are conducted in English, we don't restrict our collection to any community in particular.
\item[French:] \texttt{/r/france}, \texttt{/r/quebec} and \texttt{/r/rance}. The first two are among the biggest French speaking communities on Reddit. The third is a humor/sarcasm based offspring of \texttt{/r/france}.
\item[Japanese:] \texttt{/r/newsokur}, \texttt{/r/bakanewsjp}, \texttt{/r/newsokuvip}, \texttt{/r/lowlevelaware} and \texttt{/r/steamr}. Those are the biggest Japanese speaking communities, with over 2,000 subscribers at the time of collection.
\end{description}

We collect comments made during the 03/27/2018-03/29/3018 time period for English, 09/2018-03/2018 for French and 11/2017-03/2018 for Japanese. The large difference in collection time is due to the variance in comment throughput and relative amount of noise between the languages.

\subsection{Contrast Corpora}
\label{sec:ref_corp}

Not all comments found on Reddit exhibit noise as described in Section \ref{sec:noise}.
Because we would like to focus our data collection on noisy comments, we devise criteria that allow us to distinguish potentially noisy comments from clean ones.
Specifically, we compile a \emph{contrast} corpus composed of clean text that we can compare to, and find potentially noisy text that differs greatly from the contrast corpus. 
Given that our final goal is \ac{mt} robust to noise, we prefer that these contrast corpora consist of the same type of data that is often used to train \ac{nmt} models. We select different datasets for each language:

\begin{description}
\item[English:] The English side of the preprocessed parallel training data provided for the German-English WMT 2017 News translation task,\footnote{\url{http://www.statmt.org/wmt17/translation-task.html}} as provided on the website. This amounts to $\approx 5.85$ million sentences.
\item[French:] The entirety of the French side of the parallel training data provided for the English-French WMT 2015 translation task.\footnote{\url{http://www.statmt.org/wmt15/translation-task.html}}
This amounts to $\approx 40.86$ million sentences.
\item[Japanese:] We aggregate three small/medium sized \ac{mt} datasets: KFTT \cite{neubig11kftt}, JESC \cite{pryzant2017jesc} and TED talks \cite{cettoloEtAl:EAMT2012}, amounting to $\approx 4.19$ million sentences.
\end{description}

\subsection{Identifying Noisy Comments}

We now describe the procedure used to identify comments containing noise.

\paragraph{Pre-filtering} First, we perform three pre-processing to discard comments that do not represent natural noisy text in the language of interest:

\begin{enumerate}
\item Comments containing a URL, as detected by a regular expression.
\item Comments where the author's username contains ``bot'' or ``AutoModerator''. This mostly removes automated comments from bots.
\item Comments in another language: we run \texttt{langid.py}\footnote{\url{https://github.com/saffsd/langid.py}} \cite{lui2012langid} and discard comments where $p(\text{lang}\mid\text{comment}) > 0.5$ for any language other than the one we are interested in.
\end{enumerate}

This removes cases that are less interesting, i.e. those that could be solved by rule-based pattern matching or are not natural text created by regular users in the target language. Our third criterion in particular discards comments that are blatantly in another language while still allowing comments that exhibit code-switching or that contain proper nouns or typos that might skew the language identification. In preliminary experiments, we noticed that these criteria 14.47, 6.53 and 7.09 \% of the collected comments satisfied the above criteria respectively.

\paragraph{Normalization}

After this first pass of filtering, we pre-process the comments before running them through our noise detection procedure. We first strip Markdown\footnote{\url{https://daringfireball.net/projects/markdown}} syntax from the comments. For English and French, we normalize the punctuation, lowercase and tokenize the comments using the Moses tokenizer. For Japanese, we simply lowercase the alphabetical characters in the comments. Note that this normalization is done for the purpose of noise detection only. The collected comments are released without any kind of preprocessing.
We apply the same normalization procedure to the contrast corpora.

\paragraph{Unknown words} In the case of French and English, a clear indication of noise is the presence of \ac{oov}: we record all lowercased words encountered in our reference corpus described in Section \ref{sec:ref_corp} and only keep comments that contain at least one \ac{oov}. Since we did not use word segmentation for the Japanese reference corpus, we found this method not to be very effective to select Japanese comments and therefore skipped this step.

\paragraph{Language model scores}

The final step of our noise detection procedure consists of selecting those comments with a low probability under a language model trained on the reference monolingual corpus. This approach mirrors the one used in \citet{moore-lewis:2010:Short} and \citet{axelrod-he-gao:2011:EMNLP} to select data similar to a specific domain using language model perplexity as a metric. We search for comments that have a low probability under a sub-word language model for more flexibility in the face of \ac{oov} words. We segment the contrast corpora with \ac{bpe} using the sentencepiece\footnote{\url{https://github.com/google/sentencepiece}} implementation. We set the vocabulary sizes to $1,000$, $1,000$ and $4,000$ for English, French and Japanese respectively. We then use a 5-gram Kneser-Ney smoothed language model trained using \texttt{kenLM}\footnote{\url{https://kheafield.com/code/kenlm/}} \cite{heafield2013estimate} to calculate the log probability, normalized by the number of tokens for every sentence in the reference corpus. Given a reddit comment, we compute the normalized log probability of each of its lines under our subword language model. If for any line this score is below the 1st percentile of scores in the reference corpus, the comment is labeled as noisy and saved.

\subsection{Creating the Parallel Corpora}
\label{sec:parallel_corpora}

Once enough data has been collected, we isolate $15,000$ comments in each language by the following procedure:
\begin{itemize}
\item Remove all duplicates. In particular, this handles comments that might have been scraped twice or automatic comments from bots.
\item To further weed out outliers (comments that are too noisy, \eg ASCII art, wrong language\ldots or not noisy enough), we discard comments that are on either end of the distribution of normalized \ac{lm} scores within the set of collected comments. We only keep comments whose normalized score is within the 5-70 percentile for English (resp. 5-60 for French and 10-70 for Japanese). These numbers are chosen by manually inspecting the data. 
\item Choose $15,000$ samples at random.
\end{itemize}

We then concatenate the title of the thread where the comment was found to the text and send everything to an external vendor for manual translations. Upon reception of the translations, we noticed a certain amount of variation in the quality of translations, likely because translating social media text, with all its nuances, is difficult even for humans. In order to ensure the highest quality in the translations, we manually filter the data to segment the comments into sentences and weed out poor translations for our test data. We thereby retain around $1,000$ sentence pairs in each direction for the final test set.

We gather the samples that weren't selected for the test sets to be used for training or fine-tuning models on noisy data. We automatically split comments into sentences with a regular expression detecting sentence delimiters, and then align the source and target sentences. Should this alignment fail (\ie the source comment contains a different number of sentences than the target comment after automatic splitting), we revert back to providing the whole comment without splitting. For the training data, we do not verify the correctness of translations as closely as for the test data. Finally, we isolate $\approx 900$ samples in each direction to serve as validation data. 

Information about the size of the data can be found in Table \ref{tab:test_data_stats},  \ref{tab:train_data_stats} and  \ref{tab:valid_data_stats} for the test, training and validation sets respectively. We tokenize the English and French data with the Moses \cite{koehn2007moses} tokenizer and the Japanese data with Kytea \cite{neubig11aclshort} before counting the number of tokens in each dataset.

\begin{table*}[tb]

\caption{\label{tab:para_data_stats}Datasets numbers.}
\begin{subtable}{0.50\columnwidth}
\centering
\begin{tabular}{lccc}
& \#samples & \#src tokens & \#trg tokens \\ \hline
\texttt{en-fr} & 36,058 & 841k & 965k \\
\texttt{fr-en} & 19,161 & 661k & 634k \\
\texttt{en-ja} & 5,775 & 281k & 506k\\
\texttt{ja-en} & 6,506 & 172k & 128k\\
\hline
\end{tabular}
\caption{\label{tab:train_data_stats}Training sets numbers.}
\end{subtable}
~
\begin{subtable}{0.45\columnwidth}
\centering
\begin{tabular}{ccc}
\#samples & \#src tokens & \#trg tokens \\ \hline
 852 & 16,957 & 18,948 \\
 886 & 41,578 & 46,886 \\
 852 & 40,124 & 46,886\\
 965 & 25,010 & 23,289\\
\hline
\end{tabular}
\caption{\label{tab:valid_data_stats}Validation sets numbers.}
\end{subtable}
\begin{subtable}{0.50\columnwidth}
\centering
\begin{tabular}{lccc}
& \#samples & \#src tokens & \#trg tokens \\ \hline
\texttt{en-fr} & 1,020 & 15,919 & 18,445\\
\texttt{fr-en} & 1,022 & 16,662 & 16,038\\
\texttt{en-ja} & 1,002 & 11,040 & 20,008\\
\texttt{ja-en} & 1,020 & 23,997 & 33,429\\
\hline
\end{tabular}
\caption{\label{tab:test_data_stats}Test set numbers.
}
\end{subtable}
~
\begin{subtable}{0.45\columnwidth}
\centering
\begin{tabular}{clccc}
&&  \#samples & \#tok & \#char \\ \hline\hline
\multirow{2}{*}{\texttt{en}}& train&81,631&3,99M&18,9M\\
& dev&3,000& 146k&698k\\\hline
\multirow{2}{*}{\texttt{fr}}& train& 26,485 & 1,52M&7,49M\\
& dev&3,000&176k&867k\\\hline
\multirow{2}{*}{\texttt{ja}}& train&32,042&943k&3.9M\\
& dev&3,000&84k&351k\\
\hline
\end{tabular}
\caption{\label{tab:lm_data_stats}Monolingual data numbers.}
\end{subtable}
\end{table*}

\subsection{Monolingual Corpora}
\label{sec:lm_data}

After the creation of the parallel train and test sets, a large number of unused comments remain in each language, which we provide as monolingual corpora. This additional data has two purposes: first, it serves as a resource for in-domain training using semi-supervised methods relying on monolingual data (e.g. \citet{P16-1185,zhang-zong:2016:EMNLP2016}). Second, it provides a language modeling dataset for noisy text in three languages.

\begin{table*}[t]
\centering
\caption{\label{tab:noise_qual} Numbers, per 100 tokens, of quantifiable noise occurrences. For each language and category, the dataset with the highest amount of noise is highlighted.}
\begin{tabular}{clcccc}
 & & Spelling & Grammar & Emojis & Profanities \\
\hline
\hline
\multirow{3}{*}{\texttt{en}} & newstest2014 & 0.210 & 0.189 & 0.000 & 0.030 \\
 & newsdiscusstest2015 & 0.621 & 0.410 & 0.021 & 0.076 \\
 & \ac{mtnt} (\ac{enfr}) & \bf 2.180 & \bf 0.559 & \bf 0.289 & \bf 0.239 \\
\hline
\multirow{3}{*}{\texttt{fr}} & newstest2014 & 2.776 & 0.091 & 0.000 & 0.245 \\
 & newsdiscusstest2015 & 1.686 & 0.457 & 0.024 & 0.354 \\
 & \ac{mtnt} & \bf 4.597 & \bf 1.464 & \bf 0.252 & \bf 0.690 \\
\hline
\multirow{4}{*}{\texttt{ja}}& TED & 0.011 & 0.266 & 0.000 & 0.000 \\
& KFTT & 0.021 & 0.228 & 0.000 & 0.000 \\
& JESC & 0.096 & 0.929 & 0.090 & \bf 0.058 \\
& \ac{mtnt} & \bf 0.269 & \bf 1.527 & \bf 0.156 & 0.036 \\
\hline
\end{tabular}
\end{table*}

We select $3,000$ comments at random in each dataset to form a validation set to be used to tune hyper-parameters, and provide the rest as training data. The data is provided with one comment per line. Newlines within individual comments are replaced with spaces. Table \ref{tab:lm_data_stats} contains information on the size of the datasets. As with the parallel \ac{mt} data, we provide the number of tokens after tokenization with the Moses tokenizer for English and French and Kytea for Japanese.

\section{Dataset Analysis}
\label{sec:data_anal}

In this section, we investigate the proposed data to understand how different categories of noise are represented and to show that our test sets contain more noise overall than established \ac{mt} benchmarks.

\subsection{Quantifying Noisy Phenomena}

We run a series of tests to count the number of occurrences of some of the types of noise described in Section \ref{sec:noise}. Specifically we pass our data through spell checkers to count spelling and grammar errors. 
Due to some of these tests being impractical to run on a large scale, we limit our analysis to the test sets of \ac{mtnt}.

We use slightly different procedures depending on the tools available for each language. We test for spelling and grammar errors in English data using Grammarly\footnote{\url{https://www.grammarly.com/}}, an online resource for English spell-checking.
Due to the unavailability of an equivalent of Grammarly in French and Japanese, we test for spelling and grammar error using the integrated spell-checker in Microsoft Word 2013\footnote{\url{https://products.office.com/en-us/microsoft-word-2013}}. Note that Word seems to count proper nouns as spelling errors, giving higher numbers of spelling errors across the board in French as compared to English.

For all languages, we also count the number of profanities and emojis using custom-made lists and regular expressions\footnote{available with our code at \url{https://github.com/pmichel31415/mtnt}}. In order to compare results across datasets of different sizes, we report all counts per $100$ words.

The results are recorded in the last row of each section in Table \ref{tab:noise_qual}. In particular, for the languages with a segmental writing system, English and French, spelling errors are the dominant type of noise, followed by grammar error. Unsurprisingly, the former are much less present in Japanese.

\subsection{Comparison to Existing Machine Translation Test Sets}

Table \ref{tab:noise_qual} also provide a comparison with the relevant side of established \ac{mt} test sets. For English and French, we compare our data to newstest2014\footnote{\label{notewmt15dev}\url{http://www.statmt.org/wmt15/dev-v2.tgz}} and newsdiscusstest2015\footnote{\label{notewmt15test}\url{http://www.statmt.org/wmt15/test.tgz}} test sets. For Japanese, we compare with the test sets of the datasets described in Section \ref{sec:ref_corp}.

Overall, \ac{mtnt} contains more noise in all metrics but one (there are more profanities in JESC, a Japanese subtitle corpus). This confirms that MTNT indeed provides a more appropriate benchmark for translation of noisy or non-standard text.

Compared to synthetically created noisy test sets \cite{belinkov2017synthetic} \ac{mtnt} contains less systematic spelling errors and more varied types of noise (\eg emojis and profanities) and is thereby more representative of naturally occurring noise.
\section{Machine Translation Experiments}
\label{sec:mt_exp}

We evaluate standard \ac{nmt} models on our proposed dataset to assess its difficulty. Our goal is not to train state-of-the art models but rather to test standard off-the-shelf \ac{nmt} systems on our data, and elucidate what features of the data make it difficult.

\subsection{Model Description}

All our models are implemented in DyNet \cite{neubig2017dynet} with the XNMT toolkit \cite{neubig2018xnmt}. We use approximately the same setting for all language pairs: the encoder is a bidirectional LSTM with 2 layers, the attention mechanism is a multi layered perceptron and the decoder is a 2 layered LSTM. The embedding dimension is 512, all other dimensions are 1024. We tie the target word embeddings and the output projection weights \cite{press-wolf:2017:EACLshort}. We train with Adam \cite{Kingma2014Adam} with XNMT's default hyper-parameters, as well as dropout (with probability $0.3$). We used \ac{bpe} subwords to handle \ac{oov} words. Full configuration details as well as code to reproduce the baselines is available at \url{https://github.com/pmichel31415/mtnt}.

\subsection{Training Data}
\label{sec:mt_train_data}

We train our models on standard \ac{mt} datasets:

\begin{itemize}
\item en $\leftrightarrow$ fr: Our training data consists in the europarl-v7\footnote{\url{http://www.statmt.org/europarl/}} and news-commentary-v10\footnote{\url{http://www.statmt.org/wmt15/training-parallel-nc-v10.tgz}} corpora, totaling $2,164,140$ samples, $54,611,105$ French tokens and $51,745,611$ English tokens (non-tokenized). We use the newsdiscussdev2015\textsuperscript{\ref{notewmt15dev}} dev set from WMT15 as validation data and evaluate the model on the newsdiscusstest2015\textsuperscript{\ref{notewmt15test}} and newstest2014\textsuperscript{\ref{notewmt15dev}} test sets.
\item en $\leftrightarrow$ ja: We concatenate the respective train, validation and test sets of the three corpora mentioned in \ref{sec:ref_corp}. In particular we detokenize the Japanese part of each dataset to make sure that any tokenization we perform will be uniform (in practice we remove ASCII spaces). This amounts to $3,900,772$ training samples ($34,989,346$ English tokens without tokenization). We concatenate the dev sets associated with these corpora to serve as validation data and evaluate on each respective test set separately.
\end{itemize}

\begin{table}[ht]
\centering
\caption{\label{tab:bleu_scores} BLEU scores of \ac{nmt} models on the various datasets.}
\begin{tabular}{lcc}
& \ac{enfr}  & \ac{fren}   \\\hline\hline

newstest2014 & $33.52$ & $28.93$\\
newsdiscusstest2015 & $33.03$ & $30.76$\\ \hline
\ac{mtnt} & $21.77$ & $23.27$ \\
\ac{mtnt} (+tuning) & $29.73$ & $30.29$ \\
\hline
&\ac{enja}  & \ac{jaen} \\\hline\hline
TED&$14.51$&$13.25$\\
KFTT&$20.82$&$20.77$\\
JESC&$15.77$&$18.00$\\ \hline
\ac{mtnt}&$9.02$&$6.65$\\
\ac{mtnt} (+tuning) & $12.45$ & $9.82$ \\
\hline
\end{tabular}
\end{table}

\begin{table*}[t]
{\small
\centering
\caption{\label{tab:pre_post_finetunig_qual}Comparison of our model's output before and after fine-tuning in \ac{fren}.}
\begin{tabular}{ll}
\hline\hline
Source & Moi faire la gueule dans le métro me manque, c'est grave ? \\ \hline
Target & I miss sulking in the underground, is that bad? \\ \hline
Our model & I do not know what is going on in the metro, that is a serious matter. \\ \hline\
+ fine-tuning & I do not want to be in the metro, it's serious? \\ \hline\hline
Source & * C'est noël / pâques / pentecôte / toussaint : Pick One, je suis pas catho \\ \hline
Target &  Christmas / Easter / Pentecost / All Saints: Pick One, I'm not Catholic! \\ \hline
Our model & <unk> It is a pale/poward, a palec<unk>te d'<unk>tat: Pick One, I am not a catho! \\ \hline\
+ fine-tuning & <unk> It's no<unk>l / pesc<unk>e /pentecate /mainly: Pick One, I'm not catho! \\ \hline\hline

\end{tabular}
}
\end{table*}

\subsection{Results}
\label{sec:mt_results}

We use \texttt{sacreBLEU}\footnote{\url{https://github.com/mjpost/sacreBLEU}}, a standardized BLEU score evaluation script proposed by \citet{post2018call}, for BLEU evaluation of our benchmark dataset. It takes in detokenized references and hypotheses and performs its own tokenization before computing BLEU score. We specify the \texttt{intl} tokenization option. In the case of Japanese text, we run both hypothesis and reference through KyTea before computing BLEU score. We strongly encourage that evaluation be performed in the same manner in subsequent work, and will provide both scripts and an evaluation web site in order to facilitate reproducibility.

Table \ref{tab:bleu_scores} lists the BLEU scores for our models on the relevant test sets in the two language pairs, including the results on \ac{mtnt}.

\subsection{Analysis of the MT outputs}
\label{sec:mt:analysis}

To better understand the types of errors made by our model, we count the n-grams that are over- and under- generated with respect to the reference translation. Specifically, we compare the count ratios of all 1- to 3-grams in the output and in the reference and look for the ones with the highest (over-generated) and lowest (under-generated) ratio.

We find that in English, the model under-generates the contracted form of the negative (``do not''/``don't'') or of auxiliaries (``That is''/``I'm''). Similarly, in French, our model over generates ``de votre'' (where ``votre'' is the formal 2nd person plural for ``your'') and ``n'ai pas'' which showcases the ``ne [\ldots] pas'' negation, often dropped in spoken language. Conversely, the informal second person ``tu'' is under-generated, as is the informal and spoken contraction of ``cela'', ``ça''. In Japanese, the model under-generates, among others, the informal personal pronoun \ja{俺} (``ore'') or the casual form \ja{だ} (``da'') of the verb \ja{です} (``desu'', to be). In \ac{jaen} the results are difficult to interpret as the model seems to produce incoherent outputs (\eg~``no, no, no\ldots'') when the \ac{nmt} system encounters sentences it has not seen before. The full list of n-grams with the top 5 and bottom 5 count ratios in each language pair is displayed in Table \ref{tab:compare_ngrams}.

\begin{table}[t]
\centering
\caption{\label{tab:compare_ngrams}Over and under generated n-grams in our model's output for \ac{enfr}}
\begin{tabular}{cccc}
\ac{fren} & \ac{enfr}&\ac{jaen} & \ac{enja} \\\hline\hline
\multicolumn{4}{c}{Over generated}\\\hline
\small{<unk>}&\small{<unk>} &\small{no, no,} & \small{\ja{※}}\\
\small{it is not}&\small{qu’ils} &\small{i} & \small{\ja{が }}\\
\small{I do not}&\small{de votre} &\small{no, no, no,} & \small{\ja{か ?}}\\
\small{That is}&\small{s’il} &\small{so on and} & \small{\ja{て }}\\
\small{not have}&\small{n’ai pas} &\small{on and so} & \small{\ja{す か ?}}\\
\hline
\multicolumn{4}{c}{Under generated}\\\hline
\small{it's} &\small{tu}& \small{|} & \small{\ja{？}}\\
\small{I'm} &\small{ça}& \small{Is} & \small{\ja{よ 。}}\\
\small{I don't} &\small{que tu} & \small{>} & \small{\ja{って}}\\
\small{>} &\small{!} & \small{""The} & \small{\ja{俺}}\\
\small{doesn't}&\small{as} & \small{those} & \small{\ja{だ 。}}\\
\hline
\end{tabular}
\end{table}

\subsection{Fine-Tuning}

Finally, we test a simple domain adaptation method by fine-tuning our models on the training data described in Section \ref{sec:parallel_corpora}. We perform one epoch of training with vanilla SGD with a learning rate of $0.1$ and a batch size of $32$. We do  not use the validation data at all. As evidenced by the results in the last row of Table \ref{tab:bleu_scores}, this drives BLEU score up by 3.17 to 7.96 points depending on the language pair. However large this increase might be, our model still breaks on very noisy sentences. Table \ref{tab:pre_post_finetunig_qual} shows three examples in \ac{fren}. Although our model somewhat improves after fine-tuning, the translations remain inadequate in all cases. In the third case, our model downright fails to produce a coherent output. This shows that despite improving BLEU score, naive domain adaptation by fine-tuning doesn't solve the problem of translating noisy text.

\section{Language Modeling Experiments}
\label{sec:lm_exp}

In addition to our \ac{mt} experiments, we report character-level language modeling results on the monolingual part of our dataset. We use the data described in Section \ref{sec:lm_data} as training and validation sets. We evaluate the trained model on the source side of our \ac{enfr}, \ac{fren} and \ac{jaen} test sets for English, French and Japanese respectively.

We report results for two models: a Kneser-Ney smoothed 6-gram model (implemented with KenLM) and an implementation of the AWD-LSTM proposed in \cite{merity2017regularizing}\footnote{\url{https://github.com/salesforce/awd-lstm-lm}}. We report the Bit-Per-Character (bpc) counts in table \ref{tab:bpc_lm}. We intend these results to serve as a baseline for future work in language modeling of noisy text in either of those three languages.

\begin{table}[ht]
\centering
\caption{\label{tab:bpc_lm}Language modeling scores}
\begin{tabular}{lcccc}
& \multicolumn{2}{c}{6-gram} &\multicolumn{2}{c}{AWD LSTM}\\
& dev &test & dev& test\\\hline\hline
English &2.081&2.179& 1.706& 1.810\\
French &1.906&2.090& 1.449&1.705\\
Japanese &5.003&5.497 & 4.801&5.225\\
\hline
\end{tabular}
\end{table}

\section{MTNT in Action: the 2019 and 2020 WMT Robustness Shared Tasks}
\label{sec:mtnt_shared_task}

A year after MTNT was originally published, the dataset was featured in the Robustness shared task\footnote{\url{http://www.statmt.org/wmt19/robustness.html}} at the 2019 conference on machine translation (WMT). Participants were allowed to use large amounts of out-of-domain data from the main WMT translation task\footnote{\url{http://www.statmt.org/wmt15/translation-task.html}} or from the three \texttt{en-ja} corpora described in Section \ref{sec:mt_exp}, as well as the MTNT dataset itself as a small in-domain corpus. Submitted systems were evaluated on new, blind test sets collected using the same procedure as MTNT \textit{i.e.} scraped from Reddit, filtered out for noisy comments using a sub-word language modeling criterion and translated by professionals.

The shared task attracted 23 submissions from 11 teams. Methods used by competing teams included variants of data cleaning (removal of noisy training samples), the use of placeholders (to account for special characters that could be copied such as emojis), data augmentation strategies (such as back-translation or the addition of filtered data from external sources), domain-aware training (via the addition of domain tags), fine-tuning and ensembling. Table \ref{tab:fr_en_casing_example} showcases an example where specific handling of casing allowed some systems to outperform others on samples containing ALL CAPS (a common phenomenon in user-generated content on social media). A more detailed description of the submissions, as well as an analysis of the results, can be found in the findings paper \cite{li2019findings}.

The shared task was renewed in 2020, with a broader focus on general domain robustness\footnote{\url{http://www.statmt.org/wmt20/robustness.html}}. In this iteration, participants were asked to train models in one of two language pairs (English-German and English Japanese) using only the data available for the WMT20 news translation task \citep{barrault2020findings}. The evaluation data was aggregated from three different domains: Wikipedia comments with toxic content\footnote{\url{https://www.kaggle.com/c/jigsaw-toxic-comment-classification-challenge}}, speech recognition transcripts \citep{wang2020covost}, and a new test set of reddit comments, collected following again the same procedure as \ac{mtnt}. Evaluation proceeded in two distinct phases. In the first ``zero-shot'' phase, submitted systems were simply evaluated against the unseen test sets, thus testing the domain robustness of models trained using only out-of-domain data. In the second ``few-shot'' phase, participants were given a single week to fine-tune their systems using limited amounts of training data from each domain (including the \ac{mtnt} training data). Each system output was evaluated using both automatic metrics (BLEU) and human judgements. In the case of the latter, evaluators were not only asked to rate the machine-generated translations on a numerical scale of 1 to 5, but also to indicate the presence of ``catastrophic errors'' belonging to pre-defined categories such as ``introduction of toxicity'', ``mis-translation of named entities'' or ``change in units/time/date/numbers''

This edition of the task received 59 submissions by 11 participating teams from both industry and academic institutions. By and large, most teams used similar approaches as the previous year, specifically tagged back-translations, ensembling or fine-tuning on filtered data. In addition, some teams used adaptor modules \citep{bapna2019simple} for more efficient fine-tuning, with some level of success. The human evaluation of the results uncovered a predominance of named entity mis-translations among catastrophic errors, followed by sentiment reversal and introduction of toxicity. We refer to the findings paper \citep{specia2020findings} for a full description of the data collection process, evaluation procedure and participating systems.

\begin{table*}[t]
  \centering
  \begin{tabular}{c | p{10cm}c}
  \toprule 
 & Output & BLEU+1 \\ \midrule 
Ref & From Sri Lanka , to Russia , to the United States , to Japan I mean it 's a market THAT GOES EVERYWHERE . &  \\
CUNI & from sri lanka , to russia , to the united states , to japon I mean it 's a market QUI VA PARTOUT . & 33.0 \\
NLE & From Sri Lanka , to Russia , to the United States , to Japan I mean it 's a market THAT GOES EVERYWHERE . & 100 \\
\bottomrule 
  \end{tabular}
  \caption{An example of handling of casing in two \ac{fren} systems, CUNI \citep{helcl2019cuni} and NLE \citep{berard2019naver}. The CUNI system does not recognize the ALL CAPS text in the French sentence, and simply copies the source. On the other hand, the NLE system uses a specific \emph{case tagging} technique which enables it to properly translate the relevant phrase, while preserving the original casing.}
  \label{tab:fr_en_casing_example}
\end{table*}

{\it
\paragraph{Acknowledgement} The work discussed in this section was carried out in two different papers, \citet{li2019findings} and \citet{specia2020findings}. The present author was not first author on either of these papers. His contributions were, for \citet{li2019findings} to help with the general organization of the task, including establishing the goal of the task, planning, handling submissions and in particular to be responsible for collecting the blind test sets. For \citet{specia2020findings}, his contributions were more focused on data collection for the reddit test set.
}

\section{Related work}

Handling noisy text has received growing attention among various language processing tasks due to the abundance of user generated content on popular social media platforms \cite{crystal:01,herring2003media,danet:07}. These contents are considered as noisy when compared to news corpora which have been the main data source for language tasks \cite{baldwin-EtAl:2013:IJCNLP,eisenstein:2013:NAACL-HLT}. They pose several unique challenges because they contain a larger variety of linguistic phenomena that are absent in the news domain and that lead to degraded quality when applying an model to out-of-domain data \citep{ritter2011named, luong2015stanford}.
Additionally, they are live examples of the Cmabrigde Uinervtisy (Cambridge University) effect, where state-of-the-art models become brittle while human's language processing capability is more robust \citep{sakaguchi2017robsut, belinkov2017synthetic}. 

Efforts to address these challenges have been focused on creating in-domain datasets and annotations \citep{owoputi2013improved, kong2014dependency, blodgett2017dataset}, and domain adaptation training \citep{luong2015stanford}. In \ac{mt}, improvements were obtained for \ac{smt} \citep{formiga2012dealing}. However, the specific challenges for neural machine translation have not been studied until recently \cite{belinkov2017synthetic,sperber2017toward,cheng2018towards}. The first provides empirical evidence of non-trivial quality degradation when source sentences contain natural noise or synthetic noise within words, and the last two explore data augmentation and adversarial approaches of adding noise efficiently to training data to improve robustness. 

Our work also contributes to recent advances in evaluating neural machine translation quality with regard to specific linguistic phenomena, such as manually annotated test sentences for English to French translation, in order to identify errors due to specific linguistic divergences between the two languages \citep{isabelle2017challenge}, or automatically generated test sets to evaluate typical errors in English to German translation \citep{sennrich2017how}. Our contribution distinguishes itself from this previous work and other similar initiatives \citep{peterson2011openmt12} by providing an open test set consisting of naturally occurring text exhibiting a wide range of phenomena related to noisy input text from contemporaneous social media.

\section{Discussion}

In this chapter, we proposed a new dataset to test \ac{mt} models for robustness to the types of noise encountered in natural language on the Internet. We contribute parallel training and test data in both directions for two language pairs, English $\leftrightarrow$ French and English $\leftrightarrow$ Japanese, as well as monolingual data in those three languages. We show that this dataset contains more noise than existing \ac{mt} test sets and poses a challenge to models trained on standard \ac{mt} corpora. We further demonstrate that these challenges cannot be overcome by a simple domain adaptation approach alone. After its publication, \ac{mtnt} served as the basis for two shared tasks at the WMT conference in machine translation, fostering research on models and evaluation metrics for this specific problem \citep{li2019findings,specia2020findings}. In addition, it was used independently in a variety of subsequent work, either as a benchmark for \eg{} data augmentation techniques \citep{karpukhin2019training, vaibhav2019improving} or as an exemplar of domain shift in machine translation \citep{michel2019sixteen}.

\clearpage

\chapter{On Evaluation of Adversarial Perturbations for Sequence-to-Sequence Models}
\chaptermark{On Evaluation of Adversarial Perturbations}
\label{ch:on_evaluation_of_adversarial}

\section{Introduction}

While the previous chapter's focus was on evaluating models under distributional shift to the social media domain (a type of shift which we referred to as a ``global'' shift in Chapter \ref{ch:background}), in this chapter we take a closer look at a different type of distributional shift: adversarial perturbations. At a high level, attacking a machine learning model with adversarial perturbations is the process of making changes to its input to maximize an adversarial goal, such as mis-classification \cite{Szegedy2013IntriguingPO} or mis-translation \cite{zhao2018generating}.
These attacks provide insight into the vulnerabilities of machine learning models and their brittleness to samples outside the training distribution. Lack of robustness to these attacks poses security concerns to safety-critical applications, \eg{} self-driving cars \cite{bojarski2016end}.

Adversarial attacks were first defined and investigated for computer vision systems (\citet{Szegedy2013IntriguingPO,Goodfellow2014ExplainingAH,MoosaviDezfooli2016DeepFoolAS} inter alia), where the input space is continuous, making minuscule perturbations largely imperceptible to the human eye.
In discrete spaces such as natural language sentences, the situation is more problematic; even a flip of a single word or character is generally perceptible by a human reader.
Thus, most of the mathematical framework in previous work is not directly applicable to discrete text data.
Moreover, there is no canonical distance metric for textual data like the $\ell_p$ norm in real-valued vector spaces such as images, and evaluating the level of semantic similarity between two sentences is a field of research of its own  \cite{cer-EtAl:2017:SemEval}.
This elicits a natural question: \textit{what does the term ``adversarial perturbation'' mean in the context of \ac{nlp}}?

We propose a simple but natural criterion for adversarial examples in \ac{nlp}, particularly untargeted\footnote{Here we use the term untargeted in the same sense as \cite{Ebrahimi2018OnAE}: an attack whose goal is simply to decrease performance with respect to a reference translation.} attacks on \ac{seq2seq} models: \emph{adversarial examples should be meaning-preserving on the source side, but meaning-destroying on the target side}.
The focus on explicitly evaluating meaning preservation is in contrast to previous work on adversarial examples for \ac{seq2seq} models \cite{belinkov2018synthetic,zhao2018generating,cheng2018seq2sick,Ebrahimi2018OnAE}.
Nonetheless, this feature is extremely important; given two sentences with equivalent meaning, we would expect a good model to produce two outputs with equivalent meaning.
In other words, any meaning-preserving perturbation that results in the model output changing drastically highlights a fault of the model.

A first technical contribution of this chapter is to lay out a method for formalizing this concept of meaning-preserving perturbations (Section \ref{sec:eval_adv_attacks}).
This makes it possible to evaluate the effectiveness of adversarial attacks or defenses either using gold-standard human evaluation, or approximations that can be calculated without human intervention.
We further propose a simple method of imbuing gradient-based word substitution attacks (Section \ref{sec:attack_paradigm}) with simple constraints aimed at increasing the chance that the meaning is preserved (Section \ref{sec:constraints}).

Our experiments are designed to answer several questions about meaning preservation in \ac{seq2seq} models.
First, we evaluate our proposed ``source-meaning-preserving, target-meaning-destroying'' criterion for adversarial examples using both manual and automatic evaluation (Section \ref{sec:corr_human_auto}) and find that a less widely used evaluation metric (chrF) provides significantly better correlation with human judgments than the more widely used BLEU and METEOR metrics.
We proceed to perform an evaluation of adversarial example generation techniques, finding that chrF does help to distinguish between perturbations that are more meaning-preserving across a variety of languages and models (Section \ref{sec:attack_results}). Finally, we apply existing methods for adversarial training to the adversarial examples with these constraints and show that making adversarial inputs more semantically similar to the source is beneficial for robustness to adversarial attacks and does not decrease test performance on the original data distribution (Section \ref{sec:adv_train}). A toolkit implementing our evaluation framework is released at \teapoturl{}.

\section{A Framework for Evaluating Adversarial Attacks}
\label{sec:eval_adv_attacks}

In this section, we present a simple procedure for evaluating adversarial attacks on \ac{seq2seq} models. We will use the following notation: $x$ and $y$ refer to the source and target sentence respectively. We denote $x$'s translation by model $M$ as $y_M$. Finally, $\hat x$ and $\hat y_M$ represent an adversarially perturbed version of $x$ and its translation by $M$, respectively. The nature of $M$ and the procedure for obtaining $\hat x$ from $x$ are irrelevant to the discussion below.

\subsection{The Adversarial Trade-off}
\label{sec:adv_tradeoff}

The goal of adversarial perturbations is to produce failure cases for the model $M$. Hence, the evaluation must include some measure of the \emph{target similarity} between $y$ and $y_{M}$, which we will denote $\stgt(y, \hat y_M)$.
However, if no distinction is being made between perturbations that preserve the meaning and those that don't, a sentence like ``he's very \textit{friendly}'' is considered a valid adversarial perturbation of ``he's very \textit{adversarial}'', even though its meaning is the opposite.
Hence, it is crucial, when evaluating adversarial attacks on \ac{mt} models, that the discrepancy between the original and adversarial input sentence be quantified in a way that is sensitive to meaning. Let us denote such a \emph{source similarity} score $\ssrc(x,\hat x)$.

Based on these functions, we define the \emph{target relative score decrease} as:
\begin{equation}
\dtgt(y, y_M, \hat y_M)=
\begin{cases}
    0 \text{ if } \small\stgt(y, \hat y_M) \ge \stgt(y, y_M) \\
    \frac{\stgt(y, y_M)-\stgt(y, \hat y_M)}{\stgt(y, y_M)} \text{ otherwise.}
\end{cases}
\end{equation}

The choice to report the \emph{relative} decrease in $\stgt$ makes scores comparable across different models or languages\footnote{Note that we do not allow negative $\dtgt$ to keep all scores between 0 and 1.}. For instance, for languages that are comparatively easy to translate (\eg{} French-English), $\stgt$ will be higher in general, and so will the gap between $\stgt(y, y_M)$ and $\stgt(y, \hat{y}_M)$. However this does not necessarily mean that attacks on this language pair are more effective than attacks on a ``difficult'' language pair (\eg{} Czech-English) where $\stgt$ is usually smaller.

We recommend that both $\ssrc$ and $\dtgt$ be reported when presenting adversarial attack results. However, in some cases where a single number is needed, we suggest reporting the attack's \emph{success} $\mathcal S\coloneqq \ssrc + \dtgt $.
The interpretation is simple: $\mathcal S>1 \Leftrightarrow\dtgt>1-\ssrc$, which means that the attack has destroyed the target meaning ($\dtgt$) more than it has destroyed the source meaning ($1-\ssrc$).

Importantly, this framework can be extended beyond strictly meaning-preserving attacks. For example, for targeted keyword introduction attacks \cite{cheng2018seq2sick,Ebrahimi2018OnAE}, the same evaluation framework can be used if $\stgt$ (resp. $\ssrc$) is modified to account for the presence (resp. absence) of the keyword (or its translation in the source). Similarly this can be extended to other tasks by adapting $\stgt$  (\eg{} for classification one would use the zero-one loss, and adapt the success threshold).

\subsection{Similarity Metrics}
\label{sec:eval_metrics}

Throughout Section \ref{sec:adv_tradeoff}, we have not given an exact description of the semantic similarity scores $\ssrc$ and $\stgt$. Indeed, automatically evaluating the semantic similarity between two sentences is an open area of research and it makes sense to decouple the definition of adversarial examples from the specific method used to measure this similarity. In this section, we will discuss manual and automatic metrics that may be used to calculate it.

\subsubsection{Human Judgment}
\label{sec:human_judgement}

Judgment by speakers of the language of interest is the \textit{de facto} gold standard metric for semantic similarity. Specific criteria such as adequacy/fluency \cite{Ma2006CorpusSF}, acceptability \cite{Goto2013OverviewOT}, and 6-level semantic similarity \cite{cer-EtAl:2017:SemEval} have been used in evaluations of \ac{mt} and sentence embedding methods.
In the context of adversarial attacks, we propose the following 6-level evaluation scheme, which is motivated by previous measures, but designed to be (1) symmetric, like \citet{cer-EtAl:2017:SemEval}, (2) and largely considers meaning preservation but at the very low and high levels considers fluency of the output\footnote{This is important to rule out nonsensical sentences and distinguish between clean and ``noisy'' paraphrases (\eg{} typos, non-native speech\ldots). We did not give annotators additional instruction specific to typos.}, like \citet{Goto2013OverviewOT}:

{
\begin{center}
\framebox{
\begin{minipage}{0.9\columnwidth}
How would you rate the similarity between the meaning of these two sentences?
\begin{enumerate}[itemsep=-4pt]
\setcounter{enumi}{-1}
\item The meaning is completely different or one of the sentences is meaningless
\item The topic is the same but the meaning is different
\item Some key information is different
\item The key information is the same but the details differ
\item Meaning is essentially equal but some expressions are unnatural
\item Meaning is essentially equal and the two sentences are well-formed English\footnote{Or the language of interest.}
\end{enumerate}
\end{minipage}
}
\end{center}
}

\subsubsection{Automatic Metrics}
\label{sec:auto_metrics}

Unfortunately, human evaluation is expensive, slow and sometimes difficult to obtain, for example in the case of low-resource languages. This makes automatic metrics that do not require human intervention  appealing for experimental research.
This section describes 3 evaluation metrics commonly used as alternatives to human evaluation, in particular to evaluate translation models.%
\footnote{
Note that other metrics of similarity are certainly applicable within the overall framework of Section \ref{sec:human_judgement}, but we limit our examination in this chapter to the three noted here.
}

\textbf{BLEU:} \cite{papineni-EtAl:2002:ACL} is an automatic metric based on n-gram precision coupled with a penalty for shorter sentences. It relies on exact word-level matches and therefore cannot detect synonyms or morphological variations.

\textbf{METEOR:} \cite{denkowski:lavie:meteor-wmt:2014} first estimates alignment between the two sentences and then computes unigram F-score (biased towards recall) weighted by a penalty for longer sentences. Importantly, METEOR uses stemming, synonymy and paraphrasing information to perform alignments. On the downside, it requires language specific resources.

\textbf{chrF:} \cite{popovic:2015:WMT} is based on the character $n$-gram F-score. In particular we will use the chrF2 score (based on the F2-score --- recall is given more importance), following the recommendations from \citet{popovic:2016:WMT}. By operating on a sub-word level, it can reflect the semantic similarity between different morphological inflections of one word (for instance), without requiring language-specific knowledge which makes it a good one-size-fits-all alternative.

Because multiple possible alternatives exist, it is important to know which is the best stand-in for human evaluation.
To elucidate this, we will compare these metrics to human judgment in terms of Pearson correlation coefficient on outputs resulting from a variety of attacks in Section \ref{sec:corr_human_auto}.

\section{Gradient-Based Adversarial Attacks}
\label{sec:attacks}

In this section, we overview the adversarial attacks we will be considering in the rest of this chapter.

\subsection{Attack Paradigm}
\label{sec:attack_paradigm}

\begin{table*}[!h]
\centering
{
\caption{\label{tab:qual_constraints} Examples of different adversarial inputs. The substituted word is highlighted.}
\begin{tabular}{ll}
\hline\hline
Original & {\bf Pourquoi} faire cela ? \\
English gloss & {\bf Why} do this? \\
\unconstrained{} & {\color{red}\bf construisant} {\color{orange}(English: building)} faire cela ? \\ 
\knn{} & {\color{red}\bf interrogez} {\color{orange}(English: interrogate)} faire cela ? \\
\unkonly{} & {\color{red}\bf Puorquoi} {\color{orange}(typo)} faire cela ?\\ \hline\hline
Original& Si seulement je pouvais me muscler {\bf aussi} rapidement.\\
English gloss& If only I could build my muscle {\bf this} fast.\\
\unconstrained{} & Si seulement je pouvais me muscler {\color{red}\bf etc}  rapidement.\\
\knn{} & Si seulement je pouvais me muscler {\color{red}\bf plsu} {\color{orange}(typo for ``more'')} rapidement.\\
\unkonly{} & Si seulement je pouvais me muscler {\color{red}\bf asusi} {\color{orange}(typo)} rapidement.\\\hline\hline
\end{tabular}
}
\end{table*}

We perform gradient-based attacks that replace one word in the sentence so as to maximize an adversarial loss function $\Ladv$, similar to the substitution attacks proposed in \cite{ebrahimi2018hotflip}.

\subsubsection{General Approach}

Precisely, for a word-based translation model $M$%
\footnote{Note that this formulation is also valid for character-based models (see \citet{Ebrahimi2018OnAE}) and subword-based models. For subword-based models, additional difficulty would be introduced due to changes to the input resulting in different subword segmentations. This poses an interesting challenge that is beyond the scope of the current work.}, and given an input sentence $w_1,\ldots,w_n$, we find the position $i^*$ and word $w^*$ satisfying the following optimization problem:
\begin{equation}\label{eq:adv_optim}
    \argmax_{1\leq i\leq n, \hat w\in \mathcal V}\Ladv(w_0,\ldots,w_{i-1},\hat w, w_{i+1},\ldots,w_n)
\end{equation}
\noindent where $\Ladv$ is a differentiable function which represents our adversarial objective. Using the first order approximation of $\Ladv$ around the original word vectors $\w_1,\ldots,\w_n$\footnote{More generally we will use the bold $\w$ when talking about the embedding vector of word $w$}, this can be derived to be equivalent to optimizing 
\begin{equation}
   \argmax_{1\leq i\leq n, \hat w\in \mathcal V}\left[\hat\w-{\w}_i\right]^\intercal\nabla_{\w_i}\Ladv.
\end{equation}

The above optimization problem can be solved by brute-force in $\bigO{n\vert\mathcal V\vert}$ space complexity, whereas the time complexity is bottlenecked by a $\vert\mathcal V\vert\times d$ times $n\times d$ matrix multiplication, which is not more computationally expensive than computing logits during the forward pass of the model. Overall, this naive approach is sufficiently fast to be conducive to adversarial training.
We also found that the attacks benefited from normalizing the gradient by taking its sign.

Extending this approach to finding the optimal perturbations for more than 1 substitution would require exhaustively searching over all possible combinations.
However, previous work \cite{Ebrahimi2018OnAE} suggests that greedy search is a good enough approximation.

\subsubsection{The Adversarial Loss $\Ladv$}

We want to find an adversarial input $\hat x$ such that, assuming that the model has produced the correct output $y_1,\ldots,y_{t-1}$ up to step $t-1$ during decoding, the probability that the model makes an error at the next step $t$ is maximized. 

In the log-semiring, this translates into the following loss function:
\begin{equation}
    \Ladv(\hat x, y)=\sum_{t=1}^{\vert y\vert}\log(1-p(y_t\mid \hat x, y_1,\ldots,y_{t-1}))
\end{equation}

\subsection{Enforcing Semantically Similar Adversarial Inputs}
\label{sec:constraints}

In contrast to previous methods, which don't consider meaning preservation, we
propose simple modifications of the approach presented in Section \ref{sec:attack_paradigm} to create adversarial perturbations at the word level that are more likely to preserve meaning.
The basic idea is to restrict the possible word substitutions to similar words. We compare two sets of constraints:

\textbf{\knn:} This constraint enforces that the word be replaced only with one of its 10 nearest neighbors in the source embedding space. This has two effects: first, the replacement will be likely semantically related to the original word (if words close in the embedding space are indeed semantically related, as hinted by Table~\ref{tab:qual_constraints}). Second, it ensures that the replacement's word vector is close enough to the original word vector that the first order assumption is more likely to be satisfied.

\textbf{\unkonly:} This constraint requires that the substituted words must be obtained by swapping characters. Word internal character swaps have been shown to not affect human readers greatly \cite{mccusker1981word}, hence making them likely to be meaning-preserving. Moreover we add the additional constraint that the substitution must not be in the vocabulary, which will likely be particularly meaning-destroying on the target side for the word-based models we test here. In such cases where word-internal character swaps are not possible or can't produce \ac{oov} words, we resort to the naive strategy of repeating the last character of the word. The exact procedure used to produce this kind of perturbations is described in Listing \ref{lst:gen_char_swap}. Note that for a word-based model, every \ac{oov} will look the same (a special \unk{} token), however the choice of \ac{oov} will still have an influence on the output of the model because we use unk-replacement.

\begin{lstlisting}[language=Python,label={lst:gen_char_swap}]
def make_oov(
  word,
  vocab,
  max_scrambling,
):
  """Modify a word to make it OOV
  (while keeping the meaning)"""
  # If the word has >3 letters
  # try scrambling them
  L = len(word)
  if L > 3:
    # For a fixed number of steps
    for _ in range(max_scrambling):
      # Swap two adjacent letters
      # in the middle of the word
      pos = random.randint(1, L - 3)
      word = word[:pos] 
      word += word[pos+1] + word[pos]
      word += word[pos+2:]
      # If we got an OOV already just
      # return it
      if word not in vocab:
        return word
  # If nothing worked, or the word is
  # too short for scrambling, just
  # repeat the last letter ad nauseam
  char = word[-1]
  while word in vocab:
    word = word + char
  return word
\end{lstlisting}

In contrast, we refer the base attack without constraints as {\bf \unconstrained} hereforth. Table \ref{tab:qual_constraints} gives qualitative examples of the kind of perturbations generated under the different constraints.

For subword-based models, we apply the same procedures at the subword-level  on the original segmentation. We then de-segment and re-segment the resulting sentence (because changes at the subword or character levels are likely to change the segmentation of the resulting sentence).

\section{Experiments}
\label{sec:adv_attacks_experiments}

Our experiments serve two purposes.
First, we examine our proposed framework of evaluating adversarial attacks (Section \ref{sec:eval_adv_attacks}), and also elucidate which automatic metrics correlate better with human judgment for the purpose of evaluating adversarial attacks (Section \ref{sec:corr_human_auto}). Second, we use this evaluation framework to compare various adversarial attacks and demonstrate that adversarial attacks that are explicitly constrained to preserve meaning receive better assessment scores (Section \ref{sec:attack_results}).

\subsection{Experimental setting}

\textbf{Data:}
Following previous work on adversarial examples for \ac{seq2seq} models \cite{belinkov2018synthetic,Ebrahimi2018OnAE}, we perform all experiments on the IWSLT2016 dataset \cite{Cettolo2016TheI2} in the \{French,German,Czech\}$\rightarrow$English directions (\fren{}, \deen{} and \csen{}). We compile all previous IWSLT test sets before 2015 as validation data, and keep the 2015 and 2016 test sets as test data. The data is tokenized with the Moses tokenizer \cite{koehn2007moses}. The exact data statistics can be found in Table \ref{tab:iwslt2016_stats}.

\begin{table}[!h]
\centering
\caption{\label{tab:iwslt2016_stats}IWSLT2016 data statistics.}
\begin{tabular}{lccc}
& \#train & \#valid & \#test \\ \hline
\fren & 220.4k & 6,824 & 2,213 \\
\deen & 196.9k & 11,825 & 2,213 \\
\csen & 114.4k & 5,716 & 2,213\\
\hline
\end{tabular}
\end{table}

\textbf{\ac{mt} Models:}
We perform experiments with two common \ac{nmt} models. The first is an LSTM based encoder-decoder architecture with attention~\citep{luong-pham-manning:2015:EMNLP}. It uses 2-layer encoders and decoders, and dot-product attention. We set the word embedding dimension to 300 and all others to 500.
The second model is a self-attentional Transformer \cite{vaswani2017attention}, with 6 1024-dimensional encoder and decoder layers and 512 dimensional word embeddings. Both the models are trained with Adam \cite{Kingma2014Adam}, dropout \cite{srivastava2014dropout} of probability 0.3 and label smoothing \cite{Szegedy2016RethinkingTI} with value 0.1. We experiment with both word based models (vocabulary size fixed at 40k) and subword based models (BPE \cite{sennrich2016neural} with 30k operations). For word-based models, we perform \unk{} replacement, replacing \unk{} tokens in the translated sentences with the source words with the highest attention value during inference. The full experimental setup and source code are available at \codeurl{}.

\textbf{Automatic Metric Implementations:} To evaluate both sentence and corpus level BLEU score, we first de-tokenize the output and use {\tt sacreBLEU}\footnote{\url{https://github.com/mjpost/sacreBLEU}} \cite{post2018call} with its internal {\tt intl} tokenization, to keep BLEU scores agnostic to tokenization. We compute METEOR using the official implementation\footnote{\url{http://www.cs.cmu.edu/~alavie/METEOR/}}. ChrF is reported with the {\tt sacreBLEU} implementation on detokenized text with default parameters. A toolkit implementing the evaluation framework described in Section \ref{sec:adv_tradeoff} for these metrics is released at \teapoturl{}.

\subsection{Correlation of Automatic Metrics with Human Judgment}
\label{sec:corr_human_auto}

\begin{table}[!ht]
\centering
\caption{\label{tab:human_eval_results} Correlation of automatic metrics to human judgment of adversarial source and target sentences. ``$^*$'' indicates that the correlation is significantly better than the next-best one.}
\begin{tabular}{lccc}
Language & BLEU & METEOR & chrF\\\hline\hline
French&0.415&0.440&\bf0.586$^*$\\
English&0.357&0.478$^*$&\bf0.497\\\hline
\end{tabular}
\end{table}

We first examine which of the automatic metrics listed in Section \ref{sec:eval_metrics} correlates most with human judgment for our adversarial attacks. For this experiment, we restrict the scope to the case of the LSTM model on \fren{}. For the French side, we randomly select 900 sentence pairs $(x,\hat x)$ from the validation set, 300 for each of the \unconstrained{}, \knn{} and \unkonly{} constraints. To vary the level of perturbation, the 300 pairs contain an equal amount of perturbed input obtained by substituting 1, 2 and 3 words. On the English side, we select 900 pairs of reference translations and translations of adversarial input $(y, \hat y_M)$ with the same distribution of attacks as the source side, as well as 300 $(y, y_M)$ pairs (to include translations from original inputs). This amounts to 1,200 sentence pairs in the target side.

These sentences are sent to English and French speaking annotators to be rated according to the guidelines described in Section \ref{sec:human_judgement}. Each sample (a pair of sentences) is rated by two independent evaluators. If the two ratings differ, the sample is sent to a third rater (an auditor and subject matter expert) who makes the final decision.

Finally, we compare the human results to each automatic metric with Pearson's correlation coefficient. The correlations are reported in Table \ref{tab:human_eval_results}. As evidenced by the results, chrF exhibits higher correlation with human judgment, followed by METEOR and BLEU. This is true both on the source side ($x$ vs $\hat x$) and in the target side ($y$ vs $\hat y_M$). We evaluate the statistical significance of this result using a paired bootstrap test for $p<0.01$. Notably we find that chrF is significantly better than METEOR in French but not in English. This is not too unexpected because METEOR has access to more language-dependent resources in English (specifically synonym information) and thereby can make more informed matches of these synonymous words and phrases. Moreover the French source side contains more ``character-level'' errors (from \unkonly{} attacks) which are not picked-up well by word-based metrics like BLEU and METEOR.

We provide a breakdown of the correlation coefficients of automatic metrics with human judgment for source-side meaning-preservation, both in terms of number of perturbed words (Table \ref{tab:corr_edits_breakdown}) and constraint (Table \ref{tab:corr_constraints_breakdown}). While those coefficients are computed on a much smaller sample size, and their differences are not all statistically significant with $p<0.01$, they exhibit the same trend as the results from Table \ref{tab:human_eval_results} (BLEU $<$ METEOR $<$ chrF). In particular Table \ref{tab:corr_edits_breakdown} shows that the good correlation of chrF with human judgment is not only due to the ability to distinguish between different number of edits.

\begin{table}
\centering
\caption{\label{tab:corr_edits_breakdown} Correlation of automatic metrics to human judgment of semantic similarity between original and adversarial source sentences, broken down by number of perturbed words. ``$^*$'' indicates that the correlation is significantly better than the next-best one.}
\begin{tabular}{cccc}
\# edits & BLEU & METEOR & chrF\\\hline\hline
1&0.351&0.352&\bf0.486$^*$\\
2&0.403&0.424&\bf0.588$^*$\\
3&0.334&0.393&\bf0.560$^*$\\
\hline
\end{tabular}
\end{table}

\begin{table}
\centering
\caption{\label{tab:corr_constraints_breakdown} Correlation of automatic metrics to human judgment of semantic similarity between original and adversarial source sentences, broken down by type of constraint on the perturbation. ``$^*$'' indicates that the correlation is significantly better than the next-best one.}
\begin{tabular}{cccc}
Constraint & BLEU & METEOR & chrF\\\hline\hline
\unconstrained{} &0.274 & 0.572 & \bf0.599\\
\unkonly{}&0.274 & 0.319 & \bf0.383\\
\knn{} & 0.534 & 0.584 & \bf0.606\\
\hline
\end{tabular}
\end{table}

Thus, in the following, we report attack results both in terms of chrF in the source ($\ssrc$) and \ac{rdb} in the target ($d_{\text{tgt}}$).

\begin{table}[t]

\centering
\caption{\label{tab:all_attacks_results} Target RDchrF and source chrF scores for all the attacks on all our models (word- and subword-based LSTM and Transformer).}
\begin{tabular}{clccccccc}
 &  & \multicolumn{3}{c}{LSTM} &\ & \multicolumn{3}{c}{Transformer} \\\hline\hline
 & Language pair & \csen & \deen & \fren&\ & \csen & \deen & \fren\\\cline{3-9}
\multirow{8}{*}{Word-based}& & \multicolumn{3}{c}{Target RDChrF} & & \multicolumn{3}{c}{Target RDChrF}\\\cline{3-5}\cline{7-9}
 & Original chrF & 45.68 & 49.43 & 57.49 & & 47.66 & 51.08 & 58.04\\
 &\unconstrained{} & 25.38 & 25.54 & 25.59 & & 25.24 & 25.00 & 24.68\\
 &\unkonly{} & 24.11 & 24.94 & 23.60 & & 21.59 & 23.23 & 21.75\\
 &\knn{} & 15.00 & 15.59 & 15.22& & 20.74 & 19.97 & 18.59 \\
 &&\multicolumn{3}{c}{Source chrF}&&\multicolumn{3}{c}{Source chrF}\\\cline{3-5}\cline{7-9}
 &\unconstrained{} & 70.14 & 72.39 & 74.29 & & 69.03 & 71.93 & 73.23\\
 &\unkonly{} & 82.65 & 84.40 & 86.62 & & 84.13 & 85.97 & 87.02\\
 &\knn{} & 78.08 & 78.11 & 77.62 & & 74.94 & 77.92 & 77.88\\\hline
\multirow{8}{*}{Subword-based}& & \multicolumn{3}{c}{Target RDChrF} & & \multicolumn{3}{c}{Target RDChrF}\\\cline{3-5}\cline{7-9}
 & Original chrF &48.30 & 52.42 & 59.08 & & 49.70 & 54.01 & 59.65\\
 &\unconstrained{} & 25.79 & 26.03 & 26.96 & & 23.97 & 25.07 & 25.28\\
 &\unkonly{} & 18.65 & 19.15 & 19.75 & & 16.98 & 18.38 & 17.85\\
 &\knn{} & 15.00 & 16.26 & 17.12 & & 19.02 & 18.58 & 18.63\\
 &&\multicolumn{3}{c}{Source chrF}&&\multicolumn{3}{c}{Source chrF}\\\cline{3-5}\cline{7-9}
 &\unconstrained{} & 69.32 & 72.12 & 73.57 & & 68.66 & 71.51 & 72.65\\
 &\unkonly{} & 85.84 & 87.46 & 87.98 & & 85.79 & 87.07 & 87.99\\
 &\knn{} & 76.17 & 77.74 & 78.03 & & 73.05 & 75.91 & 76.54\\
\hline
\end{tabular}
\end{table}

\begin{figure*}[!t]
\centering
\includegraphics[width=\textwidth]{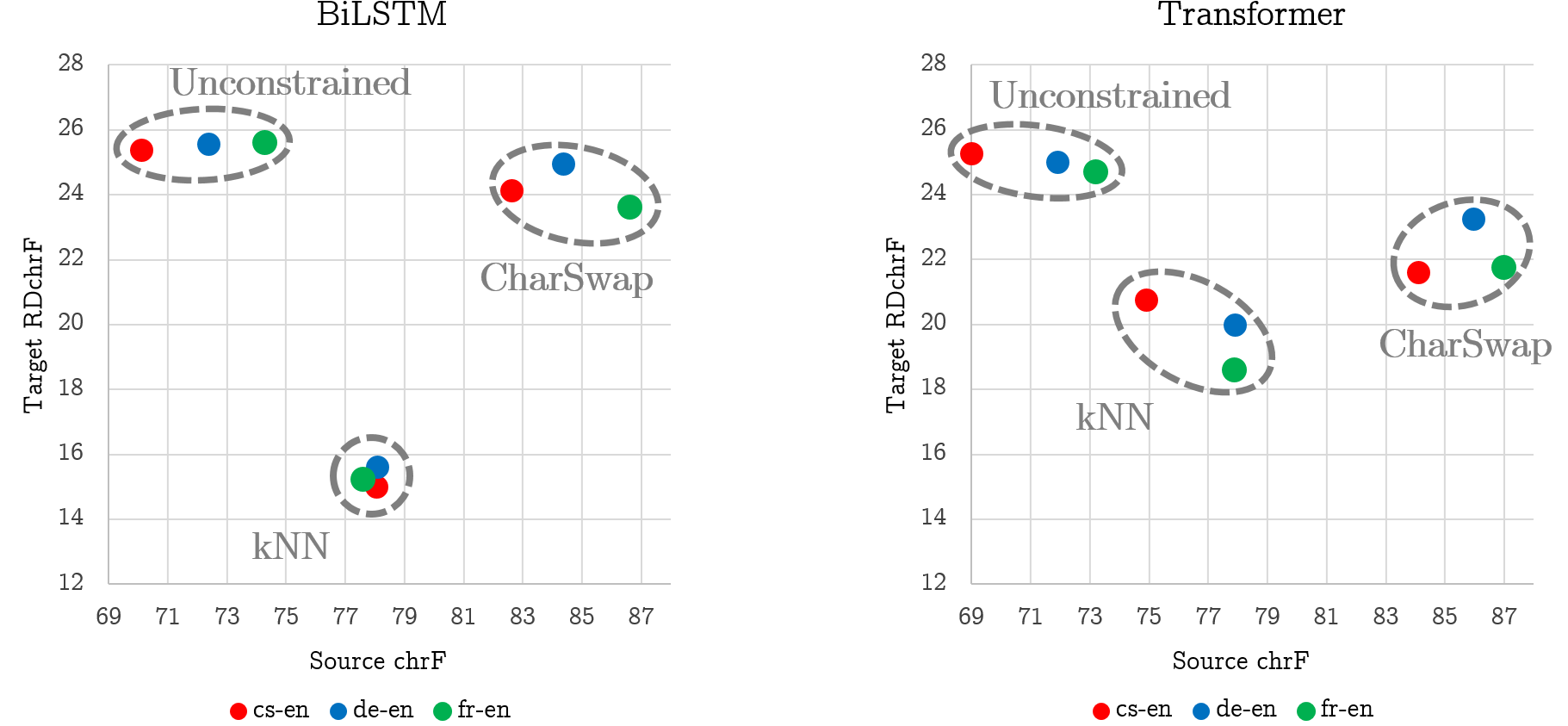}
\caption{\label{fig:chrf_plots} Graphical representation of the results in Table \ref{tab:all_attacks_results} for word-based models. High source chrF and target \ac{rdb} (upper-right corner) indicates a good attack.}
\end{figure*}

\subsection{Attack Results}
\label{sec:attack_results}

We can now compare attacks under the three constraints \unconstrained{}, \knn{} and \unkonly{} and draw conclusions on their capacity to preserve meaning in the source and destroy it in the target. Attacks are conducted on the validation set using the approach described in Section \ref{sec:attack_paradigm} with 3 substitutions (this means that each adversarial input is at edit distance at most 3 from the original input). Results (on a scale of 0 to 100 for readability) are reported in Table \ref{tab:all_attacks_results} for both word- and subword- based LSTM and Transformer models. To give a better idea of how the different variables (language pair, model, attack) affect performance, we give a graphical representation of these same results in Figure \ref{fig:chrf_plots} for the word-based models.
The rest of this section discusses the implication of these results.

\textbf{Source chrF Highlights the Effect of Adding Constraints:}
Comparing the \knn{} and \unkonly{} rows to \unconstrained{} in the ``source'' sections of Table \ref{tab:all_attacks_results} clearly shows that constrained attacks have a positive effect on meaning preservation. Beyond validating our assumptions from Section \ref{sec:constraints}, this shows that source chrF is useful to carry out the comparison in the first place\footnote{It can be argued that using chrF gives an advantage to \unkonly{} over \knn{} for source preservation (as opposed to METEOR for example). We find that this is the case for Czech and German (source METEOR is higher for \knn{}) but not French. Moreover we find (see \eg{} Table \ref{tab:corr_edits_breakdown}) that chrF correlates better with human judgement even for \knn{}.}. To give a point of reference, results from the manual evaluation carried out in Section \ref{sec:corr_human_auto} show that that $90\%$ of the French sentence pairs to which humans gave a score of 4 or 5 in semantic similarity have a chrF $>78$.

\textbf{Different Architectures are not Equal in the Face of Adversity:}
Inspection of the target-side results yields several interesting observations. First, the high \ac{rdb} of \unkonly{} for word-based model is yet another indication of their known shortcomings when presented with words out of their training vocabulary, even with \unk{}-replacement.
Second, and perhaps more interestingly, Transformer models appear to be less robust to small embedding perturbations (\knn{} attacks) compared to LSTMs. Although the exploration of the exact reasons for this phenomenon is beyond the scope of this work, this is a good example that \ac{rdb} can shed light on the different behavior of different architectures when confronted with adversarial input.
Overall, we find that the \unkonly{} constraint is the only one that consistently produces attacks with $>1$ average success (as defined in Section \ref{sec:adv_tradeoff}) according to Table \ref{tab:all_attacks_results}. Table \ref{tab:qual_unkonly_attack} contains two qualitative examples of this attack on the LSTM  model in \fren{}.

\begin{table}[!h]
\centering
{
\caption{\label{tab:qual_unkonly_attack} Example of \unkonly{} attacks on the \fren{} LSTM. The first example is a successful attack (high source chrF and target \ac{rdb}) whereas the second is not.}

\begin{tabular}{ll}
\hline\hline
\multicolumn{2}{c}{Successful attack}\\
\multicolumn{2}{c}{(source chrF $=80.89$, target \ac{rdb} $=84.06$)}\\\hline
Original & Ils le r\'{e}investissent directement en engageant plus de proc\`{e}s.\\
Adv. src& {\color{red}Ilss} le r\'{e}investissent {\color{red}dierctement} en {\color{red}engagaent} plus de proc\`{e}s.\\
Ref. & They plow it right back into filing more troll lawsuits.\\
Base output& They direct it directly by engaging more cases.\\
Adv. output& .. de plus.\\\hline\hline
\multicolumn{2}{c}{Unsuccessful attack}\\
\multicolumn{2}{c}{(source chrF $=54.46$, target \ac{rdb} $=0.00$)}\\\hline
Original & C'\'{e}tait en Juillet 1969.\\
Adv. src & C'{\color{red}\'{e}tiat} en {\color{red}Jiullet} 1969.\\
Ref. & This is from July, 1969.\\
Base output & This was in July 1969.\\
Adv. output & This is. in 1969.\\\hline\hline
\end{tabular}
}
\end{table}

\section{Adversarial Training with Meaning-Preserving Attacks}
\label{sec:adv_train}

\subsection{Adversarial Training}

Adversarial training \cite{Goodfellow2014ExplainingAH} augments the training data with adversarial examples. Formally, in place of the \ac{nll} objective on a sample $x, y$, $\mathcal{L}(x,y)=NLL(x,y)$, the loss function is replaced with an interpolation of the \ac{nll} of the original sample $x,y$ and an adversarial sample $\hat x, y$:

\begin{equation}
    \mathcal{L}'(x,y)=(1-\alpha)NLL(x,y) + \alpha NLL(\hat x,y)
\end{equation}

\citet{Ebrahimi2018OnAE} suggest that while adversarial training improves robustness to adversarial attacks, it can be detrimental to test performance on non-adversarial input.
We investigate whether this is still the case when adversarial attacks are largely meaning-preserving.

In our experiments, we generate $\hat x$ by applying 3 perturbations on the fly at each training step. To maintain training speed we do not solve Equation (\ref{eq:adv_optim}) iteratively but in one shot by replacing the argmax by top-3. Although this is less exact than iterating, this makes adversarial training time less than $2\times$ slower than normal training. We perform adversarial training with perturbations without constraints (\unconstrained{}-adv) and with the \unkonly{} constraint (\unkonly{}-adv). All experiments are conducted with the word-based LSTM model.

\subsection{Results}
\label{sec:adv_train_experiments}

Test performance on non-adversarial input is reported in Table \ref{tab:adv_train_bleu_scores}. In keeping with the rest of the paper, we primarily report chrF results, but also show the standard BLEU as well.

We observe that when $\alpha=1.0$, \ie{}\ the model only sees the perturbed input during training\footnote{This setting is reminiscent of word dropout \cite{iyyer-EtAl:2015:ACL-IJCNLP}.}, the \unconstrained{}-adv model suffers a drop in test performance, whereas \unkonly{}-adv's performance is on par with the original. This is likely attributable to the spurious training samples $(\hat x, y)$ where $y$ is not an acceptable translation of $\hat x$ introduced by the lack of constraint. This effect disappears when $\alpha=0.5$ because the model sees the original samples as well.

Not unexpectedly, Table \ref{tab:adv_train_robustness} indicates that \unkonly{}-adv is more robust to \unkonly{} constrained attacks for both values of $\alpha$, with $1.0$ giving the best results. On the other hand, \unconstrained{}-adv is similarly or more vulnerable to these attacks than the baseline.
Hence, we can safely conclude that adversarial training with \unkonly{} attacks improves robustness while not impacting test performance as much as unconstrained attacks.

\begin{table}[t]
\centering
\begin{tabular}{lccc}
Language pair & \csen & \deen & \fren\\\hline\hline
\multirow{2}{*}{Base} & 44.21 & 49.30 & 55.67  \\
& \small (22.89) & \small (28.61) & \small (35.28)\\\cline{2-4}
 & \multicolumn{3}{c}{$\alpha=1.0$} \\\cline{2-4}
\multirow{2}{*}{\unconstrained{}{}-adv} & 41.38 & 46.15 & 53.39 \\
& \small (21.51) & \small (27.06) & \small (33.96)\\
\multirow{2}{*}{\unkonly{}-adv} & 43.74 & 48.85 & 55.60 \\
& \small (23.00) & \small (28.45) & \small (35.33)\\\cline{2-4}
 & \multicolumn{3}{c}{$\alpha=0.5$} \\\cline{2-4}
\multirow{2}{*}{\unconstrained{}{}-adv} &  43.68 & 48.60 & 55.55 \\
& \small (22.93) & \small (28.30) & \small (35.25)\\
\multirow{2}{*}{\unkonly{}-adv} & 44.57 & 49.14 & 55.88 \\
& \small (23.66) & \small (28.66) & \small (35.63)\\
\hline
\end{tabular}
\caption{\label{tab:adv_train_bleu_scores}ChrF (BLEU) scores on the original test set before/after adversarial training of the word-based LSTM model.}
\end{table}

\begin{table}[t]
\centering
\begin{tabular}{lccc}
Language pair & \csen & \deen & \fren\\\hline\hline
Base &  24.11 & 24.94 & 23.60   \\\cline{2-4}
 & \multicolumn{3}{c}{$\alpha=1.0$} \\\cline{2-4}
\unconstrained{}-adv &  25.99 & 26.24 & 25.67  \\
\unkonly{}-adv &  16.46 & 17.19 & 15.72  \\\cline{2-4}
 & \multicolumn{3}{c}{$\alpha=0.5$} \\\cline{2-4}
\unconstrained{}-adv & 26.52 & 27.26 & 24.92  \\
\unkonly{}-adv & 20.41 & 20.24 & 16.08 \\
\hline
\end{tabular}
\caption{\label{tab:adv_train_robustness} Robustness to \unkonly{} attacks on the validation set with/without adversarial training (\ac{rdb}). Lower is better.}
\end{table}

\section{Related work}
\label{sec:related}

Following seminal work on adversarial attacks by \citet{Szegedy2013IntriguingPO}, 
\citet{Goodfellow2014ExplainingAH} introduced gradient-based attacks and adversarial training. Since then, a variety of attack \cite{MoosaviDezfooli2016DeepFoolAS} and defense \cite{Ciss2017ParsevalNI,Kolter2017ProvableDA} mechanisms have been proposed.
Adversarial examples for \ac{nlp} specifically have seen attacks on sentiment \cite{papernot2016crafting,samanta2017towards,ebrahimi2018hotflip}, malware \cite{grosse2016adversarial}, gender \cite{reddy-knight:2016:NLPandCSS} or toxicity \cite{hosseini2017deceiving} classification to cite a few.

In \ac{mt}, methods have been proposed to attack word-based \cite{zhao2018generating,cheng2018seq2sick} and character-based \cite{belinkov2018synthetic,Ebrahimi2018OnAE} models. However these works side-step the question of meaning preservation in the source: they mostly focus on target side evaluation. Finally there is work centered around meaning-preserving adversarial attacks for \ac{nlp} via paraphrase generation \cite{iyyer-EtAl:2018:N18-1} or rule-based approaches \cite{jia-liang:2017:EMNLP2017,ribeiro-singh-guestrin:2018:Long,naik-EtAl:2018:C18-1,alzantot-EtAl:2018:EMNLP}. However the proposed attacks are highly engineered and focused on English.

\section{Conclusion}

This chapter highlights the importance of performing \emph{meaning-preserving} adversarial perturbations for \ac{nlp} models (with a focus on \ac{seq2seq}).
We proposed a general evaluation framework for adversarial perturbations and compared various automatic metrics as proxies for human judgment to instantiate this framework.
We then confirmed that, in the context of \ac{mt}, ``naive'' attacks do not preserve meaning in general, and proposed alternatives to remedy this issue. Finally, we have shown that a careful choice of more meaning-preserving perturbations is beneficial for adversarial training.

Since the original publication of the content in this chapter in 2019, a number of works have followed-up, either by adapting the evaluation framework to other tasks, \eg{} semantic parsing in \cite{huang2021robustness}, or by building upon it for designing more imperceptible adversarial perturbations, for instance using the proposed evaluation metrics as rewards for reinforcement learning based perturbation generation \citep{zou2020reinforced}, or pushing further the concept of indistinguishable perturbations with encoding specific character substitutions \citep{boucher2021bad}.

\clearpage

\part{Making Robust Models}
\label{sec:making_robust}

\chapter{Modeling the Second Player in Distributionally Robust Optimization}
\label{ch:modeling_the_second_player_dro}

\section{Introduction}

Up to this point, we have principally been concerned with \emph{evaluating} the effect that various types of distributional shifts have on existing models. In this chapter and the next, we now tackle the problem of \emph{training} models that are more robust against distributional shifts.

The tendency of machine learning models to exhibit drops in accuracy when confronted with data from domains that are absent from or under-represented in their training data often arises from the objective function of \ac{erm}. Recall that in \ac{erm}, the parameters $\theta$ of the model are learned by minimizing the expectation of a loss function $\ell$ under a data distribution $p$ (or, specifically in practice, an associated empirical data distribution $\hat p$)
\begin{equation}\label{eq:erm_loss}
    \mathcal L_{\text {ERM}}(\theta)=\mathbb E_{(x,y)\sim \hat p}\ell(x,y, \theta).
\end{equation}
When the model encounters data sampled from a different distribution $q_\text{test}\neq p$, performance can suffer significantly. As described in Chapter \ref{ch:background}, \acf{dro} provides a natural solution to this issue by replacing the expected risk under a single distribution $p$ with the \emph{worst} expected risk over a pre-determined family of distributions $\mathcal Q$ (the ``uncertainty set'')
\begin{equation}\label{eq:dro_loss_for_pdro}
    \mathcal L_{\text {DRO}}(\theta)=\max_{q\in\mathcal Q}\E_{(x,y)\sim q}\ell(x,y, \theta).
\end{equation}
If $\mathcal Q$ contains $q_\text{test}$, the \ac{dro} objective upper bounds the expected risk under $q_\text{test}$. However, \textit{a priori} knowledge of possible test distributions is not always available or easy to acquire. For example, training a model to be robust to some demographic attributes ($\mathcal Q = \{q_{\text{demographic 1}}, q_{\text{demographic 2}},\ldots\}$) requires collecting and annotating data with the necessary information, an expensive and ethically fraught endeavour. In the absence of such information, one has to resort to defining the uncertainty set analytically, drawing on one's intuition of what constitutes a possible test distribution given the observed training distribution, such as using moment constraints \citep{delage2010distributionally,nguyen2020robust}, $f$-divergence \citep{ben2013robust,hu2013kullback,faury2020distributionally}, Wasserstein/IPM \citep{sinha2017certifying,husain2020distributional} balls, or coarse-grained mixture models \citep{oren2019distributionally,hu2018does}. However, the need for keeping the inner supremum in Eq.~(\ref{eq:dro_loss_for_pdro}) tractable limits the possible choices.

In this chapter, we propose that the uncertainty set be instead defined as a family of parametric generative models. The resulting \ac{dro} objective (\S\ref{sec:pdro_method}) is a differentiable game with two players: the original model $\ell(x,y;\theta)$ and a model of its worst-case distribution $q_\psi(x,y)$, the titular ``second player'' which we hereafter refer to as \emph{the adversary}. Using this formulation --- which we call \ac{p-dro} --- allows for more flexibility in the choice of the adversary's architecture (and so the uncertainty set). Unfortunately, finding a solution of this game via direct application of simultaneous gradient descent \citep{singh2000nash} is difficult \citep{balduzzi2018mechanics}. In particular, direct gradient descent on the uncertainty set suffers from instability due to the large variance of the gradients \citep{greensmith2004variance}, and hyper-parameter selection is not straightforward.

To address these challenges, we make two main contributions (\S\ref{sec:optimizing}): first, we propose a new relaxation of the \ac{dro} game's inner maximization problem (with KL constraints). The resulting objective is more amenable to simultaneous gradient update than the original zero-sum game and significantly improves training stability, while still yielding useful adversaries. Second, we develop a principled approach for selecting hyper-parameters: we leverage the learned adversaries to decide which of any two given models trained with \ac{p-dro} is more robust than the other.

We do an in-depth set of experiments analyzing the effect of our proposed changes on both a toy task as well as a more realistic, yet still synthetic sentiment classification task (\S\ref{sec:analysis}). Finally, we show that in the more realistic setting of toxicity detection, \ac{p-dro} yields models that are more robust to changes in demographic groups, even though these groups are unknown at training time, opening up applications in combatting dataset bias (\S\ref{sec:toxicity}). Code to reproduce our experiments can be found at \url{https://github.com/pmichel31415/P-DRO}.

\section{Parameterizing the Uncertainty Set}
\label{sec:pdro_method}

Consider a model parameterized by $\theta\in\mathbb R^{d_\text{model}}$. Minimizing the \ac{dro} objective described in Eq.~(\ref{eq:dro_loss_for_pdro}) over the uncertainty set $\mathcal Q$ turns the optimization problem into the min-max (or zero-sum) game
\begin{equation}
\label{eq:dro_game}
    \min_{\theta\in\mathbb{R}^d} \max_{q\in\mathcal Q}\E_{(x,y)\sim q}\ell(x,y, \theta).
\end{equation}
The first player controls the parameters $\theta$, whilst the second player controls the worst-case distribution $q$. In the absence of explicit information on groups of interest (such as demographics, domain, etc.), an adequate choice of the uncertainty set $\mathcal{Q}$ is critical to the success of \ac{dro}.
This is in fact very much an active area of research (\citet{sinha2017certifying,duchi2018learning,oren2019distributionally}, see \citet{rahimian2019distributionally} for a survey). $\mathcal Q$ must be sufficiently large to contain test distributions of interest, but if it is too large it may contain ``adversarial'' distributions on which no model can perform well. Moreover, the design of $\mathcal Q$ is also circumscribed by the necessity of keeping the min-max problem tractable, particularly in the context of stochastic optimization. In \citet{hu2013kullback} and \citet{duchi2016statistics} for example, the choice of $f$-divergence balls allows the use of duality arguments to reformulate (\ref{eq:dro_game}) as a more manageable min-min problem. Others, like \citet{hu2018does} or \citet{oren2019distributionally}, propose using mixture models, the simplicity of which enables them to solve the inner maximization problem efficiently.

Instead, we propose to explicitly model the second player in the \ac{dro} game as a parametric model $q_\psi$ of the data. Of course, not all parameterizations $\psi\in\mathbb R^{d_\text{adv}}$ of a given generative model represent useful distributions, and we require that the adversary stay ``close'' to the underlying true data distribution $p$. As a measure of distance between $q_\psi$ and $p$, we choose the KL \citep{kullback1951information} divergence due to its wide acceptance in the machine learning community, as well as its appealing properties in the context of \ac{dro}.\footnote{For instance: $\KL{q}{p}<+\infty$ implies that $q$ stays within the support of $p$} The KL upper bound, $\kappa$, is left as a parameter to be decided by the experimenter. We refer to the resulting \ac{dro} formulation as \acl{p-dro}
\begin{equation}
\label{eq:p_dro}
\min_{\theta}\max_{\substack{\psi\\\KL{q_\psi}{p}\leq \kappa}} \underbrace{\mathbb E_{(x,y)\sim q_\psi}\ell(x,y,\theta)}_{\mathcal L_{\text {P-DRO}}(\theta,\psi)}.
\end{equation}

\section{Optimizing P-DRO}
\label{sec:optimizing}

\begin{figure*}[!t]
\centering
\includegraphics[width=\columnwidth]{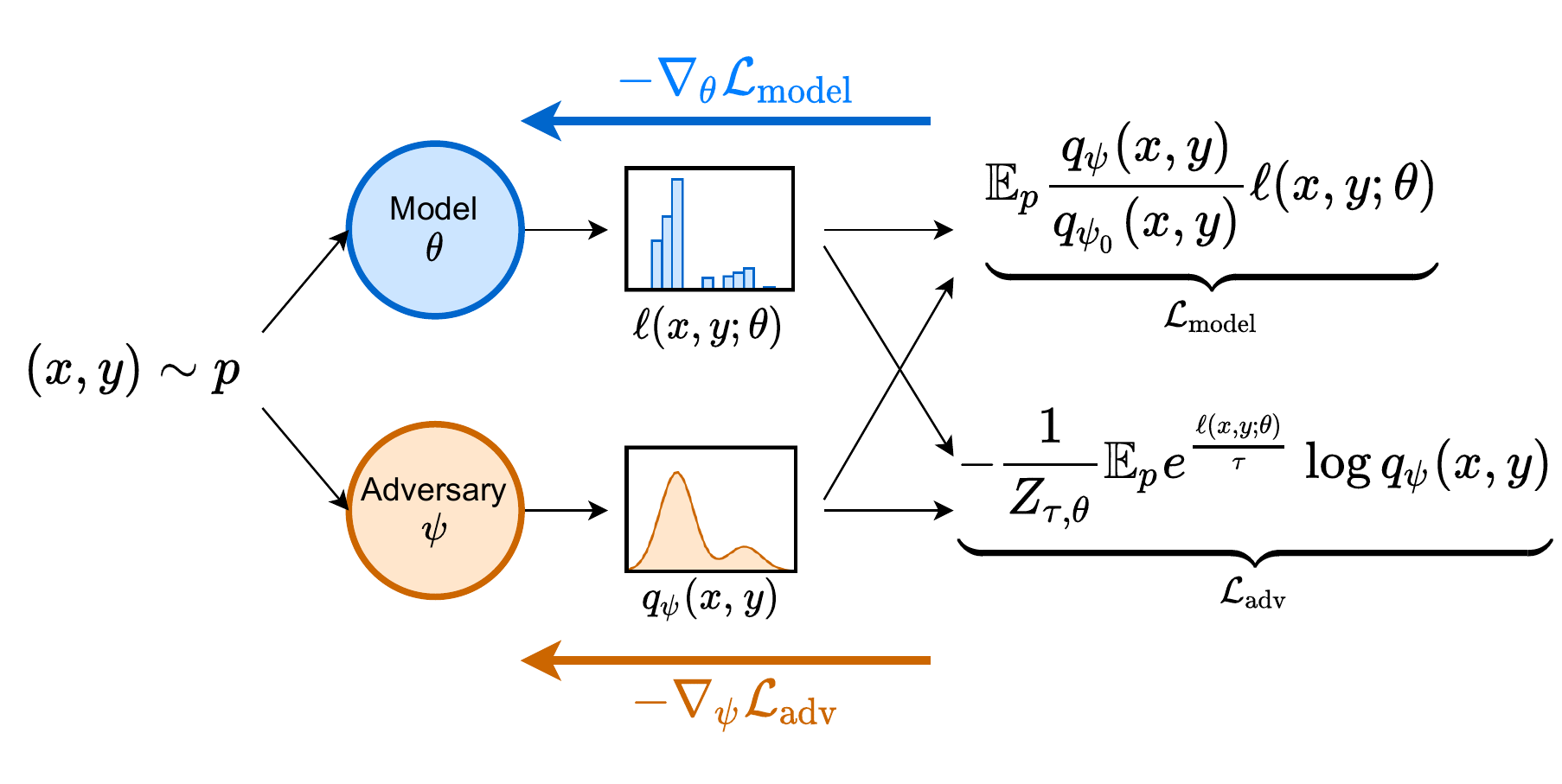}
\caption{\label{fig:pdro_diagram} Summary of \ac{p-dro}: At every step of training, $(x,y)$ pairs are sampled from the data distribution $p$ and fed to both the model $\theta$ and the adversary $\psi$. For every sample, the model produces loss values $\ell(x,y;\theta)$ and the adversary produces densities $q_\psi(x,y)$. Both are combined into $\mathcal L_{\text{model}}$ and $\mathcal L_{\text{adv}}$, which are used to update the $\theta$ and $\psi$ respectively, via simultaneous gradient updates.}
\end{figure*}

The min-max problem in Eq. (\ref{eq:p_dro}) belongs to a class of games called ``differentiable games'' (another famous representative being generative adversarial networks \citep{goodfellow2014generative}). We can search for a solution of this game with simultaneous gradient descent \citep{singh2000nash}, \ie{} by simultaneously updating $\theta$ and $\psi$ with $-\nabla_{\theta}\Ell_{\text {\ac{p-dro}}}$ and $\nabla_{\psi}\Ell_{\text {\ac{p-dro}}}$ respectively. Unfortunately, in general, there is no theoretical guarantee that simultaneous gradient descent will converge to a Nash equilibrium\footnote{Nash equilibria \citep{osborne1994course} can be thought of the game theoretic analog of global minima in optimization.} \citep{balduzzi2018mechanics}, nor that any such equilibrium even exists if the objective is non-convex in $\theta$ (or non-concave in $\psi$). The success of GANs and the follow-up literature \citep{wang2019generative} serves as an encouraging example that gradient based methods can yield useful solutions despite the pessimistic theoretical results. In this section, we discuss difficulties that arise when optimizing $\theta$ and $\psi$ jointly, and propose modifications of the objective to address them.

\subsection{Training the Model $\theta$}

We could train the model $\theta$ by taking negative gradient steps on $\E_{(x,y)\sim q_\psi}\ell(x,y;\theta)$. This gradient can be estimated by sampling examples from $q_\psi$ and averaging the gradient of their losses. Unfortunately, this objective requires that $q_\psi$ is well-behaved at all iterations, as it is the only source of supervision for $\theta$. If $q_\psi$ is initialized incorrectly or begins producing unrealistic $(x,y)$, the quality of $\theta$ degrades as it begins to learn a predictor on invalid training examples from $q_\psi.$ As an alternative, we opt to compute the gradients for $\theta$ with importance sampling, \ie{} rewriting $\Ell_{\text{ \ac{p-dro}}}$ as $\E_{(x,y)\sim p}\frac{q_\psi(x,y)}{p(x,y)}\ell(x,y;\theta)$, which ensures that all $(x,y)$ samples will be derived from the training set itself. Unfortunately, the true density $p$ is unknown to us. As an approximation, we replace $\frac{q_\psi(x,y)}{p(x,y)}$ with the likelihood ratio between $q_\psi$ and the maximum likelihood estimate of $p$, $q_{\psi_0}\coloneqq\argmax_{q_\psi} \E_{(x,y)\sim p}\log q_\psi(x,y)$. This changes the min-max problem to
\begin{equation}
\label{eq:p_dro_hat}
    \min_{\theta}\max_{\substack{\psi\\\KL{q_\psi}{p}\leq \kappa}}\underbrace{\mathbb E_{(x,y)\sim p}\frac{q_\psi(x,y)}{q_{\psi_0}(x,y)}\ell(x,y,\theta)}_{\mathcal L_{\text{model}}}.
\end{equation}
This becomes a simple expected loss objective, which we can estimate by sampling from the empirical distribution $\hat p$. In experiments, we find that with this formulation we are able to train robust $\theta$ even when $q_\psi$ is only a mediocre generative model (see Appendix \ref{sec:weak_adv}). To further stabilize training at the beginning of the optimization process, we initialize $\psi$ with $\psi_0$, making the objective exactly the same as \ac{erm} for the first gradient step.

\subsection{Training the Adversary $\psi$}
\label{sec:optimizing_psi}

According to Eq.~(\ref{eq:p_dro_hat}) the adversary $\psi$ must maximize $E_{(x,y)\sim q_\psi}\frac{p(x,y)}{q_{\psi_0}(x,y)}\ell(x,y,\theta)$ within a KL ball of fixed radius. This is challenging for several reasons: first, enforcing the bound is intractable for complex families of adversaries where \eg{} projecting onto the KL ball is another difficult optimization problem of its own. Second, maximizing the expectation with respect to the parameters of the distribution $q_\psi$ is prone to instability due to large gradient variance \citep{greensmith2004variance}.

\paragraph{Lagrangian Relaxation}
To address the first difficulty, we loosen the strict KL constraint and instead consider the Lagrangian relaxation $\mathbb{L}$
\begin{align}
    \mathbb L (\psi,\tau)&=\E_{(x,y)\sim q_\psi}\frac{p(x,y)}{q_{\psi_0}(x,y)}\ell(x,y,\theta) - \tau \left(\KL{q_\psi}{p} - \kappa\right).
\end{align}
We fix the Lagrangian multiplier $\tau>0$ as treat it as a ``temperature'' hyper-parameter. With some reorganization (which we develop in Appendix \ref{sec:lagrangian_reorg}), we can show that
\begin{align}
    \mathbb L (\psi,\tau)=-\tau\KL{q_\psi}{q^*_{\tau, \theta}} +C.
\end{align}
Where $q^*_{\tau, \theta}\propto p(x,y)e^{\frac{p(x,y)}{q_{\psi_0}(x,y)}\frac{\ell(x,y;\theta)}{\tau}}$ and $C$ is a constant in $\psi$. In other words, maximizing $\mathbb L$ in $\psi$ is equivalent to minimizing the KL divergence between $q_\psi$ and $q^*_{\tau, \theta}$. One difficulty with this objective is that $q^*_{\tau,\theta}$ depends upon the unknown probability density $p(x,y)$. We avoid this problem by treating the density ratio $\frac{p(x,y)}{q_{\psi_0}(x,y)}$ as a constant, which is closely related to assumptions that have been used successfully in past formulations of DRO~\citep{oren2019distributionally}. Empirically, we find that incorporating $q_{\psi_0}$ as a surrogate for $p$ is a serviceable approximation, as demonstrated in Section \ref{sec:analysis}.

\paragraph{Reversing the KL}
Minimizing the KL divergence in this direction is difficult for several reasons. First, it entails optimizing an expectation in $q_\psi$ over $\psi$, which is difficult due to the large variance of the gradients \citep{greensmith2004variance}. Second, computing this KL necessitates access to the true theoretical density $p(x,y)$ in order to compute $q^*_{\tau,\theta}(x,y)$ in the argument of the expectation, but this quantity is unknown in practice.\footnote{Note that substituting the empirical distribution $\hat p$ for $p$ poses issues here, because $q_\psi$ is not absolutely continuous with respect to $\hat p$.} To sidestep these issues, we elect to minimize the reverse direction $\KL{q^*_{\tau, \theta}}{q_\psi}$ instead. Due to the KL divergence being non-symmetric, this is a rather crude approximation\footnote{For instance, the optimum of the reverse KL doesn't necessarily match that of the forward KL within the parametric confusion set $\mathcal Q$}, the implications of which are discussed in \citet{norouzi2016reward}. However, we find that this approach dramatically stabilizes the gradient dynamics while still yielding good adversaries, as observed empirically in Section \ref{sec:kl_relaxation_ablation}. Discarding the entropy term (constant in $\psi$), the resulting problem is equivalent to minimizing
\begin{equation}
\label{eq:psi_objective}
 {\Ell}_{\text{adv}}(\psi, \tau)\coloneqq -\frac 1 {Z_{\tau,\theta}} \E_p e^{\frac{\ell(x,y;\theta)}{\tau}}\log q_\psi(x, y)
\end{equation}
in $\psi$, where $Z_{\tau,\theta}=\E_p e^{\frac{\ell(x,y;\theta)}{\tau}}$ is the normalizer of $q^*$. In this case, we can estimate this expectation by substituting the empirical distribution $\hat p$ for $p$ in the expectation.

\paragraph{Computing the Normalizer}
Approximating the inverse normalizer $\frac 1 {Z_{\tau,\theta}}$ in a minibatch yields a biased estimator. On the other hand, computing  ${Z_{\tau,\theta}}$ over the entire training data at each step is prohibitive since it requires computing the loss of every single example. As a middle ground, we keep a running normalizer $\Tilde{Z}_k$ computed from the average of the normalizers over a fixed number $k$ of consecutive minibatches. In other words, if $B_i$ and $\theta_i$ denote the minibatch and adversary parameters at step $i$ respectively, the normalizer at step $t$ will be
\begin{equation}
   \Tilde{Z}_k=\frac 1 {\sum_{i=t-k}^t|B_i|}\sum_{i=t-k}^t\sum_{x,y\in B_i}e^{\frac{\ell(x,y;\theta_i)}{\tau}}.
\end{equation}
If $k$ is too low, there is a risk of under-estimating the normalizer, especially if the distribution of weights contains infrequent high weight samples. On the other hand, if $k$ is too high there is a risk of using ``stale'' weights in the normalizer. In experiments, we treat $k$ as a hyper-parameter.

\subsection{Optimal Stopping}
\label{sec:optimal_stopping}

When should one stop training a model with \ac{p-dro}? In \ac{erm} it is customary to stop training after the empirical risk --- periodically evaluated on a held out validation dataset --- stops decreasing. This is particularly important to prevent over-fitting to the training data. However, it is not an appropriate criterion for \ac{p-dro}, since the model is not trained to minimize  empirical risk in the first place. A more pertinent choice is to compare the robust validation losses
\begin{equation}
    \mathcal{L}_{\text{robust},\text{valid}}(\theta)=\max_{q_\psi\in \mathcal{Q}}\underbrace{\frac 1 {|D_{\text{valid}}|} \sum_{x,y\in D_{\text{valid}}}\frac{q_\psi(x,y)}{q_{\psi_0}(x, y)}\ell(x,y;\theta)}_{\coloneqq \Ell_{\text{valid}}(\theta, \psi)}.
\end{equation}
However, finding the inner supremum for each of the $T$ evaluation checkpoints  $\theta_1\ldots\theta_T$ is expensive as it requires solving $T$ independent optimization problems. Instead, we leverage the existence of adversaries $\psi_t$ associated with each model $\theta_t$, as well as the initial adversary $\psi_0$ and take the maximum over the $T+1$ adversaries $\{\psi_0,\ldots,\psi_T\}$. Since our relaxation of the \ac{p-dro} objective loosens the KL constraint, we need weed out adversaries which might violate it. Specifically, we estimate the $\KL{q_\psi}{p}=\E_p {q_\psi}/{p}\log {q_\psi}/{p}$ on the validation set, using $q_\psi/q_{\psi_0}$ as a stand-in for $q_\psi/p$, and reject all adversaries for which the result is greater than a threshold, which we set to $\log 10$ based on preliminary experiments detailed in Appendix \ref{sec:valid_kl_threshold}.\footnote{To simplify notation, this additional constraint is implicit in the rest of this section.} We refer to this stopping criterion as \textbf{Minmax}.

Computing the full min-max necessitates keeping track of $T$ models and $T+1$ adversaries, which is ponderous when the model is large. As a solution, we propose an approximation, \textbf{Greedy-Minmax}, in which we only keep one best model $\theta^*$. At each evaluation step $T$, we compare $\theta_T$ to $\theta^*$, and update $\theta^*$ to whichever achieves lower robust validation loss over the $T+1$ adversaries $\psi_0,\ldots,\psi_T$.

By keeping track of only one additional model, and using the weights $\frac{q_{\psi_t}(x_i, y_i)}{q_{\psi_0}(x_i, y_i)}$ of individual examples in $D_\text{valid}$ as sufficient statistics for computing the loss against each adversary, Greedy-Minmax can be achieved with space complexity $2 d_{\text{model}}+T |D_\text{valid}|$, which is much more efficient than the $T (d_{\text{model}} + d_{\text{adv}})$ of Minmax.

\subsection{Hyper-parameter Selection}
\label{sec:model_selection}

Our proposed \ac{p-dro} method relies on 3 different hyper-parameters (in addition to the model's hyper-parameters): the adversary learning rate $\lambda$, the temperature $\tau$ and the size of the re-normalizing window $k$. As a consequence, we need a reliable criterion for deciding which of two configurations is better. This model comparison bears many similarities with the stopping problem described above. Therefore, we resort to a similar solution: given two models $\theta_1$, $\theta_2$ trained with \ac{p-dro}, and their respective adversaries $\{\psi^1_0,\ldots,\psi^1_T\}$,  $\{\psi^2_0,\ldots,\psi^2_T\}$ (for instance, the adversaries associated with $\theta_1$ and $\theta_2$ at periodic checkpoints during training), we select the best model following
\begin{equation}
    \theta^*=\argmin_{\theta\in\{\theta_1,\theta_2\}}\max_{\psi\in\{\psi^1_0,\ldots,\psi^1_T,\psi^2_0,\ldots,\psi^2_T\}} \mathcal{L}_{\text{valid}}(\theta, \psi).
\end{equation}

\section{Experimental Analysis of \ac{p-dro}}
\label{sec:analysis}

Before moving on to a real world scenario in Section \ref{sec:toxicity}, we first demonstrate that \ac{p-dro} is able to learn robust models in a synthetic \ac{nlp} task, and perform ablation studies to examine the importance of the various modifications described in Section \ref{sec:optimizing}.

\subsection{Experimental Setting}

For analysis purposes, we design a simple \ac{nlp} task amenable to \ac{dro}.
We specifically choose \ac{nlp} as a domain due to the striking success of language models as generative models of textual data \citep{sundermeyer2012lstm,radford2018improving}, which can be used to model the uncertainty set.
We base our task off of the binary version of the Stanford Sentiment Treebank dataset (SST-2; \citet{socher2013recursive}), which we modify to introduce spurious correlation. Specifically, we introduce a distractor token to some sentences. The distractor we use consists of prepending ``so , '' to the sentence (``i hated this movie'' $\longrightarrow$ ``so , I hated this movie''), which doesn't change the underlying sentiment. The resulting samples can be categorized in 4 ``groups'' depending on their label (positive or negative) and the presence or absence of the distractor. In particular, we add this distractor to 95\% of the negative reviews and 5\% of the positive reviews in the training and validation set, so that the presence of the distractor strongly correlates with negative sentiment (a similar construction is proposed in \citep{Utama2020TowardsDN}). In the test data, we modify 50\% of all sentences for each class equitably to ensure that there is enough data in each group, but we report ``average'' test accuracy by re-weighting the group accuracies to mimick the training distribution. We call this modified task \textbf{BiasedSST}.

\subsection{Model Settings}

In all experiments, we split the text into sub-word tokens using the tokenizer described in \citep{devlin2018bert}. For the classifier, we train a simple one layer BiLSTM model with embedding/hidden dimension 300.  During training, we sample minibatches that contain at most $64$ sentences or $2500$ tokens, whichever is greater, in order to prevent GPU memory overflow in case of long sentences. We train all models with Adam \citep{Kingma2014Adam} with an initial learning rate of $2\times 10^{-5}$, which we decay linearly at each step until the end of training. We validate the models every epoch. For BERT, we start from the \texttt{bert-base-uncased} checkpoint.

\subsection{Adversary Settings}

In all experiments, we use an auto-regressive transformer model based on the successful GPT-2 language model \citep{radford2019language} architecture but with 6 layers, a dimension of 512 and 8 attention heads (we experiment with a smaller, LSTM based adversary in Section \ref{sec:weak_adv}). In order to model the input output pair $(x,y)$, we pre-pend a special label-specific token to sentences before running them through the language model. We train the adversary with vanilla stochastic gradient descent, which we found more stable in experiments.

To initialize the adversary (to obtain $\psi_0$), we first pre-train the model on a generic, relatively large language modeling dataset, WikiText-103 \citep{merity2016pointer}. We also use a batch size of 64 samples or 2500 tokens, and train with Adam for 10 epochs, with a fixed learning rate of $3\times 10^{-4}$. Then, we fine-tune this model on each dataset, this time minimizing the negative log-likelihood of the $(x,y)$ pair, using the same hyper-parameters but a smaller learning rate ($10^{-5}$). We find that, due to the small to medium size of the datasets under consideration, this LM pretraining step helped achieve lower error on the generative modeling task.

\subsection{Baselines}
\label{sec:baseline_settings}

In addition to ERM, we also compare three other approaches. First, to appreciate how well the model could perform if the groups were known at training time, we train with Group-DRO on the oracle groups using an exponentiated-gradients based online algorithm (\textbf{Oracle DRO}; \citet{sagawa2019distributionally}). Second, we implement \textbf{Topic DRO} \citep{oren2019distributionally}, a method for \ac{dro} on \ac{nlp} where the uncertainty set is determined by mixtures of a topic model. Finally, we compare to non-parametric \ac{dro} with a \ac{kl} constrained uncertainty set \citep{hu2013kullback,hu2018does}, which we adapt to fit our online mini-batch training setting (\textbf{NonParam}). Below, we outline experimental details for the latter two.

\subsubsection{Topic DRO}

To train the topic model for Topic DRO, we first pre-process the text by removing all  punctuation, urls and user mentions (for twitter data). Importantly, we remove stop-words for our toxicity experiments but \textit{not} for our BiasedSST experiment. This is because the distractor token we use (``so'') belongs to most English stop words lists, and removing it would completely prevent the topic model from picking up on the groups of interest. We then estimate the parameters of the model with Gensim\footnote{\url{https://radimrehurek.com/gensim/}} and use similar settings as \citet{oren2019distributionally} ($\alpha=0.1$, $\beta=1.0$), setting the number of topics to 10.

For both Oracle-DRO and Topic DRO, we use the algorithm proposed in \citet{sagawa2019distributionally} to estimate the worst-case group (either oracle group or topic in Topic DRO) online during training. We perform grid-search over $\{1,0.1,0.01\}$ to find the best learning rate for the group weights update. For Oracle \ac{dro}, the best model is simply selected by robust validation accuracy. For Topic DRO, unless specified otherwise, we select the model with the lowest worst-case error over all topics.

\subsubsection{NonParam}

In the KL-constrained non-parametric setting, the min-max problem reads
\begin{equation}
\label{eq:non_param_kl_dro}
\min_{\theta}\max_{\substack{q~\text{s.t.}\\\KL{q_\psi}{p}\leq \kappa}} \mathbb E_{(x,y)\sim q}\ell(x,y,\theta).
\end{equation}
Here, $\kappa$ is the desired radius of the \ac{kl} ball, and is treated as a hyper-parameter. The solution of the inner maximum has an analytical solution of the form $q^*_{\theta}= \frac a {Z_{\theta,\tau^*}}p(x,y)e^{\frac{\ell(x,y;\theta)}{\tau^*}}$ (see \citet{hu2013kullback,hu2018does} for details) with $Z_{\theta,\tau^*}=\mathbb E_p e^{\frac{\ell(x,y;\theta)}{\tau^*}}$ and $\tau^*$ such that
\begin{equation*}
    \KL{q^*}{p}=\mathbb E_{p}\frac{e^{\frac{\ell(x,y;\theta)}{\tau^*}}}{Z_{\theta,\tau^*}}\left(\frac{\ell(x,y;\theta)}{\tau^*} - \log Z_{\theta,\tau^*}\right)=\kappa.
\end{equation*}
Note that both computing $Z_{\theta,\tau^*}$ and $\KL{q^*}{p}$ require taking expectations over $p$. In our setting, where $\ell(x,y;\theta)$ is the output of a large neural network, we cannot afford to take this expectation over the entire training data at each step. Instead, we fall back to taking the average over each mini-batch. We find $\tau^*$ with binary search in $\log_{10}$ space within the $[10^{-10},10^{10}]$ interval and clip to the lowest or highest value should the result lie outside the search interval.

In all experiments, we try 4 different values for $\kappa$: $0.01$, $0.1$, $1$ and $10$. Unless indicated otherwise, we perform early stopping and hyper-parameter selection using our Minmax criterion using the non-parametric weights as adversaries on the validation data.

\subsection{P-DRO can Learn Robust Models}
\label{sec:biased_sst_main_result}

\begin{table}[h]
\begin{center}
\caption{\label{tab:biased_sst_95_main_result} Average and robust accuracies on BiasedSST. Underlining indicates statistically significant difference compared to \ac{erm} ($p<0.05$)}
\resizebox{0.5\textwidth}{!}{
\begin{tabular}{llc}
\toprule
  & Robust & Average \\
\midrule
ERM & { 2.15}  {\scriptsize$\pm$ 0.97} & { 95.09}  {\scriptsize$\pm$ 0.16} \\\midrule
Topic DRO & \underline{ 5.18}  {\scriptsize$\pm$ 1.46} & { 95.00}  {\scriptsize$\pm$ 0.10} \\
NonParam & \underline{ 28.11}  {\scriptsize$\pm$ 2.16} & \underline{ 92.45}  {\scriptsize$\pm$ 1.55} \\
\ac{p-dro} & \underline{ 34.98}  {\scriptsize$\pm$ 9.39} & \underline{ 84.21}  {\scriptsize$\pm$ 2.11} \\\midrule
Oracle DRO & \underline{ 67.71}  {\scriptsize$\pm$ 3.03} & \underline{ 77.91}  {\scriptsize$\pm$ 4.49} \\
\bottomrule
\end{tabular}
}
\end{center}
    \end{table}
We train 7 models with \ac{p-dro} on BiasedSST using different hyper-parameters for the adversary. We start from configuration $\lambda=10^{-4}$, $\tau=0.01$, $k=5$, and for each hyper-parameter we run a configuration with a smaller and a higher value, keeping all other hyper-parameters the same. We train for 50 epochs and select the best model using the strategies described in Section \ref{sec:optimizing}.

We report the worst-case (``robust'') accuracy over all groups on the test set, as well the average accuracy in Table \ref{tab:biased_sst_95_main_result} (we report the mean and standard deviation over 5 runs). We find that both Topic DRO, NonParam and \ac{p-dro} are more robust than \ac{erm}, but the latter outperforms the former two close to 30 and 7 points respectively, achieving $52\%$ of Oracle DRO's robust accuracy, while not leveraging any information on the oracle groups.

\subsection{Optimal Stopping and Hyper-parameter Selection Ablation}
\label{sec:stopping_exps}

To understand the importance of the optimal stopping  and hyper-parameter selection strategy described in Section \ref{sec:optimal_stopping}, we perform an ablation on the BiasedSST dataset comparing 4 strategies:

\begin{itemize}
    \item \textbf{Average}: models are selected based on their average zero-one loss (\ie{} error rate) on the unmodified validation set. This is the baseline stopping criterion.
    \item \textbf{Minmax}: selection based on the adversaries (as described in Section \ref{sec:optimal_stopping}), with and without the \textbf{KL constraint}, as well as its variant \textbf{Greedy-Minmax} for stopping.
    \item \textbf{Oracle}: in this setting the groups are known (in the validation set), and models are selected based on their error rate on the worst performing group. This is the optimal criterion for the group-DRO setting we are considering.
\end{itemize}

To compare stopping criterions experiments, we only consider one set of hyper-parameters: $\lambda=10^{-4}$, $k=5$ and $\tau=0.01$. From the robust validation accuracies reported in Table \ref{tab:biased_sst_95_50epochs_optimal_stopping}, we first observe that Average stopping results in a robust accuracy of 0, highlighting the necessity for a suitable stopping criterion. We find that Minmax, especially with a KL constraint, is a much better strategy, recovering $\approx 60\%$ of the performance achievable with Oracle stopping. Notably, the Greedy-Minmax variant which we use in practice reaches very close results ($<1$ point difference) despite its requiring to keep track of only 2 out of the 50 model checkpoints at any time.

To understand the effectiveness of the Minmax strategy for selecting hyper-parameters. We take the models trained in Section \ref{sec:biased_sst_main_result}, but select the best hyper-parameters using the different strategies described above. Results, shown in Table \ref{tab:biased_sst_95_model_selection}, confirm that Minmax (with the KL constraint) is a better choice than Average for selecting hyper-parameters, even though the improvement is not as striking as for stopping.

\begin{table*}[t]

\caption{\label{tab:biased_sst_95_50epochs_stopping_and_hyper} Effect of different optimal stopping and hyper-parameter selection strategies on robust validation accuracy.}

\begin{subtable}{0.5\columnwidth}

\caption{\label{tab:biased_sst_95_50epochs_optimal_stopping} Optimal stopping}
\begin{center}
\begin{tabular}{ll}
\toprule
 Criterion  & Robust accuracy \\
\midrule

Average & {0.00} {\scriptsize$\pm$ 0.00} \\

Minmax &  {25.22} {\scriptsize$\pm$ 13.01} \\

~~+KL constraint &  {31.30} {\scriptsize$\pm$ 10.07} \\
~~+Greedy-Minmax & {\bf 32.17} $\pm$  {\scriptsize$\pm$ 11.20}  \\
\midrule

Oracle & 50.95 {\scriptsize$\pm$ 5.01} \\

\bottomrule
\end{tabular}
\end{center}
\end{subtable}
\begin{subtable}{0.5\columnwidth}

\caption{\label{tab:biased_sst_95_model_selection} Hyper-parameter selection}
\begin{center}
\begin{tabular}{ll}
\toprule
 Criterion  & Robust accuracy \\
\midrule

 Average & 31.03  {\scriptsize$\pm$ 12.16} \\
 Minmax & 28.62  {\scriptsize$\pm$ 12.37} \\
~~+KL constraint & {\bf{35.65}}  {\scriptsize$\pm$ 11.47} \\\midrule
 Oracle & {38.26}  {\scriptsize$\pm$ 13.01} \\

\bottomrule
\end{tabular}
\end{center}
\end{subtable}
\end{table*}

\subsection{Importance of $\Ell_{\text{adv}}$}
\label{sec:kl_relaxation_ablation}

\begin{wrapfigure}{R}{0.4\textwidth}
\centering
\includegraphics[width=0.4\textwidth]{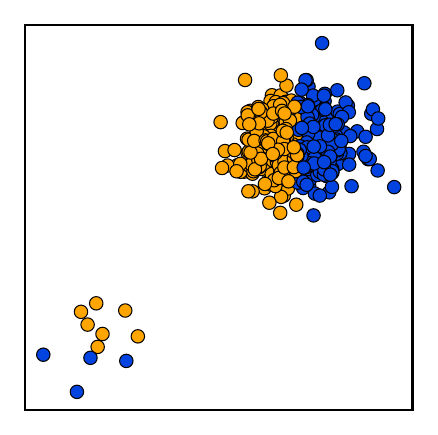}
\caption{\label{fig:toy_figure_linear} A toy classification task.}
\end{wrapfigure}

Finally, we investigate the importance of modifying the adversary's objective as described in Section \ref{sec:optimizing_psi}. For this experiment, we devise a simpler toy task on which directly training the constrained \ac{dro} objective is possible. Specifically, we consider the two-dimensional binary classification problem pictured in Figure \ref{fig:toy_figure_linear}.
The training data consists of 10,000 points partitioned in two normally distributed ``domains'' with a 1:50 sampling ratio and different classification boundaries. We train a logistic regression model, which cannot perfectly fit the training data and must trade-off between accuracy on each domain. For the sake of simplicity, we only model the input variables $x$\footnote{In other words, we set $q_\psi(x,y)=p(y\mid x)q_\psi(x)$, where $p(y\mid x)$, is the true conditional which will be canceled out in the ratio $\frac{q_\psi(x,y)}{q_{\psi_0}(x,y)}$.} as isotropic normal distributions with fixed variance: the adversaries' parameter $\psi\in\mathbb R^2$ represents the location of the Gaussian (we fix the variance to the empirical variance of the data).

We compare 3 different versions of \ac{p-dro}:  first, naive simultaneous gradient descent on the zero-sum game, without any constraint on the adversary (\textbf{bare \ac{p-dro}}), then the same, but with an approximation of the explicit KL constraint between $q_\psi$ and $q_{\psi_0}$ (\textbf{+KL constraint}. Even in this simplest setting, the exact KL between $q_\psi$ (a gaussian) and $p$ (a mixture of gaussians) does not have an analytical expression \citep{Hershey2007ApproximatingTK}. Instead, we fall back on enforcing the KL constraint between $q_\psi$ and $q_{\psi_0}$, both isotropic gaussians with the same standard deviation. Let $\mu$ and $\mu_0\in\mathbb R^2$ denote their respective mean, and $\sigma>0$ their standard deviation. In this context, their KL divergence reduces to:
\begin{equation*}
    \KL{q_\psi}{q_{\psi_0}} = \KL{q_{\psi_0}}{q_\psi} = \frac 1 {2\sigma^2}\Vert \mu-\mu_0\Vert^2
\end{equation*}
In other words, the KL divergence is equivalent to the euclidean distance between the distributions' means. We use this fact to project $\psi$ (in the KL sense) onto $B_\kappa=\{\hat\psi\mid\KL{q_{\hat \psi}}{q_{\psi_0}}<\kappa\}$:
\begin{align*}
    \text{proj}_{B_\kappa}(\psi)&\coloneqq\argmin_{\hat\psi\in B_\kappa}\KL{q_\psi}{q_{\hat \psi}}\\
    &=\psi_0 + \frac{\sqrt{2\kappa}\sigma}{\Vert\psi-\psi_0\Vert}(\psi-\psi_0)
\end{align*}

\begin{wraptable}{r}{0.5\textwidth}
    \caption{Ablation of \ac{p-dro} to train the linear model on the toy task. We report accuracy on both domains, as well as robust accuracy.}
    \label{tab:toy_mlp_results}
    \begin{tabular}{l|cc}
        \toprule
        & Average & Robust\\
        \midrule
        ERM & {\bf 84.66} {\scriptsize $\pm$ 0.10 } & 49.75 {\scriptsize $\pm$ 0.05 } \\\midrule
        bare \ac{p-dro} & 49.97 {\scriptsize $\pm$ 0.10} &  49.85 {\scriptsize $\pm$ 0.10}  \\
        ~~+kl constraint & 76.63 {\scriptsize $\pm$ 9.93} & 58.41 {\scriptsize $\pm$ 9.25} \\
        ~~+$\Ell_{\text{adv}}$ & 76.97 {\scriptsize $\pm$ 0.43} & {\bf 64.32} {\scriptsize $\pm$ 1.31} \\
        \bottomrule
    \end{tabular}
\end{wraptable}

Finally we report results using our relaxation and the KL reversal described in Section \ref{sec:optimizing_psi} (\textbf{+$\Ell_{\text{adv}}$}). For each setting, we report the average and robust accuracy with mean and standard deviation over 10 runs. For the KL constraint and the relaxation, we report the best results among 4 values of the KL bound $\kappa$ and the temperature $\tau$ respectively.

In Table \ref{tab:toy_mlp_results}, we observe that bare \ac{p-dro} is too unstable and systematically diverges. The addition of a KL constraint mitigates this behaviour, but the zero-sum objective is still unstable, as evidenced by the high standard deviations. Finally, we find that the addition of $\Ell_{\text{rev}}$ stabilizes the training process greatly, leading to consistently high robust accuracy.

\section{P-DRO in Practice: Case Study of Toxicity Detection}
\label{sec:toxicity}

In this section, we demonstrate the effectiveness of \ac{p-dro} in the more realistic setting of toxicity detection, the task of recognizing various forms of toxic language (eg. hate speech or offensive language). Identifying online abuse on the internet is a crucial challenge, and has garnered much interest in the \ac{nlp} community \citep{schmidt2017survey,fortuna2018survey}. However, recent work \citep{sap2019risk} has shown that there is strong correlation between toxic labels and the presence of certain markers of dialects of English spoken by minority groups. This correlation is in turn amplified by hate speech classifiers trained on such data, leading to biased prediction.

Our results on BiasedSST suggest that \ac{p-dro} can provide one solution to preventing models from absorbing spurious correlations present in their training data, even in the absence of protected attributes (such as language variety here).

\subsection{Experimental Setting}

Following \citet{sap2019risk} and \citet{xia2020demoting}, we perform experiments on two datasets: {\bf\davidson{}} \citep{davidson2017automated}, a corpus of 25K tweets classified in three categories: \textit{hate speech} ($6\%$), \textit{offensive} ($76\%$) and \textit{neither} ($18\%$), and {\bf\founta{}} \citep{founta2018large}, a 100k sized dataset, also collected from Twitter and annotated with an additional \textit{spam} label, with the following breakdown by categories: \textit{hateful} ($5\%$), \textit{abusive} ($27\%$), \textit{normal} ($54\%$) and \textit{spam} ($14\%$).

The released version of these datasets does not contain information on the dialect of each user. In order to be able to evaluate our models, and to train an Oracle DRO baseline, we follow \citet{sap2019risk} and use annotations provided by the dialect classifier described in \citet{blodgett2016demographic} to label each example as one of four English varieties: White-aligned, African American, Hispanic, and Other. Note that, as these are automatically obtained labels, the groups may not exactly correspond to the actual racial sociolects, however \citet{sap2019risk} does report that they correlate highly with self-reported race, and they serve as a useful proxy in the absence of manual annotation.

We formulate the group-DRO problem by separating each dataset into independent groups identified by both language variety and label, for a total of 12 and 16 groups for \davidson{} and \founta{} respectively. Some of these groups are severely under-represented in the test set. In order to make our robust accuracy results reliable yet still representative of the under-represented groups, we combine groups that contain less than 100 samples into a single group to compute robust test accuracies.

On \davidson{}, we train the same BiLSTM model as described in Section \ref{sec:stopping_exps}. To illustrate the applicability of \ac{p-dro} to other model architectures, we pick BERT \citep{devlin2018bert}, a large scale pre-trained model as a classifier on \founta{}. In both cases, we adopt the Transformer architecture described in Section \ref{sec:stopping_exps} as the adversary. We train the adversary with a temperature of $\tau=0.01$ and a normalizing window $k=10$. To demonstrate the efficacy of automatic hyper-parameter selection in the \ac{p-dro} setting, we delegate the choice of the adversary's learning rate $\lambda$ to grid-search, training 3 models with $\lambda \in\{10^{-5},10^{-4},10^{-3}\}$ and selecting the best using the Minmax criterion described in Section \ref{sec:model_selection}. We also report numbers for Oracle DRO and Topic DRO. Results are averaged over 5 runs, each with a different random seed.

\begin{table*}[t]

\caption{\label{tab:toxicity_results} Robust test accuracy on the \davidson{} and \founta{} toxicity detection tasks.}
\begin{subtable}{\textwidth}
\centering
\caption{\label{tab:toxicity_robust_results} No group information.}
\begin{tabular}{l|cc|cc}
\toprule
  & \multicolumn{2}{c}{\davidson{}} &  \multicolumn{2}{c}{\founta{}}\\
 & Robust & Average & Robust & Average\\
\midrule
ERM & { 53.19}  {\scriptsize$\pm$ 1.70} & { 69.44}  {\scriptsize$\pm$ 0.53} & { 19.57}  {\scriptsize$\pm$ 7.00} & {\bf  81.56}  {\scriptsize$\pm$ 0.26} \\
Topic DRO & \underline{ 45.26}  {\scriptsize$\pm$ 3.47} & \underline{ 61.68}  {\scriptsize$\pm$ 5.02}& { 16.48}  {\scriptsize$\pm$ 5.46} & \underline{ 80.49}  {\scriptsize$\pm$ 0.49} \\
NonParam & { 54.13}  {\scriptsize$\pm$ 1.14} & \underline{ \bf  70.54}  {\scriptsize$\pm$ 0.64} & \underline{ 17.54}  {\scriptsize$\pm$ 6.41} & \underline{ 81.20}  {\scriptsize$\pm$ 0.11} \\
\ac{p-dro} & \underline{ \bf 69.06}  {\scriptsize$\pm$ 1.70} & {69.69}  {\scriptsize$\pm$ 2.50} & { \bf 30.25}  {\scriptsize$\pm$ 10.13} & { 79.91}  {\scriptsize$\pm$ 1.41} \\\midrule
Oracle DRO & \underline{ 74.50}  {\scriptsize$\pm$ 1.74} & \underline{ 65.79}  {\scriptsize$\pm$ 0.76}  & \underline{ 55.23}  {\scriptsize$\pm$ 3.97} & \underline{ 72.43}  {\scriptsize$\pm$ 2.61} \\
\bottomrule
\end{tabular}
\end{subtable}
\begin{subtable}{\textwidth}
\centering
\caption{\label{tab:toxicity_robust_results_oracle} With group information on the validation set.}
\begin{tabular}{l|cc|cc}
 & Robust & Average & Robust & Average\\
\midrule
ERM & { 53.15}  {\scriptsize$\pm$ 0.87} & { 69.64}  {\scriptsize$\pm$ 1.01} & { 34.07}  {\scriptsize$\pm$ 3.20} & { 78.78}  {\scriptsize$\pm$ 0.38} \\
Topic DRO & { 52.02}  {\scriptsize$\pm$ 1.26} & {\bf 69.11}  {\scriptsize$\pm$ 0.49} & { 34.82}  {\scriptsize$\pm$ 3.73} & {\bf 79.59}  {\scriptsize$\pm$ 0.85} \\
NonParam & { 49.41}  {\scriptsize$\pm$ 5.60} & \underline{ 58.53}  {\scriptsize$\pm$ 5.71} & { 43.13}  {\scriptsize$\pm$ 6.97} & \underline{ 69.51}  {\scriptsize$\pm$ 3.07} \\
\ac{p-dro} & \underline{\bf 63.05}  {\scriptsize$\pm$ 4.25} & \underline{ 63.07}  {\scriptsize$\pm$ 3.92} & \underline{\bf 47.61}  {\scriptsize$\pm$ 4.53} & \underline{ 74.82}  {\scriptsize$\pm$ 1.90} \\\midrule
Oracle DRO & \underline{ 74.50}  {\scriptsize$\pm$ 1.74} & \underline{ 65.79}  {\scriptsize$\pm$ 0.76} & \underline{ 55.23}  {\scriptsize$\pm$ 3.97} & \underline{ 72.43}  {\scriptsize$\pm$ 2.61} \\
\bottomrule
\end{tabular}
\end{subtable}
\end{table*}

\subsection{Can \ac{p-dro} Produce more Robust Models?}

Table \ref{tab:toxicity_robust_results} reports the robust test accuracies of all models on both tasks. Importantly, except for Oracle DRO, none of the methods compared here necessitate any knowledge of the groups, neither in the training nor validation data. We observe that in both settings \ac{p-dro} is able to achieve higher robust accuracy than \ac{erm}, Topic DRO and NonParam.

This suggests \ac{p-dro} as a useful option in case no group information whatsoever is available. However, in practice, it may be feasible to annotate at least a small amount of data with group information. To emulate this scenario, we perform the same experiment, but assume that group annotations are available on the validation data, which we use to determine optimal stopping and hyper-parameters. Results for this setting are reported in Table \ref{tab:toxicity_robust_results_oracle}. We find that, while the use of robust validation accuracy yields more robust models even for \ac{erm} (especially on \founta{}), \ac{p-dro} is still the best alternative that doesn't require group annotation on the training data.

\section{Additional Experiments}

\subsection{Minmax Validation KL Threshold}
\label{sec:valid_kl_threshold}

The Monte-Carlo estimate of $\KL{q_\psi}{p}$ on the validation set is $\frac{1}{|D_{\text{valid}}|}\sum_{x,y\in D_{\text{valid}}}\frac{q_\psi(x,y)}{p(x, y)}\log \frac{q_\psi(x,y)}{p(x, y)}$. Similarly to Section \ref{sec:optimizing}, we approximate the (unknown) likelihood ratio $\frac{q_\psi(x,y)}{p(x, y)}$ with $\frac{q_\psi(x,y)}{q_{\psi_0}(x, y)}$.

We want to reject all adversaries where this approximated KL is greater than some threshold, $\kappa_\text{valid}$, but how do we choose a good value for $\kappa_\text{valid}$? Consider an adversary which selects a fraction of the validation data of size $\alpha|D_{\text{valid}}|$ for some $\alpha\in (0,1]$. In such a case, the likelihood ratio is $1/\alpha$ on this subset and 0 everywhere else, and the resulting KL estimate will be $\log \alpha$. In other words, choosing a threshold of $\kappa_\text{valid}$ means allowing the adversary to potentially select any subset of size at least $1/e^{\kappa_\text{valid}}$ of the original data. Our heuristic choice, $\log 10$, corresponds to allowing subsets of size at least $10\%$ of $|D_{\text{valid}}|$.

\begin{figure}
\centering
\caption{\label{fig:kl_threshold} Evolution of the robust validation accuracy of the model selected by Greedy-Minmax as a function of the \ac{kl} threshold $\kappa_\text{valid}$}
\includegraphics[width=0.7\textwidth]{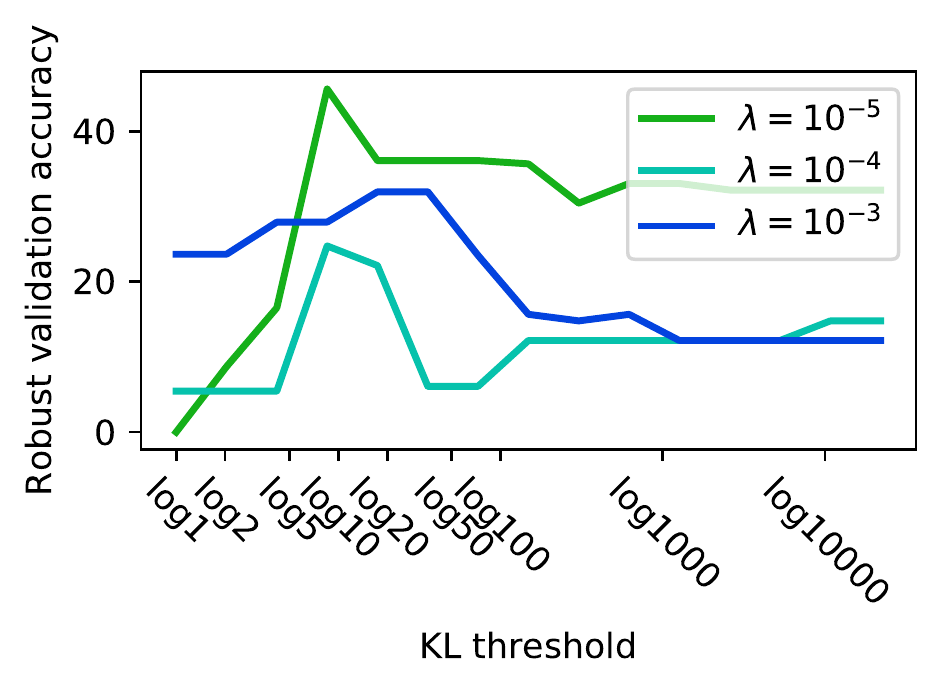}
\end{figure}

Of course, this is only a heuristic because the adversary can reweight the validation set non-uniformly. To assess the effect of $\kappa_{\text{valid}}$ on Greedy-Minmax, we compute  the average robust validation error of the selected model across 5 runs for 3 different values of the adversary's learning rate. Results on BiasedSST, depicted in Figure \ref{fig:kl_threshold}, show that adversaries with higher learning rate are more sensitive to the choice of threshold, but all values of $\kappa_{\text{valid}}$ between $\log 5$ and $\log 20$ seem to work for these settings.

\subsection{\ac{p-dro} Experiments with an LSTM Adversary}
\label{sec:weak_adv}

We replicate the experiments BiasedSST experiments in Section \ref{sec:analysis}, but this time using a smaller generative model, which is unlikely to generate good samples. Specifically, we use a one layer LSTM model \citep{hochreiter1997long} with embedding and hidden dimension 256. We only perform grid-search over $\lambda\in[10^{-5},10^{-4},10^{-3}]$ and select the best with Minmax.

Once pre-trained on the BiasedSST dataset, this model achieves a perplexity of $227.0$, more than 4 times worse than the transformer model we use in other experiments ($49.8$). However, as evidenced by its robust accuracy displayed in Table \ref{tab:biased_sst_95_lstm}, \ac{p-dro} is still able to learn a robust model. We take this as evidence that the re-weighting introduced in Section \ref{sec:optimizing} helps stabilize training even when $q_\psi$ is not a perfect model of the data.

\begin{table}
\begin{center}
\caption{\label{tab:biased_sst_95_lstm} Average and robust accuracies on BiasedSST when \ac{p-dro} is trained with an LSTM adversary. Underlining indicates statistically significant difference compared to \ac{erm} ($p<0.05$)}
\begin{tabular}{llc}
\toprule
  & Robust & Average \\
\midrule
ERM & { 2.15}  {\scriptsize$\pm$ 0.97} & { 95.09}  {\scriptsize$\pm$ 0.16} \\\midrule
Topic DRO & \underline{ 5.18}  {\scriptsize$\pm$ 1.46} & { 95.00}  {\scriptsize$\pm$ 0.10} \\
NonParam & \underline{ 28.11}  {\scriptsize$\pm$ 2.16} & \underline{ 92.45}  {\scriptsize$\pm$ 1.55} \\
\ac{p-dro} & \underline{ 43.68}  {\scriptsize$\pm$ 4.93} & \underline{ 86.58}  {\scriptsize$\pm$ 1.77} \\
\midrule
Oracle DRO & \underline{ 67.71}  {\scriptsize$\pm$ 3.03} & \underline{ 77.91}  {\scriptsize$\pm$ 4.49} \\
\bottomrule
\end{tabular}
\end{center}
\end{table}

\subsection{Influence of hyper-parameters on \ac{p-dro}}
\label{sec:hyperparams_study}

We study the influence of the 3 hyper-parameters $\tau$ (temperature), $k$ (size of the renormalization window) and $\lambda$ (learning rate of the adversary) on the performance of \ac{p-dro}. All experiments are run on the BiasedSST dataset, and the analysis proceeds as follows: we start from configuration $\tau=0.01$, $k=5$ and $\lambda=10^{-4}$ and vary each of the hyper-parameters independently.  We report two numbers for each configuration: robust accuracy of the best model (1) using Greedy-Minmax stopping and (2) using Oracle stopping. The latter is useful to disentangle the effect of the stopping criterion.

As seen in the results shown in Table \ref{tab:biased_sst_95_hyperparam_study}, we find that $\tau$ has the least effect on robust accuracies. While the renormalization window parameter $k$ has some effect on optimal stopping, the best robust accuracy achieved by the model (with oracle stopping) varies little. We observe the adversary's learning rate $\lambda$ to be the most sensitive hyper-parameter, which is why we restrict our grid-search to $\lambda$ in Section \ref{sec:toxicity}.

\begin{table}
\begin{center}
\caption{\label{tab:biased_sst_95_hyperparam_study} Effect of hyper-parameters on robust validation accuracy on BiasedSST}
\begin{tabular}{llc}
\toprule
& \multicolumn{2}{c}{Robust accuracy}\\
  & Minmax stopping & Oracle stopping \\
\midrule
$\lambda=10^{-5}$ & { 28.62}  {\scriptsize$\pm$ 12.37} & { 45.10}  {\scriptsize$\pm$ 4.50} \\
$\lambda=10^{-4}$ & { 44.74}  {\scriptsize$\pm$ 3.24} & { 50.43}  {\scriptsize$\pm$ 5.05} \\
$\lambda=10^{-3}$ & { 25.57}  {\scriptsize$\pm$ 10.33} & { 38.70}  {\scriptsize$\pm$ 2.97} \\
\midrule
$\tau=0.1$ & { 39.72}  {\scriptsize$\pm$ 5.55} & { 50.00}  {\scriptsize$\pm$ 4.98} \\
$\tau=0.01$ & { 44.74}  {\scriptsize$\pm$ 3.24} & { 50.43}  {\scriptsize$\pm$ 5.05} \\
$\tau=0.001$ & { 44.74}  {\scriptsize$\pm$ 3.24} & { 50.87}  {\scriptsize$\pm$ 5.09} \\
\midrule
$k=1$ & { 41.98}  {\scriptsize$\pm$ 4.48} & { 49.60}  {\scriptsize$\pm$ 5.39} \\
$k=5$ & { 44.74}  {\scriptsize$\pm$ 3.24} & { 50.43}  {\scriptsize$\pm$ 5.05} \\
$k=10$ & { 32.17}  {\scriptsize$\pm$ 11.20} & { 50.95}  {\scriptsize$\pm$ 5.01} \\
\bottomrule
\end{tabular}
\end{center}
\end{table}

\section{Conclusion}

In this chapter, we have shown that there is promise in using parametric families of neural generative models for defining the uncertainty set in distributionally robust optimization. While we only perform experiments on \ac{nlp} tasks, this approach can, in theory, be applied in any modality. However, the reliance of \ac{p-dro} on good quality generative models limits its adoption in applications where good quality generative models are unavailable, or when such model cannot produce exact densities efficiently. 
In the following chapter, we will address these issues by parameterizing the likelihood ratio $q_\psi/p$ directly, an alternative formulation which poses different implementation challenges.

\clearpage

\chapter{Distributionally Robust Models with Parametric Likelihood Ratios}
\label{ch:r_pdro}

In chapter \ref{ch:modeling_the_second_player_dro} we proposed \ac{p-dro}, a promising approach where the uncertainty set is defined by a parametric family of generative models, which allows for more flexibility in defining the uncertainty set. Despite \ac{p-dro}'s demonstrated ability to train models that are robust to distributional shift, it suffers from several drawbacks: first, it necessitates training an auxiliary generative model of the data. This can be difficult for several reasons. First, this limits the applicability of the method to domains with generative models that allow for exact probability computations. Moreover, even when such generative models are available, they are often more computationally demanding than their discriminative counterparts. In language models for instance, probabilities for sequences of text are obtained by iteratively producing conditional probabilities over all tokens in the vocabulary. This additional step results in considerable computational overhead compared to discriminative models. Second, it is challenging to use in practice due to its reliance on a number of hyperparameters and approximations to the objective function.

In this chapter, we propose a new approach for \ac{dro}, called R-PDRO, based on a key modification of the \ac{p-dro} algorithm: instead of modeling the worst-case distributions directly, we parametrize the likelihood ratio between the training distribution and the worst-case distribution. This removes the dependency on an unwieldy generative model, making the method useful for more applications. 
While likelihood-ratio formulations of DRO have been tried in prior work \citep{sagawa2019distributionally}, we show that they are particularly effective for parametric, neural-network based adversaries. Our approach relies on two simple ideas: a mini-batch level normalization strategy to enforce likelihood ratio constraints and a KL divergence penalty to limit the scope of the uncertainty set. R-PDRO consistently achieves equal or better robust sub-population accuracy compared to \ac{p-dro} and other baselines on a variety of standard benchmarks in image and text classification. In addition, we find it is both faster than \ac{p-dro} and depends on fewer hyper-parameters. Additional ablation and analysis experiments demonstrate that our minibatch normalization strategy is necessary for high performance, and that the parametric adversaries enabled by R-PDRO are substantially more resistant to label noise compared to traditional non-parametric approaches to DRO.

\section{Parametric Likelihood Ratio}
\label{sec:rpdro_method}

\subsection{DRO as a Likelihood Ratio Optimization Problem}
\label{sec:likelihood_ratio}
In the situation that all distributions in $\mathcal Q$ are absolutely continuous with respect to $p$ (i.e. for all $q\in \mathcal Q$, $q(x,y)>0$ only if $p(x, y)>0$) the inner maximum in Equation \ref{eq:dro_problem} can be rewritten purely as a function of the likelihood ratio $\frac{q}{p}$
\begin{equation}
\E_{(x,y)\sim q}\ell_\theta(x, y)=\E_{(x,y)\sim p}\frac{q(x,y)}{p(x,y)}\ell_\theta(x,y).
\end{equation}
Such absolute continuity assumptions are standard in $f$-divergence and group DRO methods, which both rely upon re-weighting the training distributions. In fact, the KL divergence constraint in \ac{p-dro} presupposes absolute continuity.

This suggests that the inner maximum can be re-written as an optimization problem on functions $r: \mathcal{X}\times\mathcal{Y}\longrightarrow\mathbb R_+$ within the uncertainty set $\mathcal R\in \{r\mid pr\in \mathcal Q\}$
\begin{equation}\label{eq:ratio_dro_problem}
    \min_{\theta}\max_{r\in\mathcal R}\E_{(x,y)\sim p}r(x,y)\ell_\theta(x, y).
\end{equation}
This reparametrization of the problem will allow us to replace the parametric family of generative models with a parametric family over probability ratios.

\subsection{Ratio-based \ac{p-dro}}

The likelihood-ratio formulation described above is appealing for \ac{p-dro} because it enables the use of discriminative style neural architectures for parameterizing the ratio $r$, which opens up many more options for defining the parametric uncertainty set. Specifically, we can set the adversary to be any parametric function $r_\psi : \mathcal X\times Y\longrightarrow \mathbb R^+$ verifying $\E_{x,y\sim p}r_\psi(x, y)=1$. The key insight that we use to realize our proposed method is that we do not need to limit the choice of adversaries to those that implicitly satisfy this normalization condition (\ie{} generative models). Instead, we can pick any adversary and treat normalization as an additional constraint (the ``normalization constraint'').

Note that in this case, the KL constraint takes the simple form $\KL{pr_\psi}{p}=\E_{pr_\psi}\log\frac{pr_\psi}{p}=\E_{p}r_\psi\log r_\psi$. The final min-max problem, which we dub ratio-based \ac{p-dro} (R-PDRO), is as follows:
\begin{equation}
    \min_{\theta}\max_{\substack{\psi\\\E_{p}r_\psi\log r_\psi\leq \kappa\\\mathbb E_{p} r_\psi=1}} \underbrace{\mathbb E_{(x,y)\sim {p}}r_\psi(x,y)\ell_\theta(x,y)}_{\mathcal L_{\text{R-PDRO}}}.
\end{equation}
As in \ac{p-dro}, we can look for equilibria of this differentiable min-max game by performing simultaneous gradient updates \citep{singh2000nash} to $\theta$ and $\psi$ in directions $-\nabla_\theta \mathcal{L}_{\text{R-PRDO}}$ and $+\nabla_\psi \mathcal{L}_{\text{R-PRDO}}$ respectively. Although finding global equilibria is not guaranteed in high dimensional non-convex settings \citep{balduzzi2018mechanics}, empirical evidence from Chapter \ref{ch:modeling_the_second_player_dro} suggests that models trained in this manner still reach useful solutions.

In experiments, we adopt an exponential parameterization $r_\psi(x,y)=e^{f_\psi(x,y)}$ where $f_\psi$ is the output of any parametric model with values in $\mathbb R$. Similarly to \ac{p-dro}, we do not explicitly enforce the KL constraint (due to the difficulty of projecting onto the KL ball), and instead relax it in the form of a KL term $\tau\E_{p}r_\psi\log r_\psi$ added to the loss function. The regularization strength $\tau$ is treated as a hyper-parameter.

\subsection{Enforcing the normalization constraint}
\label{sec:normalization}

In addition to the KL constraint, R-PDRO necessitates that $r_\psi$ satisfies a normalization constraint $\E_{p} r_\psi=1$ to ensure that ${p}r_\psi$ is a proper probability distribution over $\mathcal D_\text{train}$. If that were not the case, the adversary $r_\psi$ could artificially increase the weighted expectation $\E_{p}r_\psi\ell_\theta$ by assigning a total weight greater than 1 to the entire dataset.

Existing methods for ratio based \ac{dro} such as \cite{sagawa2019distributionally} achieve this by either projecting $r_\psi$ onto the simplex $\{r\mid \E_pr=1\}$ after each gradient update on $\psi$, or by directly parametrizing the ratio as $r_\psi(x, y)=e^{f_\psi(x, y)}/\E_pe^{f_\psi}$. Unfortunately, these solutions are computationally infeasible in practical scenarios with large datasets. Indeed, they necessitate computing the entire expectation over $p$, which can be costly when each $f_\psi(x, y)$ is the output of a neural model.

We propose two simple, yet effective solutions for addressing this issue in the context of minibatch training where we can only compute $f_\psi$ for small number of samples ${(x_1, y_1),\ldots,(x_n, y_n)}$ at each step of training.

\paragraph{Self-normalization} is inspired by the idea of ``self-normalization'' developed for globally normalized structured prediction models \citep{andreas2015accuracy,goyal2019empirical}. It consists in adding a relaxation of the normalization constraint to the objective. Specifically, following \citet{goyal2019empirical} we add a squared penalty on the log normalizer at the minibatch level. Ignoring the KL penalty, this regularized objective takes the following form:
\begin{equation}
    \mathcal{\mathcal{L}_{\text{self-norm}}}(\theta, \psi)=\frac 1 n\sum_{i=1}^nr_\psi(x_i, y_i)\ell_\theta(x_i, y_i) - \beta\log\left(\frac 1 n\sum_{i=1}^nr_\psi(x_i, y_i)\right)^2.
\end{equation}
The hyper-parameter $\beta$ controls the regularization strength. Intuitively, this penalizes adversary that assign too much (or too little) total weight to the minibatch. However, the choice of an optimal $\beta$ adds an additional degree of freedom to R-PDRO, which suggests our second option as a simpler alternative.

\paragraph{Batch-level normalization} consists of using the normalized ratio at the minibatch level by setting
\begin{equation}
    \tilde r_\psi(x_i,y_i) = \frac{e^{f_\psi(x_i, y_i)}}{\sum_{j=1}^ne^{f_\psi(x_j, y_j)}}
\end{equation}
for each sample $(x_i, y_i)$ in the minibatch. An obvious downside of this approach is that the weight of each sample now depends on the minibatch it was sampled from. This can be problematic for small batch sizes: as an extreme example, for a batch size of 1, this normalization scheme assigns the same weight of 1 to every sample, making the objective equivalent to ERM.

However, minibatch approximations have proven effective for other forms of DRO \citep{hu2018does,levy2020large} and there is some evidence that they can yield accurate estimates for higher batch sizes \citep{cortes2010learning}. In practice we find that this approach yields good results for commonly used batch sizes, is generally more stable than the self-normalization penalty, and does not introduce any additional hyper-parameters. In most of our experiments, we adopt this approach unless specified otherwise. In that case, the R-PDRO objective on a minibatch becomes
\begin{equation}
    \mathcal{L}_{\text{batch-level norm}}(\theta, \psi)=\underbrace{\sum_{i=1}^n\tilde{r}_\psi(x_i, y_i)\ell_\theta(x_i, y_i)}_{\text{expected loss under }pr_\psi} - \tau \underbrace{\sum_{i=1}^n \tilde{r}_\psi(x_i, y_i)\log \tilde{r}_\psi(x_i, y_i)}_{\text{KL penalty}}.
\end{equation}
The KL term serves to penalize ratios that deviate too much from~$1$.%
\footnote{Note that the penalty is subtracted because we are maximizing with respect to $\psi$} 
The only hyper-parameter that needs to be tuned is the KL regularization strength $\tau$.

\section{Experiments}
\label{sec:rpdro_experiments}

\subsection{Datasets}
\label{sec:datasets}

We perform experiments on four datasets: two text classification datasets used in chapter \ref{ch:modeling_the_second_player_dro}, Biased SST and \founta{} and two image classification datasets from \citet{sagawa2019distributionally}, Waterbirds and CelebA. Specific details for each dataset follow these previous works, as described below:

\paragraph{Biased SST} is based on the SST-2 sentiment classification dataset \citep{radford2018improving}, but modified to introduce spurious correlation between a distractor token (``So,'') and positive labels in around 95\% of the dataset. In this setting models trained with ERM can very easily learn this spurious correlation, which hurts performance on the small sub-population that doesn't suffer from this bias.
\paragraph{\founta{}} A toxicity detection dataset of tweets labeled as  \textit{hateful} ($5\%$), \textit{abusive} ($27\%$), \textit{normal} ($54\%$) and \textit{spam} ($14\%$). The group-\ac{dro} problem is formulated by partitioning the evaluation data along labels as dialectal annotation obtained automatically with an off-the shelf classifier \citep{blodgett2016demographic,sap2019risk}. In particular these dialects align closely with self-reported race, and \citet{sap2019risk} found that machine learning models trained on such toxicity detection datasets tend to exhibit bias towards certain labels depending on the demographics of the tweets' authors, particularly with minorities. In order to report more reliable accuracy numbers, all groups containing less than 100 samples are aggregated when computing test accuracies. 

\paragraph{Waterbirds} An image classification dataset where the task is to predict ``land bird'' or ``water bird'' with the confounding factor of the background; most water (resp. land) bird pictures have water (resp. land) on the background.

\paragraph{CelebA} A popular face recognition dataset originally published by \citet{liu2015faceattributes}. The group-\ac{dro} problem is formulated as a task of predicting the hair color (``Blond'' or ``Dark'') across groups formed by the combination of the label and the (binary) gender of the subject. Due to the spurious correlation between blond hair/female and dark hair/male, models trained with ERM tend to achieve lower accuracies on underrepresented groups such as ``blond-haired male'' which totals only 0.85\% of the training data.

\subsection{Experimental Setting}

On BiasedSST and \founta{} we follow the experimental setting of chapter \ref{ch:modeling_the_second_player_dro} and train a BiLSTM and BERT-base model \cite{devlin2018bert} respectively. On the image classification datasets we train Resnet-50 architectures \citep{he2016deep} pre-trained on ImageNet \citep{deng2009imagenet} as in \citet{sagawa2019distributionally}.

\begin{table*}[t]

\caption{\label{tab:results_text} Robust and average test accuracies on the Biased SST and \founta{} datasets.  Underlined numbers indicates statistically significant difference compared to \ac{erm} ($p<0.05$). Bold numbers indicates the best number in each column (barring Oracle DRO).}

\centering
\begin{tabular}{l|cc|cc}
\toprule
  & \multicolumn{2}{c}{Biased SST} &\multicolumn{2}{c}{\founta{}}\\
 & Robust & Average & Robust & Average\\
\midrule

ERM & { 2.15}  {\scriptsize$\pm$ 0.97} & { \bf 95.09}  {\scriptsize$\pm$ 0.16} & { 19.57}  {\scriptsize$\pm$ 7.00} & {\bf  81.56}  {\scriptsize$\pm$ 0.26}\\
Topic DRO & \underline{ 5.18}  {\scriptsize$\pm$ 1.46} & { 95.00}  {\scriptsize$\pm$ 0.10} & { 16.48}  {\scriptsize$\pm$ 5.46} & \underline{ 80.49}  {\scriptsize$\pm$ 0.49} \\
NonParam & \underline{ 28.11}  {\scriptsize$\pm$ 2.16} & \underline{ 92.45}  {\scriptsize$\pm$ 1.55} & \underline{ 17.54}  {\scriptsize$\pm$ 6.41} & \underline{ 81.20}  {\scriptsize$\pm$ 0.11} \\
P-DRO & \underline{ 34.98}  {\scriptsize$\pm$ 9.39} & \underline{ 84.21}  {\scriptsize$\pm$ 2.11} & {30.25}  {\scriptsize$\pm$ 10.13} & { 79.91}  {\scriptsize$\pm$ 1.41} \\
R-PDRO & \underline{\bf 50.70}  {\scriptsize$\pm$ 7.33} & \underline{ 86.60}  {\scriptsize$\pm$ 2.96} & \underline{ \bf 53.52}  {\scriptsize$\pm$ 1.66} & \underline{ 76.62}  {\scriptsize$\pm$ 1.43} \\\midrule
Oracle DRO & \underline{ 67.71}  {\scriptsize$\pm$ 3.03} & \underline{ 77.91}  {\scriptsize$\pm$ 4.49} & \underline{ 55.23}  {\scriptsize$\pm$ 3.97} & \underline{ 72.43}  {\scriptsize$\pm$ 2.61} \\
\bottomrule
\end{tabular}
\end{table*}


\begin{table*}[t]

\caption{\label{tab:results_images} Robust and average test accuracies on the Waterbirds and CelebA datasets. Underlined numbers indicates statistically significant difference compared to \ac{erm} ($p<0.05$). Bold numbers indicates the best number in each column (barring Oracle DRO).}

\centering
\begin{tabular}{l|cc|cc}
\toprule
  & \multicolumn{2}{c}{Waterbirds} &\multicolumn{2}{c}{CelebA}\\
 & Robust & Average & Robust & Average\\
\midrule

ERM & { 68.32}  {\scriptsize$\pm$ 2.02} & { 89.23}  {\scriptsize$\pm$ 0.36} & { 40.33}  {\scriptsize$\pm$ 2.29} & { \bf 95.89}  {\scriptsize$\pm$ 0.05} \\
NonParam & \underline{ 72.21}  {\scriptsize$\pm$ 0.95} & \underline{ \bf 90.54}  {\scriptsize$\pm$ 0.72} & { 43.33}  {\scriptsize$\pm$ 3.58} & \underline{ 95.72}  {\scriptsize$\pm$ 0.10} \\
R-PDRO & \underline{ \bf 73.49}  {\scriptsize$\pm$ 3.01} & { 90.15}  {\scriptsize$\pm$ 0.74} & \underline{ \bf 55.78}  {\scriptsize$\pm$ 9.15} & { 93.10}  {\scriptsize$\pm$ 3.87} \\\midrule
Oracle DRO & \underline{ 85.60}  {\scriptsize$\pm$ 0.95} & { 89.12}  {\scriptsize$\pm$ 1.20} & \underline{ 89.22}  {\scriptsize$\pm$ 0.90} & \underline{ 92.59}  {\scriptsize$\pm$ 0.40} \\
\bottomrule
\end{tabular}
\end{table*}

Since the adversary in R-PDRO can be any discriminative architecture, we opt for the natural solution of using a similar architecture for this model. For instance on Biased SST, we take $f_\psi(x,y)$ as the raw logit output by a BiLSTM architecture identical to that of the classifier (without the final softmax layer). We do the same for the other datasets, with the exception of \founta{} where we use a smaller DistillBERT model \citep{sanh2019distilbert} for efficiency. We use the same learning rate and optimizer for both model and adversary and only vary the KL penalty weight $\tau\in\{0.001,0.01,0.1,1.0\}$. We perform optimal stopping using the Minmax criterion proposed in chapter \ref{ch:modeling_the_second_player_dro}: every epoch $T$, we determine the best model by explicitly solving a greedy approximation of the min-max game between all previously checkpointed adversaries and models on the validation dataset $\mathcal D_{\text{valid}}$.
\begin{equation}\label{eq:valid_minmax}
    \theta^*=\argmin_{\theta_{i=1\ldots T}}\max_{\psi_{j=1\ldots T}} \frac 1 {|D_{\text{valid}}|} \sum_{x,y\in D_{\text{valid}}}\tilde{r}_{\psi_j}(x, y)\ell_{\theta_i}(x,y)
\end{equation}
A similar strategy is applied for hyper-parameter selection. Importantly, we substitute the 0-1 loss for $\ell_\theta$ in Equation \ref{eq:valid_minmax} as we found in preliminary experiments on BiasedSST that it consistently produced better results.

We compare our results to 5 different methods experimented with in chapter \ref{ch:modeling_the_second_player_dro}:

\begin{itemize}
    \item \textbf{ERM}: standard training to minimize the average loss.
    \item \textbf{NonParam}: Non-parametric DRO with a KL constrained uncertainty set \citep{hu2013kullback,hu2018does}. In this particular case, the inner maximization problem has an analytical solution of the form $q(x,y)\propto e^{\frac {\ell_\theta(x,y)}{\tau^*}}$ where $\tau^*$ is chosen so that the KL divergence with the training distribution is smaller than a pre-defined bound $\kappa$. As a result, examples with high losses are up-weighted.
    \item \textbf{Topic-DRO}: a variation on Topic CVaR \citep{oren2019distributionally} using the online algorithm from \citet{sagawa2019distributionally} to minimize the worst case loss on a collection of pseudo domains automatically generated via topic modeling.\footnote{This baseline was inaccurately referred to as ``Topic CVaR'' in chapter \ref{ch:modeling_the_second_player_dro}} We use this baseline for the text datasets only (Biased SST and \founta{}). 
    \item \textbf{\ac{p-dro}}: the parametric DRO approach proposed in chapter \ref{ch:modeling_the_second_player_dro}. For image datasets, preliminary experiments using auto-regressive models for image modeling \citep{van2016pixel} proved to be prohibitively slow. Therefore, we only report \ac{p-dro} on text datasets as in chapter \ref{ch:modeling_the_second_player_dro}.
    \item \textbf{Oracle DRO}: an online algorithm for minimizing the worst-case loss on the true uncertainty set \citep{sagawa2019distributionally}. Contrary to all other methods, Oracle DRO presupposes that we know the groups of interest at training time. As such, it is not directly comparable and serves to provide an upper bound of the robust accuracy that can be achieved.
\end{itemize}

Across all experiments, we report results with mean and standard deviation across 5 reruns with different seeds. 

\subsubsection{Dataset-specific Hyper-parameters}

All hyper-parameters listed below are constant across all methods:

\paragraph{Text Datasets} The input data is tokenized using the \texttt{bert-base-uncased} sub-word tokenizer from \citet{devlin2018bert}. We train both classifier and adversary with Adam \citep{Kingma2014Adam} using a learning rate of $2\times 10^{-5}$, linearly decay the learning rate to 0 at each step. We train with batches of size 64 (or containing up to 2500 tokens, whichever is lower) for 50 and 20 epochs for BiasedSST and \founta{} respectively, evaluating model on the validation data every epoch.

\paragraph{Image Datasets} On both datasets, images are rescaled to $224\times224$ pixels and pixel values are normalized to have mean 0 and variance 1 across all 3 color channels on the training data. At training time, we augment the data by randomly cropping or flipping the images horizontally. We train using regular stochastic gradient descent using a constant learning rate of $10^{-3}$ and a batch size of 32. We train for 75 and 13 epochs on Waterbirds and CelebA respectively (those numbers were chosen to match the number of steps trained to \citet{sagawa2019distributionally} despite the smaller batch size), and validate every 100 (for Waterbirds) and 1000 (for CelebA) training steps.

\subsubsection{Method Specific Hyper-parameters}

For NonParam we follow the adaptation of \citet{hu2018does} used in chapter \ref{ch:modeling_the_second_player_dro} and choose the optimal temperature $\tau^*$ based on minibatch level estimates of the KL divergence. We treat the KL bound $\kappa$ as a hyper-parameter. We adapt the Minmax stopping criterion of \ac{p-dro} to the non-parametric adversaries as we found it yielded more robust models than those selected with average validation accuracy. We sweep over $\kappa\in\{0.01,0.1,1.0,10.0\}$

For R-PDRO we perform optimal stopping using the Minmax criterion with a KL threshold of $\log 10$ in all experiments, to match the value recommended for \ac{p-dro}. Specifically, we estimate the KL divergence of checkpointed adversaries $\psi_i$ on the validation data as follows:
\begin{equation}
    \frac 1 {|\mathcal{D}_{\text{valid}}|}\sum_{x,y\in \mathcal{D}_{\text{valid}}}\hat{r}_\psi(x, y)\log \hat{r}_\psi(x, y)
\end{equation}
and reject adversaries for which this quantity exceeds $\log 10$.

\subsection{Results}

For all methods, we report the worst accuracy over all groups in the test set (the ``robust accuracy''). Models that are robust to sub-population shift will have higher robust accuracies. Since there is often a trade-off between being robust to distribution shift and performing well on the training distribution, we also report the standard accuracy on the original test set (the ``average accuracy'')

As shown in Table \ref{tab:results_text}, R-PDRO works particularly well on the two text classification tasks, beating all baselines by a large margin on both BiasedSST and \founta{}. In fact, its robust accuracy almost matches that of Oracle DRO on \founta{}, despite the fact that the former doesn't use an any group information at training time. Compared to \ac{p-dro}, we find that results are not only better, but also more consistent as evidenced by the lower standard deviation.

Results are also generally good on image datasets (see Table \ref{tab:results_images}). On Waterbirds both the NonParam baseline and R-PDRO perform much better than ERM, with a slight edge for R-PDRO (although the latter exhibits a higher variance across reruns). On CelebA, R-PDRO largely outperforms the baselines.

\section{Analysis and Ablations}

We perform a series of analysis experiments and ablation studies to better (1) identify how the parametric representation of the ratio provides improvements over non-parametric alternatives and (2) understand the importance of the various choices made with regards to the renormalization scheme described in Section \ref{sec:rpdro_method}. Throughout this section, experiments are performed on the BiasedSST dataset.

\begin{figure}[t!]
    \centering
    \includegraphics[width=0.95\textwidth]{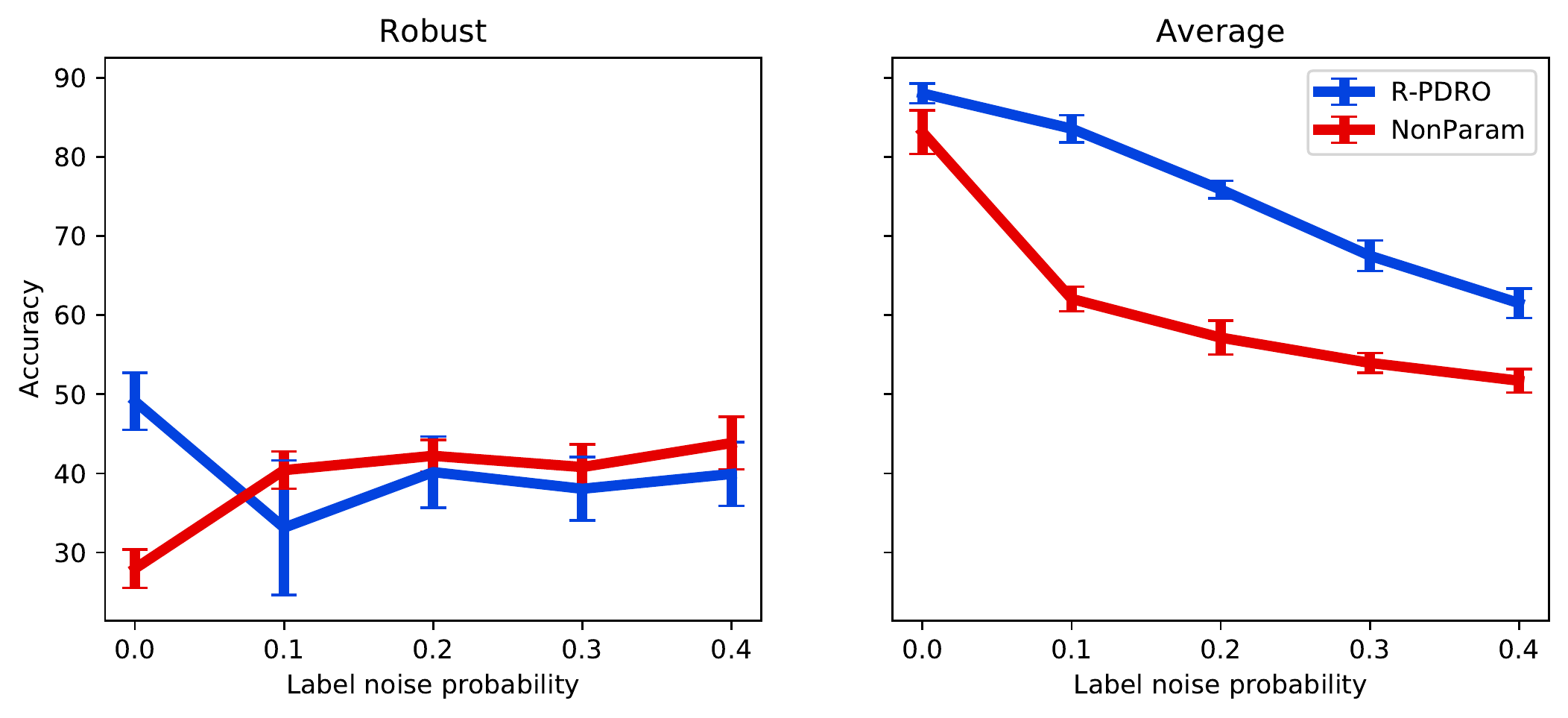}
    \caption{\label{fig:noise}Effect of label noise on parametric and non parametric DRO.}
    
\end{figure}

\subsection{Why are Parametric Adversaries Better? The Case of Label Noise}

Our experimental results in Section \ref{sec:rpdro_experiments} shows that parametric approaches such as \ac{p-dro} and R-PDRO consistently outperform their non-parametric counterparts. A possible explanation for this phenomenon is that for non-parametric approaches, the optimal weights generally only depends on the loss of the model. This can be problematic because the non-parametric worst-case distribution will indiscriminately up-weight noisy samples that have high loss. On the other hand, we hypothesize that it is more difficult for the parametric adversary to ``fit to the noise'' and that it tends to focus more on systematic patterns of failures of the model.

To corroborate this hypothesis, we perform experiments by adding increasing amounts of noise to the Biased SST. Specifically, for each example in the training set we replace the original label with a random label with probability $p_{\text{noise}}= 0, 0.1, \ldots,0.5$.\footnote{$p_\text{noise}=0$ corresponds to the original dataset.} We then train models on these increasingly noisy datasets using both a parametric (R-PDRO) and non-parametric (NonParam) approach. To simplify experiments we only run one hyper-parameter configuration for each ($\tau=0.1$ and $\kappa=1$ for R-PDRO and NonParam respectively) and report the test accuracies of the model with the highest robust accuracy on the validation set. Results are averaged over 5 runs with different random seeds.

As showcased in Figure \ref{fig:noise}, while R-PDRO's robust accuracy seems to suffer under the effect of noise, we find that its average accuracy is a lot more resilient to the introduction of label noise compared to NonParam, losing around $12$ points when $p_{\text{noise}}=0.2$ ($88.04\rightarrow 75.89$), versus $26$ for NonParam ($83.14\rightarrow 57.16$). 
This further supports our hypothesis that non-parametric adversaries tend to fit to these noisy examples, which decreases the overall quality of the resulting classifier.

\subsection{Optimization with Simultaneous Gradient Updates Plays a Crucial Role}

Despite the aforementionned results, it remains unclear \emph{why} R-PDRO learns re-weightings that are less sensitive to label noise or difficult examples compared to R-PDRO. Indeed, since the non-parametric adversary is the optimal solution of the inner minimization problem in Eq. \ref{eq:ratio_dro_problem}, it stands to reason that (large, sometimes over-parameterized) parametric adversaries from R-PDRO would converge to the same solutions as NonParam.

Our hypothesis is that the simultaneous updates to both model and adversary parameters prevent the parametric adversary from converging towards such non-parametric solutions, and provides some implicit regularization against up-weighting examples that are noisy or too difficult.

To verify this hypothesis, we conduct a toy experiment where we allow the adversary to take additional gradient steps in-between each update to the classifier. At the limit, this would allow the adversary to find an optimum of the inner maximization problem at each step of training the classifier (which for large enough adversaries, might come close to the non-parametric solution).

\begin{figure}[t!]
    \centering
    \includegraphics[width=0.6\textwidth]{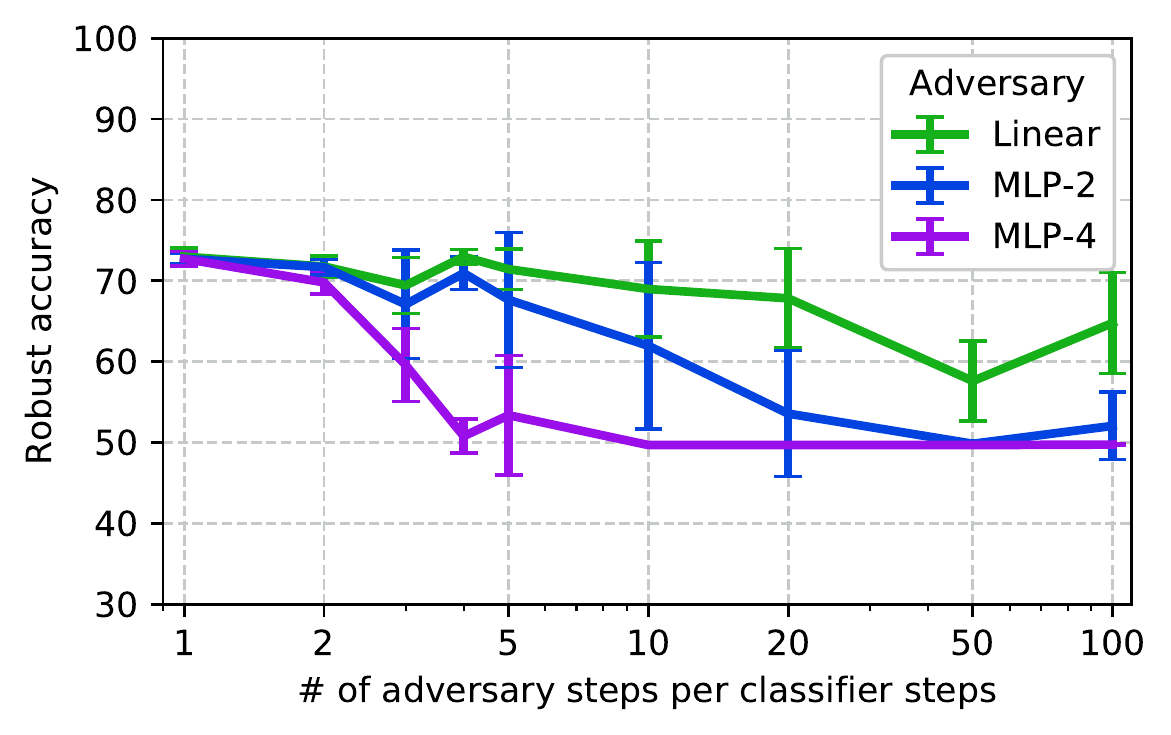}
    \caption{\label{fig:adv_steps}Evolution of R-PDRO's robust accuracy as the adversary is allowed to take more gradient steps than the classifier in a toy setting.}
    
\end{figure}

For computational efficiency, these experiments are performed on the toy setting described in Chapter \ref{ch:modeling_the_second_player_dro}, Section \ref{sec:kl_relaxation_ablation}: a linear model is trained on a binary classification problem with two domains, one of which is severely under-represented. For our adversary, we experiment with a linear adversary, as well as larger multilayer perceptrons with one hidden layer and 2 (MLP-2) and 4 (MLP-4) hidden units. In Figure \ref{fig:adv_steps}, we report the average robust accuracy (across 5 reruns) for classifiers trained with R-PDRO when the adversary is allowed to take more steps than the classifier.

We observe that R-PDRO’s robust accuracy suffers from giving the adversary too much time to catch up with the classifier: as the number of updates to the adversary increases, robust accuracy decreases. This effect is amplified in larger adversaries (\eg{} MLP-4).

This experiment underlines the importance of the simultaneous gradient updates, which prevent large, over-parameterized adversaries from converging to the sub-optimal non-parametric solution.

\begin{figure}[!t]
\centering
\begin{subfigure}[t]{\columnwidth}
\centering
\includegraphics[width=\columnwidth]{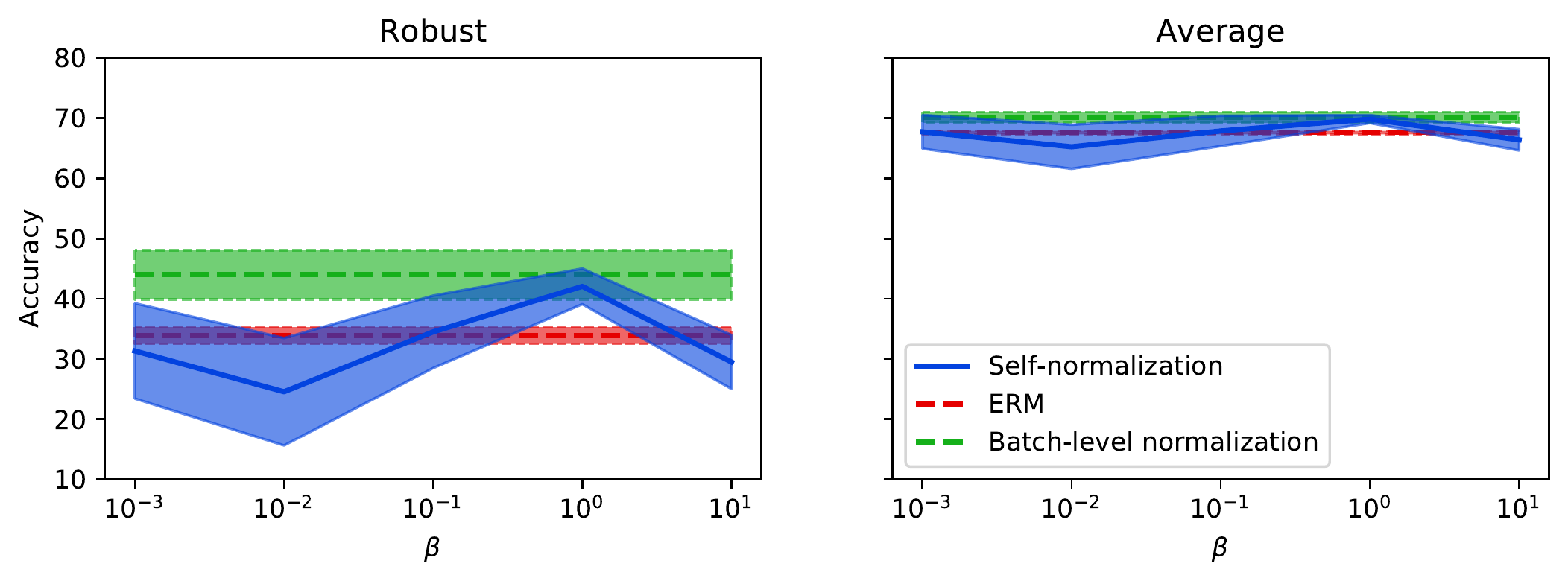}
\caption{\label{fig:self_norm_minmax} Minmax stopping}
\end{subfigure}

\begin{subfigure}[t]{\columnwidth}

\centering
\includegraphics[width=\columnwidth]{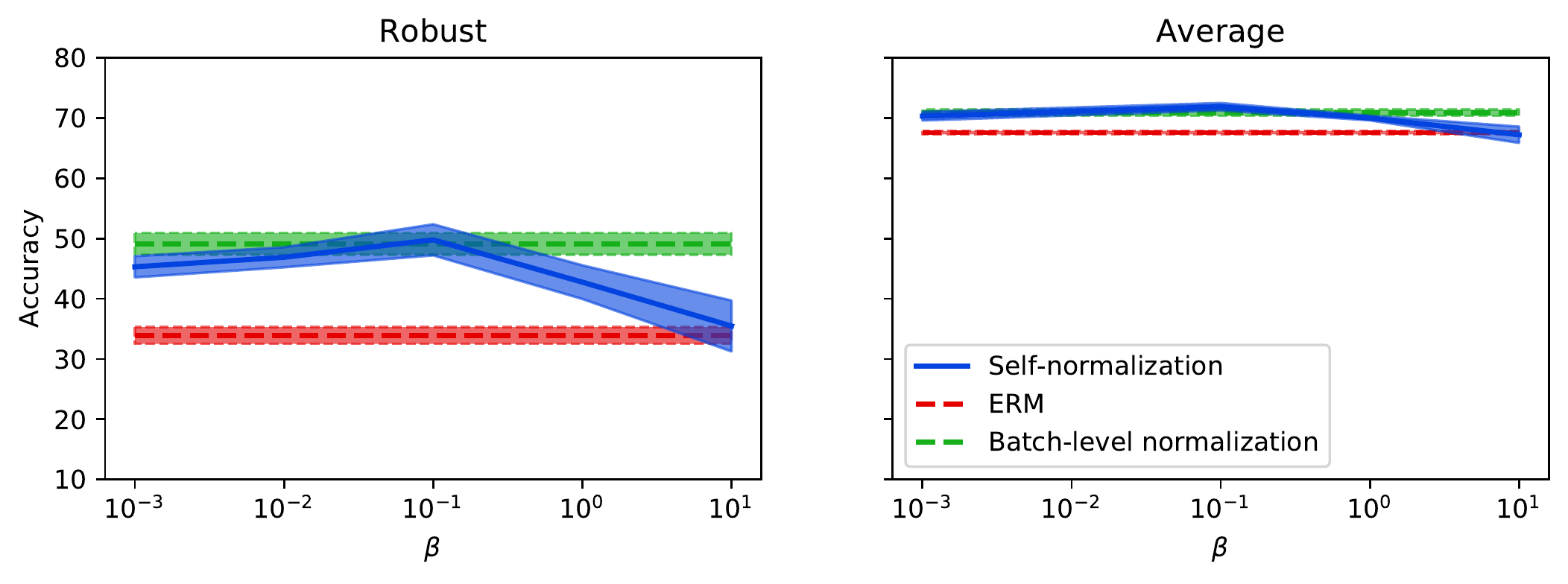}
\caption{\label{fig:self_norm_oracle} Oracle stopping}
\end{subfigure}
\caption{\label{fig:self_norm} Effect of self-normalization coefficient $\beta$ on robust and average accuracy. For two different stopping strategy: Minmax stopping (used in experiments in Section \ref{sec:rpdro_experiments}) and Oracle stopping. We report results of ERM (which corresponds to $\beta=\infty$) and batch level renormalization for comparison.}
\end{figure}

\subsection{Batch-level Normalization vs Self-normalization}

In Section \ref{sec:normalization}, we discussed two alternatives for enforcing the normalization constraint on the ratios ($\mathbb E_{p}r_\psi=1$): a regularization-based approach (``self-normalization'') and batch level renormalization. In Figure \ref{fig:self_norm_minmax}, we show the effect of self-normalization with different values of the regularization weight $\beta$ for a fixed value of the KL penalty $\tau=0.1$. We find that batch-level normalization achieves a slightly better robust accuracy than the best self-normalization configuration.

In chapter \ref{ch:modeling_the_second_player_dro}, the Minmax stopping criterion was identified as a major contributing factor to the performance of parametric DRO approaches. To understand how it affects each normalization strategy, we perform the same experiment as above but this time \emph{without} the Minmax criterion, selecting instead the model with the highest robust accuracy on the validation set (``Oracle stopping''), which provides an upper bound to the robust accuracy that can be achieved. Results in Figure \ref{fig:self_norm_oracle} show that although accuracies are generally closer, we again find that batch-level normalization matches the best self-normalization penalty. This indicates that batch-level normalization not only performs as well as self-normalization (without the need for tuning the additional hyper-parameter $\beta$), but also that it interacts better with the Minmax stopping criterion, making it a preferable alternative overall.

\subsection{Effect of Batch Size on Re-normalization}

As pointed out in Section \ref{sec:normalization}, a potential downside of the minibatch-level normalization approach is that the effective weight of each sample then depends on the other samples within the minibatch. For example, consider an adversary that assigns high weight to only 5\% of the training data. With a small enough batch size, it is likely that some batches may not contain any example of the high weight subpopulation, in which case minibatch level renormalization will overestimate the weight of the sample in the minibatch.

To assess the severity of this issue, we run R-PDRO on Biased SST with $\tau=0.1$ and vary the batch size in $\{4, 8, 16, 32, 64, 128\}$. Each configuration is run 3 times, and we report average and standard deviation of the robust and average test accuracies in Table \ref{tab:bsz_ablations}. Results suggest that while robust accuracy indeed deteriorates for lower batch sizes (4 and 8), results are consistently good for batch sizes upwards of 16, a reasonable number considering that larger batch sizes are often preferred in the literature \citep{popel2018training,goyal2017accurate}.

\begin{table}
\begin{center}
\caption{\label{tab:bsz_ablations} Effect of batch size on R-PDRO performance.}
\begin{tabular}{rrr}
\toprule
Batch size & Robust & Average \\
\midrule
64 (ERM) & { 32.97}  {\scriptsize$\pm$ 2.34} & { 92.42}  {\scriptsize$\pm$ 0.38} \\
\midrule
4 & { 37.50}  {\scriptsize$\pm$ 2.37} & { 92.32}  {\scriptsize$\pm$ 0.75} \\
8 & { 38.08}  {\scriptsize$\pm$ 2.06} & { 91.61}  {\scriptsize$\pm$ 0.48} \\
16 &  41.67  {\scriptsize$\pm$ 3.53} & {91.24}  {\scriptsize$\pm$ 0.34} \\
32 & 42.32  {\scriptsize$\pm$ 0.72} & { 89.77}  {\scriptsize$\pm$ 1.42} \\
64 & { 44.15}  {\scriptsize$\pm$ 6.83} & { 88.30}  {\scriptsize$\pm$ 2.33} \\
128 & { 42.25}  {\scriptsize$\pm$ 7.90} & { 88.21}  {\scriptsize$\pm$ 2.02} \\
\bottomrule
\end{tabular}
\end{center}
\end{table}

\section{Conclusion}

In this chapter we have proposed a parametric, likelihood ratio based approach to distributionally robust optimization of machine learning models. With the proposed method, we can use any type of parametric function estimator to define the uncertainty set of the DRO min-max game. We showed that with a careful renormalization strategy, the proposed method (R-PDRO) can be used to train robust models. It depends on very few hyper-parameters and consistently performs well on a number of benchmarks, making it an appealing off-the-shelf option. Finally we have shown that such parametric approaches are more resilient to the presence of noise in the training data when compared to their non-parametric alternatives.

The main downside of R-PDRO is the computational overhead of jointly training a second neural model. An interesting direction for future work is to improve its efficiency through parallel computation or by sharing parameters between the classifier and the adversary.

\clearpage

\part{Adaptation}
\label{sec:adaptation}

\chapter{Regularizing Trajectories to Mitigate Catastrophic Forgetting}
\chaptermark{Regularizing Trajectories}
\label{ch:regularizing_trajectories}

\setlength\epigraphwidth{.42\textwidth}
\epigraph{
It is good to have an end to journey toward; but it is the journey that matters, in the end.}{\textit{Ursula K. Le Guin}}

\section{Introduction}

The methods presented in previous chapters are able to train models that are more robust to distributional shifts because they can identify sub-population of the training data corresponding to potential domains where the model performs poorly. However, there are cases where the model might be confronted with a completely unseen test distribution, or even with a new task. In such cases where the test environment cannot be anticipated, an alternative solution is to \emph{adapt} the model, using the limited amounts of data that are available in the target domain or task.

However, continual learning of multiple tasks is hampered by catastrophic forgetting \citep{mccloskey1989catastrophic,ratcliff1990connectionist}, the tendency of previously acquired knowledge to be overwritten when learning a new task.

Modern techniques to mitigate catastrophic forgetting can be roughly categorized into 3 lines of work (see \citet{parisi2019continual} for a comprehensive overview):
1.~regularization-based approaches, where forgetting is mitigated by the addition of a penalty term in the learning objective (\citet{kirkpatrick2017overcoming,chaudhry2018riemannian}, \textit{inter alia}),
2.~dynamic architectures approaches, which incrementally increase the model's capacity to accomodate the new tasks \citep{rusu2016progressive}, and
3.~memory-based approaches, which retain data from learned tasks for later reuse \citep{lopez2017gradient,chaudhry2018efficient,chaudhry2019continual}. 
Among these, regularization-based approaches are particularly appealing because they do not increase the model size and do not require access to past data. 
This is particularly relevant to real-world scenarios where keeping data from previous training tasks may be impractical because of infrastructural or privacy-related reasons. Moreover, they are of independent intellectual interest because of their biological inspiration rooted in the idea of synaptic consolidation \citep{kirkpatrick2017overcoming}.

\begin{figure*}[t!]
\centering
\includegraphics[width=\textwidth]{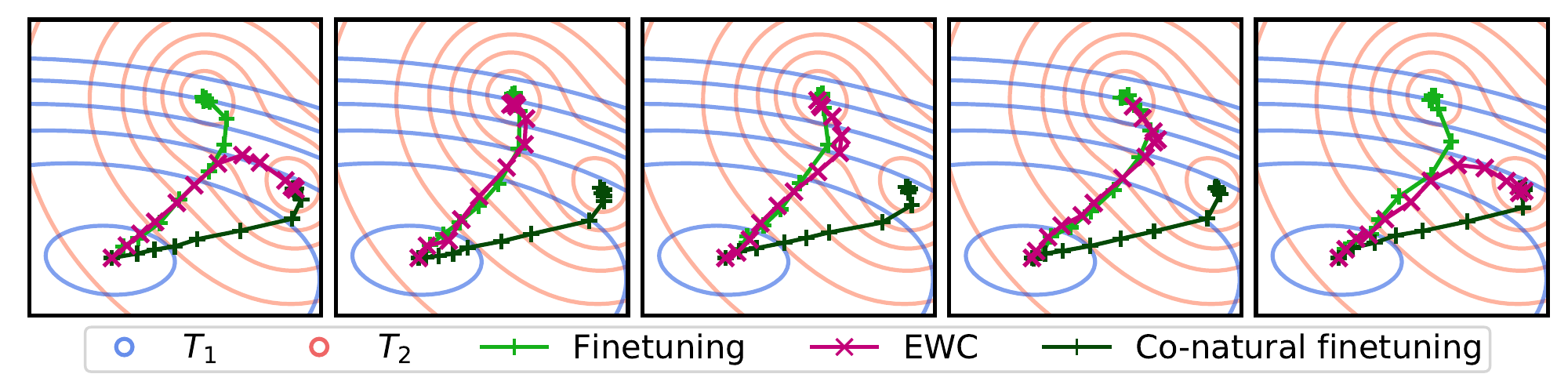}
\caption{\label{fig:trajectories_explanation} {\bf On the importance of trajectories: an example with 2-dimensional logistic regression.} Having learned task $T_1$, the model is trained on $T_2$ with two different objectives: minimizing the loss on $T_2$ (Finetuning) and a regularized objective (EWC; \citet{kirkpatrick2017overcoming}). We add a small amount of Gaussian noise to gradients in order to simulate the stochasticity of the trajectory. Plain finetuning and EWC often converge to a solution with high loss for $T_1$, but the co-natural optimization trajectory \textit{consistently} converges towards the optimum with lowest loss for $T_1$. Each individual plot corresponds to a separate run using a different random seed.}
\end{figure*}

In these regularization-based approaches, a good regularizer will ensure that, when learning a new task, gradient descent will ultimately converge to parameters that yield good results on the new task while preserving performance on previously learned tasks.
Critically, this is predicated upon successful optimization of the regularized objective, a fact that has been largely taken for granted in previous work.
Non-convexity of the loss function, along with noise in the data (due to small or biased datasets) or in the gradients (due to stochastic gradient descent), can yield optimization trajectories --- and ultimately convergence points --- that are highly non-deterministic, even for the same starting parameters.
As we demonstrate in this chapter, this can cause unintended catastrophic forgetting along the optimization path. This is illustrated in a toy setting in Figure \ref{fig:trajectories_explanation}: a two parameter model is trained to perform task $T_2$ (an arbitrary bi-modal loss function) after having learned task $T_1$ (a logistic regression task). Standard finetuning, even in the presence of a regularized objective (EWC; \citet{kirkpatrick2017overcoming}), quickly changes the loss of $T_1$ and tends to converge to a solution with high $T_1$ loss.

We propose to remedy this issue by regularizing not the objective function but \emph{the optimization trajectory itself}, specifically by preconditioning gradient descent with the empirical Fisher information of previously learned tasks (\S\ref{sec:approach}).
This yields what we refer to as a \emph{co-natural} gradient, an update rule inspired by the natural gradient \citep{amari1997neural}, but taking the Fisher information of \emph{previous tasks} as a natural Riemannian metric\footnote{Informally, the reader can think of a Riemannian metric as a function that assigns an inner product $u,v\mapsto g_{\mathbf{x}}(u,v)$ to each point $x$ in the space, thus inducing a localized notion of distance and curvature.} of the parameter space, instead of the Fisher information of the task being optimized for. When we introduce our proposed co-natural gradient for the toy example of Figure \ref{fig:trajectories_explanation}, the learning trajectory follows a path that changes the loss on $T_1$ much more slowly, and tends to converges to the optimum that incurs the lowest performance degradation on $T_1$.

We test the validity of our approach in a continual learning scenario (\S\ref{sec:cl}). We show that the co-natural gradient consistently reduces forgetting in a variety of existing continual learning approaches by a large factor, and greatly improves performance over simple finetuning, without modification to the training objective. As an additional contribution, we assemble a new collection of 13 tasks for evaluating continual learning on text classification.

We further investigate the special case of transfer learning in a two-task, low-resource scenario. In this specific case, control over the optimization trajectory is particularly useful because the optimizer has to rely on early stopping to prevent overfitting to the meager amount of training data in the target task. We show that the co-natural gradient yields the best trade-offs between source and target domain performance over a variety of hyper-parameters (\S\ref{sec:low_resource}).

The work presented in this chapter bears many similarities to previous related works on Bayesian formulations of continual learning which used similar inverse-Fisher conditioning to reduce forgetting: NVCL \citep{tseran2018natural} and GNG \citep{chen2018facilitating}. However, we our work makes significant contributions over these previous approaches

First, these prior works were developed in the framework of Bayesian continual learning. While the central idea of using the Fisher to modulate the update is similar, we adopt a broader approach and successfully apply this technique to a variety of other techniques outside the Bayesian scaffolding. Second, we introduce a damping coefficient, which allows control over the regularization effect of the co-natural gradient, as evidenced by our analysis in Appendix A.4 specifically, and more generally by the results in Section 4 and 5, where our results are obtained by performing grid-search over this additional hyper-parameter. Third, we provide empirical evidence of the co-natural gradients' effectiveness in a variety of experimental settings from relatively small, toy-like models and datasets (\textit{e.g.} Split CIFAR or Omniglot) to more realistic scenarios (\textit{e.g.} MiniImageNet, BERT on text classification and our MT experiments). In comparison, the aforementioned previous work only test small 2 layers MLP models on MNIST and FashionMNIST. We believe that this establishes our work as a strong, independent contribution.

\section{Background and Notations}

We first give a brief overview of the continual learning paradigm and existing approaches for overcoming catastrophic forgetting.

\subsection{Notation}

Let us define a task as a triplet containing an input space $\mathcal X$ and an output space $\mathcal Y$, both measurable spaces, as well as a distribution $\mathcal{D}$ over $\mathcal X\times \mathcal Y$. 
In general, learning a task will consist of training a model to approximate the conditional distribution $p(y\mid x)$ induced by $\mathcal D$.

Consider a probabilistic model $p_{\theta}$ parametrized by $\theta\in\mathbb R^d$ where $d$ is the size of the model, trained to perform a \textit{source} task $S = \langle \mathcal{X}_S, \mathcal{Y}_S, \mathcal{D}_S \rangle$ to some level of performance, yielding parameters $\theta_S$. 
In the most simple instance of continual learning, we are tasked with learning a second \textit{target} task $T=\langle \mathcal{X}_T, \mathcal{Y}_T, \mathcal{D}_T \rangle$. 
In general in a multitask setting, it is not the case that the input or output spaces are the same. 
The discrepancy between input/output space can be addressed in various ways, \eg{} by adding a minimal number of task-specific parameters (for example, different softmax layers for different label sets). 
To simplify exposition, we set these more specific considerations aside for now, and assume that $\mathcal X_S=\mathcal X_T$ and $\mathcal Y_S=\mathcal Y_T$. 

At any given point during training for task $T$, our objective will be to minimize the loss function $\mathcal L_T(\theta)$ -- generally the expected negative log-likelihood $\mathbb E_{x,y\sim\mathcal D_T} [-\log p_\theta(y \mid x)]$. 
Typically, this will be performed by iteratively adding incremental update vectors $\delta\in\mathbb{R}^d$ to the parameters $\theta \longleftarrow \theta + \delta$.

\subsection{Existing Approaches for Continual Learning}

In this chapter, we focus on those models that have a fixed architecture over the course of continual learning. The study of continual learning for models of fixed capacity can be split into two distinct (but often overlapping) streams of work:

\textbf{Regularization-based approaches} introduce a penalty in the loss function $\mathcal L_T$, often quadratic, pulling the weights $\theta$ back towards $\theta_S$:

\begin{equation}
    \mathcal{L}_T(\theta) = \underbrace{\mathbb E_{\mathcal D_T} -\log p_\theta(y \mid x)}_{\text{NLL on task }T} + \underbrace{\lambda (\theta-\theta_S)^T\Omega_S(\theta-\theta_S)}_{\text{Regularization term}}
\end{equation}

where $\Omega_S$ is a matrix, generally diagonal, that encodes the respective importance of each parameter with respect to task $S$, and $\lambda$ is a regularization strength hyper-parameter. Various choices have been proposed for $\Omega_S$; the diagonal empirical Fisher information matrix \citep{kirkpatrick2017overcoming}, or path-integral based importance measures \citep{zenke2017continual,chaudhry2018riemannian}. More elaborate regularizers have been proposed based on \eg{} a Bayesian formulation of continual learning \citep{nguyen2017variational,ahn2019uncertainty} or a distillation term \citep{li2017learning,dhar2019learning}. The main advantage of these approaches is that they do not rely on having access to training data of previous tasks.

\textbf{Memory-based approaches} store data from previously seen tasks for re-use in continual learning, either as a form of constraint, by \eg{} ensuring that training on the new task doesn't increase the loss on previous tasks \citep{lopez2017gradient,chaudhry2018efficient}, or for replay \ie{} by retraining on instances from previous tasks \citep{rebuffi2017icarl,chaudhry2019continual,aljundi2019gradient,aljundi2019online}. Various techniques have been proposed for the selection of samples to store in the memory \citep{chaudhry2019continual,aljundi2019gradient} or for retrieval of the samples to be used for replay \cite{aljundi2019online}.

All of these methods rely on stochastic gradient descent to optimize their regularized objective or to perform experience replay, with the notable exception of GEM \citep{lopez2017gradient,chaudhry2018efficient}, where the gradients are projected onto the orthogonal complement of previous task's gradients. However, this method has been shown to perform poorly in comparison with simple replay \citep{chaudhry2019continual}, and it still necessitates access to data from previous tasks.
\section{Regularizing the Trajectory}
\label{sec:approach}

After briefly recalling how the usual update is obtained in gradient descent, we derive a new, \emph{co-natural} update designed to better preserve the distribution induced by the model over previous tasks.

\subsection{Warm up: the Standard Gradient Descent Update}

At point $\theta$ in the parameter space, gradient descent finds the optimal update $\delta$ that is (1) small and (2) locally minimizes the difference in loss $\mathcal L(\theta+\delta) - \mathcal L(\theta)$ ($\approx \delta^\intercal\nabla_\theta\mathcal L$ at the first order). 
Traditionally this can be formulated as minimizing the Lagrangian:

\begin{equation}\label{eqn:lagrangian_std}
    \mathbb L(\delta)= \underbrace{\delta^\intercal\nabla_\theta\mathcal L_T}_{\substack{\text{first order} \\ \text{loss minimization term}}} + \underbrace{\mu\Vert\delta\Vert^2}_{\text{``small update'' term}}
\end{equation}

with Lagrangian multiplier $\mu>0$. 
Minimizing $\mathbb L$ for $\delta$ yields the well-known optimal update $\delta^*$:

\begin{equation}\label{eqn:std_update}
    \delta^*=-\frac 1 {2\mu}\nabla_\theta\mathcal L_T
\end{equation}

where $\frac 1 {2\mu}$ corresponds to the learning rate (see Appendix \ref{ch:supplemental_07} for the full derivation).

\subsection{KL Regularization of Trajectories}

The $\Vert\delta\Vert^2$ term in $\mathbb L$ implicitly expresses the underlying assumption that the best measure of distance between parameters $\theta$ and $\theta+\delta$ is the Euclidean distance. In a continual learning setting however, the quantity we are most interested in preserving is the probability distribution that $\theta$ models on the source task $S$:

\begin{equation}
    p^S_{\theta}(x, y)=p_\theta(y\mid x) p^S(x)
\end{equation}

Therefore, a more \emph{natural} distance between $\theta$ and $\theta+\delta$ is the Kullback-Leibler divergence $\text{KL}(p^S_{\theta}\Vert p^S_{\theta+\delta})$ \citep{kullback1951information}. For preventing catastrophic forgetting along the optimization path, we incorporate incorporate this KL term into the Lagrangian $\mathbb L$ itself:

\begin{equation}\label{eqn:lagrangian_kl}
    \mathbb L(\delta)= \delta^\intercal\nabla_\theta\mathcal L_T + \mu\Vert\delta\Vert^2 + \nu \text{KL}(p^S_{\theta}\Vert p^S_{\theta+\delta})
\end{equation}

Doing so means that the optimization trajectory will tend to follow the direction that changes the distribution of the model the least. Notably, this is not a function of the previous objective $\mathcal{L}_S$, so knowledge of the original training objective is not necessary during continual learning (which is typically the case in path-integral based regularization methods \citep{zenke2017continual} or experience replay \citep{chaudhry2019continual}).

\subsection{Co-natural Gradient Optimization}

Presuming that $\delta$ is small, we can perform a second order Taylor approximation of the function $\delta \mapsto \text{KL}(p^S_{\theta}\Vert p^S_{\theta+\delta})$ around 0. 
Considering that both the zeroeth and first order terms are null because $0$ is a global minimizer of $\delta\mapsto \text{KL}(p^S_{\theta}\Vert p^S_{\theta+\delta})$, this reduces the Lagrangian to a quadratic optimization problem (we refer the reader to \citet{pascanu2013revisiting} for a more detailed derivation.):

\begin{equation}\label{eqn:lagrangian_fisher}
    \mathbb L(\delta)= \delta^\intercal\nabla_\theta\mathcal L_T + \mu\Vert\delta\Vert^2 + \frac 1 2 \nu\delta^\intercal F^S_\theta \delta
\end{equation}

where $F^S_\theta$ is the Hessian of the KL divergence around $\theta$. A crucial, well-known property of this matrix is that it coincides with the Fisher information matrix\footnote{Hence our use of the letter $F$ to designate the Hessian} $\mathbb E_{x,y\sim p_\theta}[(\nabla \log p^S_\theta) (\nabla \log p^S_\theta)^T]$ (the expectation being taken over the model's distribution $p_\theta$; see Appendix \ref{ch:supplemental_07} for details). 
This is appealing from a computational perspective because the Fisher can be computed by means of first order derivatives only.

Minimizing for $\delta$ yields the following optimal update:

\begin{equation}\label{eqn:conatural_gradient}
    \delta^*=-\lambda\left[F^S_{\theta} + \alpha I\right]^{-1}\nabla_\theta\mathcal L_T
\end{equation}

where coefficients $\mu$ and $\nu$ are folded into two hyper-parameters: the learning rate $\lambda$ and a damping coefficient $\alpha$ (the step-by-step derivation can be found in Appendix \ref{ch:supplemental_07}). 
In practice, especially with low damping coefficients, it is common to obtain updates that are too large (typically when some parameters have no effect on the KL divergence).
To address this, we re-normalize $\delta^*$ to have the same norm as the original gradient,  $\Vert\nabla\mathcal{L}_T\Vert$. Moreover, to improve numerical stability in the case of degenerate Fisher matrices, we add a very small damping term of $10^{-12}$ to all experiments.

For computational reasons, we will make 3 key practical approximations to the Fisher:

\begin{enumerate}
    \item $F^S_{\theta}\approx F^S_{\theta_S}$: we maintain the Fisher computed at $\theta_S$, instead of recomputing $F_S$ at every step of training. This relieves us of the computational burden of updating the Fisher for every new value of $\theta$. This approximation (shared by previous work, \eg{} \citet{kirkpatrick2017overcoming};\citet{chaudhry2018riemannian}) is only valid insofar as $\theta_S$ and $\theta$ are close. Empirically we observe that this still leads to good results.
    \item $F^S$ is diagonal: this is a common approximation in practice with two appealing properties. First, this makes it realistic to store the $d$ diagonal Fisher coefficients in memory. Second, this trivializes the inverse operation (simply invert the diagonal elements).
    \item Empirical Fisher: this common approximation replaces the expectation under the model's distribution by the expected log-likelihood of the \emph{true} distribution: $\mathbb \mathbb E_{x, y\sim p^S}[(\nabla \log p^S_\theta) (\nabla \log p^S_\theta)^T]$ (mind the subscript). This is particularly useful in tasks with a large or unbounded number of classes (\eg{} structured prediction), where summing over all possible outputs is intractable. We can then compute the diagonal of the empirical Fisher using Monte Carlo sampling: $\frac 1 N\sum_{i=1}^N[\nabla
    \log p^S_{\theta}(y_i\mid x_i)]^2$ with $(x_i, y_i)$ sampled from $\mathcal D_S$ (we use $N=1000$ for all experiments).
\end{enumerate}

This formulation bears many similarities with the natural gradient from \citet{amari1997neural}, which also uses the KL divergence as a metric for choosing the optimal update $\delta^*$ in gradient descent.
There is a however a crucial difference, both in execution and purpose: where the natural gradient uses knowledge of the curvature of the KL divergence of $\mathcal{D}_T$ to \emph{speed-up} convergence, our proposed method leverages the curvature of the KL divergence on $\mathcal{D}_S$ to \emph{slow-down} divergence from $p^S_{\theta_S}$. To highlight the resemblance and complementarity between these two concepts, we refer to the new update as the \emph{co-natural} gradient.

\subsection{Beyond Two Tasks}
\label{sec:beyond_two_tasks}

In a continual learning scenario, 
we are confronted with a large number of tasks $T_1\ldots T_n$ presented in sequential order. When learning $T_n$, we can change the Lagrangian $\mathbb L$ from \ref{eqn:lagrangian_kl} to incorporate the constraints for all previous tasks $T_1\ldots T_{n-1}$:

\begin{equation}\label{eqn:lagrangian_kl_n}
    \mathbb L(\delta)= \delta^\intercal\nabla_\theta\mathcal L_{T_n} + \mu\Vert\delta\Vert^2 + \sum_{i=1}^{n-1} \nu_i \text{KL}(p^{T_i}_{\theta}\Vert p^{T_i}_{\theta+\delta})
\end{equation}

This in turn changes the Fisher in Eq.~\ref{eqn:lagrangian_kl_n} to $\Tilde{F}_{n-1}:=\frac 1 2\sum_{i=1}^{n-1}\nu_i F^{T_{i}}$. The choice of the coefficients $\nu_i$ is crucial. Setting all $\nu_i$ to the same value, \ie{} assigning the same importance to all tasks is suboptimal for a few reasons. First and foremost, it is unreasonable to expect of a model with finite capacity to remember an unbounded number of tasks (as tasks ``fill-up'' the model capacity, $\Tilde{F}_{n-1}$ is likely to become more ``homogeneous''). Second, as training progresses and $\theta$ changes, our approximation that $F^{T_i}_{\theta}\approx F^{T_i}_{\theta_{T_i}}$ is less and less likely to hold.

We address this issue in the same fashion as \citet{schwarz2018progress}, by keeping a rolling exponential average of the Fisher matrices:

\begin{equation}\label{eqn:rolling_fisher}
    \Tilde{F}^{\gamma}_{n} = \gamma F_{T_{n}}+ (1-\gamma) \Tilde{F}^{\gamma}_{n-1}
\end{equation}

In this case, previous tasks are gracefully forgotten at an exponential rate controlled by $\gamma$. We account for the damping $\alpha$ term in Eq.~\ref{eqn:conatural_gradient} by setting $\Tilde{F}_{0}:=\frac{\alpha}{\gamma}I$. In preliminary experiments, we have found $\gamma=0.9$ to yield consistently good results, and use this value in all presented experiments.

\section{Continual Learning Experiment Setting}
\label{sec:cl}

\begin{table*}[!t]
{
\caption{\label{tab:cl_average_acc_and_f} Average accuracies and forgetting after all tasks have been learnt, with and without the co-natural gradient. Results are reported in percentages ($\pm$ the standard deviation over 5 re-runs). Bold print indicates statistically significant difference between standard and co-natural ($p<0.05$).}
\setlength\tabcolsep{5pt}
\begin{subtable}{0.55\columnwidth}
\centering
\caption{\label{tab:split_cifar_acc_and_f}\bf Split CIFAR}
\noindent\begin{tabular}{@{}c|ccc}
 & Finetuning & EWC & ER \\\hline\hline
 & \multicolumn{3}{c}{Average accuracy $\uparrow$} \\
Standard & { 31.56} {\scriptsize$\pm$2.11} & { 50.63} {\scriptsize$\pm$2.30} & { 59.30} {\scriptsize$\pm$0.92} \\
Co-natural & {\bf 55.97} {\scriptsize$\pm$1.48} & {\bf 54.81} {\scriptsize$\pm$2.06} & {\bf 64.72} {\scriptsize$\pm$0.85} \\\hline
 & \multicolumn{3}{c}{Forgetting $\downarrow$} \\
Standard & { 35.48} {\scriptsize$\pm$2.46} & { 11.56} {\scriptsize$\pm$1.93} & { 9.20} {\scriptsize$\pm$0.99} \\
Co-natural & {\bf 5.04} {\scriptsize$\pm$1.24} & {\bf 4.77} {\scriptsize$\pm$1.11} & {\bf 2.52} {\scriptsize$\pm$0.58} \\\hline
\end{tabular}
\end{subtable}
\vspace{1em}
\begin{subtable}{0.4\columnwidth}
\centering
\caption{\label{tab:omniglot_acc_and_f}\bf Omniglot}
\noindent\begin{tabular}{@{}c|ccc}
 & Finetuning & EWC & ER \\\hline\hline
 & \multicolumn{3}{c}{Average accuracy $\uparrow$} \\
 & { 20.21} {\scriptsize$\pm$3.97} & {\bf 71.16} {\scriptsize$\pm$0.51} & { 70.41} {\scriptsize$\pm$3.05} \\
 & {\bf 67.84} {\scriptsize$\pm$4.85} & { 67.39} {\scriptsize$\pm$1.45} & {\bf 78.04} {\scriptsize$\pm$1.79} \\\hline
 & \multicolumn{3}{c}{Forgetting $\downarrow$} \\
 & { 73.52} {\scriptsize$\pm$3.80} & { 22.27} {\scriptsize$\pm$0.76} & { 12.60} {\scriptsize$\pm$2.76} \\
 & {\bf 12.09} {\scriptsize$\pm$4.69} & {\bf 8.61} {\scriptsize$\pm$1.93} & {\bf 4.95} {\scriptsize$\pm$0.39} \\\hline
\end{tabular}
\end{subtable}

\begin{subtable}{0.55\columnwidth}
\centering
\caption{\label{tab:miniimagenet_acc_and_f}\bf MiniImageNet}
\noindent\begin{tabular}{@{}c|ccc}
 & Finetuning & EWC & ER \\\hline\hline
 & \multicolumn{3}{c}{Average accuracy $\uparrow$} \\
Standard & { 35.36} {\scriptsize$\pm$1.75} & { 62.19} {\scriptsize$\pm$1.49} & { 65.93} {\scriptsize$\pm$0.49} \\
Co-natural & {\bf 63.91} {\scriptsize$\pm$1.54} & { 63.89} {\scriptsize$\pm$1.11} & {\bf 71.00} {\scriptsize$\pm$0.67} \\\hline
 & \multicolumn{3}{c}{Forgetting $\downarrow$} \\
Standard & { 42.08} {\scriptsize$\pm$1.56} & { 6.98} {\scriptsize$\pm$1.65} & { 12.57} {\scriptsize$\pm$0.61} \\
Co-natural & {\bf 8.86} {\scriptsize$\pm$2.03} & { 6.96} {\scriptsize$\pm$1.28} & {\bf 4.72} {\scriptsize$\pm$0.89} \\\hline
\end{tabular}
\end{subtable}
\begin{subtable}{0.4\columnwidth}
\centering
\caption{\label{tab:text_classification_acc_and_f}\bf Text Classification}
\noindent\begin{tabular}{@{}c|ccc}
 & Finetuning & EWC & ER \\\hline\hline
 & \multicolumn{3}{c}{Average accuracy $\uparrow$} \\
 & { 62.13} {\scriptsize$\pm$2.90} & { 67.79} {\scriptsize$\pm$1.18} & {\bf 73.57} {\scriptsize$\pm$0.42} \\
 & {\bf 68.74} {\scriptsize$\pm$1.49} & { 68.07} {\scriptsize$\pm$1.30} & { 69.35} {\scriptsize$\pm$1.45} \\\hline
 & \multicolumn{3}{c}{Forgetting $\downarrow$} \\
 & { 13.79} {\scriptsize$\pm$2.84} & { 0.75} {\scriptsize$\pm$0.39} & { 2.16} {\scriptsize$\pm$0.23} \\
 & {\bf 0.73} {\scriptsize$\pm$0.18} & { 0.50} {\scriptsize$\pm$0.37} & {\bf 0.50} {\scriptsize$\pm$0.25} \\\hline
\end{tabular}
\end{subtable}
}
\end{table*}

\subsection{Setup, Evaluation and Baselines}

To examine our hypothesis that controlling the optimization trajectory with the co-natural gradient reduces catastrophic forgetting, we follow the experimental procedure from \citet{chaudhry2019continual}: given a collection of tasks, we create a ``validation set'' of 3 tasks used to select the best hyper-parameters, and keep the remaining tasks for evaluation. This split is chosen at random and kept the same across all experiments. In most settings, the nature and possibly the number of classes changes from task to task. We account for this by training a separate ``task head'' for each task: an affine transform projecting the features onto the number of classes, followed by a softmax layer. We apply continual learning only to the remaining, ``feature-extraction'' part of the model.

We report results using two common metrics for continual learning: \textbf{average accuracy}, the accuracy at the end of training averaged over all tasks, and \textbf{forgetting}. Forgetting is defined in \citet{chaudhry2018riemannian} as the difference in performance on a task between the current model and the best performing model on this task. Formally if $A^T_t$ represents the accuracy on task $T$ at step $t$ of training, the forgetting $F^T_t$ at step $t$ is defined as $F^T_t=\max_{\tau<t}A^T_\tau - A^T_t$. Low forgetting means that the model tend to keep the same level of performance on a task it has previously learned.

\begin{figure*}[!t]
\centering
\begin{subfigure}[t]{0.4\textwidth}
\centering
\includegraphics[width=\columnwidth]{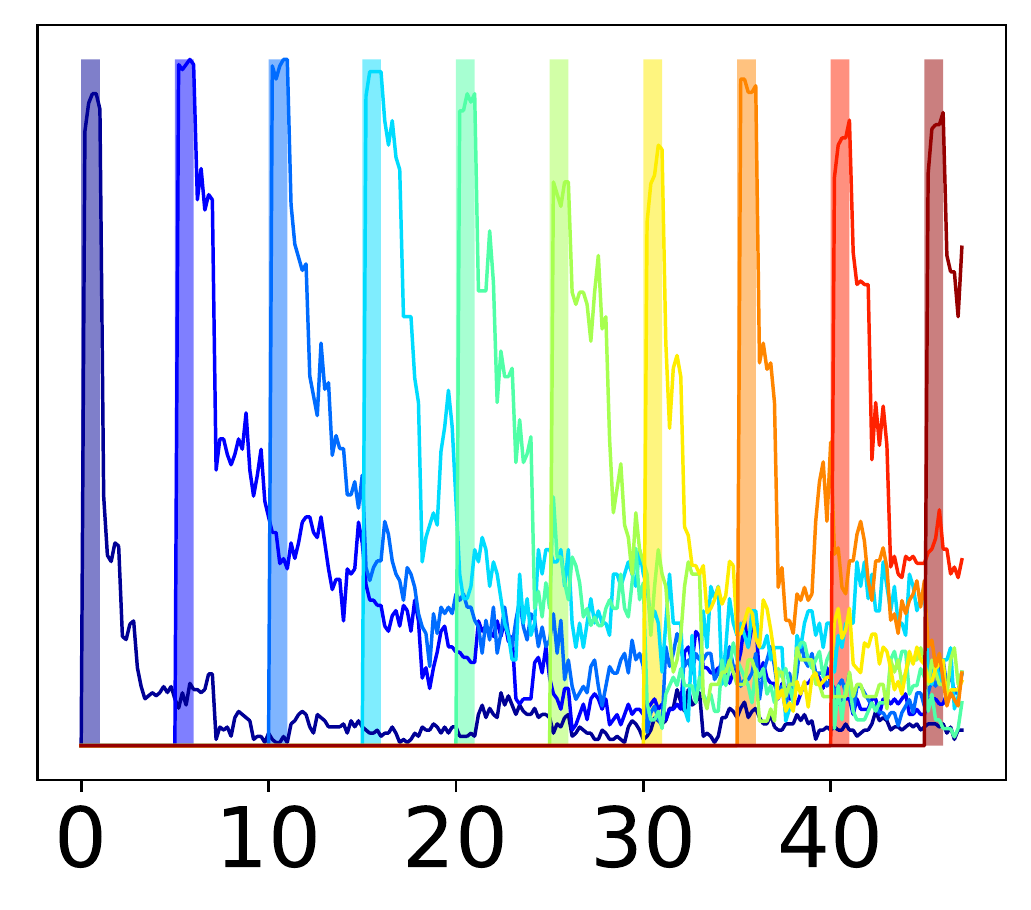}
\caption{\label{fig:finetuning} Finetuning}
\end{subfigure}\hspace{2em}%
\begin{subfigure}[t]{0.4\textwidth}
\centering
\includegraphics[width=\columnwidth]{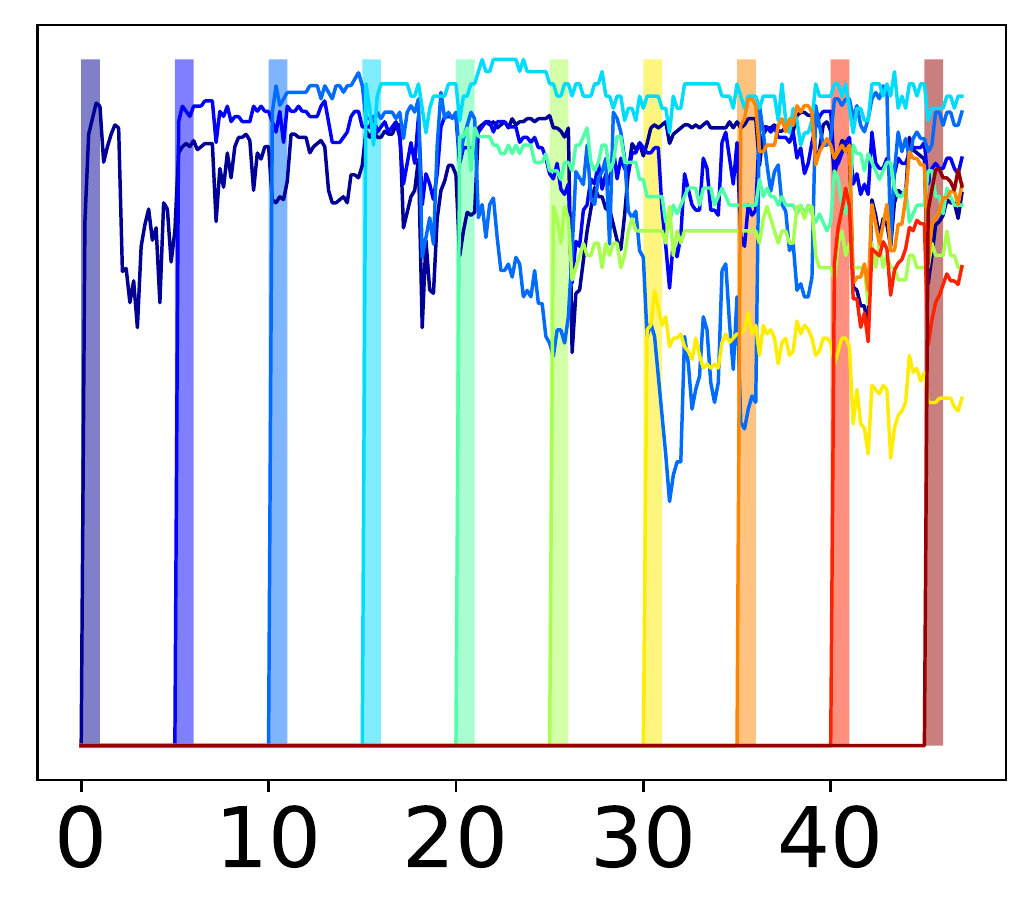}
\caption{\label{fig:ewc} EWC}
\end{subfigure}

\begin{subfigure}[t]{0.4\textwidth}
\centering
\includegraphics[width=\columnwidth]{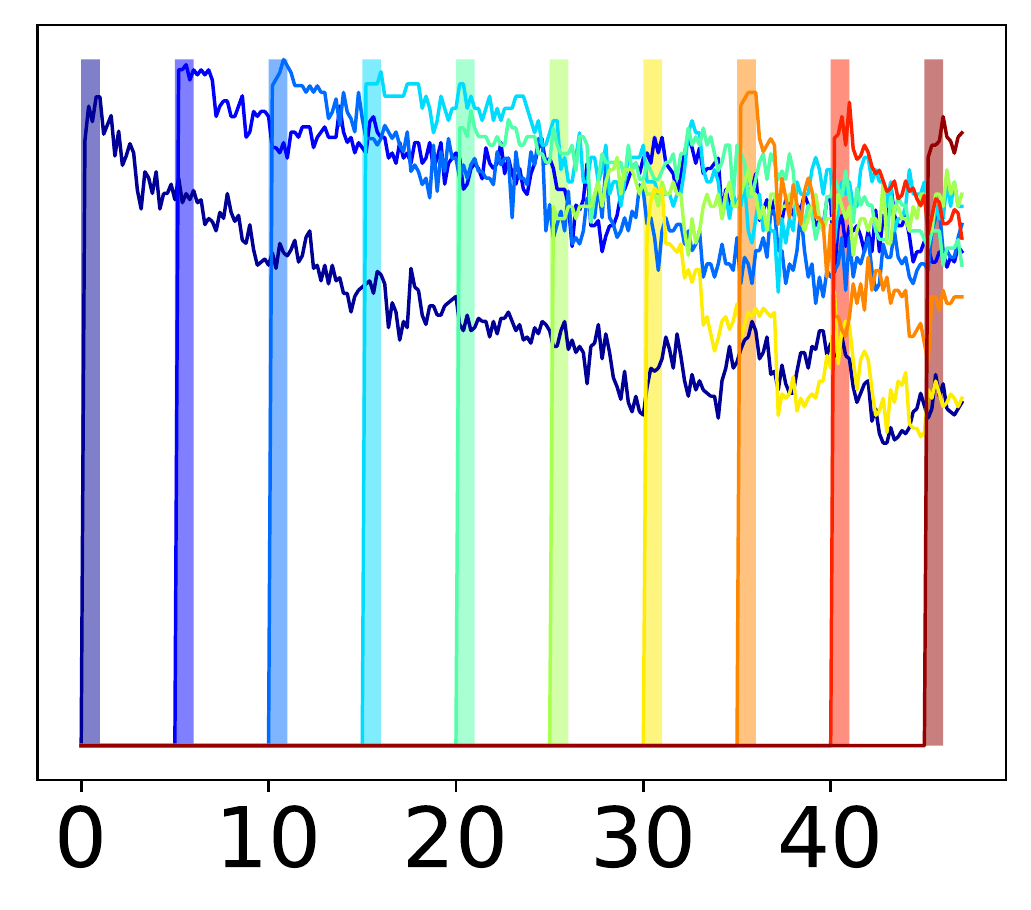}
\caption{\label{fig:er} ER}
\end{subfigure}\hspace{2em}%
\begin{subfigure}[t]{0.4\textwidth}
\centering
\includegraphics[width=\columnwidth]{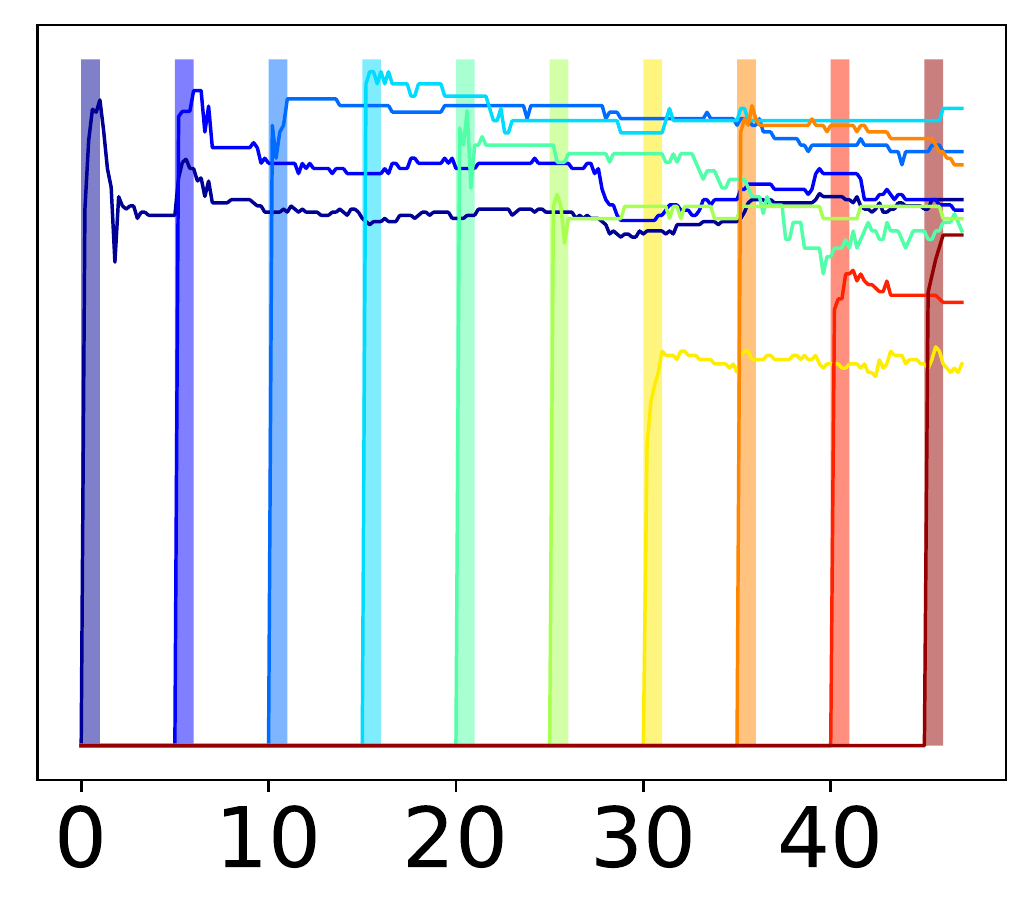}
\caption{\label{fig:conatural_finetuning} Co-natural Finetuning}
\end{subfigure}
\caption{\label{fig:viz} Evolution of task performance over the course of continual learning on one ordering of Omniglot. For visibility we only show accuracies for every fifth task. The rectangular shaded regions delineate the period during which each task is being trained upon; with the exception of ER, this is the only period the model has access to the data for this task.}
\end{figure*}

We implement the proposed co-natural update rule on top of 3 baselines:

\begin{itemize}
    \item {\bf Finetuning}: Simply train the model on the task at hand, without any form of regularization.
    \item {\bf EWC}: Proposed by \citet{kirkpatrick2017overcoming}, it is a simple but effective quadratic regularization approach. While neither the most recent nor sophisticate regularization technique, it is a natural baseline for us to compare to in that it also consists in a Fisher-based penalty --- albeit in the loss function instead of the optimization dynamics. We also use the rolling Fisher described in Section \ref{sec:beyond_two_tasks}, making our EWC baseline equivalent to the superior online EWC introduced by \citet{schwarz2018progress}.
    \item {\bf ER}: Experience replay with a fixed sized episodic memory proposed by \citet{chaudhry2019continual}. While not directly comparable to EWC in that it presupposes access to data from previous tasks, ER is a simple approach that boasts the best performances on a variety of benchmarks \citep{chaudhry2019continual}. In all experiments, we use a memory of size 1,000 populated with reservoir sampling.
\end{itemize}

Training proceeds as follows: we perform exhaustive search on all possible hyper-parameter configurations using the validation tasks. Every configuration is reran 5 times (3 for text classification) with a different random seed and assigned a score based on the average task score at the end of training (averaged over reruns). We then evaluate the best hyper-parameters by continual training on the evaluation tasks. Results are reported over 5 random restarts, and we control for statistical significance using a paired t-test (we pair together runs with the same task ordering). For all experiments we train with plain stochastic gradient descent.

\subsection{Datasets}

We use the four following task suites in our experiments:

\subsubsection{Split CIFAR}

The CIFAR100 dataset, split into 20 independent 5-way classification tasks with 2500 training examples and 500 test examples each. This split, performed at random, is kept the same across all experiments, only the order of these tasks is changed. Following \citet{chaudhry2018efficient}, we use a smaller version of the ResNet architecture \citep{he2016deep} train on each task for 1 epoch with batch size 10.

\subsubsection{Omniglot}

The Omniglot dataset \citep{lake2015human} consists of 50 independent character recognition datasets on different alphabet. We adopt the setting of \citet{schwarz2018progress} and consider each alphabet as a separate task, and split each task such that every character is present 12, 4 and 4 times in the training, validation and test set respectively (out of the 20 images for each character).\footnote{Note that this is a different setting than the usual meta-learning scenario that Omniglot is used for.}

On this dataset we use the same small CNN architecture as \citet{schwarz2018progress}. We augment the training data by randomly shifting and rotating images and train on each task for 2500 steps (120 to 417 epochs depending on the alphabet) with batch size 32. We ignore the validation data and simply evaluate on the test set at the end of training.

\subsubsection{Split MiniImageNet}

The MiniImageNet dataset (a subset of the popular ImageNet \citep{deng2009imagenet} dataset\footnote{Similarly to Omniglot, MiniImageNet was originally intended as a meta-learning benchmark and therefore its standard train/validation/test split consists of disjoint classes. We perform a custom transversal split so that the dataset can be used as a standard 100-way classification task. The accuracies reported here are not to be compared with the meta-learning literature.}; \citet{vinyals2016matching}). We split the dataset into 20 disjoint 5-way classification tasks, similarly to Split CIFAR, and use the same smaller ResNet. We train on each task for 500 steps with batch size 32 ($= 6.4$ epochs). We ignore the validation data and simply evaluate on the test set at the end of training.

\subsubsection{Text Classification}

The most recent work on continual learning of language tasks \citep{de2019episodic} relies on a relatively small set of tasks (5 classification tasks from \citet{zhang2015character}). This small number of tasks is not amenable to our experimental setup described above, where 3 tasks are reserved for validation. Therefore we assemble a larger collection of text classification datasets from three sources: the GLUE benchmark \citep{wang2018glue}, the SuperGLUE benchmark \citep{wang2019superglue} and the text classification datasets used in \citet{de2019episodic}. To keep things relatively simple, we only keep tasks that 1.~are single sentence or sentence pair classification tasks and 2.~have more than 1000 training examples. Specifically, we use the following tasks from each dataset (reported with the training data size):

\begin{itemize}
    \item \textbf{GLUE}
    \begin{itemize}
        \item CoLA \citep{Warstadt2018NeuralNA}: 8.6K
        \item MultiNLI \citep{williams18multinli}: 392.7K
        \item MRPC \citep{brocket2005automatically}: 3.7K
        \item QNLI \citep{rajpurkar-etal-2016-squad}: 108.4K
        \item QQP \citep{iyer2017first}: 363.8K
        \item RTE \citep{dagan2005pascal,haim2006second,giampiccolo2007third,bentivogli2009fifth}: 2.5K
        \item SST-2 \citep{socher2013recursive}: 67.3K
    \end{itemize}
    \item \textbf{SuperGLUE}
    \begin{itemize}
        \item BoolQ \citep{clark-etal-2019-boolq}: 9.4K
    \end{itemize}
    \item \textbf{\citet{de2019episodic}}
    \begin{itemize}
        \item AG News \citep{gulli2017agnews} (115.0K)
        \item Amazon Reviews Full \citep{mcauley2013hidden} (115.0K)
        \item DBPedia \citep{lehmann2015dbpedia} (115.0K)
        \item Yahoo Answers \citep{zhang2015character} (115.0K)
        \item Yelp Reviews Full \citep{yelp2015dataset} (115.0K)
    \end{itemize}
\end{itemize}

Note that the RTE dataset from SuperGLUE also satisfies our task type and dataset size constraint, however according to \citet{wang2019superglue} it is an exact duplicate of GLUE's RTE dataset, therefore we don't include it. \citet{de2019episodic} performed some mild preprocessing and subsampling of the datasets in \citet{zhang2015character}, however they did not release the final data. We perform a similar preprocessing step: for datasets where each sample consists in several pieces of text (typically review title and body in the Amazon and Yelp datasets), we concatenate all into one utterance, joined with a period and a space (``\texttt{. }''). We subsample the datasets to 115K training samples, 5K validation samples and 7.6K test samples.

We randomly select three out of these 13 tasks to serve as the validation split upon which to perform hyper-parameter search, namely: \textbf{BoolQ}, \textbf{MRPC} and \textbf{SST-2}. Since test sets are not available for the GLUE and SuperGLUE benchmark, we compute the final scores on the validation datasets. Note that in all our experiments, validation data is not used during training, therefore there is no leakage of the ultimate evaluation dataset in the training procedure 

Instructions to download and code to preprocess the data will be made available at \codeurl{} to facilitate reproduction of our results and future work on continual learning for text classification. Following recent practice in text classification, we use a large model that has already been pre-trained in an unsupervised fashion (specifically  \texttt{bert-base-uncased}\footnote{We use the open-source implementation from \citet{Wolf2019HuggingFacesTS} with the default configuration.} from \citet{devlin2018bert}), and fine-tune the model on the supervised classification tasks. We train on each task for 1000 steps with batch size 16 (from 7 to $< 1$ epochs due to the large variance in dataset sizes).

\subsection{Grid-search Parameters}

For each all image-related tasks, we perform grid-search over the following parameter values:

\begin{itemize}
    \item Learning rate (all methods): 0.1, 0.03, 0.01
    \item EWC regularization strength (EWC, Co-natural EWC): 0.5, 1, 5
    \item Fisher damping coefficient (Co-natural finetuning, Co-natural EWC, Co-natural ER):  0, 1, 0.1
\end{itemize}

For text classification with BERT specifically, preliminary experiments showed that all methods benefitted from lower learning rate as well as lower regularization (likely due to the fact that optimization is starting from a pretrained model). We therefore use the following values:

\begin{itemize}
    \item Learning rate (all methods): 0.05, 0.01, 0.005
    \item EWC regularization strength (EWC, Co-natural EWC): 0.5, 0.1, 0.05
    \item Fisher damping coefficient (Co-natural finetuning, Co-natural EWC, Co-natural ER):  1, 10, 100
\end{itemize}

For ER, we simply set the replay batch size to the same value as standard training (10 and 32 for Split CIFAR and Omniglot respectively). Note that whenever applicable, we re-normalize the diagonal Fisher so that the sum of its weights is equal to the number of parameters in the model. This is so that the hyper-parameter choice is less dependent on the size of the model. In particular this means that the magnitude of each diagonal element is much bigger, which is why we do grid-search over smaller regularization parameters for EWC than is common in the literature.

\section{Continual Learning Results}

\subsection{The Co-natural Gradient Helps Prevent Forgetting}

The upper half of Tables \ref{tab:split_cifar_acc_and_f}, \ref{tab:omniglot_acc_and_f}, \ref{tab:miniimagenet_acc_and_f} and \ref{tab:text_classification_acc_and_f} reports the average accuracy of all the tasks at the end of training (higher is better). We observe that the co-natural gradient always improves greatly over simple finetuning, and occasionally over EWC and ER. We note that simple co-natural finetuning (without any other form of regularization) sometimes matches or exceeds the performance of EWC even though it requires strictly fewer resources (there is no need to store the previous parameters as in EWC, or data in ER).

Even more appreciable is the effect of the co-natural trajectories on forgetting, as shown in the lower half of Table \ref{tab:cl_average_acc_and_f}. As evidenced by the results in the lowest rows, using the co-natural gradient on top of finetuning and ER systematically results in large drops in forgetting across all datasets. With EWC, the conclusion is more nuanced. While the co-natural gradient never increases forgetting, the effect is not always statistically significant.

To get a qualitative assessment of the learning trajectories that yield such results, we visualize the accuracy curves of 10 out of the 47 evaluation tasks of Omniglot in Figure \ref{fig:viz}. We observe that previous approaches do poorly at keeping stable levels of performance over a long period of time (especially for tasks learned early in training), a problem that is largely alleviated by the co-natural preconditioning. This seems to come at the cost of more intransigence \citep{chaudhry2018riemannian}, \ie{} some of the later tasks are not being learnt properly. In models of fixed capacity, there is a natural trade-off between intransigence and forgetting (see also the ``stability-plasticity'' dilemma in neuroscience \citet{grossberg1982studies}). Our results position the co-natural gradient as a strong low-forgetting/moderate intransigence basis for future work.

\subsection{Sensitivity to Damping}
\label{sec:damp_anal}

The main hyper-parameter for the co-natural gradient is the damping parameter $\alpha$ from Eq. \ref{eqn:conatural_gradient}. In our previous experiments, the value of $\alpha$ is chosen according to grid search on a small number of tasks. While this is a realistic setting for practical applications, in this section we perform a smaller, targeted experiment to examine the effect of $\alpha$ on catastrophic forgetting.

We focus on the 17 evaluation tasks of the Split CIFAR dataset and set the learning rate to 0.1. We observe how much the model ``forgets'' about the first task it has observed after training on all others. Specifically, we train on the first task, compute its Fisher $F_1$, and then proceed to train on the 16 remaining tasks with the co-natural gradient using $F_1+\alpha$ (in particular we do not regularize for the other tasks). 
We observe how the value of alpha affects the final forgetting of the first task at the end of training, as well as the model's ability to learn new tasks (sometimes referred to as ``intransigence'' in the literature \citet{chaudhry2018riemannian}). As a measure of the latter, we report the maximum accuracy achieved by the model averaged over the 16 remaining tasks.

We evaluate values of $\alpha$ from $10^{-7}$ to $10^2$ (following a geometric progression), as well as the two extremal values $\alpha=0$ (``pure'' co-natural gradient\footnote{As mentioned in Section \ref{sec:approach}, we do actually add a small damping term $\varepsilon=10^{-12}$ to all experiments.}) and $\alpha=\infty$ (simple finetuning). All results are averaged over 10 random restarts (with different model initialization and task order).

\begin{figure}[!t]
\centering
\begin{subfigure}[t]{0.48\columnwidth}
\centering
\includegraphics[width=\columnwidth]{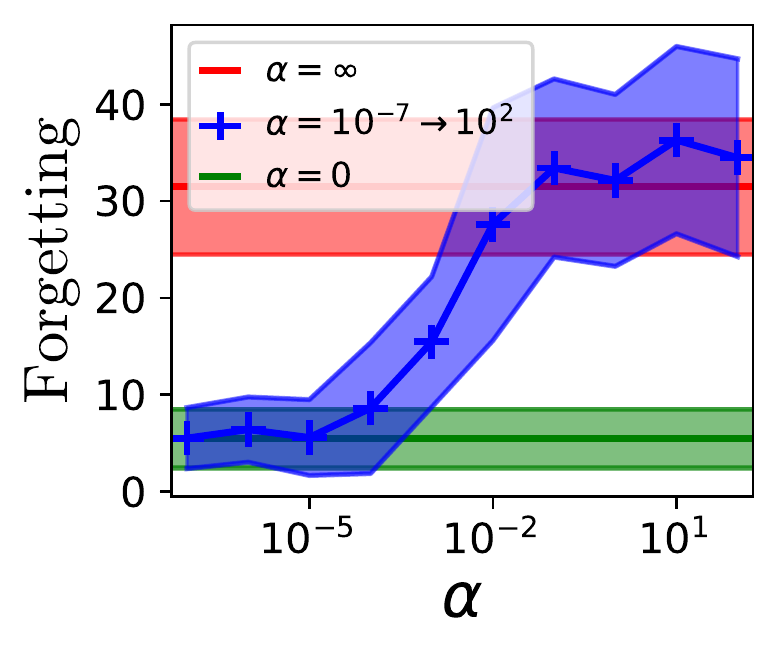}
\caption{\label{fig:damping_forgetting} Forgetting}
\end{subfigure}
~
\begin{subfigure}[t]{0.48\columnwidth}
\centering
\includegraphics[width=\columnwidth]{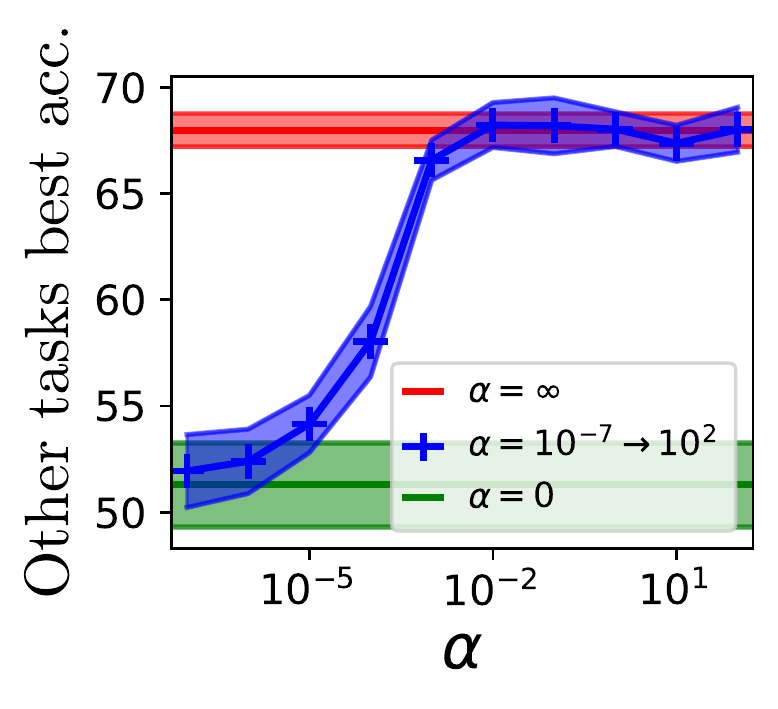}
\caption{\label{fig:damping_intransigence} Intransigence}
\end{subfigure}
\caption{\label{fig:damp_anal} Effect of damping on forgetting and intransigence. The horizontal axis (measuring $\alpha$) follows a logarithmic scale. The width of the bands on each side of the curves represent the standard deviation over the 10 re-runs.}
\end{figure}

We observe in Figure \ref{fig:damping_forgetting} that forgetting monotonically increases with the damping coefficient. Similarly, Figure \ref{fig:damping_intransigence} shows how increasing $\alpha$ results in lower intransigence. While these two observations are expected, it is interesting to note the existence of a ``sweet spot'' around $\alpha\in[10^{-5},10^{-4}]$ where damped co-natural gradient is significantly less intransigent than the un-damped co-natural gradient while simultaneously not being significantly less robust to forgetting (all hypotheses are tested for with $p<0.05$).

\section{Low-Resource Adaptation Experiments}
\label{sec:low_resource}

\begin{figure*}[t!]
\centering
\begin{subfigure}[t]{0.48\textwidth}
\centering
\includegraphics[width=\textwidth]{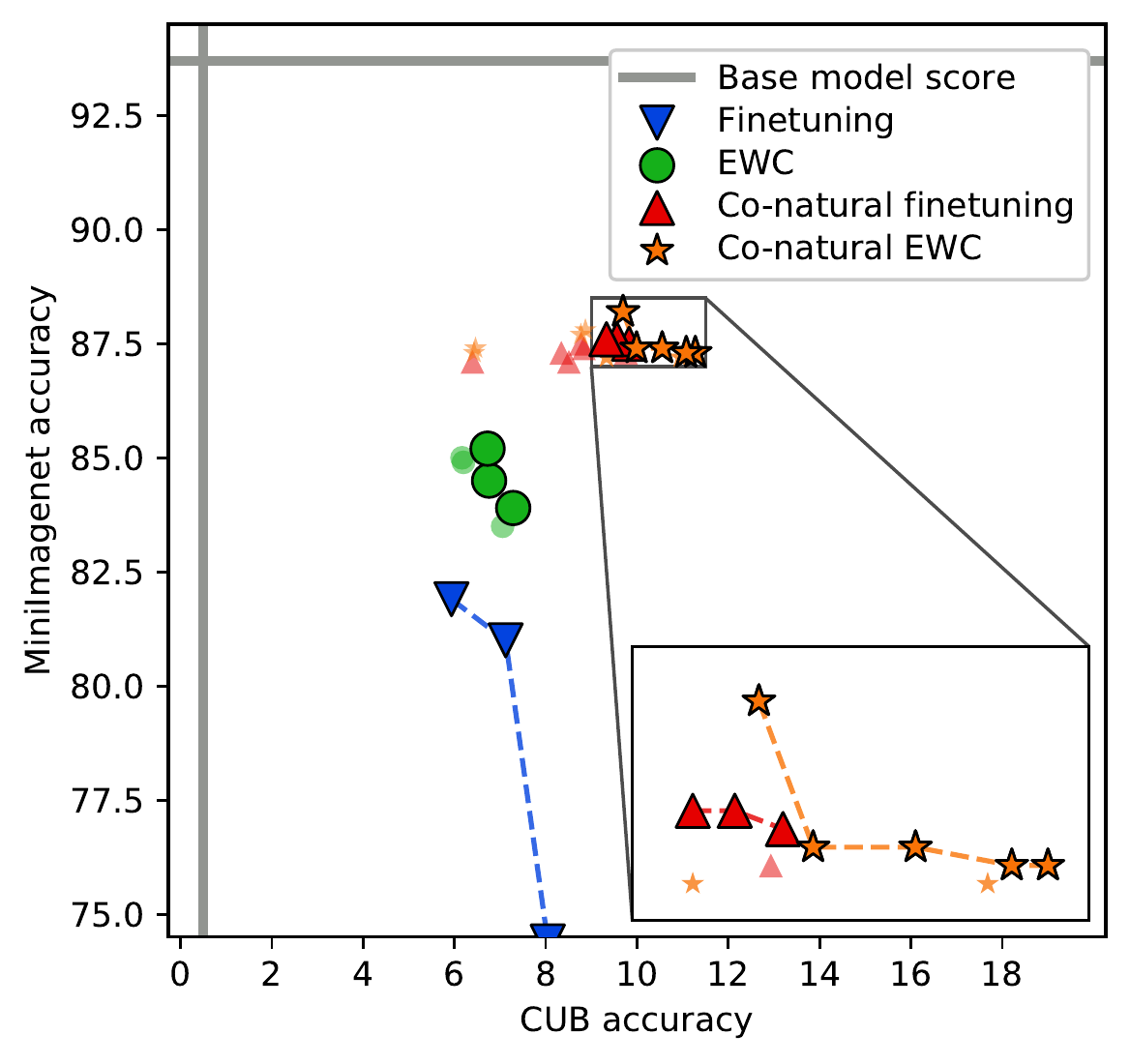}
\caption{\label{fig:imagenet_to_cub} MiniImageNet to CUB adaptation
}
\end{subfigure}
~
\begin{subfigure}[t]{0.48\textwidth}
\centering
\includegraphics[width=\columnwidth]{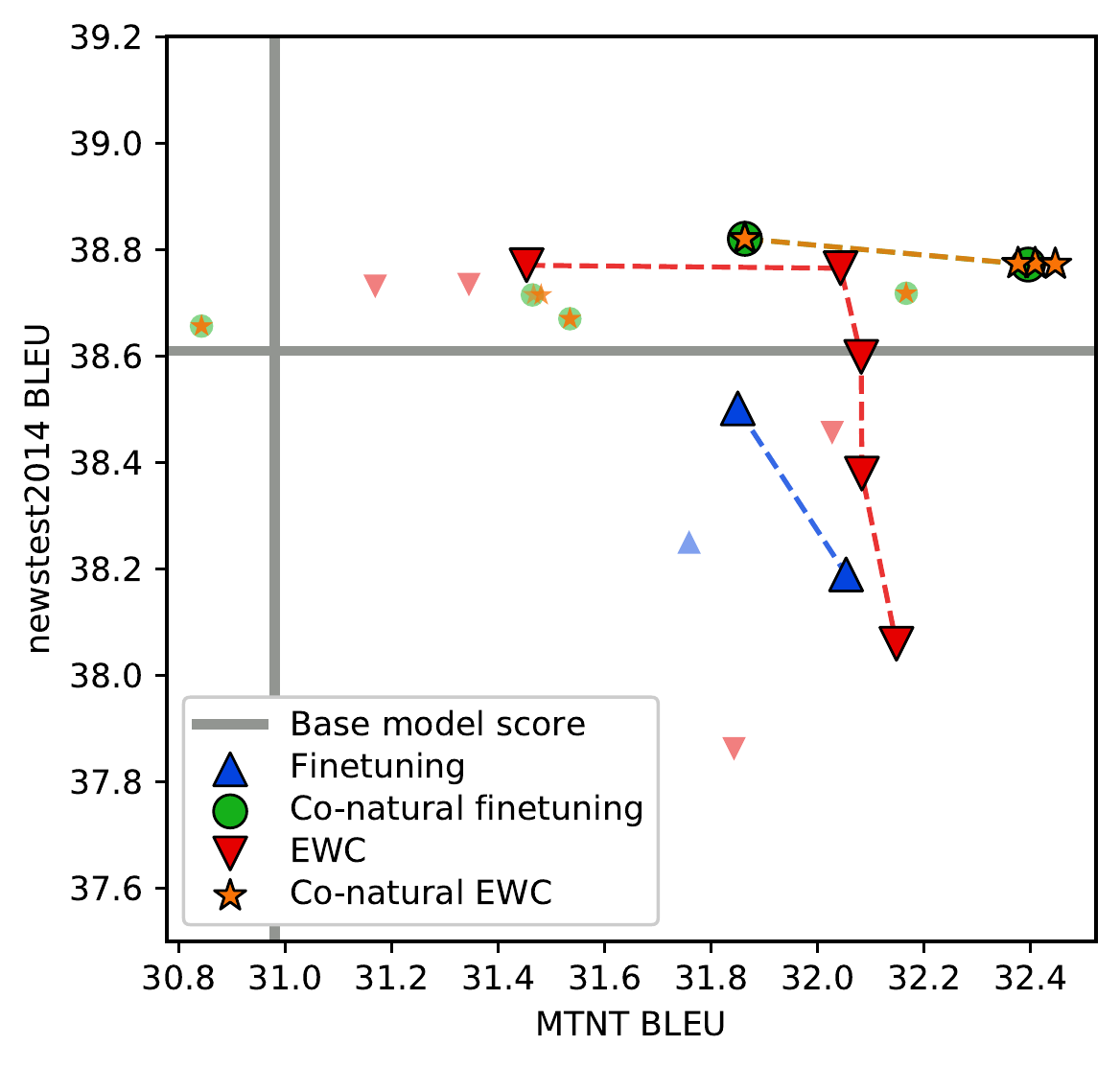}
\caption{\label{fig:wmt_to_mtnt} WMT to MTNT fine-tuning
}
\end{subfigure}
\caption{\label{fig:low_resource_adaptation} Low-resource adaptation results. The source (resp. target) task performance is represented on the vertical (resp. horizontal) axis. Pareto optimal configurations for each method are highlighted and the frontier is represented with dashed lines. The solid gray lines indicate the score of the original model trained on the source task.}
\end{figure*}

In this section we take a closer look at the specific case of adapting a model from a single task to another, when we only have access to a minimal amount of data in the target task. In this case, controlling the learning trajectory is particularly important because the model is being trained on an unreliable sample of the true distribution of the target task, and we have to rely on early-stopping to prevent overfitting. We show that using the co-natural gradient during adaptation helps both at preserving source task performance and reach higher overall target task performance.

\subsection{Experimental Setting}

We perform experiments on two different scenarios:

\paragraph{Image classification} We take MiniImagenet as a source task and CUB (a 200-way birds species classification dataset; \citet{welinder2009cub}) as a target task. To guarantee a strong base model despite the small size of MiniImageNet, we start off from a ResNet18 model \citep{he2016deep} pretrained on the full ImageNet, which we retrofit to MiniImageNet by replacing the last fully connected layer with a separate linear layer regressed over the MiniImageNet training data. To simulate a low-resource setting, we sub-sample the CUB training set to 200 images ($\approx 1$ per class). Scores for these tasks are reported in terms of accuracy.

\paragraph{Machine translation} We consider adaptation of an English to French model trained on WMT15 (a dataset of parallel sentences crawled from parliamentary proceedings, news commentary and web page crawls; \citet{bojar2015findings}) to MTNT (a dataset of Reddit comments; \citet{michel18mtnt}). Our model is a Transformer \citep{vaswani2017attention} pretrained on WMT15. Similarly to CUB, we simulate a low-resource setting by taking a sub-sample of 1000 sentence pairs as a training set. Scores for these two datasets are reported in terms of BLEU score \citep{papineni-EtAl:2002:ACL}.\footnote{We use sacrebleu \citep{post-2018-call} with \texttt{-tok intl} as recommended by \citet{michel18mtnt}.}

Here we do not allow any access to data in the source task when training on the target task. We compare four methods \textbf{Finetuning} (our baseline), \textbf{Co-natural finetuning}, \textbf{EWC} (which has been proven effective for domain adaptation, see \citet{thompson-etal-2019-overcoming}) and \textbf{Co-natural EWC}.

Given that different methods might lead to different trade-offs between source and target task performance, with some variation depending on the hyper-parameters (\eg{} learning rate, regularization strength\ldots), we take inspiration from \citet{thompson-etal-2019-overcoming} and graphically report results for all hyper-parameter configuration of each method on the 2 dimensional space defined by the score on source and target tasks\footnote{For CUB in particular we report the average accuracy of every configuration over 5 runs, each with a different 200-sized random subset of the data.}. Additionally, we highlight the Pareto frontier of each method \ie{} the set of configurations that are not strictly worse than any other configuration for the same model.

\subsection{Results}

The adaptation results for both scenarios are reported in Figure \ref{fig:low_resource_adaptation}. We find that in both cases, the co-natural gradient not only helps preserving the source task performance, but to some extent it also allows the model to reach better performance on the target task as well. We take this to corroborate our starting hypothesis: while introducing a regularizer does help, controlling the optimization dynamics actively helps counteract overfitting to the very small amount of training data, because the co-natural pre-conditioning makes it harder for stochastic gradient descent to push the model towards directions that would also hurt the source task.

\section{Conclusion}
\label{sec:conclusion}

We have presented the co-natural gradient, a technique that regularizes the optimization trajectory of models trained in a continual setting. We have shown that the co-natural gradient stands on its own as an efficient approach for overcoming catastrophic forgetting, and that it effectively complements and stabilizes other existing techniques at a minimal cost. 
We believe that the co-natural gradient --- and more generally, trajectory regularization --- can serve as a solid bedrock for building agents that learn without forgetting.
\clearpage

\chapter{Conclusion}
\label{ch:conclusion}

In this thesis, we have explored ways to mitigate the difficulty of learning neural models for natural language processing when confronted with changes in the data distribution. We presented an overview of the problem and the current state of the literature, and made contributions along three core axes: evaluation, robustness and adaptation.

Because benchmarks and evaluation metrics are our compass towards successfully addressing distributional shift, in Part \ref{sec:measuring} we first developed a new dataset (\ac{mtnt}, Chapter \ref{ch:mtnt}) for evaluating the ability of machine translation models (usually trained on news articles or parliamentary proceedings) to handle a shift to the social media domain. This confirmed that common neural architectures suffer significant performance drops when confronted with text data from social media sources. In Chapter \ref{ch:on_evaluation_of_adversarial}, we tackled the distinct, but equally important issue of evaluating adversarial perturbation on textual inputs, with a focus on sequence-to-sequence models. Specifically, we proposed an evaluation framework which we validated against human evaluations, and showed its benefits for the purpose of carrying out effective adversarial training strategies. Collecting datasets and designing evaluation metrics is a time-consuming, often thankless and ultimately sisyphean endeavour: new models inevitably ``overfit'' to the available benchmarks on which they are evaluated, which necessitates creating new datasets, new metrics \etc{}\ldots But as our work illustrates, every contribution helps: for example, our \ac{mtnt} dataset helped assess the robustness of modern \ac{mt} models, and was used as the basis for two subsequent robust machine translation shared task, helping to drive innovation in that area.

In Part \ref{sec:making_robust} we focused on the central problem of training models that are robust to distributional shifts. Our approach took inspiration from existing work in \acl{dro}, wherein models are trained to minimize their expected risk against the most difficult data distribution within a pre-determined family (the ``uncertainty set''). In our work we confronted the difficulty of defining this uncertainty set appropriately for practical applications. Our approach (called \acl{p-dro}) revolved around modeling the uncertainty set as a parametric family of generative models (Chapter \ref{ch:modeling_the_second_player_dro}) or likelihood ratios (Chapter \ref{ch:r_pdro}) to strike a better balance between the extreme pessimism or the limited expressivity of existing alternatives in the literature. In \ac{p-dro}, a robust classifier learns to minimize its expected loss under a parametric adversary which is itself trained to model the sub-population on which the classifier performs the worst. We empirically demonstrated that both variants could be used to derive models that were significantly more robust to sub-population shifts compared with relevant baselines. While the work presented in these two chapters is more recent, we hope that it can serve as a strong foundation towards more distributionally robust models in \ac{nlp}.

Finally, due to the inherent limitations of our \ac{dro} approach, which anticipates distributional shifts using only a static dataset, in Part \ref{sec:adaptation} we turned to the problem of \emph{adapting} existing models to new data distributions or tasks. We used ideas from information geometry to derive a gradient update rule (dubbed the ``co-natural gradient'') for mitigating the catastrophic forgetting effect, where adapting neural models to new tasks or domains causes significant drops in performance in previously learned tasks (Chapter \ref{ch:regularizing_trajectories}). The effectiveness of this method was validated on a diverse selection of continual-learning benchmarks, as well as a machine translation domain adaptation problem using our previously proposed \ac{mtnt} dataset as a test bed.

Beyond the conclusions directly derived from our contributions described above, we identify two higher level takeaways from our time pursuing research on the specific topic of distributional shift. The first is that \emph{proper evaluation is of paramount importance}. This idea has permeated every chapter of this thesis. Of course, this was the central motivation for the entirety of Part \ref{sec:measuring}, but we have also made a conscious effort to conduct fair and consistent evaluation in chapters \ref{ch:modeling_the_second_player_dro}, \ref{ch:r_pdro} and \ref{ch:regularizing_trajectories}. This is particularly important in the specialised sub-areas that are \acl{dro} and continual learning, because due to their cross-sectional nature, they tend to attract practitioners with diverse research backgrounds. This often results in inconsistent evaluation standards, which can can hamper progress \citep{gulrajani2020search,farquhar2018towards}.

Our second key takeaway is that, for practical applications of machine learning (such as \ac{nlp}), \emph{research progress should be guided by, but not limited to what is predicted by theory}. This is especially relevant in the context of deep learning, where theory lags behind the state of the art \citep{nagarajan2019uniform}\footnote{And of course, advancing theory is a worthwhile and crucial pursuit in and of itself.}. In our work on \acl{p-dro} in Part \ref{sec:making_robust}, we experimented with highly non-convex-concave differentiable games which were not guaranteed to even converge. Furthermore, we used a number of (sometimes extreme) approximations to make our methods useable in practical \ac{nlp} scenarios. Despite this, we found that they outperformed more principled, but perhaps less flexible baselines from the non-parametric \ac{dro} literature. This is not to say that research should proceed by blindly throwing ideas on the metaphorical wall and ``seeing what sticks'' (an inefficient, not to mention computationally unsustainable process). But there is a balance to strike, and we can achieve significant progress by taking inspiration from more conservative lines of work and porting them over to more applied research areas, without being restrained by mathematical rigour.

\subsection*{Future Work}

During our investigations throughout the course of producing the work described in this thesis, we have come to identify open questions which we believe would provide fertile ground for future research. In the next few paragraphs, we elaborate on these possible future directions towards close the gap between human and machine learning in \ac{nlp} in the face of distributional shift

In our own work, we have mostly been concerned with ``interpolative'' versions of distributional robustness, wherein we assume that our training data contains enough diversity for us to simulate potential distributional shifts. This setting has its uses, as demonstrated by our positive results in a variety of realistic benchmarks in Part \ref{sec:robustness}. But there is value in considering a more difficult scenario where we might have to extrapolate and guess at what unseen domains might look like. While there is some existing work in generating difficult or out-of-domain examples in computer vision \citep{yue2019domain,zhou2020deep}, this remains an under-explored area of the literature, especially when it comes to natural language processing. Leveraging the generative capabilities of strong parametric adversaries trained with \ac{p-dro} to hallucinate unseen, yet difficult examples could be a viable path to progress in this challenging setting.

With respect to the adaptation aspect, we believe that it is time for the community to move beyond the episodic setting commonly considered in continual learning and start tackling the ``streaming'' setting where the test (or training!) distribution is liable to change continuously over time. This scenario is more representative of how we as humans interact with our non-stationary environment, and it poses exciting new challenges, not only terms of the learning algorithms needed in such environments, but also with regards to how these same algorithms should be evaluated. How do ``episodic'' continual learning algorithm such as the co-natural gradient (Chapter \ref{ch:regularizing_trajectories}) fare against continuous distribution shift? And can the variety of data distributions observed over time in turn guide the construction of more accurate uncertainty sets in \ac{dro}? These are only some of the many open questions in this so far relatively uncharted research area. While there has been preliminary work on this general topic very recently \citep{lazaridou2021pitfalls,hombaiah2021dynamic}, there is still a long road ahead, as it were.

Finally, as large models pre-trained with a language modeling objective have become ever more prominent in \ac{nlp} research over the past few years \citep{radford2019language,dai2019transformer,raffel2019exploring}, the distinction between domain and task has become blurrier. For example, in \citet{radford2019language} and \citet{raffel2019exploring}, multiple tasks are reformulated as a single, unified language modeling or question answering task. We hypothesize that this trend is only going to increase because, as its name indicates, natural language is a natural communication protocol for us to interact with and query our models. With this in mind, we posit that creating models that are both robust and able to adapt to shift in the data distribution is a critical requirement for developing artificial agents that are capable of tackling a variety of tasks in diverse environments. We hope that the work presented in this thesis towards training models that are better equipped to deal with distributional shift provides a useful step in this direction.

\clearpage
\bibliography{bibliography/myplain,bibliography/references}

\begin{thebibliography}{265}
\providecommand{\natexlab}[1]{#1}
\providecommand{\url}[1]{\texttt{#1}}
\expandafter\ifx\csname urlstyle\endcsname\relax
  \providecommand{\doi}[1]{doi: #1}\else
  \providecommand{\doi}{doi: \begingroup \urlstyle{rm}\Url}\fi

\bibitem[Ahn et~al.(2019)Ahn, Lee, Cha, and Moon]{ahn2019uncertainty}
Ahn, H., Lee, D., Cha, S., and Moon, T.
\newblock Uncertainty-based continual learning with adaptive regularization.
\newblock In \emph{Proceedings of the 33rd Annual Conference on Neural
  Information Processing Systems (NeurIPS)}, 2019.

\bibitem[Aljundi et~al.(2019{\natexlab{a}})Aljundi, Caccia, Belilovsky, Caccia,
  Charlin, and Tuytelaars]{aljundi2019online}
Aljundi, R., Caccia, L., Belilovsky, E., Caccia, M., Charlin, L., and
  Tuytelaars, T.
\newblock Online continual learning with maximally interfered retrieval.
\newblock In \emph{Proceedings of the 33rd Annual Conference on Neural
  Information Processing Systems (NeurIPS)}, 2019{\natexlab{a}}.

\bibitem[Aljundi et~al.(2019{\natexlab{b}})Aljundi, Lin, Goujaud, and
  Bengio]{aljundi2019gradient}
Aljundi, R., Lin, M., Goujaud, B., and Bengio, Y.
\newblock Gradient based sample selection for online continual learning.
\newblock In \emph{Proceedings of the 33rd Annual Conference on Neural
  Information Processing Systems (NeurIPS)}, 2019{\natexlab{b}}.

\bibitem[Alzantot et~al.(2018)Alzantot, Sharma, Elgohary, Ho, Srivastava, and
  Chang]{alzantot-EtAl:2018:EMNLP}
Alzantot, M., Sharma, Y., Elgohary, A., Ho, B.-J., Srivastava, M., and Chang,
  K.-W.
\newblock Generating natural language adversarial examples.
\newblock In \emph{Proceedings of the 2018 Conference on Empirical Methods in
  Natural Language Processing}, pp.\  2890--2896, 2018.

\bibitem[Amari(1997)]{amari1997neural}
Amari, S.-i.
\newblock Neural learning in structured parameter spaces-natural riemannian
  gradient.
\newblock In \emph{Proceedings of the 9th Annual Conference on Neural
  Information Processing Systems (NIPS)}, pp.\  127--133, 1997.

\bibitem[Amodei et~al.(2016)Amodei, Ananthanarayanan, Anubhai, Bai, Battenberg,
  Case, Casper, Catanzaro, Cheng, Chen, Chen, Chen, Chen, Chrzanowski, Coates,
  Diamos, Ding, Du, Elsen, Engel, Fang, Fan, Fougner, Gao, Gong, Hannun, Han,
  Johannes, Jiang, Ju, Jun, LeGresley, Lin, Liu, Liu, Li, Li, Ma, Narang, Ng,
  Ozair, Peng, Prenger, Qian, Quan, Raiman, Rao, Satheesh, Seetapun, Sengupta,
  Srinet, Sriram, Tang, Tang, Wang, Wang, Wang, Wang, Wang, Wang, Wu, Wei,
  Xiao, Xie, Xie, Yogatama, Yuan, Zhan, and Zhu]{amodei2016deep}
Amodei, D., Ananthanarayanan, S., Anubhai, R., Bai, J., Battenberg, E., Case,
  C., Casper, J., Catanzaro, B., Cheng, Q., Chen, G., Chen, J., Chen, J., Chen,
  Z., Chrzanowski, M., Coates, A., Diamos, G., Ding, K., Du, N., Elsen, E.,
  Engel, J., Fang, W., Fan, L., Fougner, C., Gao, L., Gong, C., Hannun, A.,
  Han, T., Johannes, L., Jiang, B., Ju, C., Jun, B., LeGresley, P., Lin, L.,
  Liu, J., Liu, Y., Li, W., Li, X., Ma, D., Narang, S., Ng, A., Ozair, S.,
  Peng, Y., Prenger, R., Qian, S., Quan, Z., Raiman, J., Rao, V., Satheesh, S.,
  Seetapun, D., Sengupta, S., Srinet, K., Sriram, A., Tang, H., Tang, L., Wang,
  C., Wang, J., Wang, K., Wang, Y., Wang, Z., Wang, Z., Wu, S., Wei, L., Xiao,
  B., Xie, W., Xie, Y., Yogatama, D., Yuan, B., Zhan, J., and Zhu, Z.
\newblock Deep speech 2 : End-to-end speech recognition in english and
  mandarin.
\newblock In \emph{Proceedings of the 33rd International Conference on Machine
  Learning (ICML)}, pp.\  173--182, 2016.
\newblock URL \url{http://proceedings.mlr.press/v48/amodei16.html}.

\bibitem[Andreas et~al.(2015)Andreas, Rabinovich, Jordan, and
  Klein]{andreas2015accuracy}
Andreas, J., Rabinovich, M., Jordan, M.~I., and Klein, D.
\newblock On the accuracy of self-normalized log-linear models.
\newblock In \emph{Proceedings of the 29th Annual Conference on Neural
  Information Processing Systems (NIPS)}, pp.\  1783--1791, 2015.

\bibitem[Axelrod et~al.(2011)Axelrod, He, and Gao]{axelrod-he-gao:2011:EMNLP}
Axelrod, A., He, X., and Gao, J.
\newblock Domain adaptation via pseudo in-domain data selection.
\newblock In \emph{Proceedings of the 2011 Conference on Empirical Methods in
  Natural Language Processing (EMNLP)}, pp.\  355--362, 2011.

\bibitem[Bahdanau et~al.(2014)Bahdanau, Cho, and Bengio]{bahdanau2014neural}
Bahdanau, D., Cho, K., and Bengio, Y.
\newblock Neural machine translation by jointly learning to align and
  translate.
\newblock In \emph{International Conference on Learning Representations}, 2014.

\bibitem[Balduzzi et~al.(2018)Balduzzi, Racaniere, Martens, Foerster, Tuyls,
  and Graepel]{balduzzi2018mechanics}
Balduzzi, D., Racaniere, S., Martens, J., Foerster, J., Tuyls, K., and Graepel,
  T.
\newblock The mechanics of n-player differentiable games.
\newblock In \emph{Proceedings of the 35th International Conference on Machine
  Learning (ICML)}, pp.\  354--363, 2018.
\newblock URL
  \url{http://proceedings.mlr.press/v80/balduzzi18a/balduzzi18a.pdf}.

\bibitem[Baldwin et~al.(2013)Baldwin, Cook, Lui, MacKinlay, and
  Wang]{baldwin-EtAl:2013:IJCNLP}
Baldwin, T., Cook, P., Lui, M., MacKinlay, A., and Wang, L.
\newblock How noisy social media text, how diffrnt social media sources?
\newblock In \emph{Proceedings of the 6th International Joint Conference on
  Natural Language Processing (IJCNLP)}, pp.\  356--364, 2013.
\newblock URL \url{http://www.aclweb.org/anthology/I13-1041}.

\bibitem[Bapna \& Firat(2019)Bapna and Firat]{bapna2019simple}
Bapna, A. and Firat, O.
\newblock Simple, scalable adaptation for neural machine translation.
\newblock In \emph{Proceedings of the 2019 Conference on Empirical Methods in
  Natural Language Processing (EMNLP)}, 2019.

\bibitem[Barrault et~al.(2020)Barrault, Biesialska, Bojar, Costa-juss{\`a},
  Federmann, Graham, Grundkiewicz, Haddow, Huck, Joanis, Kocmi, Koehn, Lo,
  Ljube{\v{s}}i{\'c}, Monz, Morishita, Nagata, Nakazawa, Pal, Post, and
  Zampieri]{barrault2020findings}
Barrault, L., Biesialska, M., Bojar, O., Costa-juss{\`a}, M.~R., Federmann, C.,
  Graham, Y., Grundkiewicz, R., Haddow, B., Huck, M., Joanis, E., Kocmi, T.,
  Koehn, P., Lo, C.-k., Ljube{\v{s}}i{\'c}, N., Monz, C., Morishita, M.,
  Nagata, M., Nakazawa, T., Pal, S., Post, M., and Zampieri, M.
\newblock Findings of the 2020 conference on machine translation ({WMT}20).
\newblock In \emph{Proceedings of the 5th Conference on Machine Translation
  (WMT)}, 2020.

\bibitem[Belinkov \& Bisk(2018{\natexlab{a}})Belinkov and
  Bisk]{belinkov2017synthetic}
Belinkov, Y. and Bisk, Y.
\newblock Synthetic and natural noise both break neural machine translation.
\newblock \emph{International Conference on Learning Representations},
  2018{\natexlab{a}}.

\bibitem[Belinkov \& Bisk(2018{\natexlab{b}})Belinkov and
  Bisk]{belinkov2018synthetic}
Belinkov, Y. and Bisk, Y.
\newblock Synthetic and natural noise both break neural machine translation.
\newblock In \emph{Proceedings of the International Conference on Learning
  Representations (ICLR)}, 2018{\natexlab{b}}.

\bibitem[Ben-David et~al.(2007)Ben-David, Blitzer, Crammer, Pereira,
  et~al.]{ben2007analysis}
Ben-David, S., Blitzer, J., Crammer, K., Pereira, F., et~al.
\newblock Analysis of representations for domain adaptation.
\newblock \emph{Proceedings of the 20th Annual Conference on Neural Information
  Processing Systems (NIPS)}, 2007.

\bibitem[Ben-David et~al.(2010)Ben-David, Blitzer, Crammer, Kulesza, Pereira,
  and Vaughan]{ben2010theory}
Ben-David, S., Blitzer, J., Crammer, K., Kulesza, A., Pereira, F., and Vaughan,
  J.~W.
\newblock A theory of learning from different domains.
\newblock \emph{Machine learning}, 79\penalty0 (1-2):\penalty0 151--175, 2010.

\bibitem[Ben-Tal et~al.(2013)Ben-Tal, Den~Hertog, De~Waegenaere, Melenberg, and
  Rennen]{ben2013robust}
Ben-Tal, A., Den~Hertog, D., De~Waegenaere, A., Melenberg, B., and Rennen, G.
\newblock Robust solutions of optimization problems affected by uncertain
  probabilities.
\newblock \emph{Management Science}, 59\penalty0 (2):\penalty0 341--357, 2013.

\bibitem[Bentivogli et~al.(2009)Bentivogli, Clark, Dagan, and
  Giampiccolo]{bentivogli2009fifth}
Bentivogli, L., Clark, P., Dagan, I., and Giampiccolo, D.
\newblock The fifth pascal recognizing textual entailment challenge.
\newblock In \emph{TAC}, 2009.

\bibitem[B{\'e}rard et~al.(2019)B{\'e}rard, Calapodescu, and
  Roux]{berard2019naver}
B{\'e}rard, A., Calapodescu, I., and Roux, C.
\newblock Naver labs europe’s systems for the wmt19 machine translation
  robustness task.
\newblock In \emph{Proceedings of the Fourth Conference on Machine Translation
  (Volume 2: Shared Task Papers, Day 1)}, pp.\  526--532, 2019.

\bibitem[Bickel et~al.(2009)Bickel, Br{\"u}ckner, and
  Scheffer]{bickel2009discriminative}
Bickel, S., Br{\"u}ckner, M., and Scheffer, T.
\newblock Discriminative learning under covariate shift.
\newblock \emph{Journal of Machine Learning Research}, 10\penalty0
  (Sep):\penalty0 2137--2155, 2009.

\bibitem[Blank \& Koch(2013)Blank and Koch]{blank2013historical}
Blank, A. and Koch, P.
\newblock \emph{Historical semantics and cognition}, volume~13.
\newblock Walter de Gruyter, 2013.

\bibitem[Blitzer \& Pereira(2007)Blitzer and Pereira]{blitzer2007domain}
Blitzer, J. and Pereira, F.
\newblock Domain adaptation of natural language processing systems.
\newblock \emph{University of Pennsylvania}, pp.\  1--106, 2007.

\bibitem[Blodgett et~al.(2016)Blodgett, Green, and
  O{'}Connor]{blodgett2016demographic}
Blodgett, S.~L., Green, L., and O{'}Connor, B.
\newblock Demographic dialectal variation in social media: A case study of
  {A}frican-{A}merican {E}nglish.
\newblock In \emph{Proceedings of the 54th Annual Meeting of the Association
  for Computational Linguistics (ACL)}, pp.\  1119--1130, 2016.
\newblock URL \url{https://www.aclweb.org/anthology/D16-1120}.

\bibitem[Blodgett et~al.(2017)Blodgett, Wei, and O'Connor]{blodgett2017dataset}
Blodgett, S.~L., Wei, J., and O'Connor, B.
\newblock A dataset and classifier for recognizing social media english.
\newblock In \emph{Proceedings of the 3rd Workshop on Noisy User-generated
  Text}, pp.\  56--61, 2017.

\bibitem[Bloomfield(1933)]{bloomfield1933language}
Bloomfield, L.
\newblock \emph{Language}.
\newblock Allen \& Unwin, 1933.

\bibitem[Bojar et~al.(2015)Bojar, Chatterjee, Federmann, Haddow, Huck, Hokamp,
  Koehn, Logacheva, Monz, Negri, Post, Scarton, Specia, and
  Turchi]{bojar2015findings}
Bojar, O., Chatterjee, R., Federmann, C., Haddow, B., Huck, M., Hokamp, C.,
  Koehn, P., Logacheva, V., Monz, C., Negri, M., Post, M., Scarton, C., Specia,
  L., and Turchi, M.
\newblock Findings of the 2015 workshop on statistical machine translation.
\newblock In \emph{Proceedings of the 10th Workshop on Statistical Machine
  Translation (WMT)}, pp.\  1--46, 2015.

\bibitem[Bojarski et~al.(2016)Bojarski, Del~Testa, Dworakowski, Firner, Flepp,
  Goyal, Jackel, Monfort, Muller, Zhang, et~al.]{bojarski2016end}
Bojarski, M., Del~Testa, D., Dworakowski, D., Firner, B., Flepp, B., Goyal, P.,
  Jackel, L.~D., Monfort, M., Muller, U., Zhang, J., et~al.
\newblock End to end learning for self-driving cars.
\newblock \emph{arXiv preprint arXiv:1604.07316}, 2016.

\bibitem[Boucher et~al.(2021)Boucher, Shumailov, Anderson, and
  Papernot]{boucher2021bad}
Boucher, N., Shumailov, I., Anderson, R., and Papernot, N.
\newblock Bad characters: Imperceptible nlp attacks.
\newblock \emph{arXiv preprint arXiv:2106.09898}, 2021.

\bibitem[Cer et~al.(2017)Cer, Diab, Agirre, Lopez-Gazpio, and
  Specia]{cer-EtAl:2017:SemEval}
Cer, D., Diab, M., Agirre, E., Lopez-Gazpio, I., and Specia, L.
\newblock Semeval-2017 task 1: Semantic textual similarity multilingual and
  crosslingual focused evaluation.
\newblock In \emph{Proceedings of the 11th International Workshop on Semantic
  Evaluation (SemEval-2017)}, pp.\  1--14, 2017.

\bibitem[Cettolo et~al.(2012)Cettolo, Girardi, and
  Federico]{cettoloEtAl:EAMT2012}
Cettolo, M., Girardi, C., and Federico, M.
\newblock Wit$^3$: Web inventory of transcribed and translated talks.
\newblock In \emph{Proceedings of the 16$^{th}$ Conference of the European
  Association for Machine Translation (EAMT)}, pp.\  261--268, 2012.

\bibitem[Cettolo et~al.(2016)Cettolo, Niehues, St{\"u}ker, Bentivogli, Cattoni,
  and Federico]{Cettolo2016TheI2}
Cettolo, M., Niehues, J., St{\"u}ker, S., Bentivogli, L., Cattoni, R., and
  Federico, M.
\newblock The iwslt 2016 evaluation campaign.
\newblock In \emph{Proceedings of the 2016 International Workshop on Spoken
  Language Translation (IWSLT)}, 2016.

\bibitem[Chaudhry et~al.(2018{\natexlab{a}})Chaudhry, Dokania, Ajanthan, and
  Torr]{chaudhry2018riemannian}
Chaudhry, A., Dokania, P.~K., Ajanthan, T., and Torr, P.~H.
\newblock Riemannian walk for incremental learning: Understanding forgetting
  and intransigence.
\newblock In \emph{Proceedings of the 16th European Conference on Computer
  Vision (ECCV)}, pp.\  532--547, 2018{\natexlab{a}}.

\bibitem[Chaudhry et~al.(2018{\natexlab{b}})Chaudhry, Ranzato, Rohrbach, and
  Elhoseiny]{chaudhry2018efficient}
Chaudhry, A., Ranzato, M., Rohrbach, M., and Elhoseiny, M.
\newblock Efficient lifelong learning with a-gem.
\newblock \emph{Proceedings of the International Conference on Learning
  Representations (ICLR)}, 2018{\natexlab{b}}.

\bibitem[Chaudhry et~al.(2019)Chaudhry, Rohrbach, Elhoseiny, Ajanthan, Dokania,
  Torr, and Ranzato]{chaudhry2019continual}
Chaudhry, A., Rohrbach, M., Elhoseiny, M., Ajanthan, T., Dokania, P.~K., Torr,
  P.~H., and Ranzato, M.
\newblock Continual learning with tiny episodic memories.
\newblock \emph{arXiv preprint arXiv:1902.10486}, 2019.

\bibitem[Chen et~al.(2018)Chen, Diethe, and Lawrence]{chen2018facilitating}
Chen, Y., Diethe, T., and Lawrence, N.
\newblock Facilitating bayesian continual learning by natural gradients and
  stein gradients.
\newblock In \emph{Continual Learning Workshop@ NeurIPS}, 2018.

\bibitem[Cheng et~al.(2018{\natexlab{a}})Cheng, Yi, Zhang, Chen, and
  Hsieh]{cheng2018seq2sick}
Cheng, M., Yi, J., Zhang, H., Chen, P.-Y., and Hsieh, C.-J.
\newblock Seq2sick: Evaluating the robustness of sequence-to-sequence models
  with adversarial examples.
\newblock \emph{arXiv preprint arXiv:1803.01128}, 2018{\natexlab{a}}.

\bibitem[Cheng et~al.(2016)Cheng, Xu, He, He, Wu, Sun, and Liu]{P16-1185}
Cheng, Y., Xu, W., He, Z., He, W., Wu, H., Sun, M., and Liu, Y.
\newblock Semi-supervised learning for neural machine translation.
\newblock In \emph{Proceedings of the 54th Annual Meeting of the Association
  for Computational Linguistics (ACL)}, pp.\  1965--1974, 2016.

\bibitem[Cheng et~al.(2018{\natexlab{b}})Cheng, Tu, Meng, Zhai, and
  Liu]{cheng2018towards}
Cheng, Y., Tu, Z., Meng, F., Zhai, J., and Liu, Y.
\newblock Towards robust neural machine translation.
\newblock In \emph{Proceedings of the 56th Annual Meeting of the Association
  for Computational Linguistics (ACL)}, 2018{\natexlab{b}}.

\bibitem[Chu et~al.(2017)Chu, Dabre, and
  Kurohashi]{chu-dabre-kurohashi:2017:Short}
Chu, C., Dabre, R., and Kurohashi, S.
\newblock An empirical comparison of domain adaptation methods for neural
  machine translation.
\newblock In \emph{Proceedings of the 55th Annual Meeting of the Association
  for Computational Linguistics (ACL)}, pp.\  385--391, 2017.

\bibitem[Ciss{\'e} et~al.(2017)Ciss{\'e}, Bojanowski, Grave, Dauphin, and
  Usunier]{Ciss2017ParsevalNI}
Ciss{\'e}, M., Bojanowski, P., Grave, E., Dauphin, Y., and Usunier, N.
\newblock Parseval networks: Improving robustness to adversarial examples.
\newblock In \emph{International Conference on Machine Learning}, 2017.

\bibitem[Clark et~al.(2019)Clark, Lee, Chang, Kwiatkowski, Collins, and
  Toutanova]{clark-etal-2019-boolq}
Clark, C., Lee, K., Chang, M.-W., Kwiatkowski, T., Collins, M., and Toutanova,
  K.
\newblock {B}ool{Q}: Exploring the surprising difficulty of natural yes/no
  questions.
\newblock In \emph{Proceedings of the 2019 Conference of the North American
  Chapter of the Association for Computational Linguistics: Human Language
  Technologies (NAACL-HLT)}, pp.\  2924--2936, 2019.

\bibitem[Cortes et~al.(2010)Cortes, Mansour, and Mohri]{cortes2010learning}
Cortes, C., Mansour, Y., and Mohri, M.
\newblock Learning bounds for importance weighting.
\newblock In \emph{Proceedings of the 23rd Annual Conference on Neural
  Information Processing Systems (NIPS)}, volume~10, pp.\  442--450. Citeseer,
  2010.

\bibitem[Cover(1999)]{cover1999elements}
Cover, T.~M.
\newblock \emph{Elements of information theory}.
\newblock John Wiley \& Sons, 1999.

\bibitem[Crystal(2001)]{crystal:01}
Crystal, D.
\newblock \emph{Language and the Internet}.
\newblock Cambridge University Press, 2001.

\bibitem[Curi et~al.(2020)Curi, Levy, Jegelka, and Krause]{curi2020adaptive}
Curi, S., Levy, K.~Y., Jegelka, S., and Krause, A.
\newblock Adaptive sampling for stochastic risk-averse learning.
\newblock In Larochelle, H., Ranzato, M., Hadsell, R., Balcan, M.~F., and Lin,
  H. (eds.), \emph{Proceedings of the 34th Annual Conference on Neural
  Information Processing Systems (NeurIPS)}, volume~33, pp.\  1036--1047.
  Curran Associates, Inc., 2020.
\newblock URL
  \url{https://proceedings.neurips.cc/paper/2020/file/0b6ace9e8971cf36f1782aa982a708db-Paper.pdf}.

\bibitem[Dagan et~al.(2005)Dagan, Glickman, and Magnini]{dagan2005pascal}
Dagan, I., Glickman, O., and Magnini, B.
\newblock The pascal recognising textual entailment challenge.
\newblock In \emph{Machine Learning Challenges Workshop}, pp.\  177--190.
  Springer, 2005.

\bibitem[Dai et~al.(2019)Dai, Yang, Yang, Cohen, Carbonell, Le, and
  Salakhutdinov]{dai2019transformer}
Dai, Z., Yang, Z., Yang, Y., Cohen, W.~W., Carbonell, J., Le, Q.~V., and
  Salakhutdinov, R.
\newblock Transformer-xl: Attentive language models beyond a fixed-length
  context.
\newblock \emph{arXiv preprint arXiv:1901.02860}, 2019.

\bibitem[Danet \& Herring(2007)Danet and Herring]{danet:07}
Danet, B. and Herring, S.
\newblock \emph{The Multilingual Internet: Language, Culture, and Communication
  Online}.
\newblock Oxford University Press., New York, 2007.

\bibitem[Davidson et~al.(2017)Davidson, Warmsley, Macy, and
  Weber]{davidson2017automated}
Davidson, T., Warmsley, D., Macy, M., and Weber, I.
\newblock Automated hate speech detection and the problem of offensive
  language.
\newblock In \emph{Proceedings of the 11th International AAAI Conference on
  Weblogs and Social Media (ICWSM)}, 2017.

\bibitem[de~Masson~d'Autume et~al.(2019)de~Masson~d'Autume, Ruder, Kong, and
  Yogatama]{de2019episodic}
de~Masson~d'Autume, C., Ruder, S., Kong, L., and Yogatama, D.
\newblock Episodic memory in lifelong language learning.
\newblock In \emph{Advances in Neural Information Processing Systems}, pp.\
  13122--13131, 2019.

\bibitem[Delage \& Ye(2010)Delage and Ye]{delage2010distributionally}
Delage, E. and Ye, Y.
\newblock Distributionally robust optimization under moment uncertainty with
  application to data-driven problems.
\newblock \emph{Operations research}, 58\penalty0 (3):\penalty0 595--612, 2010.

\bibitem[Deng et~al.(2009)Deng, Dong, Socher, Li, Li, and
  Fei-Fei]{deng2009imagenet}
Deng, J., Dong, W., Socher, R., Li, L.-J., Li, K., and Fei-Fei, L.
\newblock Imagenet: A large-scale hierarchical image database.
\newblock In \emph{Proceedings of the 22nd IEEE Conference on Computer Vision
  and Pattern Recognition (CVPR)}, pp.\  248--255, 2009.

\bibitem[Denkowski \& Lavie(2014)Denkowski and
  Lavie]{denkowski:lavie:meteor-wmt:2014}
Denkowski, M. and Lavie, A.
\newblock Meteor universal: Language specific translation evaluation for any
  target language.
\newblock In \emph{Proceedings of the EACL 2014 Workshop on Statistical Machine
  Translation}, 2014.

\bibitem[Devlin et~al.(2018)Devlin, Chang, Lee, and Toutanova]{devlin2018bert}
Devlin, J., Chang, M.-W., Lee, K., and Toutanova, K.
\newblock Bert: Pre-training of deep bidirectional transformers for language
  understanding.
\newblock In \emph{Proceedings of the 2019 Conference of the North American
  Chapter of the Association for Computational Linguistics: Human Language
  Technologies (NAACL-HLT)}, 2018.

\bibitem[Dhar et~al.(2019)Dhar, Singh, Peng, Wu, and
  Chellappa]{dhar2019learning}
Dhar, P., Singh, R.~V., Peng, K.-C., Wu, Z., and Chellappa, R.
\newblock Learning without memorizing.
\newblock In \emph{Proceedings of the IEEE Conference on Computer Vision and
  Pattern Recognition}, pp.\  5138--5146, 2019.

\bibitem[Dixon et~al.(2018)Dixon, Li, Sorensen, Thain, and
  Vasserman]{dixon2018measuring}
Dixon, L., Li, J., Sorensen, J., Thain, N., and Vasserman, L.
\newblock Measuring and mitigating unintended bias in text classification.
\newblock In \emph{Proceedings of the 2018 AAAI/ACM Conference on AI, Ethics,
  and Society}, pp.\  67--73, 2018.

\bibitem[Dolan \& Brockett(2005)Dolan and Brockett]{brocket2005automatically}
Dolan, W.~B. and Brockett, C.
\newblock Automatically constructing a corpus of sentential paraphrases.
\newblock In \emph{Proceedings of the The 3rd International Workshop on
  Paraphrasing (IWP)}, 2005.
\newblock URL \url{http://aclweb.org/anthology/I05-5002}.

\bibitem[Duchi \& Namkoong(2018)Duchi and Namkoong]{duchi2018learning}
Duchi, J. and Namkoong, H.
\newblock Learning models with uniform performance via distributionally robust
  optimization.
\newblock \emph{arXiv preprint arXiv:1810.08750}, 2018.
\newblock URL \url{https://arxiv.org/pdf/1810.08750.pdf}.

\bibitem[Duchi et~al.(2016)Duchi, Glynn, and Namkoong]{duchi2016statistics}
Duchi, J., Glynn, P., and Namkoong, H.
\newblock Statistics of robust optimization: A generalized empirical likelihood
  approach.
\newblock \emph{arXiv preprint arXiv:1610.03425}, 2016.

\bibitem[Duchi et~al.(2020)Duchi, Hashimoto, and Namkoong]{DuHa20dis}
Duchi, J., Hashimoto, T., and Namkoong, H.
\newblock Distributionally robust losses for latent covariate mixtures.
\newblock \emph{arXiv preprint arXiv:2007.13982}, 2020.

\bibitem[Duchi et~al.(2019)Duchi, Hashimoto, and
  Namkoong]{duchi2019distributionally}
Duchi, J.~C., Hashimoto, T., and Namkoong, H.
\newblock Distributionally robust losses against mixture covariate shifts.
\newblock \emph{Under revision in Operations Research}, 2019.

\bibitem[Dudley(2018)]{dudley2018real}
Dudley, R.~M.
\newblock \emph{Real analysis and probability}.
\newblock CRC Press, 2018.

\bibitem[Dupa{\v{c}}ov{\'a}(1987)]{dupavcova1987minimax}
Dupa{\v{c}}ov{\'a}, J.
\newblock The minimax approach to stochastic programming and an illustrative
  application.
\newblock \emph{Stochastics: An International Journal of Probability and
  Stochastic Processes}, 20\penalty0 (1):\penalty0 73--88, 1987.

\bibitem[Dupoux(2018)]{dupoux2018cognitive}
Dupoux, E.
\newblock Cognitive science in the era of artificial intelligence: A roadmap
  for reverse-engineering the infant language-learner.
\newblock \emph{Cognition}, 173:\penalty0 43--59, 2018.

\bibitem[Dvijotham et~al.(2018)Dvijotham, Gowal, Stanforth, Arandjelovic,
  O'Donoghue, Uesato, and Kohli]{dvijotham2018training}
Dvijotham, K., Gowal, S., Stanforth, R., Arandjelovic, R., O'Donoghue, B.,
  Uesato, J., and Kohli, P.
\newblock Training verified learners with learned verifiers.
\newblock \emph{arXiv preprint arXiv:1805.10265}, 2018.

\bibitem[Ebrahimi et~al.(2018{\natexlab{a}})Ebrahimi, Lowd, and
  Dou]{Ebrahimi2018OnAE}
Ebrahimi, J., Lowd, D., and Dou, D.
\newblock On adversarial examples for character-level neural machine
  translation.
\newblock In \emph{COLING}, 2018{\natexlab{a}}.

\bibitem[Ebrahimi et~al.(2018{\natexlab{b}})Ebrahimi, Rao, Lowd, and
  Dou]{ebrahimi2018hotflip}
Ebrahimi, J., Rao, A., Lowd, D., and Dou, D.
\newblock Hotflip: White-box adversarial examples for text classification.
\newblock In \emph{Proceedings of the 56th Annual Meeting of the Association
  for Computational Linguistics (Volume 2: Short Papers)}, pp.\  31--36,
  2018{\natexlab{b}}.

\bibitem[Eisenstein(2013)]{eisenstein:2013:NAACL-HLT}
Eisenstein, J.
\newblock What to do about bad language on the internet.
\newblock In \emph{Proceedings of the 2013 Conference of the North American
  Chapter of the Association for Computational Linguistics: Human Language
  Technologies (NAACL-HLT)}, pp.\  359--369, 2013.
\newblock URL \url{http://www.aclweb.org/anthology/N13-1037}.

\bibitem[Elsahar \& Gall{\'e}(2019)Elsahar and Gall{\'e}]{elsahar2019annotate}
Elsahar, H. and Gall{\'e}, M.
\newblock To annotate or not? predicting performance drop under domain shift.
\newblock In \emph{Proceedings of the 2019 Conference on Empirical Methods in
  Natural Language Processing (EMNLP)}, pp.\  2163--2173, 2019.

\bibitem[Esfahani \& Kuhn(2018)Esfahani and Kuhn]{esfahani2018data}
Esfahani, P.~M. and Kuhn, D.
\newblock Data-driven distributionally robust optimization using the
  wasserstein metric: Performance guarantees and tractable reformulations.
\newblock \emph{Mathematical Programming}, 171\penalty0 (1):\penalty0 115--166,
  2018.

\bibitem[Fan et~al.(2017)Fan, Lyu, Ying, and Hu]{fan2017learning}
Fan, Y., Lyu, S., Ying, Y., and Hu, B.-G.
\newblock Learning with average top-k loss.
\newblock In \emph{Proceedings of the 31st Annual Conference on Neural
  Information Processing Systems (NIPS)}, pp.\  497--505, 2017.

\bibitem[Farquhar \& Gal(2018)Farquhar and Gal]{farquhar2018towards}
Farquhar, S. and Gal, Y.
\newblock Towards robust evaluations of continual learning.
\newblock \emph{arXiv preprint arXiv:1805.09733}, 2018.

\bibitem[Faury et~al.(2020)Faury, Tanielian, Dohmatob, Smirnova, and
  Vasile]{faury2020distributionally}
Faury, L., Tanielian, U., Dohmatob, E., Smirnova, E., and Vasile, F.
\newblock Distributionally robust counterfactual risk minimization.
\newblock In \emph{Proceedings of the 34th Meeting of the Association for
  Advancement of Artificial Intelligence (AAAI)}, volume~34, pp.\  3850--3857,
  2020.

\bibitem[Formiga \& Fonollosa(2012)Formiga and Fonollosa]{formiga2012dealing}
Formiga, L. and Fonollosa, J. A.~R.
\newblock Dealing with input noise in statistical machine translation.
\newblock In \emph{Proceedings of the 24th International Conference on
  Computational Linguistics (COLING)}, pp.\  319--328, 2012.
\newblock URL \url{http://www.aclweb.org/anthology/C12-2032}.

\bibitem[Fortuna \& Nunes(2018)Fortuna and Nunes]{fortuna2018survey}
Fortuna, P. and Nunes, S.
\newblock A survey on automatic detection of hate speech in text.
\newblock \emph{ACM Computing Surveys (CSUR)}, 51\penalty0 (4):\penalty0 1--30,
  2018.

\bibitem[Founta et~al.(2018)Founta, Djouvas, Chatzakou, Leontiadis, Blackburn,
  Stringhini, Vakali, Sirivianos, and Kourtellis]{founta2018large}
Founta, A.-M., Djouvas, C., Chatzakou, D., Leontiadis, I., Blackburn, J.,
  Stringhini, G., Vakali, A., Sirivianos, M., and Kourtellis, N.
\newblock Large scale crowdsourcing and characterization of twitter abusive
  behavior.
\newblock In \emph{Proceedings of the 12th International AAAI Conference on
  Weblogs and Social Media (ICWSM)}, 2018.

\bibitem[Ganin \& Lempitsky(2015)Ganin and Lempitsky]{ganin15unsupervised}
Ganin, Y. and Lempitsky, V.
\newblock Unsupervised domain adaptation by backpropagation.
\newblock In \emph{Proceedings of the 32nd International Conference on Machine
  Learning (ICML)}, pp.\  1180--1189. PMLR, 07--09 Jul 2015.
\newblock URL \url{http://proceedings.mlr.press/v37/ganin15.pdf}.

\bibitem[Gao \& Kleywegt(2016)Gao and Kleywegt]{gao2016distributionally}
Gao, R. and Kleywegt, A.~J.
\newblock Distributionally robust stochastic optimization with wasserstein
  distance.
\newblock \emph{arXiv preprint arXiv:1604.02199}, 2016.

\bibitem[Giampiccolo et~al.(2007)Giampiccolo, Magnini, Dagan, and
  Dolan]{giampiccolo2007third}
Giampiccolo, D., Magnini, B., Dagan, I., and Dolan, B.
\newblock The third pascal recognizing textual entailment challenge.
\newblock In \emph{Proceedings of the ACL-PASCAL workshop on textual entailment
  and paraphrasing}, pp.\  1--9. Association for Computational Linguistics,
  2007.

\bibitem[Glorot et~al.(2011)Glorot, Bordes, and Bengio]{glorot2011domain}
Glorot, X., Bordes, A., and Bengio, Y.
\newblock Domain adaptation for large-scale sentiment classification: A deep
  learning approach.
\newblock In \emph{Proceedings of the 28th International Conference on Machine
  Learning (ICML)}, 2011.

\bibitem[Goodfellow et~al.(2014{\natexlab{a}})Goodfellow, Pouget-Abadie, Mirza,
  Xu, Warde-Farley, Ozair, Courville, and Bengio]{goodfellow2014generative}
Goodfellow, I., Pouget-Abadie, J., Mirza, M., Xu, B., Warde-Farley, D., Ozair,
  S., Courville, A., and Bengio, Y.
\newblock Generative adversarial nets.
\newblock In \emph{Proceedings of the 28th Annual Conference on Neural
  Information Processing Systems (NIPS)}, pp.\  2672--2680, 2014{\natexlab{a}}.
\newblock URL
  \url{http://papers.nips.cc/paper/5423-generative-adversarial-nets.pdf}.

\bibitem[Goodfellow et~al.(2014{\natexlab{b}})Goodfellow, Shlens, and
  Szegedy]{Goodfellow2014ExplainingAH}
Goodfellow, I.~J., Shlens, J., and Szegedy, C.
\newblock Explaining and harnessing adversarial examples.
\newblock \emph{Proceedings of the International Conference on Learning
  Representations (ICLR)}, 2014{\natexlab{b}}.

\bibitem[Goto et~al.(2013)Goto, Chow, Lu, Sumita, and
  T'sou]{Goto2013OverviewOT}
Goto, I., Chow, K.-P., Lu, B., Sumita, E., and T'sou, B. K.-Y.
\newblock Overview of the patent machine translation task at the ntcir-10
  workshop.
\newblock In \emph{NII Test Collection for IR Systems}, 2013.

\bibitem[Goyal et~al.(2019)Goyal, Dyer, and
  Berg-Kirkpatrick]{goyal2019empirical}
Goyal, K., Dyer, C., and Berg-Kirkpatrick, T.
\newblock An empirical investigation of global and local normalization for
  recurrent neural sequence models using a continuous relaxation to beam
  search.
\newblock In \emph{Proceedings of the 2019 Conference of the North American
  Chapter of the Association for Computational Linguistics: Human Language
  Technologies (NAACL-HLT)}, pp.\  1724--1733, 2019.

\bibitem[Goyal et~al.(2017)Goyal, Doll{\'a}r, Girshick, Noordhuis, Wesolowski,
  Kyrola, Tulloch, Jia, and He]{goyal2017accurate}
Goyal, P., Doll{\'a}r, P., Girshick, R., Noordhuis, P., Wesolowski, L., Kyrola,
  A., Tulloch, A., Jia, Y., and He, K.
\newblock Accurate, large minibatch sgd: Training imagenet in 1 hour.
\newblock \emph{arXiv preprint arXiv:1706.02677}, 2017.

\bibitem[Greensmith et~al.(2004)Greensmith, Bartlett, and
  Baxter]{greensmith2004variance}
Greensmith, E., Bartlett, P.~L., and Baxter, J.
\newblock Variance reduction techniques for gradient estimates in reinforcement
  learning.
\newblock \emph{Journal of Machine Learning Research}, 5\penalty0
  (Nov):\penalty0 1471--1530, 2004.

\bibitem[Grossberg(1982)]{grossberg1982studies}
Grossberg, S.~T.
\newblock \emph{Studies of Mind and Brain: Neural Principles of Learning,
  Perception, Development, Cognition, and Motor Control}, volume~70.
\newblock Springer Science \& Business Media, 1982.

\bibitem[Grosse et~al.(2016)Grosse, Papernot, Manoharan, Backes, and
  McDaniel]{grosse2016adversarial}
Grosse, K., Papernot, N., Manoharan, P., Backes, M., and McDaniel, P.
\newblock Adversarial perturbations against deep neural networks for malware
  classification.
\newblock \emph{arXiv preprint arXiv:1606.04435}, 2016.

\bibitem[Grzega \& Schoener(2007)Grzega and Schoener]{grzega2007english}
Grzega, J. and Schoener, M.
\newblock English and general historical lexicology.
\newblock \emph{Eichst{\"a}tt-Ingolstadt: Katholische Universit{\"a}t}, 2007.

\bibitem[Gulli(2005)]{gulli2017agnews}
Gulli, A.
\newblock Ag's corpus of news articles.
\newblock
  \url{http://groups.di.unipi.it/~gulli/AG_corpus_of_news_articles.html}, 2005.

\bibitem[Gulrajani \& Lopez-Paz(2020)Gulrajani and
  Lopez-Paz]{gulrajani2020search}
Gulrajani, I. and Lopez-Paz, D.
\newblock In search of lost domain generalization.
\newblock In \emph{Proceedings of the International Conference on Learning
  Representations (ICLR)}, 2020.

\bibitem[Gururangan et~al.(2018)Gururangan, Swayamdipta, Levy, Schwartz,
  Bowman, and Smith]{gururangan2018annotation}
Gururangan, S., Swayamdipta, S., Levy, O., Schwartz, R., Bowman, S., and Smith,
  N.~A.
\newblock Annotation artifacts in natural language inference data.
\newblock In \emph{Proceedings of the 2018 Conference of the North American
  Chapter of the Association for Computational Linguistics: Human Language
  Technologies (NAACL-HLT)}, pp.\  107--112, 2018.

\bibitem[Gururangan et~al.(2020)Gururangan, Marasovi{\'c}, Swayamdipta, Lo,
  Beltagy, Downey, and Smith]{gururangan-etal-2020-dont}
Gururangan, S., Marasovi{\'c}, A., Swayamdipta, S., Lo, K., Beltagy, I.,
  Downey, D., and Smith, N.~A.
\newblock Don{'}t stop pretraining: Adapt language models to domains and tasks.
\newblock In \emph{Proceedings of the 58th Annual Meeting of the Association
  for Computational Linguistics}, pp.\  8342--8360, Online, July 2020.
  Association for Computational Linguistics.
\newblock \doi{10.18653/v1/2020.acl-main.740}.
\newblock URL \url{https://www.aclweb.org/anthology/2020.acl-main.740}.

\bibitem[Haim et~al.(2006)Haim, Dagan, Dolan, Ferro, Giampiccolo, Magnini, and
  Szpektor]{haim2006second}
Haim, R.~B., Dagan, I., Dolan, B., Ferro, L., Giampiccolo, D., Magnini, B., and
  Szpektor, I.
\newblock The second pascal recognising textual entailment challenge.
\newblock In \emph{Proceedings of the Second PASCAL Challenges Workshop on
  Recognising Textual Entailment}, 2006.

\bibitem[Hand(2006)]{hand2006classifier}
Hand, D.~J.
\newblock Classifier technology and the illusion of progress.
\newblock \emph{Statistical science}, 21\penalty0 (1):\penalty0 1--14, 2006.

\bibitem[Hashimoto et~al.(2018)Hashimoto, Srivastava, Namkoong, and
  Liang]{hashimoto2018fairness}
Hashimoto, T., Srivastava, M., Namkoong, H., and Liang, P.
\newblock Fairness without demographics in repeated loss minimization.
\newblock In \emph{Proceedings of the 35th International Conference on Machine
  Learning (ICML)}, pp.\  1929--1938. PMLR, 2018.

\bibitem[He et~al.(2016)He, Zhang, Ren, and Sun]{he2016deep}
He, K., Zhang, X., Ren, S., and Sun, J.
\newblock Deep residual learning for image recognition.
\newblock In \emph{Proceedings of the 29th IEEE Conference on Computer Vision
  and Pattern Recognition (CVPR)}, pp.\  770--778, 2016.

\bibitem[Heafield et~al.(2013)Heafield, Pouzyrevsky, Clark, and
  Koehn]{heafield2013estimate}
Heafield, K., Pouzyrevsky, I., Clark, J.~H., and Koehn, P.
\newblock Scalable modified {Kneser-Ney} language model estimation.
\newblock In \emph{Proceedings of the 51st Annual Meeting of the Association
  for Computational Linguistics (ACL)}, pp.\  690--696, 2013.
\newblock URL \url{https://kheafield.com/papers/edinburgh/estimate\_paper.pdf}.

\bibitem[Helcl et~al.(2019)Helcl, Libovick{\`y}, and Popel]{helcl2019cuni}
Helcl, J., Libovick{\`y}, J., and Popel, M.
\newblock Cuni system for the wmt19 robustness task.
\newblock In \emph{Proceedings of the 4th Conference on Machine Translation
  (WMT)}, 2019.

\bibitem[Hern(2017)]{hern2017facebook}
Hern, A.
\newblock Facebook translates 'good morning' into 'attack them', leading to
  arrest.
\newblock
  https://www.theguardian.com/technology/2017/oct/24/facebook-palestine-israel-translates-good-morning-attack-them-arrest,
  2017.
\newblock URL
  \url{https://www.theguardian.com/technology/2017/oct/24/facebook-palestine-israel-translates-good-morning-attack-them-arrest}.
\newblock Accessed: 2020-04-17.

\bibitem[Herring(2003)]{herring2003media}
Herring, S.~C.
\newblock Media and language change: Introduction.
\newblock \emph{Journal of Historical Pragmatics}, 4\penalty0 (1):\penalty0
  1--17, 2003.

\bibitem[Hershey \& Olsen(2007)Hershey and Olsen]{Hershey2007ApproximatingTK}
Hershey, J. and Olsen, P.
\newblock Approximating the kullback leibler divergence between gaussian
  mixture models.
\newblock \emph{Proceedings of the International Conference on Acoustics,
  Speech, and Signal Processing (ICASSP)}, 4:\penalty0 IV--317--IV--320, 2007.

\bibitem[Hochreiter \& Schmidhuber(1997)Hochreiter and
  Schmidhuber]{hochreiter1997long}
Hochreiter, S. and Schmidhuber, J.
\newblock Long short-term memory.
\newblock \emph{Neural computation}, 9\penalty0 (8):\penalty0 1735--1780, 1997.

\bibitem[Hombaiah et~al.(2021)Hombaiah, Chen, Zhang, Bendersky, and
  Najork]{hombaiah2021dynamic}
Hombaiah, S.~A., Chen, T., Zhang, M., Bendersky, M., and Najork, M.
\newblock Dynamic language models for continuously evolving content.
\newblock \emph{arXiv preprint arXiv:2106.06297}, 2021.

\bibitem[Hosseini et~al.(2017)Hosseini, Kannan, Zhang, and
  Poovendran]{hosseini2017deceiving}
Hosseini, H., Kannan, S., Zhang, B., and Poovendran, R.
\newblock Deceiving google's perspective api built for detecting toxic
  comments.
\newblock \emph{arXiv preprint arXiv:1702.08138}, 2017.

\bibitem[Hovy \& S{\o}gaard(2015)Hovy and S{\o}gaard]{hovy2015tagging}
Hovy, D. and S{\o}gaard, A.
\newblock Tagging performance correlates with author age.
\newblock In \emph{Proceedings of the 53rd Annual Meeting of the Association
  for Computational Linguistics (ACL)}, pp.\  483--488, 2015.
\newblock URL \url{https://www.aclweb.org/anthology/P15-2079}.

\bibitem[Hu et~al.(2018)Hu, Niu, Sato, and Sugiyama]{hu2018does}
Hu, W., Niu, G., Sato, I., and Sugiyama, M.
\newblock Does distributionally robust supervised learning give robust
  classifiers?
\newblock In \emph{Proceedings of the 35th International Conference on Machine
  Learning (ICML)}, pp.\  2029--2037, 2018.
\newblock URL \url{http://proceedings.mlr.press/v80/hu18a/hu18a.pdf}.

\bibitem[Hu \& Hong(2013)Hu and Hong]{hu2013kullback}
Hu, Z. and Hong, L.~J.
\newblock Kullback-leibler divergence constrained distributionally robust
  optimization.
\newblock \emph{Available at Optimization Online}, 2013.

\bibitem[Huang et~al.(2021)Huang, Li, Qu, and Pan]{huang2021robustness}
Huang, S., Li, Z., Qu, L., and Pan, L.
\newblock On robustness of neural semantic parsers.
\newblock In \emph{Proceedings of the 16th European Chapter of the Association
  for Computational Linguistics (EACL)}, pp.\  3333--3342, 2021.

\bibitem[Husain(2020)]{husain2020distributional}
Husain, H.
\newblock Distributional robustness with ipms and links to regularization and
  gans.
\newblock In \emph{Proceedings of the 34th Annual Conference on Neural
  Information Processing Systems (NeurIPS)}, 2020.

\bibitem[Isabelle et~al.(2017)Isabelle, Cherry, and
  Foster]{isabelle2017challenge}
Isabelle, P., Cherry, C., and Foster, G.
\newblock A challenge set approach to evaluating machine translation.
\newblock In \emph{Proceedings of the 2017 Conference on Empirical Methods in
  Natural Language Processing (EMNLP)}, pp.\  2486--2496, 2017.
\newblock URL \url{https://www.aclweb.org/anthology/D17-1263}.

\bibitem[Iyer et~al.(2017)Iyer, Dandekar, and Csernai]{iyer2017first}
Iyer, S., Dandekar, N., and Csernai, K.
\newblock First quora dataset release: Question pairs.
\newblock
  \url{https://www.quora.com/q/quoradata/First-Quora-Dataset-Release-Question-Pairs},
  2017.

\bibitem[Iyyer et~al.(2015)Iyyer, Manjunatha, Boyd-Graber, and
  Daum\'{e}~III]{iyyer-EtAl:2015:ACL-IJCNLP}
Iyyer, M., Manjunatha, V., Boyd-Graber, J., and Daum\'{e}~III, H.
\newblock Deep unordered composition rivals syntactic methods for text
  classification.
\newblock In \emph{Proceedings of the 53rd Annual Meeting of the Association
  for Computational Linguistics and the 7th International Joint Conference on
  Natural Language Processing (Volume 1: Long Papers)}, pp.\  1681--1691, 2015.

\bibitem[Iyyer et~al.(2018)Iyyer, Wieting, Gimpel, and
  Zettlemoyer]{iyyer-EtAl:2018:N18-1}
Iyyer, M., Wieting, J., Gimpel, K., and Zettlemoyer, L.
\newblock Adversarial example generation with syntactically controlled
  paraphrase networks.
\newblock In \emph{Proceedings of the 2018 Conference of the North American
  Chapter of the Association for Computational Linguistics: Human Language
  Technologies, Volume 1 (Long Papers)}, pp.\  1875--1885, 2018.

\bibitem[Jia \& Liang(2017)Jia and Liang]{jia-liang:2017:EMNLP2017}
Jia, R. and Liang, P.
\newblock Adversarial examples for evaluating reading comprehension systems.
\newblock In \emph{Proceedings of the 2017 Conference on Empirical Methods in
  Natural Language Processing}, pp.\  2021--2031, 2017.

\bibitem[Jia et~al.(2019)Jia, Raghunathan, G{\"o}ksel, and
  Liang]{jia2019certified}
Jia, R., Raghunathan, A., G{\"o}ksel, K., and Liang, P.
\newblock Certified robustness to adversarial word substitutions.
\newblock In \emph{Proceedings of the 2019 Conference of the North American
  Chapter of the Association for Computational Linguistics: Human Language
  Technologies (NAACL-HLT)}, 2019.

\bibitem[Kalchbrenner \& Blunsom(2013)Kalchbrenner and
  Blunsom]{kalchbrenner-blunsom:2013:EMNLP}
Kalchbrenner, N. and Blunsom, P.
\newblock Recurrent continuous translation models.
\newblock In \emph{Proceedings of the 2013 Conference on Empirical Methods in
  Natural Language Processing (EMNLP)}, pp.\  1700--1709, 2013.

\bibitem[Karpukhin et~al.(2019)Karpukhin, Levy, Eisenstein, and
  Ghazvininejad]{karpukhin2019training}
Karpukhin, V., Levy, O., Eisenstein, J., and Ghazvininejad, M.
\newblock Training on synthetic noise improves robustness to natural noise in
  machine translation.
\newblock In \emph{Proceedings of the 5th Workshop on Noisy User-generated Text
  (W-NUT)}, pp.\  42--47, 2019.

\bibitem[Kashyap et~al.(2021)Kashyap, Hazarika, Kan, and
  Zimmermann]{kashyap2021domain}
Kashyap, A.~R., Hazarika, D., Kan, M.-Y., and Zimmermann, R.
\newblock Domain divergences: A survey and empirical analysis.
\newblock In \emph{Proceedings of the 2019 Conference of the North American
  Chapter of the Association for Computational Linguistics: Human Language
  Technologies (NAACL-HLT)}, 2021.

\bibitem[Katz et~al.(2017)Katz, Barrett, Dill, Julian, and
  Kochenderfer]{katz2017reluplex}
Katz, G., Barrett, C., Dill, D.~L., Julian, K., and Kochenderfer, M.~J.
\newblock Reluplex: An efficient smt solver for verifying deep neural networks.
\newblock In \emph{International Conference on Computer Aided Verification},
  pp.\  97--117. Springer, 2017.

\bibitem[Kelly et~al.(1999)Kelly, Hand, and Adams]{kelly1999impact}
Kelly, M.~G., Hand, D.~J., and Adams, N.~M.
\newblock The impact of changing populations on classifier performance.
\newblock In \emph{Proceedings of the 5th ACM SIGKDD international conference
  on Knowledge discovery and data mining (KDD)}, 1999.

\bibitem[Khayrallah \& Koehn(2018)Khayrallah and Koehn]{khayrallah2018noise}
Khayrallah, H. and Koehn, P.
\newblock On the impact of various types of noise on neural machine
  translation.
\newblock In \emph{Proceedings of the 2nd Workshop on Neural Machine
  Translation and Generation (WNMT)}, pp.\  74--83, 2018.

\bibitem[Kifer et~al.(2004)Kifer, Ben-David, and Gehrke]{kifer2004detecting}
Kifer, D., Ben-David, S., and Gehrke, J.
\newblock Detecting change in data streams.
\newblock In \emph{Proceedings of the 30th International Conference on Very
  Large Data Bases}, 2004.

\bibitem[Kingma \& Ba(2014)Kingma and Ba]{Kingma2014Adam}
Kingma, D.~P. and Ba, J.
\newblock Adam: A method for stochastic optimization.
\newblock In \emph{Proceedings of the International Conference on Learning
  Representations (ICLR)}, 2014.

\bibitem[Kirkpatrick et~al.(2017)Kirkpatrick, Pascanu, Rabinowitz, Veness,
  Desjardins, Rusu, Milan, Quan, Ramalho, Grabska-Barwinska,
  et~al.]{kirkpatrick2017overcoming}
Kirkpatrick, J., Pascanu, R., Rabinowitz, N., Veness, J., Desjardins, G., Rusu,
  A.~A., Milan, K., Quan, J., Ramalho, T., Grabska-Barwinska, A., et~al.
\newblock Overcoming catastrophic forgetting in neural networks.
\newblock \emph{Proceedings of the National Academy of Sciences}, 114\penalty0
  (13):\penalty0 3521--3526, 2017.

\bibitem[Koehn et~al.(2007)Koehn, Hoang, Birch, Callison-Burch, Federico,
  Bertoldi, Cowan, Shen, Moran, Zens, Dyer, Bojar, Constantin, and
  Herbst]{koehn2007moses}
Koehn, P., Hoang, H., Birch, A., Callison-Burch, C., Federico, M., Bertoldi,
  N., Cowan, B., Shen, W., Moran, C., Zens, R., Dyer, C., Bojar, O.,
  Constantin, A., and Herbst, E.
\newblock Moses: Open source toolkit for statistical machine translation.
\newblock In \emph{Proceedings of the 45th Annual Meeting of the Association
  for Computational Linguistics (ACL)}, pp.\  177--180, 2007.

\bibitem[Koh et~al.(2020)Koh, Sagawa, Marklund, Xie, Zhang, Balsubramani, Hu,
  Yasunaga, Phillips, Gao, et~al.]{koh2020wilds}
Koh, P.~W., Sagawa, S., Marklund, H., Xie, S.~M., Zhang, M., Balsubramani, A.,
  Hu, W., Yasunaga, M., Phillips, R.~L., Gao, I., et~al.
\newblock Wilds: A benchmark of in-the-wild distribution shifts.
\newblock \emph{arXiv preprint arXiv:2012.07421}, 2020.

\bibitem[Kolter \& Wong(2017)Kolter and Wong]{Kolter2017ProvableDA}
Kolter, J.~Z. and Wong, E.
\newblock Provable defenses against adversarial examples via the convex outer
  adversarial polytope.
\newblock \emph{arXiv preprint arXiv:1711.00851}, 2017.

\bibitem[Kong et~al.(2014)Kong, Schneider, Swayamdipta, Bhatia, Dyer, and
  Smith]{kong2014dependency}
Kong, L., Schneider, N., Swayamdipta, S., Bhatia, A., Dyer, C., and Smith,
  N.~A.
\newblock A dependency parser for tweets.
\newblock In \emph{Proceedings of the 2014 Conference on Empirical Methods in
  Natural Language Processing (EMNLP)}, pp.\  1001--1012, 2014.
\newblock URL \url{http://www.aclweb.org/anthology/D14-1108}.

\bibitem[Kullback \& Leibler(1951)Kullback and
  Leibler]{kullback1951information}
Kullback, S. and Leibler, R.~A.
\newblock On information and sufficiency.
\newblock \emph{The annals of mathematical statistics}, 22\penalty0
  (1):\penalty0 79--86, 1951.

\bibitem[Kutuzov et~al.(2018)Kutuzov, {\O}vrelid, Szymanski, and
  Velldal]{kutuzov2018diachronic}
Kutuzov, A., {\O}vrelid, L., Szymanski, T., and Velldal, E.
\newblock Diachronic word embeddings and semantic shifts: a survey.
\newblock In \emph{Proceedings of the 27th International Conference on
  Computational Linguistics (COLING)}, 2018.

\bibitem[Lake et~al.(2015)Lake, Salakhutdinov, and Tenenbaum]{lake2015human}
Lake, B.~M., Salakhutdinov, R., and Tenenbaum, J.~B.
\newblock Human-level concept learning through probabilistic program induction.
\newblock \emph{Science}, 350:\penalty0 1332--1338, 2015.

\bibitem[Lample et~al.(2016)Lample, Ballesteros, Subramanian, Kawakami, and
  Dyer]{lample2016neural}
Lample, G., Ballesteros, M., Subramanian, S., Kawakami, K., and Dyer, C.
\newblock Neural architectures for named entity recognition.
\newblock In \emph{Proceedings of the 2016 Conference of the North American
  Chapter of the Association for Computational Linguistics: Human Language
  Technologies (NAACL-HLT)}, pp.\  260--270, 2016.

\bibitem[Lazaridou et~al.(2021)Lazaridou, Kuncoro, Gribovskaya, Agrawal, Liska,
  Terzi, Gimenez, d'Autume, Ruder, Yogatama, et~al.]{lazaridou2021pitfalls}
Lazaridou, A., Kuncoro, A., Gribovskaya, E., Agrawal, D., Liska, A., Terzi, T.,
  Gimenez, M., d'Autume, C. d.~M., Ruder, S., Yogatama, D., et~al.
\newblock Pitfalls of static language modelling.
\newblock \emph{arXiv preprint arXiv:2102.01951}, 2021.

\bibitem[Lehmann et~al.(2015)Lehmann, Isele, Jakob, Jentzsch, Kontokostas,
  Mendes, Hellmann, Morsey, Van~Kleef, Auer, et~al.]{lehmann2015dbpedia}
Lehmann, J., Isele, R., Jakob, M., Jentzsch, A., Kontokostas, D., Mendes,
  P.~N., Hellmann, S., Morsey, M., Van~Kleef, P., Auer, S., et~al.
\newblock Dbpedia--a large-scale, multilingual knowledge base extracted from
  wikipedia.
\newblock \emph{Semantic Web – Interoperability, Usability, Applicability},
  6\penalty0 (2):\penalty0 167--195, 2015.

\bibitem[Levy et~al.(2020)Levy, Carmon, Duchi, and Sidford]{levy2020large}
Levy, D., Carmon, Y., Duchi, J.~C., and Sidford, A.
\newblock Large-scale methods for distributionally robust optimization.
\newblock \emph{Proceedings of the 34th Annual Conference on Neural Information
  Processing Systems (NeurIPS)}, 33, 2020.

\bibitem[Li et~al.(2010)Li, Zhao, Zhang, and Zhou]{li-EtAl:2010:PAPERS3}
Li, M., Zhao, Y., Zhang, D., and Zhou, M.
\newblock Adaptive development data selection for log-linear model in
  statistical machine translation.
\newblock In \emph{Proceedings of the 23rd International Conference on
  Computational Linguistics (COLING)}, pp.\  662--670, 2010.

\bibitem[Li et~al.(2019)Li, Michel, Anastasopoulos, Belinkov, Durrani, Firat,
  Koehn, Neubig, Pino, and Sajjad]{li2019findings}
Li, X., Michel, P., Anastasopoulos, A., Belinkov, Y., Durrani, N., Firat, O.,
  Koehn, P., Neubig, G., Pino, J., and Sajjad, H.
\newblock Findings of the first shared task on machine translation robustness.
\newblock In \emph{Proceedings of the 4th Conference on Machine Translation
  (WMT)}, pp.\  91--102, 2019.
\newblock URL \url{https://www.aclweb.org/anthology/W19-5303}.

\bibitem[Li \& Hoiem(2016)Li and Hoiem]{li2017learning}
Li, Z. and Hoiem, D.
\newblock Learning without forgetting.
\newblock \emph{Proceedings of the 14th European Conference on Computer Vision
  (ECCV)}, 2016.

\bibitem[Lipton et~al.(2018)Lipton, Wang, and Smola]{lipton2018detecting}
Lipton, Z.~C., Wang, Y.-X., and Smola, A.
\newblock Detecting and correcting for label shift with black box predictors.
\newblock In \emph{Proceedings of the 35th International Conference on Machine
  Learning (ICML)}, 2018.

\bibitem[Liu et~al.(2015)Liu, Luo, Wang, and Tang]{liu2015faceattributes}
Liu, Z., Luo, P., Wang, X., and Tang, X.
\newblock Deep learning face attributes in the wild.
\newblock In \emph{Proceedings of the IEEE/CVF International Conference on
  Computer Vision (ICCV)}, December 2015.

\bibitem[Lopez-Paz \& Ranzato(2017)Lopez-Paz and Ranzato]{lopez2017gradient}
Lopez-Paz, D. and Ranzato, M.
\newblock Gradient episodic memory for continual learning.
\newblock In \emph{Proceedings of the 31st Annual Conference on Neural
  Information Processing Systems (NIPS)}, pp.\  6467--6476, 2017.

\bibitem[Lui \& Baldwin(2012)Lui and Baldwin]{lui2012langid}
Lui, M. and Baldwin, T.
\newblock langid.py: An off-the-shelf language identification tool.
\newblock In \emph{Proceedings of the 50th Annual Meeting of the Association
  for Computational Linguistics (ACL)}, pp.\  25--30, 2012.
\newblock URL \url{http://www.aclweb.org/anthology/P12-3005}.

\bibitem[Luong \& Manning(2015)Luong and Manning]{luong2015stanford}
Luong, M.-T. and Manning, C.~D.
\newblock Stanford neural machine translation systems for spoken language
  domains.
\newblock In \emph{Proceedings of the 2015 International Workshop on Spoken
  Language Translation (IWSLT)}, 2015.

\bibitem[Luong et~al.(2015)Luong, Pham, and
  Manning]{luong-pham-manning:2015:EMNLP}
Luong, T., Pham, H., and Manning, C.~D.
\newblock Effective approaches to attention-based neural machine translation.
\newblock In \emph{Proceedings of the 2015 Conference on Empirical Methods in
  Natural Language Processing}, pp.\  1412--1421, 2015.

\bibitem[Ma \& Cieri(2006)Ma and Cieri]{Ma2006CorpusSF}
Ma, X. and Cieri, C.
\newblock Corpus support for machine translation at ldc.
\newblock In \emph{Language Resources and Evaluation Conference}, 2006.

\bibitem[McAuley \& Leskovec(2013)McAuley and Leskovec]{mcauley2013hidden}
McAuley, J. and Leskovec, J.
\newblock Hidden factors and hidden topics: understanding rating dimensions
  with review text.
\newblock In \emph{Proceedings of the 7th ACM Conference on Recommender Systems
  (RecSys)}, pp.\  165--172, 2013.

\bibitem[McCloskey \& Cohen(1989)McCloskey and
  Cohen]{mccloskey1989catastrophic}
McCloskey, M. and Cohen, N.~J.
\newblock Catastrophic interference in connectionist networks: The sequential
  learning problem.
\newblock In \emph{Psychology of learning and motivation}, volume~24, pp.\
  109--165. Elsevier, 1989.

\bibitem[McCoy et~al.(2019)McCoy, Pavlick, and Linzen]{mccoy2019right}
McCoy, T., Pavlick, E., and Linzen, T.
\newblock Right for the wrong reasons: Diagnosing syntactic heuristics in
  natural language inference.
\newblock In \emph{Proceedings of the 57th Annual Meeting of the Association
  for Computational Linguistics (ACL)}, pp.\  3428--3448, 2019.

\bibitem[McCulloch(2020)]{mcculloch2020because}
McCulloch, G.
\newblock \emph{Because internet: Understanding the new rules of language}.
\newblock Riverhead Books, 2020.

\bibitem[McCusker et~al.(1981)McCusker, Gough, and Bias]{mccusker1981word}
McCusker, L.~X., Gough, P.~B., and Bias, R.~G.
\newblock Word recognition inside out and outside in.
\newblock \emph{Journal of Experimental Psychology: Human Perception and
  Performance}, 7\penalty0 (3):\penalty0 538, 1981.

\bibitem[Merity et~al.(2017)Merity, Xiong, Bradbury, and
  Socher]{merity2016pointer}
Merity, S., Xiong, C., Bradbury, J., and Socher, R.
\newblock Pointer sentinel mixture models.
\newblock \emph{Proceedings of the International Conference on Learning
  Representations (ICLR)}, 2017.

\bibitem[Merity et~al.(2018)Merity, Keskar, and Socher]{merity2017regularizing}
Merity, S., Keskar, N.~S., and Socher, R.
\newblock Regularizing and optimizing lstm language models.
\newblock \emph{Proceedings of the International Conference on Learning
  Representations (ICLR)}, 2018.

\bibitem[Miceli~Barone et~al.(2017)Miceli~Barone, Haddow, Germann, and
  Sennrich]{micelibarone-EtAl:2017:EMNLP2017}
Miceli~Barone, A.~V., Haddow, B., Germann, U., and Sennrich, R.
\newblock Regularization techniques for fine-tuning in neural machine
  translation.
\newblock In \emph{Proceedings of the 2017 Conference on Empirical Methods in
  Natural Language Processing (EMNLP)}, pp.\  1490--1495, 2017.

\bibitem[Michel \& Neubig(2018{\natexlab{a}})Michel and Neubig]{michel18mtnt}
Michel, P. and Neubig, G.
\newblock {MTNT}: A testbed for machine translation of noisy text.
\newblock In \emph{Proceedings of the 2018 Conference on Empirical Methods in
  Natural Language Processing (EMNLP)}, pp.\  543--553, 2018{\natexlab{a}}.

\bibitem[Michel \& Neubig(2018{\natexlab{b}})Michel and
  Neubig]{michel2018extreme}
Michel, P. and Neubig, G.
\newblock Extreme adaptation for personalized neural machine translation.
\newblock In \emph{Proceedings of the 56th Annual Meeting of the Association
  for Computational Linguistics (ACL)}, 2018{\natexlab{b}}.
\newblock URL \url{https://www.aclweb.org/anthology/P18-2050.pdf}.

\bibitem[Michel et~al.(2019{\natexlab{a}})Michel, Levy, and
  Neubig]{michel2019sixteen}
Michel, P., Levy, O., and Neubig, G.
\newblock Are sixteen heads really better than one?
\newblock In \emph{Proceedings of the 33rd Annual Conference on Neural
  Information Processing Systems (NeurIPS)}, 2019{\natexlab{a}}.

\bibitem[Michel et~al.(2019{\natexlab{b}})Michel, Li, Neubig, and
  Pino]{michel19evaluation}
Michel, P., Li, X., Neubig, G., and Pino, J.
\newblock On evaluation of adversarial perturbations for sequence-to-sequence
  models.
\newblock In \emph{Proceedings of the 2019 Conference of the North American
  Chapter of the Association for Computational Linguistics: Human Language
  Technologies (NAACL-HLT)}, pp.\  3103--3114, 2019{\natexlab{b}}.
\newblock URL \url{https://www.aclweb.org/anthology/N19-1314}.

\bibitem[Michel et~al.(2021)Michel, Hashimoto, and Neubig]{michel2021modeling}
Michel, P., Hashimoto, T., and Neubig, G.
\newblock Modeling the second player in distributionally robust optimization.
\newblock In \emph{Proceedings of the International Conference on Learning
  Representations (ICLR)}, 2021.

\bibitem[Mikolov et~al.(2010)Mikolov, Karafi{\'a}t, Burget, {\v{C}}ernock{\`y},
  and Khudanpur]{mikolov2010recurrent}
Mikolov, T., Karafi{\'a}t, M., Burget, L., {\v{C}}ernock{\`y}, J., and
  Khudanpur, S.
\newblock Recurrent neural network based language model.
\newblock In \emph{Proceedings of the 11th Annual Conference of the
  International Speech Communication Association (InterSpeech)}, 2010.

\bibitem[Mikolov et~al.(2013)Mikolov, Chen, Corrado, and
  Dean]{mikolov2013efficient}
Mikolov, T., Chen, K., Corrado, G., and Dean, J.
\newblock Efficient estimation of word representations in vector space.
\newblock In \emph{Proceedings of the International Conference on Learning
  Representations (ICLR)}, 2013.

\bibitem[Moore \& Lewis(2010)Moore and Lewis]{moore-lewis:2010:Short}
Moore, R.~C. and Lewis, W.
\newblock Intelligent selection of language model training data.
\newblock In \emph{Proceedings of the 48th Annual Meeting of the Association
  for Computational Linguistics (ACL)}, pp.\  220--224, 2010.

\bibitem[Moosavi-Dezfooli et~al.(2016)Moosavi-Dezfooli, Fawzi, and
  Frossard]{MoosaviDezfooli2016DeepFoolAS}
Moosavi-Dezfooli, S.-M., Fawzi, A., and Frossard, P.
\newblock Deepfool: A simple and accurate method to fool deep neural networks.
\newblock \emph{2016 IEEE Conference on Computer Vision and Pattern Recognition
  (CVPR)}, pp.\  2574--2582, 2016.

\bibitem[Nagarajan \& Kolter(2019)Nagarajan and Kolter]{nagarajan2019uniform}
Nagarajan, V. and Kolter, J.~Z.
\newblock Uniform convergence may be unable to explain generalization in deep
  learning.
\newblock \emph{Proceedings of the 33rd Annual Conference on Neural Information
  Processing Systems (NeurIPS)}, 2019.

\bibitem[Naik et~al.(2018)Naik, Ravichander, Sadeh, Rose, and
  Neubig]{naik-EtAl:2018:C18-1}
Naik, A., Ravichander, A., Sadeh, N., Rose, C., and Neubig, G.
\newblock Stress test evaluation for natural language inference.
\newblock In \emph{Proceedings of the 27th International Conference on
  Computational Linguistics}, pp.\  2340--2353, 2018.

\bibitem[Neubig(2011)]{neubig11kftt}
Neubig, G.
\newblock The {Kyoto} free translation task.
\newblock http://www.phontron.com/kftt, 2011.

\bibitem[Neubig et~al.(2011)Neubig, Nakata, and Mori]{neubig11aclshort}
Neubig, G., Nakata, Y., and Mori, S.
\newblock Pointwise prediction for robust, adaptable japanese morphological
  analysis.
\newblock In \emph{Proceedings of the 49th Annual Meeting of the Association
  for Computational Linguistics (ACL)}, pp.\  529--533, 2011.

\bibitem[Neubig et~al.(2017)Neubig, Dyer, Goldberg, Matthews, Ammar,
  Anastasopoulos, Ballesteros, Chiang, Clothiaux, Cohn,
  et~al.]{neubig2017dynet}
Neubig, G., Dyer, C., Goldberg, Y., Matthews, A., Ammar, W., Anastasopoulos,
  A., Ballesteros, M., Chiang, D., Clothiaux, D., Cohn, T., et~al.
\newblock Dynet: The dynamic neural network toolkit.
\newblock \emph{arXiv preprint arXiv:1701.03980}, 2017.

\bibitem[Neubig et~al.(2018)Neubig, Sperber, Wang, Felix, Matthews,
  Padmanabhan, Qi, Sachan, Arthur, Godard, Hewitt, Riad, and
  Wang]{neubig2018xnmt}
Neubig, G., Sperber, M., Wang, X., Felix, M., Matthews, A., Padmanabhan, S.,
  Qi, Y., Sachan, D.~S., Arthur, P., Godard, P., Hewitt, J., Riad, R., and
  Wang, L.
\newblock {XNMT}: The extensible neural machine translation toolkit.
\newblock In \emph{Conference of the Association for Machine Translation in the
  Americas (AMTA) Open Source Software Showcase}, 2018.

\bibitem[Nguyen et~al.(2017)Nguyen, Li, Bui, and Turner]{nguyen2017variational}
Nguyen, C.~V., Li, Y., Bui, T.~D., and Turner, R.~E.
\newblock Variational continual learning.
\newblock In \emph{Proceedings of the International Conference on Learning
  Representations (ICLR)}, 2017.

\bibitem[Nguyen et~al.(2020)Nguyen, Si, and Blanchet]{nguyen2020robust}
Nguyen, V.~A., Si, N., and Blanchet, J.
\newblock Robust bayesian classification using an optimistic score ratio.
\newblock In \emph{Proceedings of the 37th International Conference on Machine
  Learning (ICML)}, 2020.

\bibitem[Norouzi et~al.(2016)Norouzi, Bengio, Jaitly, Schuster, Wu, Schuurmans,
  et~al.]{norouzi2016reward}
Norouzi, M., Bengio, S., Jaitly, N., Schuster, M., Wu, Y., Schuurmans, D.,
  et~al.
\newblock Reward augmented maximum likelihood for neural structured prediction.
\newblock In \emph{Proceedings of the 30th Annual Conference on Neural
  Information Processing Systems (NIPS)}, volume~29, pp.\  1723--1731, 2016.

\bibitem[Oren et~al.(2019)Oren, Sagawa, Hashimoto, and
  Liang]{oren2019distributionally}
Oren, Y., Sagawa, S., Hashimoto, T., and Liang, P.
\newblock Distributionally robust language modeling.
\newblock In \emph{Proceedings of the 2019 Conference on Empirical Methods in
  Natural Language Processing (EMNLP)}, pp.\  4227--4237, 2019.
\newblock URL \url{https://www.aclweb.org/anthology/D19-1432}.

\bibitem[Osborne \& Rubinstein(1994)Osborne and Rubinstein]{osborne1994course}
Osborne, M.~J. and Rubinstein, A.
\newblock \emph{A Course in Game Theory}.
\newblock The MIT Press, 1994.
\newblock ISBN 0262150417.

\bibitem[Owoputi et~al.(2013)Owoputi, O{'}Connor, Dyer, Gimpel, Schneider, and
  Smith]{owoputi2013improved}
Owoputi, O., O{'}Connor, B., Dyer, C., Gimpel, K., Schneider, N., and Smith,
  N.~A.
\newblock Improved part-of-speech tagging for online conversational text with
  word clusters.
\newblock In \emph{Proceedings of the 2013 Conference of the North American
  Chapter of the Association for Computational Linguistics: Human Language
  Technologies (NAACL-HLT)}, pp.\  380--390, 2013.
\newblock URL \url{https://www.aclweb.org/anthology/N13-1039}.

\bibitem[Papernot et~al.(2016)Papernot, McDaniel, Swami, and
  Harang]{papernot2016crafting}
Papernot, N., McDaniel, P., Swami, A., and Harang, R.
\newblock Crafting adversarial input sequences for recurrent neural networks.
\newblock In \emph{Military Communications Conference, MILCOM 2016-2016 IEEE},
  pp.\  49--54. IEEE, 2016.

\bibitem[Papineni et~al.(2002)Papineni, Roukos, Ward, and
  Zhu]{papineni-EtAl:2002:ACL}
Papineni, K., Roukos, S., Ward, T., and Zhu, W.-J.
\newblock Bleu: a method for automatic evaluation of machine translation.
\newblock In \emph{Proceedings of the 40th Annual Meeting of the Association
  for Computational Linguistics (ACL)}, pp.\  311--318, 2002.

\bibitem[Parisi et~al.(2019)Parisi, Kemker, Part, Kanan, and
  Wermter]{parisi2019continual}
Parisi, G.~I., Kemker, R., Part, J.~L., Kanan, C., and Wermter, S.
\newblock Continual lifelong learning with neural networks: A review.
\newblock \emph{Neural Networks}, 2019.

\bibitem[Pascanu \& Bengio(2013)Pascanu and Bengio]{pascanu2013revisiting}
Pascanu, R. and Bengio, Y.
\newblock Revisiting natural gradient for deep networks.
\newblock In \emph{Proceedings of the International Conference on Learning
  Representations (ICLR)}, 2013.

\bibitem[Peterson(2011)]{peterson2011openmt12}
Peterson, K.
\newblock Openmt12 evaluation, 2011.

\bibitem[Popel \& Bojar(2018)Popel and Bojar]{popel2018training}
Popel, M. and Bojar, O.
\newblock Training tips for the transformer model.
\newblock \emph{arXiv preprint arXiv:1804.00247}, 2018.

\bibitem[Popovi\'{c}(2015)]{popovic:2015:WMT}
Popovi\'{c}, M.
\newblock chrf: character n-gram f-score for automatic mt evaluation.
\newblock In \emph{Proceedings of the Tenth Workshop on Statistical Machine
  Translation}, pp.\  392--395, 2015.

\bibitem[Popovi{\'{c}}(2016)]{popovic:2016:WMT}
Popovi{\'{c}}, M.
\newblock chrf deconstructed: beta parameters and n-gram weights.
\newblock In \emph{Proceedings of the First Conference on Machine Translation:
  Volume 2, Shared Task Papers}, pp.\  499--504, 2016.
\newblock \doi{10.18653/v1/W16-2341}.

\bibitem[Post(2018{\natexlab{a}})]{post-2018-call}
Post, M.
\newblock A call for clarity in reporting {BLEU} scores.
\newblock In \emph{Proceedings of the 3rd Conference on Machine Translation
  (WMT)}, pp.\  186--191, 2018{\natexlab{a}}.

\bibitem[Post(2018{\natexlab{b}})]{post2018call}
Post, M.
\newblock A call for clarity in reporting bleu scores.
\newblock \emph{arXiv preprint arXiv:1804.08771}, 2018{\natexlab{b}}.

\bibitem[Press \& Wolf(2017)Press and Wolf]{press-wolf:2017:EACLshort}
Press, O. and Wolf, L.
\newblock Using the output embedding to improve language models.
\newblock In \emph{Proceedings of the 15th European Chapter of the Association
  for Computational Linguistics (EACL)}, pp.\  157--163, 2017.

\bibitem[{Pryzant} et~al.(2017){Pryzant}, {Chung}, {Jurafsky}, and
  {Britz}]{pryzant2017jesc}
{Pryzant}, R., {Chung}, Y., {Jurafsky}, D., and {Britz}, D.
\newblock Jesc: Japanese-english subtitle corpus.
\newblock \emph{ArXiv e-prints}, 2017.

\bibitem[Qui{\~n}onero-Candela et~al.(2009)Qui{\~n}onero-Candela, Sugiyama,
  Lawrence, and Schwaighofer]{quinonero2009dataset}
Qui{\~n}onero-Candela, J., Sugiyama, M., Lawrence, N.~D., and Schwaighofer, A.
\newblock \emph{Dataset shift in machine learning}.
\newblock Mit Press, 2009.

\bibitem[Radford et~al.(2018)Radford, Narasimhan, Salimans, and
  Sutskever]{radford2018improving}
Radford, A., Narasimhan, K., Salimans, T., and Sutskever, I.
\newblock Improving language understanding with unsupervised learning.
\newblock Technical report, Technical report, OpenAI, 2018.

\bibitem[Radford et~al.(2019)Radford, Wu, Child, Luan, Amodei, and
  Sutskever]{radford2019language}
Radford, A., Wu, J., Child, R., Luan, D., Amodei, D., and Sutskever, I.
\newblock Language models are unsupervised multitask learners.
\newblock \emph{OpenAI Blog}, 1:\penalty0 8, 2019.

\bibitem[Raffel et~al.(2019)Raffel, Shazeer, Roberts, Lee, Narang, Matena,
  Zhou, Li, and Liu]{raffel2019exploring}
Raffel, C., Shazeer, N., Roberts, A., Lee, K., Narang, S., Matena, M., Zhou,
  Y., Li, W., and Liu, P.~J.
\newblock Exploring the limits of transfer learning with a unified text-to-text
  transformer.
\newblock \emph{arXiv preprint arXiv:1910.10683}, 2019.
\newblock URL \url{https://arxiv.org/pdf/1910.10683.pdf}.

\bibitem[Raghunathan et~al.(2018)Raghunathan, Steinhardt, and
  Liang]{raghunathan2018certified}
Raghunathan, A., Steinhardt, J., and Liang, P.
\newblock Certified defenses against adversarial examples.
\newblock In \emph{Proceedings of the International Conference on Learning
  Representations (ICLR)}, 2018.

\bibitem[Rahimian \& Mehrotra(2019)Rahimian and
  Mehrotra]{rahimian2019distributionally}
Rahimian, H. and Mehrotra, S.
\newblock Distributionally {R}obust {O}ptimization: A {R}eview.
\newblock \emph{arXiv preprint arXiv:1908.05659}, 2019.

\bibitem[Rahman(2012)]{rahman2012n}
Rahman, J.
\newblock The n word: Its history and use in the african american community.
\newblock \emph{Journal of English Linguistics}, 40\penalty0 (2):\penalty0
  137--171, 2012.

\bibitem[Rai et~al.(2010)Rai, Saha, Daum{\'e}~III, and
  Venkatasubramanian]{rai2010domain}
Rai, P., Saha, A., Daum{\'e}~III, H., and Venkatasubramanian, S.
\newblock Domain adaptation meets active learning.
\newblock In \emph{Proceedings of the NAACL HLT 2010 Workshop on Active
  Learning for Natural Language Processing}, 2010.

\bibitem[Rajpurkar et~al.(2016)Rajpurkar, Zhang, Lopyrev, and
  Liang]{rajpurkar-etal-2016-squad}
Rajpurkar, P., Zhang, J., Lopyrev, K., and Liang, P.
\newblock {SQ}u{AD}: 100,000+ questions for machine comprehension of text.
\newblock In \emph{Proceedings of the 2016 Conference on Empirical Methods in
  Natural Language Processing (EMNLP)}, pp.\  2383--2392, 2016.

\bibitem[Rajpurkar et~al.(2020)Rajpurkar, Jia, and Liang]{squadleaderboard2020}
Rajpurkar, P., Jia, R., and Liang, P.
\newblock Squad 2.0 benchmark leaderboard.
\newblock https://rajpurkar.github.io/SQuAD-explorer/, 2020.
\newblock URL \url{https://rajpurkar.github.io/SQuAD-explorer/}.
\newblock Accessed: 2020-04-16.

\bibitem[Ratcliff(1990)]{ratcliff1990connectionist}
Ratcliff, R.
\newblock Connectionist models of recognition memory: constraints imposed by
  learning and forgetting functions.
\newblock \emph{Psychological review}, 97\penalty0 (2):\penalty0 285, 1990.

\bibitem[Rebuffi et~al.(2017)Rebuffi, Kolesnikov, Sperl, and
  Lampert]{rebuffi2017icarl}
Rebuffi, S.-A., Kolesnikov, A., Sperl, G., and Lampert, C.~H.
\newblock icarl: Incremental classifier and representation learning.
\newblock In \emph{Proceedings of the 30th IEEE Conference on Computer Vision
  and Pattern Recognition (CVPR)}, pp.\  2001--2010, 2017.

\bibitem[Reddy \& Knight(2016)Reddy and Knight]{reddy-knight:2016:NLPandCSS}
Reddy, S. and Knight, K.
\newblock Obfuscating gender in social media writing.
\newblock In \emph{Proceedings of the First Workshop on NLP and Computational
  Social Science}. Association for Computational Linguistics, 2016.

\bibitem[Ribeiro et~al.(2018)Ribeiro, Singh, and
  Guestrin]{ribeiro-singh-guestrin:2018:Long}
Ribeiro, M.~T., Singh, S., and Guestrin, C.
\newblock Semantically equivalent adversarial rules for debugging nlp models.
\newblock In \emph{Proceedings of the 56th Annual Meeting of the Association
  for Computational Linguistics (Volume 1: Long Papers)}, pp.\  856--865, 2018.

\bibitem[Ritter et~al.(2011)Ritter, Clark, Etzioni, et~al.]{ritter2011named}
Ritter, A., Clark, S., Etzioni, O., et~al.
\newblock Named entity recognition in tweets: an experimental study.
\newblock In \emph{Proceedings of the 2011 Conference on Empirical Methods in
  Natural Language Processing (EMNLP)}, pp.\  1524--1534. Association for
  Computational Linguistics, 2011.

\bibitem[Rockafellar et~al.(2000)Rockafellar, Uryasev,
  et~al.]{rockafellar2000optimization}
Rockafellar, R.~T., Uryasev, S., et~al.
\newblock Optimization of conditional value-at-risk.
\newblock \emph{Journal of risk}, 2:\penalty0 21--42, 2000.

\bibitem[Rudolph \& Blei(2018)Rudolph and Blei]{rudolph2018dynamic}
Rudolph, M. and Blei, D.
\newblock Dynamic embeddings for language evolution.
\newblock In \emph{Proceedings of the 27th International Conference on World
  Wide Web (WWW)}, 2018.

\bibitem[Rusu et~al.(2016)Rusu, Rabinowitz, Desjardins, Soyer, Kirkpatrick,
  Kavukcuoglu, Pascanu, and Hadsell]{rusu2016progressive}
Rusu, A.~A., Rabinowitz, N.~C., Desjardins, G., Soyer, H., Kirkpatrick, J.,
  Kavukcuoglu, K., Pascanu, R., and Hadsell, R.
\newblock Progressive neural networks.
\newblock \emph{arXiv preprint arXiv:1606.04671}, 2016.

\bibitem[Sagawa et~al.(2020)Sagawa, Koh, Hashimoto, and
  Liang]{sagawa2019distributionally}
Sagawa, S., Koh, P.~W., Hashimoto, T.~B., and Liang, P.
\newblock Distributionally robust neural networks for group shifts: On the
  importance of regularization for worst-case generalization.
\newblock In \emph{Proceedings of the International Conference on Learning
  Representations (ICLR)}, 2020.
\newblock URL \url{https://arxiv.org/pdf/1911.08731.pdf}.

\bibitem[Sakaguchi et~al.(2017)Sakaguchi, Duh, Post, and
  Van~Durme]{sakaguchi2017robsut}
Sakaguchi, K., Duh, K., Post, M., and Van~Durme, B.
\newblock Robsut wrod reocginiton via semi-character recurrent neural network.
\newblock In \emph{Proceedings of the 31st Meeting of the Association for
  Advancement of Artificial Intelligence (AAAI)}, pp.\  3281--3287, 2017.

\bibitem[Saks(1937)]{saks1937theory}
Saks, S.
\newblock Theory of the integral monografie matematyczne.
\newblock \emph{Vol. VII., Warszawa-Lwow}, 1937.

\bibitem[Samanta \& Mehta(2017)Samanta and Mehta]{samanta2017towards}
Samanta, S. and Mehta, S.
\newblock Towards crafting text adversarial samples.
\newblock \emph{arXiv preprint arXiv:1707.02812}, 2017.

\bibitem[Sanh et~al.(2019)Sanh, Debut, Chaumond, and Wolf]{sanh2019distilbert}
Sanh, V., Debut, L., Chaumond, J., and Wolf, T.
\newblock Distilbert, a distilled version of bert: smaller, faster, cheaper and
  lighter.
\newblock \emph{arXiv preprint arXiv:1910.01108}, 2019.

\bibitem[Sap et~al.(2019)Sap, Card, Gabriel, Choi, and Smith]{sap2019risk}
Sap, M., Card, D., Gabriel, S., Choi, Y., and Smith, N.~A.
\newblock The risk of racial bias in hate speech detection.
\newblock In \emph{Proceedings of the 57th Annual Meeting of the Association
  for Computational Linguistics (ACL)}, pp.\  1668--1678, 2019.
\newblock URL \url{https://www.aclweb.org/anthology/P19-1163}.

\bibitem[Scarf(1957)]{scarf1957minmax}
Scarf, H.~E.
\newblock \emph{A Min-Max Solution of an Inventory Problem}.
\newblock RAND Corporation, Santa Monica, CA, 1957.

\bibitem[Schmidt \& Wiegand(2017)Schmidt and Wiegand]{schmidt2017survey}
Schmidt, A. and Wiegand, M.
\newblock A survey on hate speech detection using natural language processing.
\newblock In \emph{Proceedings of the 5th International Workshop on Natural
  Language Processing for Social Media (SocialNLP)}, pp.\  1--10, 2017.
\newblock URL \url{https://www.aclweb.org/anthology/W17-1101}.

\bibitem[Schwarz et~al.(2018)Schwarz, Czarnecki, Luketina, Grabska-Barwinska,
  Teh, Pascanu, and Hadsell]{schwarz2018progress}
Schwarz, J., Czarnecki, W., Luketina, J., Grabska-Barwinska, A., Teh, Y.~W.,
  Pascanu, R., and Hadsell, R.
\newblock Progress \& compress: A scalable framework for continual learning.
\newblock In \emph{Proceedings of the 35th International Conference on Machine
  Learning (ICML)}, pp.\  4535--4544, 2018.

\bibitem[Sennrich(2017)]{sennrich2017how}
Sennrich, R.
\newblock How grammatical is character-level neural machine translation?
  assessing mt quality with contrastive translation pairs.
\newblock In \emph{Proceedings of the 15th European Chapter of the Association
  for Computational Linguistics (EACL)}, pp.\  376--382, 2017.
\newblock URL \url{http://www.aclweb.org/anthology/E17-2060}.

\bibitem[Sennrich et~al.(2016)Sennrich, Haddow, and Birch]{sennrich2016neural}
Sennrich, R., Haddow, B., and Birch, A.
\newblock Neural machine translation of rare words with subword units.
\newblock In \emph{Proceedings of the 54th Annual Meeting of the Association
  for Computational Linguistics (ACL)}, pp.\  1715--1725, 2016.
\newblock URL \url{http://www.aclweb.org/anthology/P16-1162}.

\bibitem[Shi et~al.(2020)Shi, Zhang, Chang, Huang, and
  Hsieh]{shi2020robustness}
Shi, Z., Zhang, H., Chang, K.-W., Huang, M., and Hsieh, C.-J.
\newblock Robustness verification for transformers.
\newblock In \emph{Proceedings of the International Conference on Learning
  Representations (ICLR)}, 2020.

\bibitem[Shimodaira(2000)]{shimodaira2000improving}
Shimodaira, H.
\newblock Improving predictive inference under covariate shift by weighting the
  log-likelihood function.
\newblock \emph{Journal of statistical planning and inference}, 90\penalty0
  (2):\penalty0 227--244, 2000.

\bibitem[Singh et~al.(2000)Singh, Kearns, and Mansour]{singh2000nash}
Singh, S.~P., Kearns, M.~J., and Mansour, Y.
\newblock Nash convergence of gradient dynamics in general-sum games.
\newblock In \emph{Proceedings of the 16th Conference on Uncertainty in
  Artificial Intelligence}, UAI '00, pp.\  541–548, San Francisco, CA, USA,
  2000. Morgan Kaufmann Publishers Inc.
\newblock ISBN 1558607099.

\bibitem[Sinha et~al.(2018)Sinha, Namkoong, and Duchi]{sinha2017certifying}
Sinha, A., Namkoong, H., and Duchi, J.
\newblock Certifying some distributional robustness with principled adversarial
  training.
\newblock In \emph{Proceedings of the International Conference on Learning
  Representations (ICLR)}, 2018.

\bibitem[Socher et~al.(2013)Socher, Perelygin, Wu, Chuang, Manning, Ng, and
  Potts]{socher2013recursive}
Socher, R., Perelygin, A., Wu, J., Chuang, J., Manning, C.~D., Ng, A., and
  Potts, C.
\newblock Recursive deep models for semantic compositionality over a sentiment
  treebank.
\newblock In \emph{Proceedings of the 2013 Conference on Empirical Methods in
  Natural Language Processing (EMNLP)}, pp.\  1631--1642, 2013.

\bibitem[Specia et~al.(2020)Specia, Li, Pino, Chaudhary, Guzm{\'a}n, Neubig,
  Durrani, Belinkov, Koehn, Sajjad, et~al.]{specia2020findings}
Specia, L., Li, Z., Pino, J., Chaudhary, V., Guzm{\'a}n, F., Neubig, G.,
  Durrani, N., Belinkov, Y., Koehn, P., Sajjad, H., et~al.
\newblock Findings of the wmt 2020 shared task on machine translation
  robustness.
\newblock In \emph{Proceedings of the 5th Conference on Machine Translation
  (WMT)}, 2020.

\bibitem[Sperber et~al.(2017)Sperber, Niehues, and Waibel]{sperber2017toward}
Sperber, M., Niehues, J., and Waibel, A.
\newblock Toward robust neural machine translation for noisy input sequences.
\newblock In \emph{Proceedings of the 2017 International Workshop on Spoken
  Language Translation (IWSLT)}, pp.\  90--96, 2017.
\newblock URL
  \url{http://workshop2017.iwslt.org/downloads/iwslt2017_proceeding_v2.pdf}.

\bibitem[Srivastava et~al.(2014)Srivastava, Hinton, Krizhevsky, Sutskever, and
  Salakhutdinov]{srivastava2014dropout}
Srivastava, N., Hinton, G.~E., Krizhevsky, A., Sutskever, I., and
  Salakhutdinov, R.
\newblock Dropout: a simple way to prevent neural networks from overfitting.
\newblock \emph{Journal of Machine Learning Research}, 15\penalty0
  (1):\penalty0 1929--1958, 2014.

\bibitem[Storkey(2009)]{storkey2009training}
Storkey, A.
\newblock When training and test sets are different: characterizing learning
  transfer.
\newblock \emph{Dataset shift in machine learning}, pp.\  3--28, 2009.

\bibitem[Sugiyama \& M{\"u}ller(2005)Sugiyama and
  M{\"u}ller]{sugiyama2005input}
Sugiyama, M. and M{\"u}ller, K.-R.
\newblock Input-dependent estimation of generalization error under covariate
  shift.
\newblock \emph{Statistics \& Decisions}, 23\penalty0 (4/2005):\penalty0
  249--279, 2005.

\bibitem[Sundermeyer et~al.(2012)Sundermeyer, Schl{\"u}ter, and
  Ney]{sundermeyer2012lstm}
Sundermeyer, M., Schl{\"u}ter, R., and Ney, H.
\newblock Lstm neural networks for language modeling.
\newblock In \emph{Proceedings of the 13th Annual Conference of the
  International Speech Communication Association (InterSpeech)}, 2012.

\bibitem[Sutskever et~al.(2014)Sutskever, Vinyals, and
  Le]{sutskever2014sequence}
Sutskever, I., Vinyals, O., and Le, Q.~V.
\newblock Sequence to sequence learning with neural networks.
\newblock In \emph{Proceedings of the 28th Annual Conference on Neural
  Information Processing Systems (NIPS)}, pp.\  3104--3112, 2014.

\bibitem[Szegedy et~al.(2013)Szegedy, Zaremba, Sutskever, Bruna, Erhan,
  Goodfellow, and Fergus]{Szegedy2013IntriguingPO}
Szegedy, C., Zaremba, W., Sutskever, I., Bruna, J., Erhan, D., Goodfellow,
  I.~J., and Fergus, R.
\newblock Intriguing properties of neural networks.
\newblock \emph{arXiv preprint arXiv:1312.6199}, 2013.

\bibitem[Szegedy et~al.(2016)Szegedy, Vanhoucke, Ioffe, Shlens, and
  Wojna]{Szegedy2016RethinkingTI}
Szegedy, C., Vanhoucke, V., Ioffe, S., Shlens, J., and Wojna, Z.
\newblock Rethinking the inception architecture for computer vision.
\newblock \emph{2016 IEEE Conference on Computer Vision and Pattern Recognition
  (CVPR)}, pp.\  2818--2826, 2016.

\bibitem[Tan et~al.(2016)Tan, Niculae, Danescu-Niculescu-Mizil, and
  Lee]{tan2016winning}
Tan, C., Niculae, V., Danescu-Niculescu-Mizil, C., and Lee, L.
\newblock Winning arguments: Interaction dynamics and persuasion strategies in
  good-faith online discussions.
\newblock In \emph{Proceedings of the 25th International Conference on World
  Wide Web (WWW)}, pp.\  613--624. International World Wide Web Conferences
  Steering Committee, 2016.

\bibitem[Thompson et~al.(2019)Thompson, Gwinnup, Khayrallah, Duh, and
  Koehn]{thompson-etal-2019-overcoming}
Thompson, B., Gwinnup, J., Khayrallah, H., Duh, K., and Koehn, P.
\newblock Overcoming catastrophic forgetting during domain adaptation of neural
  machine translation.
\newblock In \emph{Proceedings of the 2019 Conference of the North American
  Chapter of the Association for Computational Linguistics: Human Language
  Technologies (NAACL-HLT)}, 2019.

\bibitem[Tseran et~al.(2018)Tseran, Khan, Harada, and Bui]{tseran2018natural}
Tseran, H., Khan, M.~E., Harada, T., and Bui, T.~D.
\newblock Natural variational continual learning.
\newblock In \emph{Continual Learning Workshop@ NeurIPS}, 2018.

\bibitem[Utama et~al.(2020)Utama, Moosavi, and Gurevych]{Utama2020TowardsDN}
Utama, P.~A., Moosavi, N., and Gurevych, I.
\newblock Towards debiasing nlu models from unknown biases.
\newblock \emph{arXiv preprint arXiv:2009.12303}, 2020.

\bibitem[Vaibhav et~al.(2019)Vaibhav, Singh, Stewart, and
  Neubig]{vaibhav2019improving}
Vaibhav, V., Singh, S., Stewart, C., and Neubig, G.
\newblock Improving robustness of machine translation with synthetic noise.
\newblock In \emph{Proceedings of the 2019 Conference of the North American
  Chapter of the Association for Computational Linguistics: Human Language
  Technologies (NAACL-HLT)}, 2019.

\bibitem[Van~Erven \& Harremos(2014)Van~Erven and Harremos]{van2014renyi}
Van~Erven, T. and Harremos, P.
\newblock R{\'e}nyi divergence and kullback-leibler divergence.
\newblock \emph{IEEE Transactions on Information Theory}, 60\penalty0
  (7):\penalty0 3797--3820, 2014.

\bibitem[Van~Oord et~al.(2016)Van~Oord, Kalchbrenner, and
  Kavukcuoglu]{van2016pixel}
Van~Oord, A., Kalchbrenner, N., and Kavukcuoglu, K.
\newblock Pixel recurrent neural networks.
\newblock In \emph{Proceedings of the 33rd International Conference on Machine
  Learning (ICML)}, pp.\  1747--1756. PMLR, 2016.

\bibitem[Vaswani et~al.(2017)Vaswani, Shazeer, Parmar, Uszkoreit, Jones, Gomez,
  Kaiser, and Polosukhin]{vaswani2017attention}
Vaswani, A., Shazeer, N., Parmar, N., Uszkoreit, J., Jones, L., Gomez, A.~N.,
  Kaiser, {\L}., and Polosukhin, I.
\newblock Attention is all you need.
\newblock In \emph{Proceedings of the 31st Annual Conference on Neural
  Information Processing Systems (NIPS)}, pp.\  5998--6008, 2017.

\bibitem[Vinyals et~al.(2016)Vinyals, Blundell, Lillicrap, Wierstra,
  et~al.]{vinyals2016matching}
Vinyals, O., Blundell, C., Lillicrap, T., Wierstra, D., et~al.
\newblock Matching networks for one shot learning.
\newblock In \emph{Proceedings of the 30th Annual Conference on Neural
  Information Processing Systems (NIPS)}, pp.\  3630--3638, 2016.

\bibitem[Wang et~al.(2019{\natexlab{a}})Wang, Pruksachatkun, Nangia, Singh,
  Michael, Hill, Levy, and Bowman]{wang2019superglue}
Wang, A., Pruksachatkun, Y., Nangia, N., Singh, A., Michael, J., Hill, F.,
  Levy, O., and Bowman, S.
\newblock Superglue: A stickier benchmark for general-purpose language
  understanding systems.
\newblock In \emph{Proceedings of the 33rd Annual Conference on Neural
  Information Processing Systems (NeurIPS)}, pp.\  3261--3275,
  2019{\natexlab{a}}.

\bibitem[Wang et~al.(2019{\natexlab{b}})Wang, Singh, Michael, Hill, Levy, and
  Bowman]{wang2018glue}
Wang, A., Singh, A., Michael, J., Hill, F., Levy, O., and Bowman, S.~R.
\newblock Glue: A multi-task benchmark and analysis platform for natural
  language understanding.
\newblock In \emph{Proceedings of the International Conference on Learning
  Representations (ICLR)}, 2019{\natexlab{b}}.

\bibitem[Wang et~al.(2020{\natexlab{a}})Wang, Singh, Michael, Hill, Levy, and
  Bowman]{glueleaderboard2020}
Wang, A., Singh, A., Michael, J., Hill, F., Levy, O., and Bowman, S.~R.
\newblock Glue benchmark leaderboard.
\newblock https://gluebenchmark.com/leaderboard, 2020{\natexlab{a}}.
\newblock URL \url{https://gluebenchmark.com/leaderboard}.
\newblock Accessed: 2020-04-16.

\bibitem[Wang et~al.(2020{\natexlab{b}})Wang, Pino, Wu, and Gu]{wang2020covost}
Wang, C., Pino, J., Wu, A., and Gu, J.
\newblock Covost: A diverse multilingual speech-to-text translation corpus.
\newblock pp.\  4197--4203, 2020{\natexlab{b}}.

\bibitem[Wang et~al.(2017)Wang, Utiyama, Liu, Chen, and
  Sumita]{wang-EtAl:2017:EMNLP20174}
Wang, R., Utiyama, M., Liu, L., Chen, K., and Sumita, E.
\newblock Instance weighting for neural machine translation domain adaptation.
\newblock In \emph{Proceedings of the 2017 Conference on Empirical Methods in
  Natural Language Processing (EMNLP)}, pp.\  1483--1489, 2017.

\bibitem[Wang et~al.(2019{\natexlab{c}})Wang, She, and
  Ward]{wang2019generative}
Wang, Z., She, Q., and Ward, T.~E.
\newblock Generative adversarial networks in computer vision: A survey and
  taxonomy.
\newblock \emph{arXiv preprint arXiv:1906.01529}, 2019{\natexlab{c}}.

\bibitem[Warstadt et~al.(2018)Warstadt, Singh, and
  Bowman]{Warstadt2018NeuralNA}
Warstadt, A., Singh, A., and Bowman, S.~R.
\newblock Neural network acceptability judgments.
\newblock \emph{arXiv preprint arXiv:1805.12471}, 2018.

\bibitem[Welinder et~al.(2010)Welinder, Branson, Mita, Wah, Schroff, Belongie,
  and Perona]{welinder2009cub}
Welinder, P., Branson, S., Mita, T., Wah, C., Schroff, F., Belongie, S., and
  Perona, P.
\newblock {Caltech-UCSD Birds 200}.
\newblock Technical Report CNS-TR-2010-001, California Institute of Technology,
  2010.

\bibitem[Widmer \& Kubat(1996)Widmer and Kubat]{widmer1996learning}
Widmer, G. and Kubat, M.
\newblock Learning in the presence of concept drift and hidden contexts.
\newblock \emph{Machine learning}, 23\penalty0 (1):\penalty0 69--101, 1996.

\bibitem[Williams et~al.(2018)Williams, Nangia, and Bowman]{williams18multinli}
Williams, A., Nangia, N., and Bowman, S.
\newblock A broad-coverage challenge corpus for sentence understanding through
  inference.
\newblock In \emph{Proceedings of the 2018 Conference of the North American
  Chapter of the Association for Computational Linguistics: Human Language
  Technologies (NAACL-HLT)}, pp.\  1112--1122, 2018.

\bibitem[Wolf et~al.(2019)Wolf, Debut, Sanh, Chaumond, Delangue, Moi, Cistac,
  Rault, Louf, Funtowicz, and Brew]{Wolf2019HuggingFacesTS}
Wolf, T., Debut, L., Sanh, V., Chaumond, J., Delangue, C., Moi, A., Cistac, P.,
  Rault, T., Louf, R., Funtowicz, M., and Brew, J.
\newblock Huggingface's transformers: State-of-the-art natural language
  processing.
\newblock \emph{ArXiv preprint arXiv:1910.03771}, 2019.

\bibitem[Wong \& Kolter(2018)Wong and Kolter]{wong2018provable}
Wong, E. and Kolter, Z.
\newblock Provable defenses against adversarial examples via the convex outer
  adversarial polytope.
\newblock In \emph{International Conference on Machine Learning}, pp.\
  5286--5295. PMLR, 2018.

\bibitem[Wu et~al.(2016)Wu, Schuster, Chen, Le, Norouzi, Macherey, Krikun, Cao,
  Gao, Macherey, et~al.]{wu2016google}
Wu, Y., Schuster, M., Chen, Z., Le, Q.~V., Norouzi, M., Macherey, W., Krikun,
  M., Cao, Y., Gao, Q., Macherey, K., et~al.
\newblock Google's neural machine translation system: Bridging the gap between
  human and machine translation.
\newblock \emph{arXiv preprint arXiv:1609.08144}, 2016.

\bibitem[Xia et~al.(2020)Xia, Field, and Tsvetkov]{xia2020demoting}
Xia, M., Field, A., and Tsvetkov, Y.
\newblock Demoting racial bias in hate speech detection.
\newblock In \emph{Proceedings of the 9th International Workshop on Natural
  Language Processing for Social Media (SocialNLP)}, pp.\  7--14, 2020.
\newblock URL \url{https://www.aclweb.org/anthology/2020.socialnlp-1.2}.

\bibitem[Xiong et~al.(2018)Xiong, Wu, Alleva, Droppo, Huang, and
  Stolcke]{xiong2017microsoft}
Xiong, W., Wu, L., Alleva, F., Droppo, J., Huang, X., and Stolcke, A.
\newblock The microsoft 2017 conversational speech recognition system.
\newblock In \emph{Proceedings of the International Conference on Acoustics,
  Speech, and Signal Processing (ICASSP)}, pp.\  5934--5938, 2018.

\bibitem[Yelp(2015)]{yelp2015dataset}
Yelp.
\newblock Yelp dataset challenge.
\newblock \url{https://www.yelp.com/dataset/challenge}, 2015.

\bibitem[Yue et~al.(2019)Yue, Zhang, Zhao, Sangiovanni-Vincentelli, Keutzer,
  and Gong]{yue2019domain}
Yue, X., Zhang, Y., Zhao, S., Sangiovanni-Vincentelli, A., Keutzer, K., and
  Gong, B.
\newblock Domain randomization and pyramid consistency: Simulation-to-real
  generalization without accessing target domain data.
\newblock In \emph{Proceedings of the IEEE/CVF International Conference on
  Computer Vision (ICCV)}, 2019.

\bibitem[Zenke et~al.(2017)Zenke, Poole, and Ganguli]{zenke2017continual}
Zenke, F., Poole, B., and Ganguli, S.
\newblock Continual learning through synaptic intelligence.
\newblock In \emph{Proceedings of the 34th International Conference on Machine
  Learning (ICML)}, pp.\  3987--3995, 2017.

\bibitem[Zhang \& Zong(2016)Zhang and Zong]{zhang-zong:2016:EMNLP2016}
Zhang, J. and Zong, C.
\newblock Exploiting source-side monolingual data in neural machine
  translation.
\newblock In \emph{Proceedings of the 2016 Conference on Empirical Methods in
  Natural Language Processing (EMNLP)}, pp.\  1535--1545, 2016.

\bibitem[Zhang et~al.(2013)Zhang, Sch{\"o}lkopf, Muandet, and
  Wang]{zhang2013domain}
Zhang, K., Sch{\"o}lkopf, B., Muandet, K., and Wang, Z.
\newblock Domain adaptation under target and conditional shift.
\newblock In \emph{Proceedings of the 30th International Conference on Machine
  Learning (ICML)}, pp.\  819--827, 2013.

\bibitem[Zhang et~al.(2015)Zhang, Zhao, and LeCun]{zhang2015character}
Zhang, X., Zhao, J., and LeCun, Y.
\newblock Character-level convolutional networks for text classification.
\newblock In \emph{Proceedings of the 29th Annual Conference on Neural
  Information Processing Systems (NIPS)}, pp.\  649--657, 2015.

\bibitem[Zhao et~al.(2018)Zhao, Dua, and Singh]{zhao2018generating}
Zhao, Z., Dua, D., and Singh, S.
\newblock Generating natural adversarial examples.
\newblock In \emph{Proceedings of the International Conference on Learning
  Representations (ICLR)}, 2018.

\bibitem[Zhou et~al.(2021)Zhou, Ma, Michel, and Neubig]{zhou2021examining}
Zhou, C., Ma, X., Michel, P., and Neubig, G.
\newblock Examining and combating spurious features under distribution shift.
\newblock In \emph{Proceedings of the 38th International Conference on Machine
  Learning (ICML)}, 2021.

\bibitem[Zhou et~al.(2020)Zhou, Yang, Hospedales, and Xiang]{zhou2020deep}
Zhou, K., Yang, Y., Hospedales, T., and Xiang, T.
\newblock Deep domain-adversarial image generation for domain generalisation.
\newblock In \emph{Proceedings of the 34th Meeting of the Association for
  Advancement of Artificial Intelligence (AAAI)}, 2020.

\bibitem[Zou et~al.(2020)Zou, Huang, Xie, Dai, and Chen]{zou2020reinforced}
Zou, W., Huang, S., Xie, J., Dai, X., and Chen, J.
\newblock A reinforced generation of adversarial examples for neural machine
  translation.
\newblock In \emph{Proceedings of the 8th Annual Meeting of the Association for
  Computational Linguistics (ACL)}, pp.\  3486--3497, 2020.

\end{thebibliography}
\bibliographystyle{bibliography/icml2020}

\clearpage
\appendix
\part*{Appendix}
\addcontentsline{toc}{part}{Appendix}

\chapter{Supplemental Material for Chapter \ref{ch:modeling_the_second_player_dro}}
\label{ch:supplemental_pdro}

\section{Reorganizing the Lagrangian $\mathbb L (\psi,\tau)$}
\label{sec:lagrangian_reorg}

Let us write the Lagrangian $\mathbb L$ explicitly:
\begin{align}
    \mathbb L (\psi,\tau)&=\E_{(x,y)\sim q_\psi}\frac{p(x,y)}{q_{\psi_0}(x,y)}\ell(x,y,\theta) - \tau \left(\KL{q_\psi}{p} - \kappa\right)\\
    &=\E_{(x,y)\sim q_\psi}\frac{p(x,y)}{q_{\psi_0}(x,y)}\ell(x,y,\theta) - \tau \E_{(x,y)\sim q_\psi}\log\frac{q_\psi(x,y)}{p(x,y)} + \tau\kappa\\
    &=\tau \E_{(x,y)\sim q_\psi}\log \left(\frac{p(x,y)e^{\frac{p(x,y)}{q_{\psi_0}(x,y)}\frac{\ell(x,y,\theta)}{\tau}}}{q_\psi(x,y)}\right) + \tau\kappa\\
    &=\tau (\kappa - \KL{q_\psi}{q^*_{\tau, \theta}}) + \log\left(\E_{(x,y)\sim p}e^{\frac{p(x,y)}{q_{\psi_0}(x,y)}\frac{\ell(x,y,\theta)}{\tau}}\right)
\end{align}
This last step requires that the log moment generating function of $\ell$ under $p$ exist for $\tau$. In most scenarios we consider, $\ell$ is typically the negative log likelihood of a neural network model, which is generally bounded. Therefore the moment generating function is defined everywhere.

Note that the KL term is the only one dependent on $\psi$, therefore maximizing $\mathbb L$ for $\psi$ is equivalent to maximizing $- \KL{q_\psi}{q^*_{\tau, \theta}}$, in other words minimizing $\KL{q_\psi}{q^*_{\tau, \theta}}$

\chapter{Supplemental Material for Chapter \ref{ch:regularizing_trajectories}}
\label{ch:supplemental_07}
\section{The Standard Gradient Update (Equation \ref{eqn:std_update})}
\label{sec:derivation_std}

We derive the standard update in Eq. \ref{eqn:std_update} by solving the Lagrangian $\mathbb L$ in Eq. \ref{eqn:lagrangian_std} for $\delta$. Given that its first and second derivatives with respect to $\delta$ are:

\begin{equation*}
    \begin{split}
        \nabla\mathbb L&=\nabla \mathcal L_T + 2\mu\delta\\
        \nabla^2\mathbb L &=2\mu I
    \end{split}
\end{equation*}

the problem is trivially strictly convex and its global minimizer $\delta^*$ satisfies:

\begin{equation*}
\evalat{\nabla \mathbb L}{\delta^*}=0 \Longleftrightarrow \delta^*=-\frac 1 {2\mu}\nabla\mathcal L_T
\end{equation*}
$\square$

\section{Equivalence of the Hessian of the KL Divergence and the Fisher Information Matrix}

To simplify notation, let us perform the change of variables $\theta+\delta\rightarrow x$. We show that the Hessian of the KL coincides with the Fisher on $\theta$: in other words, $\evalat{\nabla^2 \text{KL}(p_\theta\Vert p_x)}{x=\theta}=\mathbb E_{p_\theta}[(\nabla\log p_\theta)(\nabla\log p_\theta)^\intercal]$. Under mild regularity assumptions\footnote{Essentially allowing us to interchange derivatives and integrals.}, we can write (all derivatives are taken with respect to variable $x$):

\begin{equation*}
    \begin{split}
        \nabla^2 \text{KL}(p_\theta\Vert p_x)&=\underbrace{\nabla^2 \mathbb E_{p_\theta} [\log p_\theta]}_{=0} - \nabla^2 \mathbb E_{p_\theta} [\log p_x]\\
        &= -\mathbb E_{p_\theta} [\nabla^2\log p_x]
    \end{split}
\end{equation*}

Now note that $\nabla^2 \log p_x$ can be rewritten via standard derivatives manipulations as $\frac{\nabla^2 p_x}{p_x} - \frac{(\nabla p_x)(\nabla p_x)^\intercal}{p_x^2}$. This leads to:

\begin{equation*}
\nabla^2 \text{KL}(p_\theta\Vert p_x)=-\mathbb E_{p_\theta}\left[\frac{\nabla^2 p_x}{p_x}\right] + \mathbb E_{p_\theta}\left[\left(\frac{\nabla p_x}{p_x}\right)\left(\frac{\nabla p_x}{p_x}\right)^\intercal\right]
\end{equation*}

When taken at $\theta$, the first term evaluates to\footnote{We abuse notation and write $\evalat{\nabla p_x}{x=\theta}$ as $\nabla p_\theta$}~:

\begin{equation*}
    \begin{split}
        \mathbb E_{p_\theta}\left[\frac{\nabla^2 p_x}{p_x}\right]\biggr\vert_{x=\theta}&=\left[\int p_\theta(z)\frac{\nabla^2 p_x(z)}{p_x(z)}dz\right]\biggr\vert_{x=\theta}\\
        &=\int p_\theta(z)\frac{\nabla^2 p_\theta(z)}{p_\theta(z)}dz\\
        &= \int \nabla^2 p_\theta(z)dz\\
        &= \nabla^2\underbrace{\int p_\theta(z)dz}_{=1}=0\\
    \end{split}
\end{equation*}

By using the identity $\frac{\nabla p_x}{p_x}=\nabla \log p_x$ and evaluating at $x=\theta$, the second term gives us:

\begin{equation*}
\evalat{\nabla^2 \text{KL}(p_\theta\Vert p_x)}{x=\theta}=\mathbb E_{p_\theta}[(\nabla\log p_\theta)(\nabla\log p_\theta)^\intercal]
\end{equation*}

$\square$

\section{Obtaining the Co-natural Update (Equation \ref{eqn:conatural_gradient})}

We solve the Lagrangian from Eq. \ref{eqn:lagrangian_fisher} in a similar fashion as in \ref{sec:derivation_std}. First we compute its gradient and Hessian with respect to $\delta$:

\begin{equation*}
    \begin{split}
        \nabla\mathbb L&=\nabla \mathcal L_T + 2\mu\delta + \nu F^S_\theta\delta\\
        &=\nabla \mathcal L_T + (\nu F^S_\theta + 2\mu I)\delta\\
        \nabla^2\mathbb L &=(\nu F^S_\theta + 2\mu I)
    \end{split}
\end{equation*}

While not as straightforwardly as the one in \ref{sec:derivation_std}, this problem is also strongly convex: indeed $F^S_\theta$ is positive semi-definite (as an expectation of PSD matrices) and the addition of $\mu I$ ensures that 
$\nabla^2\mathbb L$ is positive definite. We find the unique solution by solving:

\begin{equation*}
    \begin{split}
        \evalat{\nabla\mathbb L}{\delta^*}=0&\Longleftrightarrow \nabla \mathcal L_T + (\nu F^S_\theta + 2\mu I)\delta^*=0\\
        &\Longleftrightarrow\delta^*=-[\nu F^S_\theta+2\mu I]^{-1} \nabla \mathcal L_T\\
    \end{split}
\end{equation*}

Set $\lambda:=\frac 1 {\nu}$ and $\alpha:=\frac \mu \nu$ to get Eq. \ref{eqn:conatural_gradient} $\square$

\end{document}